\documentclass[10pt,journal,compsoc]{IEEEtran}
\usepackage{amsthm}
\usepackage{mathtools}
\usepackage{ragged2e}
\usepackage{adjustbox}
\usepackage[colorlinks=true,allcolors=cyan,pagebackref=true]{hyperref}

\usepackage{comment}
\usepackage{siunitx}
\usepackage{relsize}
\usepackage{ifthen}

\usepackage[caption=false]{subfig}

\usepackage[vlined,ruled,linesnumbered]{algorithm2e}
\usepackage{graphics} 
\usepackage{rotating}
\usepackage{color}
\usepackage{enumerate}
\usepackage[T1]{fontenc}
\usepackage{psfrag}
\usepackage{epsfig} 
\usepackage{booktabs}
\usepackage{graphicx,url}
\usepackage{array}
\usepackage{latexsym}
\usepackage{amsfonts}
\usepackage{amsmath}
\usepackage{amssymb}
\usepackage{xstring}
\usepackage[noend]{algorithmic}
\usepackage{xcolor}
\usepackage{prettyref}
\usepackage{flexisym}
\usepackage{bigdelim}
\usepackage{breqn} 
\usepackage{listings}

\usepackage{enumitem}
\usepackage{xspace}
\usepackage{bm}
\graphicspath{{./figures/}}
\usepackage{tikz}
\usetikzlibrary{matrix,calc}

\usepackage{mdwlist}

\let\itemize\undefined
\makecompactlist{itemize}{stditemize}

\newrefformat{prob}{Problem\,\ref{#1}}
\newrefformat{def}{Definition\,\ref{#1}}
\newrefformat{sec}{Section\,\ref{#1}}
\newrefformat{sub}{Section\,\ref{#1}}
\newrefformat{prop}{Proposition\,\ref{#1}}
\newrefformat{app}{Appendix\,\ref{#1}}
\newrefformat{alg}{Algorithm\,\ref{#1}}
\newrefformat{cor}{Corollary\,\ref{#1}}
\newrefformat{thm}{Theorem\,\ref{#1}}
\newrefformat{lem}{Lemma\,\ref{#1}}
\newrefformat{fig}{Fig.\,\ref{#1}}
\newrefformat{tab}{Table\,\ref{#1}}

\newtheorem{theorem}{Theorem}

\newtheorem{assumption}[theorem]{Assumption}
\newtheorem{definition}[theorem]{Definition}
\newtheorem{proposition}[theorem]{Proposition}
\newtheorem{remark}[theorem]{Remark}
\newtheorem{example}[theorem]{Example}

\newcommand{\cf}{\emph{cf.}\xspace}

\newcommand{\bdmath}{\begin{dmath}}
\newcommand{\edmath}{\end{dmath}}
\newcommand{\beq}{\begin{equation}}
\newcommand{\eeq}{\end{equation}}
\newcommand{\bdm}{\begin{displaymath}}
\newcommand{\edm}{\end{displaymath}}
\newcommand{\bea}{\begin{eqnarray}}
\newcommand{\eea}{\end{eqnarray}}
\newcommand{\beal}{\beq \begin{array}{ll}}
\newcommand{\eeal}{\end{array} \eeq}
\newcommand{\beas}{\begin{eqnarray*}}
\newcommand{\eeas}{\end{eqnarray*}}
\newcommand{\ba}{\begin{array}}
\newcommand{\ea}{\end{array}}
\newcommand{\bit}{\begin{itemize}}
\newcommand{\eit}{\end{itemize}}
\newcommand{\ben}{\begin{enumerate}}
\newcommand{\een}{\end{enumerate}}

\newcommand{\calA}{{\cal A}}
\newcommand{\calB}{{\cal B}}
\newcommand{\calC}{{\cal C}}

\newcommand{\calE}{{\cal E}}
\newcommand{\calF}{{\cal F}}
\newcommand{\calG}{{\cal G}}
\newcommand{\calH}{{\cal H}}
\newcommand{\calI}{{\cal I}}
\newcommand{\calJ}{{\cal J}}
\newcommand{\calK}{{\cal K}}

\newcommand{\calN}{{\cal N}}
\newcommand{\calO}{{\cal O}}
\newcommand{\calP}{{\cal P}}

\newcommand{\calR}{{\cal R}}
\newcommand{\calS}{{\cal S}}

\newcommand{\calV}{{\cal V}}

\newcommand{\calX}{{\cal X}}
\newcommand{\calZ}{{\cal Z}}

\newcommand{\M}[1]{{\bm #1}} 
\renewcommand{\boldsymbol}[1]{{\bm #1}}

\newcommand{\wrt}{w.r.t.\xspace}

\newcommand{\grayout}[1]{{\color{gray} #1}}

\newcommand{\hiddenText}{{\color{gray} hidden text.}}
\newcommand{\hideWithText}[1]{\hiddenText}

\newcommand{\kron}{\otimes}

\newcommand{\one}{ {\mathbf{1}} }

\DeclareMathOperator*{\argmax}{arg\,max}
\DeclareMathOperator*{\argmin}{arg\,min}

\newcommand{\norm}[1]{\left\| #1 \right\|}

\newcommand{\tran}{^{\mathsf{T}}}

\newcommand{\trace}[1]{\mathrm{tr}\left(#1\right)}

\newcommand{\rank}[1]{\mathrm{rank}\left(#1\right)}

\newcommand{\inv}{^{-1}}

\newcommand{\zero}{{\mathbf 0}}
\newcommand{\eye}{{\mathbf I}}

\newcommand{\Real}[1]{ { {\mathbb R}^{#1} } }

\newcommand{\SOtwo}{\ensuremath{\mathrm{SO}(2)}\xspace}
\newcommand{\SOthree}{\ensuremath{\mathrm{SO}(3)}\xspace}

\newcommand{\MA}{\M{A}}
\newcommand{\MB}{\M{B}}
\newcommand{\MC}{\M{C}}

\newcommand{\MM}{\M{M}}

\newcommand{\MR}{\M{R}}
\newcommand{\MS}{\M{S}}

\newcommand{\MX}{\M{X}}

\newcommand{\MZ}{\M{Z}}

\newcommand{\va}{\boldsymbol{a}} 
 
\newcommand{\vb}{\boldsymbol{b}}
\newcommand{\vc}{\boldsymbol{c}}

\newcommand{\vp}{\boldsymbol{p}}
\newcommand{\vq}{\boldsymbol{q}}
\newcommand{\vr}{\boldsymbol{r}}

\newcommand{\vu}{\boldsymbol{u}}
\newcommand{\vv}{\boldsymbol{v}}
\newcommand{\vt}{\boldsymbol{t}}
\newcommand{\vxx}{\boldsymbol{x}} 
\newcommand{\vy}{\boldsymbol{y}}
\newcommand{\vw}{\boldsymbol{w}}

\newcommand{\vgamma}{\boldsymbol{\gamma}}  

\newcommand{\vtheta}{\boldsymbol{\theta}}
\newcommand{\valpha}{\boldsymbol{\alpha}}

\newcommand{\scenario}[1]{{\smaller \sf#1}\xspace}

\newcommand{\blue}[1]{{\color{blue}#1}}

\newcommand{\red}[1]{{\color{red}#1}}

\newcommand{\linkToPdf}[1]{\href{#1}{\blue{(pdf)}}}
\newcommand{\linkToPpt}[1]{\href{#1}{\blue{(ppt)}}}
\newcommand{\linkToCode}[1]{\href{#1}{\blue{(code)}}}
\newcommand{\linkToWeb}[1]{\href{#1}{\blue{(web)}}}
\newcommand{\linkToVideo}[1]{\href{#1}{\blue{(video)}}}
\newcommand{\linkToMedia}[1]{\href{#1}{\blue{(media)}}}
\newcommand{\award}[1]{\xspace} 

\newcommand{\vz}{\boldsymbol{z}}

\newcommand{\vthetastar}{{\vtheta}^{\star}}

\newcommand{\gitlink}{https://github.com/MIT-SPARK/CertifiablyRobustPerception}
\newcommand{\sym}[1]{\mathbb{S}^{#1}}
\newcommand{\pd}[1]{\sym{#1}_{++}}
\newcommand{\psd}[1]{\sym{#1}_{+}}
\newcommand{\ball}{\calB}
\renewcommand{\int}{\mathbb{Z}}
\newcommand{\nnint}{\int_{+}}
\newcommand{\pint}{\int_{++}}
\renewcommand{\deg}[1]{\mathrm{deg}\parentheses{#1}}
\newcommand{\bbX}{\mathbb{X}}
\newcommand{\bbD}{\mathbb{D}}
\newcommand{\bbQ}{\mathbb{Q}}
\newcommand{\dimrot}{q}
\newcommand{\mySOd}{\mathrm{SO}(\dimrot)}

\newcommand{\MXstar}{\MX^\star}
\newcommand{\vystar}{\vy^\star}

\newcommand{\MSstar}{\MS^\star}
\newcommand{\parentheses}[1]{\left(#1\right)}

\newcommand{\MRstar}{\MR^{\star}}
\newcommand{\vtstar}{\vt^{\star}}
\newcommand{\MRgt}{\MR^{\circ}}
\newcommand{\vtgt}{\vt^{\circ}}
\newcommand{\vcgt}{\vc^{\circ}}
\newcommand{\MRin}{\MR_{\mathrm{in}}}
\newcommand{\MRnoise}{\MR_{\varepsilon}}
\newcommand{\vvarepsilon}{\boldsymbol{\varepsilon}}

\newcommand{\bmat}{\left[\begin{array}}
\newcommand{\emat}{\end{array}\right]}

\newcommand{\sign}{\mathrm{sgn}}

\newcommand{\half}{\frac{1}{2}}
\newcommand{\supp}{Supplementary Material\xspace}

\newcommand{\tldMR}{\widetilde{\MR}}

\newcommand{\meq}{\mathrm{eq}}
\newcommand{\mathmom}{\mathrm{mom}}
\newcommand{\mathreq}{\mathrm{req}}
\newcommand{\mathloc}{\mathrm{loc}}

\newcommand{\ie}{\emph{i.e.},\xspace}
\newcommand{\eg}{\emph{e.g.},\xspace}
\newcommand{\ceil}[1]{\left\lceil #1 \right\rceil}
\newcommand{\abs}[1]{\left|#1\right|}
\newcommand{\hatMX}{\widehat{\MX}}
\newcommand{\hatMS}{\widehat{\MS}}
\newcommand{\hatvy}{\widehat{\vy}}
\newcommand{\hatvxx}{\widehat{\vxx}}
\newcommand{\hatvtheta}{\widehat{\vtheta}}
\renewcommand{\vthetastar}{\vtheta^\star}

\newcommand{\bbR}{\mathbb{R}}
\renewcommand{\norm}[1]{\left\lVert #1 \right\rVert}
\newcommand{\inprod}[2]{\left\langle #1, #2 \right\rangle}

\newcommand{\calAadj}{\calA^{*}}

\newcommand{\SOd}{\ensuremath{\mathrm{SO}(d)}\xspace}

\newcommand{\hatp}{\widehat{p}}

\newcommand{\plb}{\underline{p}}

\newcommand{\stride}{\scenario{STRIDE}}
\newcommand{\strideplus}{\scenario{STRIDE}}
\newcommand{\fpfh}{\scenario{FPFH}}
\newcommand{\robin}{\scenario{ROBIN}}
\newcommand{\gsnet}{\scenario{GSNet}}

\newcommand{\subcapsize}{\smaller}

\newcommand{\normalize}{\texttt{normalize}}

\newcommand{\namelong}{SpecTrahedral pRojected gradIent Descent along vErtices}

\newcommand{\cbrace}[1]{\left\{ #1\right\}}
\newcommand{\sbracket}[1]{\left[ #1\right]}

\newcommand{\lbfgs}{L-BFGS}
\newcommand{\admmplus}{\scenario{ADMM+}}

\newcommand{\tldvxx}{\widetilde{\vxx}}
\newcommand{\tldvxxstar}{\tldvxx^\star}

\newcommand{\sdpnal}{\scenario{SDPNAL+}}
\newcommand{\sdpt}{\scenario{SDPT3}}
\newcommand{\mosek}{\scenario{MOSEK}}
\newcommand{\cdcs}{\scenario{CDCS}}

\newcommand{\tssos}{\scenario{TSSOS}}

\newcommand{\ransac}{\scenario{RANSAC}}
\newcommand{\gnc}{\scenario{GNC}}
\newcommand{\threedmatch}{\scenario{3DMatch}}
\newcommand{\speed}{\scenario{SPEED}}
\newcommand{\apollo}{\scenario{ApolloScape}}
\newcommand{\pascal}{\scenario{PASCAL3D+}}
\newcommand{\bunny}{\scenario{Bunny}}
\newcommand{\teddybear}{\scenario{TeddyBear}}
\newcommand{\homebrew}{\scenario{HomebrewedDB}}
\newcommand{\mtls}{\mathrm{TLS}}
\newcommand{\mmc}{\mathrm{MC}}
\newcommand{\mgm}{\mathrm{GM}}
\newcommand{\mtb}{\mathrm{TB}}
\newcommand{\mlone}{\mathrm{L1}}
\newcommand{\mhuber}{\mathrm{HB}}
\newcommand{\madaptive}{\mathrm{ADT}}
\newcommand{\mono}[1]{\mathrm{mono}\parentheses{#1}}

\newcommand{\nchoosek}[2]{\left(\substack{#1 \\ #2} \right)}

\newcommand{\mmom}{m_{\mathrm{mom}}}
\newcommand{\mmomi}{m_{\mathrm{mom}_i}}

\newcommand{\mathovlp}{\mathrm{ovlp}}
\newcommand{\pfeas}{\eta_p}
\newcommand{\dfeas}{\eta_d}
\newcommand{\gap}{\eta_g}
\newcommand{\kkt}{\eta_{\max}}
\newcommand{\subopt}{\eta_s}
\newcommand{\ee}[1]{\mathrm{e}{#1}}
\newcommand{\fmincon}{\texttt{fmincon}}
\newcommand{\rounding}{\texttt{rounding}}

\newcommand{\usphere}[1]{\calS^{#1}}

\newcommand{\manopt}{\scenario{Manopt}}

\newcommand{\fstar}{f^\star}

\newcommand{\vxxstar}{\vxx^{\star}}

\newcommand{\pstar}{p^\star}

\newcommand{\pgm}{PGD}
\newcommand{\trinum}{\mathfrak{t}}

\newcommand{\nlp}{\texttt{nlp}}

\newcommand{\bbP}{\mathbb{P}}
\newcommand{\bbA}{\mathbb{A}}
\newcommand{\polyring}[1]{\mathbb{R}[#1]}
\newcommand{\binomial}[2]{\underline{#1}_{#2}}
\newcommand{\binomialc}[2]{(\underline{#1})_{#2}}

\newcommand{\vcat}{\,;\,} 
\newcommand{\barMX}{\widebar{\MX}}
\newcommand{\barMS}{\widebar{\MS}}
\newcommand{\barvxx}{\widebar{\vxx}}
\newcommand{\barvtheta}{\widebar{\vtheta}}
\newcommand{\lowvxx}{\widehat{\vxx}}
\newcommand{\lowvtheta}{\widehat{\vtheta}}
\newcommand{\lowxtheta}{\widehat{\vxx\vtheta}}

\newcommand{\setsdpp}{\calF_{\mathrm{P}}}

\newcommand{\tol}{\texttt{tol}}

\newcommand{\nrCosts}{{seven}\xspace}
\newcommand{\regularizer}{\psi(\vxx)} 
\newcommand{\revise}[1]{#1}

\newcommand{\practical}{scalable\xspace}
\newcommand{\maybeOmit}[1]{} 

\ifCLASSOPTIONcompsoc
  \usepackage[nocompress]{cite}
\else
  \usepackage{cite}
\fi
\ifCLASSINFOpdf
\else
\fi
\makeatletter
\let\save@mathaccent\mathaccent
\newcommand*\if@single[3]{%
  \setbox0\hbox{${\mathaccent"0362{#1}}^H$}%
  \setbox2\hbox{${\mathaccent"0362{\kern0pt#1}}^H$}%
  \ifdim\ht0=\ht2 #3\else #2\fi
  }
\newcommand*\rel@kern[1]{\kern#1\dimexpr\macc@kerna}
\newcommand*\widebar[1]{\@ifnextchar^{{\wide@bar{#1}{0}}}{\wide@bar{#1}{1}}}
\newcommand*\wide@bar[2]{\if@single{#1}{\wide@bar@{#1}{#2}{1}}{\wide@bar@{#1}{#2}{2}}}
\newcommand*\wide@bar@[3]{%
  \begingroup
  \def\mathaccent##1##2{%
    \let\mathaccent\save@mathaccent
    \if#32 \let\macc@nucleus\first@char \fi
    \setbox\z@\hbox{$\macc@style{\macc@nucleus}_{}$}%
    \setbox\tw@\hbox{$\macc@style{\macc@nucleus}{}_{}$}%
    \dimen@\wd\tw@
    \advance\dimen@-\wd\z@
    \divide\dimen@ 3
    \@tempdima\wd\tw@
    \advance\@tempdima-\scriptspace
    \divide\@tempdima 10
    \advance\dimen@-\@tempdima
    \ifdim\dimen@>\z@ \dimen@0pt\fi
    \rel@kern{0.6}\kern-\dimen@
    \if#31
      \overline{\rel@kern{-0.6}\kern\dimen@\macc@nucleus\rel@kern{0.4}\kern\dimen@}%
      \advance\dimen@0.4\dimexpr\macc@kerna
      \let\final@kern#2%
      \ifdim\dimen@<\z@ \let\final@kern1\fi
      \if\final@kern1 \kern-\dimen@\fi
    \else
      \overline{\rel@kern{-0.6}\kern\dimen@#1}%
    \fi
  }%
  \macc@depth\@ne
  \let\math@bgroup\@empty \let\math@egroup\macc@set@skewchar
  \mathsurround\z@ \frozen@everymath{\mathgroup\macc@group\relax}%
  \macc@set@skewchar\relax
  \let\mathaccentV\macc@nested@a
  \if#31
    \macc@nested@a\relax111{#1}%
  \else
    \def\gobble@till@marker##1\endmarker{}%
    \futurelet\first@char\gobble@till@marker#1\endmarker
    \ifcat\noexpand\first@char A\else
      \def\first@char{}%
    \fi
    \macc@nested@a\relax111{\first@char}%
  \fi
  \endgroup
}
\makeatother

\begin{document}

%

\title{\LARGE \revise{Certifiably Optimal Outlier-Robust Geometric Perception: \\ Semidefinite Relaxations and Scalable Global Optimization}}
%
%
%
%

\author{Heng Yang,~\IEEEmembership{Student Member,~IEEE,}
        and~Luca~Carlone,~\IEEEmembership{Senior Member,~IEEE}
\IEEEcompsocitemizethanks{\IEEEcompsocthanksitem H. Yang and L. Carlone are with the Laboratory for Information and Decision Systems (LIDS), Massachusetts Institute of Technology, Cambridge,
MA, 02139.\protect
\ E-mail: {\tt \{hankyang,lcarlone\}@mit.edu}
}
\thanks{Code: \url{\gitlink}}
}

%
%

\markboth{}%
{Yang \MakeLowercase{\textit{and}} Carlone: Certifiably Optimal Outlier-Robust Geometric Perception}
\IEEEtitleabstractindextext{%
\vspace{-4mm}
\justify
\begin{abstract}

We propose the first general and \practical framework to design \emph{certifiable} algorithms for robust geometric perception in the presence of outliers.
Our first contribution is to show that estimation using common robust costs, such as truncated least squares (TLS), maximum consensus, Geman-McClure, Tukey's biweight, among others, can be reformulated as \emph{polynomial optimization problems} (POPs). 
By focusing on the TLS cost, our second contribution is to exploit \emph{sparsity} in the POP and propose a sparse \emph{semidefinite programming} (SDP) relaxation that is much smaller than the standard Lasserre's hierarchy while preserving \revise{empirical} \emph{exactness}, {\ie} the SDP recovers the optimizer of the nonconvex POP with an \emph{optimality certificate}. 
Our third contribution is to solve the SDP relaxations at an unprecedented scale and accuracy by presenting {\strideplus}, a solver that blends \emph{global descent} on the convex SDP with fast \emph{local search} on the nonconvex POP. 
Our fourth contribution is an evaluation of the proposed framework on six geometric perception problems including single and multiple rotation averaging, point cloud and mesh registration, absolute pose estimation, and category-level object pose and shape estimation. Our experiments demonstrate that (i) our sparse SDP relaxation is \revise{empirically} exact with up to $60\%$--$90\%$ outliers across applications; (ii) while still being far from real-time, {\strideplus} is up to $100$ times faster than existing SDP solvers on medium-scale problems, and is the only solver that can solve large-scale SDPs with hundreds of thousands of constraints to high accuracy; (iii) {\strideplus} \revise{safeguards} existing fast heuristics for robust estimation (\eg~{\ransac} or {Graduated Non-Convexity}), \ie it certifies global optimality if the heuristic estimates are optimal, or detects and allows escaping local optima when the heuristic estimates are suboptimal.

\end{abstract}

}

\maketitle

\begin{tikzpicture}[overlay, remember picture]
\path (current page.north east) ++(-2.0,-0.4) node[below left] {
This paper has been accepted for publication in the IEEE Transactions on Pattern Analysis and Machine Intelligence.
};
\end{tikzpicture}
\begin{tikzpicture}[overlay, remember picture]
\path (current page.north east) ++(-2.5,-0.8) node[below left] {
Please cite the paper as: H. Yang and L. Carlone, ``Certifiably Optimal Outlier-Robust Geometric Perception:
};
\end{tikzpicture}
\begin{tikzpicture}[overlay, remember picture]
\path (current page.north east) ++(-1.2,-1.2) node[below left] {
Semidefinite Relaxations and Scalable Global Optimization'', \emph{IEEE Transactions on Pattern Analysis and Machine Intelligence}, 2022.
};
\end{tikzpicture}

\IEEEdisplaynontitleabstractindextext

%
\IEEEpeerreviewmaketitle

\IEEEraisesectionheading{
\section{Introduction}
\label{sec:introduction}}
\IEEEPARstart{G}EOMETRIC perception, the task of estimating unknown geometric models ({\eg} object poses, rotations, 3D structure, robot trajectory) from sensor measurements ({\eg} images, point clouds, relative poses), is a fundamental problem in computer vision and robotics. It finds extensive applications to object detection and localization \cite{Yang19rss-teaser}, motion estimation and 3D reconstruction \cite{Choi15cvpr-robustReconstruction}, simultaneous localization and mapping (SLAM) \cite{Rosen19IJRR-sesync} and structure from motion (SfM)~\cite{Schonberger16cvpr-SfMRevisited}, 
virtual and augmented reality \cite{Klein07ismar-PTAM}, and medical imaging \cite{Audette00mia-surveyMedical}, to name a few.

A modern machine perception pipeline includes a \emph{perception front-end} that extracts, describes, and matches relevant features from raw sensor data, and a \emph{perception back-end} that estimates the geometric models of interest given the putative feature matches. In practice, due to various sources of imperfections and uncertainties ({\eg} sensor failures, incorrect detections and matchings by hand-crafted or deep-learned features), a large amount of \emph{outliers} ---measurements that tell no or little information about the underlying geometric models--- are generated by the front-end. Therefore, designing an \emph{outlier-robust} back-end that can tolerate large amounts of outliers, also known as \emph{robust fitting} \cite{Chin17slcv-maximumConsensusAdvances} in computer vision and \emph{robust state estimation} \cite{Barfoot17book} in robotics, has been a longstanding quest in both communities. 

Unfortunately, from a theoretical standpoint, performing robust estimation by discerning \emph{inliers} ({\ie} the correct and useful measurements) from outliers, is known to be NP-hard and \emph{inapproximable} due to its combinatorial nature \cite{Chin17slcv-maximumConsensusAdvances,Antonante20TRO-outlier,Enqvist15IJCV-tractableRobustEstimation,Chin18eccv-robustFitting}. Consequently, existing algorithms for outlier-robust estimation are mostly divided into \emph{fast heuristics}, {\eg} \ransac \cite{Fischler81,Chum03-LORANSAC,Barath18-gcRANSAC} and graduated non-convexity (\gnc) \cite{Yang20ral-gnc,Black96ijcv-unification,Blake1987book-visualReconstruction}, that are efficient but offer no optimality guarantees, and \emph{global solvers}, {\eg} Branch-and-Bound \cite{Yang16pami-goicp} and mixed-integer programming \cite{Izatt17isrr-MIPregistration,Li09cvpr-robustFitting}, that guarantee optimality but run in worst-case exponential time. Although in some cases it is acceptable to trade off \emph{optimality} (hence robustness) for \emph{efficiency}, real-time safety-critical applications ---such as autonomous driving and space robotics--- pose high demands for \emph{efficient global optimality}. 

The conflict between the {fundamental intractability} of robust estimation and the demand for computational efficiency
calls for a paradigm shift: since it is impossible to solve all robust estimation problems in polynomial time, we argue that a useful goal is to design algorithms that perform well in typical instances and are able to \emph{certify} optimality of the resulting estimates, but at the same time can declare ``failure'' on worst-case instances rather than blindly returning an incorrect estimate. Inspired by related works \cite{Bandeira16crm,Yang20tro-teaser}, we formalize the notion of a \emph{certifiable algorithm} below.

\begin{definition}[Certifiable Algorithm]\label{def:certifiablealg}
Given an optimization problem $\mathbb{P}(\bbD)$ with input data $\bbD$, an algorithm $\mathbb{A}$ is said to be {certifiable} if (i) $\mathbb{A}$ runs in polynomial time; and after solving $\mathbb{P}(\bbD)$, $\mathbb{A}$ \revise{(ii)} either returns the global optimizer of $\mathbb{P}$ together with a certificate of optimality \revise{for common instances of $\bbD$ (empirically or provably)}, or \revise{(iii)} fails to do so \revise{for the worst instances of $\bbD$} but provides a measure of suboptimality ({\eg} a  bound on the objective value, or the distance to the global optimizer).
\end{definition}
\revise{A certifiable algorithm respects the theoretical intractability of robust estimation \cite{Chin17slcv-maximumConsensusAdvances,Antonante20TRO-outlier} in that it does \emph{not} globally optimize $\bbP(\bbD)$ for \emph{all} instances of $\bbD$ and it is allowed to fail in the worst cases (\cf (iii) of Definition \ref{def:certifiablealg}).}
\revise{However}, our notion of a certifiable algorithm is stricter than that of \cite{Yang20tro-teaser}, as it requires $\bbA$ to solve $\bbP(\bbD)$ to global optimality for common $\bbD$ \revise{(at least empirically, \cf (ii) of Definition \ref{def:certifiablealg})}. 
This requirement rules out algorithms that seldomly attain global optimality but provide suboptimality guarantees (\eg approximation algorithms \cite{Vazirani13book-approximation}).

\emph{Semidefinite relaxations} are a natural choice for designing certifiable algorithms. If the problem $\bbP$ is a \emph{polynomial optimization problem} (POP, {\ie} both its objective and constraints are polynomials), then there exists a standard semidefinite relaxation \emph{hierarchy}, known as Lasserre's hierarchy \cite{Lasserre01siopt-lasserrehierarchy}, that relaxes $\bbP$ into a hierarchy of {convex} semidefinite programs (SDPs) of increasing size. 
Each relaxation in this hierarchy can be solved in polynomial time~\cite{todd1998nesterov} and provides a measure of {suboptimality} for the resulting estimate.
Moreover, under\,mild\,technical\,conditions, 
 the suboptimality  of these relaxations becomes zero when their size is large enough, in which case we 
say the relaxation is \emph{exact}, or \emph{tight}.\footnote{ 
\revise{Lasserre's hierarchy respects the worst-case NP-hardness of POPs because one may need an SDP relaxation whose size grows exponentially with the dimension of the POP to attain certifiable optimality.}}
We provide an {accessible introduction to POPs and their relaxations in Section~\ref{sec:preliminaries}.}

Semidefinite relaxations have been successfully used to design certifiable algorithms for many geometric perception problems. The pioneering work by Kahl and Henrion \cite{Kahl07IJCV-GlobalOptGeometricReconstruction} applies Lasserre's hierarchy to solve several early perception problems including camera resectioning, homography estimation, and fundamental matrix estimation. Since then, certifiable algorithms have been designed for modern applications such as pose graph optimization \cite{Carlone16TRO-planarPGO,Rosen19IJRR-sesync}, rotation averaging \cite{Eriksson18cvpr-strongDuality,Fredriksson12accv}, triangulation \cite{Cifuentes21SIMAA-rankdeficient,Aholt12eccv-qcqptriangulation}, 3D registration \cite{Briales17cvpr-registration,Maron16tog-PMSDP,Chaudhury15Jopt-multiplePointCloudRegistration,Iglesias20cvpr-PSRGlobalOptimality}, absolute pose estimation \cite{Agostinho2019arXiv-cvxpnpl}, relative pose estimation \cite{Briales18cvpr-global2view,Zhao20pami-relativepose,Garcia21IVC-certifiablerelativepose}, hand-eye calibration \cite{Heller14icra-handeyePOP,Giamou19ral-SDPExtrinsicCalibration,Wise20MFI-certifiablyhandeye}, and category-level object perception \cite{Yang20cvpr-perfectshape,Shi21rss-pace}. Although the original formulations of the problems mentioned above are {nonconvex}, semidefinite relaxations at the \emph{lowest} relaxation order in the hierarchy are shown to be exact in practical applications. Since the SDP resulting from the lowest relaxation order can usually be solved efficiently ({\eg} below one second) by off-the-shelf SDP solvers ({\eg} \sdpt \cite{tutuncu03MP-SDPT3}, \mosek~\cite{mosek}) or the Burer-Monteiro (B-M) low-rank factorization method \cite{Burer03mp,Boumal16nips,Rosen20wafr-scalableLowRankSDP}, both efficiency and (certifiable) optimality can be obtained.

However, these successful examples of certifiable algorithms are underpinned by the restrictive assumption that \emph{the measurements are free of outliers}, which seldomly holds in practice.  Heuristics like \ransac and \gnc are typically used to filter out outliers, but it is precisely the use of such heuristics that breaks the optimality guarantee and makes the system prone to undetected failures. Although several works have attempted to design certifiable algorithms for \emph{outlier-robust} geometric perception~\cite{Wang13ima,Lajoie19ral-DCGM,Carlone18ral-convexHuber,Yang19iccv-quasar,Speciale17cvpr-MaxconLMI}, most approaches (i) are problem-specific, (ii) cannot tolerate high outlier rates ({\eg} above $70\%$) \cite{Wang13ima,Lajoie19ral-DCGM,Carlone18ral-convexHuber}, or (iii) become too large to be solved by existing SDP solvers \cite{Yang19iccv-quasar}.

{\bf Contributions}. In this paper, we propose a {general} and \practical framework for designing certifiable outlier-robust estimation algorithms that are empirically \emph{exact} with up to {$90\%$ outliers}, and present a fast SDP solver that can solve the tight relaxations at an unprecedented scale. We now describe our four contributions in detail.

{\bf (I) {Robust estimation as polynomial optimization} (Section~\ref{sec:robustandpop})}.
We investigate outlier-robust estimation with common {robust cost functions}, including truncated least squares (TLS), maximum consensus, Geman-McClure, Tukey's Biweight, L1, Huber, and Barron's adaptive kernel \cite{Barron19cvpr-adaptRobustLoss}. Our first contribution is to show that robust estimation using these  costs can be equivalently reformulated as POPs, even though the robust costs themselves are not polynomials. This result is established by introducing additional variables and manipulating the original costs to polynomials.


{\bf (II) {A sparse, but exact, semidefinite relaxation} (Section~\ref{sec:sdprelax})}.
With the POP reformulation, it is tempting to apply the standard Lasserre's hierarchy to develop certifiable algorithms for robust estimation. Nevertheless, due to the additional variables (one or two variables per measurement), even for small estimation problems with fewer than 20 measurements, the \emph{lowest-order relaxation} can already lead to SDPs that are too large for existing SDP solvers. Therefore, our second contribution is to focus on the TLS cost and show that it allows us to exploit \emph{term sparsity} of the polynomials in the POP and design a much smaller semidefinite relaxation using \emph{basis reduction}. \revise{Although exploiting sparsity of POPs is a known idea in applied mathematics \cite{Wang21SIOPT-tssos,Lasserre06siopt-correlativesparsity}, our method is more effective than existing generic-purpose techniques since it leverages the special structure of our perception problems.} Compared to the standard Lasserre's hierarchy, our sparse semidefinite relaxation leads to $100$ times reduction in the size of the SDP. Unfortunately, even with our sparse relaxation, solving the SDP using off-the-shelf SDP solvers (\eg~\mosek) is still too slow, and we can only demonstrate empirical exactness of our relaxation on small estimation problems ({\eg} $30$ measurements). 

{\bf (III) {A scalable and robust SDP solver} (Section~\ref{sec:scalableopt})}.
The limitations of existing SDP solvers lead to our third contribution, a scalable SDP solver that can \emph{certifiably optimally} solve robust estimation problems of moderate but realistic sizes ({\eg} $100$ measurements). Our solver, called \emph{SpecTrahedral pRojected gradIent Descent along vErtices} (\strideplus), 
 blends fast \emph{local search} on the nonconvex POP with \emph{global descent} on the convex SDP. Specifically, {\strideplus} follows a globally convergent trajectory driven by a \emph{projected gradient descent method} for solving the SDP, while simultaneously probing long, but \emph{safeguarded}, \emph{rank-one} “strides”, generated by fast nonlinear programming algorithms on the POP, to seek rapid descent. Notably,  fast heuristics such as {\ransac} and {\gnc} can be readily used to bootstrap {\strideplus}. Particularly, when {\ransac} and {\gnc} succeed in finding the globally optimal solution (which happens frequently in the low-outlier regime), {\strideplus} serves to certify global optimality. Otherwise, when fast heuristics converge to local minima, {\strideplus} detects suboptimality and escapes such minima. 

{\bf (IV) {Evaluation on six geometric perception problems} (Section~\ref{sec:experiments})}.
Our last contribution is to apply our framework and solver to six perception problems: single and multiple rotation averaging, point cloud and mesh registration, absolute pose estimation, and category-level object perception. With extensive experiments on synthetic and real datasets, we demonstrate (i) our sparse SDP relaxation is exact in the presence of up to $60\%$--$90\%$ outliers, (ii) while 
 still being far from real-time, {\strideplus} is up to 100 times faster than existing SDP solvers on medium-scale problems, and is the only solver than can solve large-scale SDPs with hundreds of thousands of constraints to high accuracy, (iii) {\strideplus} safeguards existing fast heuristics, \ie it certifies global optimality if the heuristic estimates are already optimal, or detects and escapes local minima otherwise.
 We showcase real examples of {\strideplus} \emph{certifiably} performing scan matching on \threedmatch \cite{Zeng17cvpr-3dmatch}, mesh registration on \homebrew \cite{Kaskman19-homebrewedDB}, satellite pose estimation on \speed \cite{Sharma19arxiv-SPEED}, and vehicle pose and shape estimation on \apollo \cite{Wang19pami-apolloscape}.

{\bf Novelty with respect to~\cite{Yang20neurips-onering,Yang21arxiv-stride}}. This paper extends and unifies the contributions presented in our previous conference papers~\cite{Yang20neurips-onering,Yang21arxiv-stride}. More in detail, we expand on~\cite{Yang20neurips-onering} by 
(i)~showing that other robust costs (beyond TLS) can be rephrased as POPs, 
(ii)~providing a more extensive comparison between (and discussion about)  Lasserre's hierarchy and the proposed sparse relaxations, 
(iii)~going beyond certification (in this paper we propose a \emph{solver}, rather than a certification approach), 
(iv)~considering a broader set of applications. 
 We also extend~\cite{Yang21arxiv-stride}, which introduced \stride, by 
 (i)~generalizing \stride to work on multi-block SDPs arising from the proposed relaxations,
 (ii)~tailoring \stride to use fast heuristics (\eg~\ransac or \gnc) as a warmstart, and
 (iii)~testing \stride on a broader range of problems. 

We remark that the main goal of this paper is \emph{not} to produce a method that outperforms problem-specific state-of-the-art algorithms in terms of robustness or efficiency. Our key contribution is instead to show that a broad 
class of robust estimation problems in geometric perception can be solved to certifiable optimality in 
polynomial time (despite their hardness), and lay out a scalable framework to build SDP relaxations, that we believe ---with further advancement of SDP solvers--- will eventually run in real time.


\subsection*{Notation} 

{\bf Scalars, vectors, matrices}. We use lowercase characters (\eg $a$) to denote real scalars, bold lowercase characters (\eg $\va$) for real (column) vectors, and bold uppercase characters (\eg $\MA$) for real matrices. $\eye_d$ denotes the identity matrix of size $d \times d$, and $\zero$ denotes the all-zero vector or matrix.
Given $\MA \in \Real{m \times n}$, $a_{ij}$ denotes the $(i,j)$-th entry of $\MA$, and $[\MA]_{\calI,\calJ}$ denotes the submatrix of $\MA$ formed by indexing rows $\calI \subseteq [m]$ and columns $\calJ \subseteq [n]$, where $[n] \triangleq \{1,\dots,n \}$ is the set of positive integers up to $n$. For a vector $\vv \in \Real{n}$, we shorthand $v_i$ for its $i$-th entry and $\vv_\calI$ for its entries indexed by $\calI \subseteq [n]$.
For $\MA,\MB \in \Real{m \times n}$, $\inprod{\MA}{\MB} \triangleq \sum_{i=1}^m \sum_{j=1}^n a_{ij} b_{ij}$ denotes the usual inner product between real matrices. $\trace{\MA} \triangleq \sum_{i=1}^n a_{ii}$ denotes the trace of a square matrix $\MA \in \Real{n \times n}$.  
We use $\norm{\cdot}$ to denote the $\ell_2$ norm of a vector and the Frobenious norm of a matrix, 
\ie $\norm{\va} \triangleq \sqrt{\inprod{\va}{\va}}$ for any $\va \in \Real{n}$ and $\norm{\MA} \triangleq \sqrt{\inprod{\MA}{\MA}}$ for any $\MA \in \Real{m \times n}$. 
$\norm{\va}_1 \triangleq \sum_{i=1}^n \abs{a_i}$ denotes the $\ell_1$ norm of a vector. $[\MA,\MB]$ denotes the \emph{horizontal} concatenation, while $[\MA \vcat \MB]$ denotes the \emph{vertical} concatenation, for proper $\MA,\MB$. For $a \in \Real{}$, the symbol $\ceil{a}$ returns the smallest integer above $a$.

{\bf Sets}. We use $\sym{n}$ to denote the space of $n \times n$ real symmetric matrices, and $\psd{n}$ (resp. $\pd{n}$) to denote the set of matrices in $\sym{n}$ that are \emph{positive semidefinite} (resp. definite). We also write $\MX \succeq 0$ (resp. $\MX \succ 0$) to indicate $\MX$ is positive semidefinite (resp. definite).  
$\usphere{d-1} \triangleq \{ \vv \in \Real{d} \mid \norm{\vv} = 1 \}$ denotes the $d$-dimensional unit sphere. 
We denote by $\SOd \triangleq \{ \MR \in \revise{\Real{d\times d}} \mid \MR\tran \MR = \eye_d, \det\parentheses{\MR} = +1 \}$ the $d$-dimensional \emph{special orthogonal group} (rotation matrices). 
$\abs{\calA}$ denotes the cardinality of a finite set $\calA$. $\nnint$ (resp. $\pint$) denotes the set of nonnegative (resp. positive) integers, and $\bbQ$ denotes the set of rational numbers. 

\section{Preliminaries}
\label{sec:preliminaries}
This section reviews key facts about multi-block semidefinite programming \cite{tutuncu03MP-SDPT3} (Section \ref{sec:pre-sdp}), 
and provides an introduction to polynomial optimization and Lasserre’s semidefinite relaxation hierarchy~\cite{Lasserre01siopt-lasserrehierarchy} (Section \ref{sec:pre-pop}).
While somewhat mathematically dense, these preliminaries are designed as a pragmatic introduction for the non-expert reader.

\subsection{Semidefinite Programming}
\label{sec:pre-sdp}
A \emph{multi-block} semidefinite programming (SDP) problem
is an optimization problem in the following \emph{primal}  form \cite{tutuncu03MP-SDPT3}:
\begin{equation}\label{eq:primalSDP}
\min_{\MX \in \bbX} \cbrace{\inprod{\MC}{\MX} \mid \calA (\MX) = \vb,\ \MX \in \calK} \tag{P}.
\end{equation}
where the variable $\MX = (\MX_1,\dots,\MX_l)$ is a collection of $l$ square matrices (the ``blocks'') with $\MX_i \in \Real{n_i \times n_i}$ for $i=1,\ldots,l$ (conveniently ordered such that $n_1\geq \dots \geq n_l$); 
the domain $\bbX \triangleq \sym{n_1} \times \dots \times \sym{n_l}$ restricts the matrices to be symmetric. 
 The objective is a linear combination of the matrices in $\MX$, \ie 
$\inprod{\MC}{\MX} \triangleq \sum_{i=1}^l \inprod{\MC_i}{\MX_i}$ (for given matrices $\MC_i\in \sym{n_i}, i=1,\ldots,l$).
The problem includes independent linear constraints $\calA (\MX) = \vb$ on $\MX$, where:
\beq
\calA (\MX) \triangleq
\sbracket{
\sum_{i=1}^l  \inprod{\MA_{i1}}{\MX_i} \vcat
\dots \vcat
\sum_{i=1}^l  \inprod{\MA_{im}}{\MX_i}
} \in \Real{m}
\eeq
for given matrices $\MA_{ij}\!\in\!\sym{n_i}, i\!=\!1,\ldots,l$ and $j\!=\!1,\dots,m$,
and $\vb\!\in\!\Real{m}$ is a given vector.
Finally, the constraint $\MX\!\in\!\calK$ enforces that each matrix in $\MX$ is positive semidefinite 
(\ie $\calK\!\triangleq\!\psd{n_1}\!\times\!\dots\!\times\!\psd{n_l}$ is a product of $l$ positive semidefinite cones). \revise{We also write $\MX \succeq 0$ to indicate each matrix in $\MX$ is positive semidefinite when $\MX$ is a collection of matrices (note that we need the notation $\MX \in \calK$ for describing details of our SDP solver).}
The feasible set of \eqref{eq:primalSDP}, denoted by $\setsdpp\!\triangleq\!\{\!\MX\!\in\!\bbX\!\mid\!\calA(\MX)\!=\!\vb,\MX\!\in\!\calK\}$, \mbox{is called a \emph{spectrahedron} \cite{Blekherman12Book-sdpandConvexAlgebraicGeometry}.} 

The Lagrangian \emph{dual} of \eqref{eq:primalSDP} is another multi-block SDP:
\begin{equation}\label{eq:dualSDP}
\max_{\vy \in \Real{m}, \MS \in \bbX} \cbrace{\inprod{\vb}{\vy} \mid \calAadj(\vy) + \MS = \MC,\ \MS \in \calK} \tag{D}
\end{equation}
where $\calAadj: \Real{m} \rightarrow \bbX$ is the adjoint of $\calA$ and is defined as:
\beq \label{eq:adjointAmultiblk}
\calAadj(\vy) \triangleq \left( \sum_{j=1}^m y_j \MA_{1j},\dots,\sum_{j=1}^m y_j \MA_{lj} \right) \in \bbX
\eeq
and the equality $\calAadj(\vy) + \MS = \MC$ is enforced block-wise.


Under mild assumptions (\eg Slater's condition \cite{Boyd04book}), \emph{strong duality} holds between \eqref{eq:primalSDP} and \eqref{eq:dualSDP} (\ie the minimum of \eqref{eq:primalSDP} equals the maximum of \eqref{eq:dualSDP}). In this case, $(\MXstar,\vystar,\MSstar) \in \bbX \times \Real{m} \times \bbX$ is simultaneously \emph{optimal} for \eqref{eq:primalSDP}-\eqref{eq:dualSDP} if and only if the following KKT conditions hold
\beal\label{eq:sdpKKT}
\text{\grayout{primal feasibility}}: & \calA(\MXstar) = \vb, \MXstar \in \calK, \\
\text{\grayout{dual feasibility}}: & \calAadj(\vystar) + \MSstar = \MC, \MSstar \in \calK, \\
\text{\grayout{complementarity}}: & \inprod{\MXstar}{\MSstar} = 0.
\eeal
The KKT conditions \eqref{eq:sdpKKT} imply strong duality because
\begin{equation}
\begin{split}
0 & = \inprod{\MXstar}{\MSstar} = \inprod{\MXstar}{\MC - \calAadj(\vystar)} \\
& = \inprod{\MC}{\MXstar} - \inprod{\calA(\MXstar)}{\vystar}  = \inprod{\MC}{\MXstar} - \inprod{\vb}{\vystar} .
\end{split}
\end{equation}
Given $(\MX,\vy,\MS) \in \calK \times \Real{m} \times \calK$, we measure its feasibility and optimality using the standard relative KKT residuals
\beal\label{eq:KKTresiduals}
\pfeas  \triangleq & \Vert \calA(\MX) - \vb \Vert  / ( 1+\norm{\vb}), \\
\dfeas  \triangleq &  \Vert {\calAadj(\vy) + \MS - \MC} \Vert  / ( 1+\norm{\MC} ),\\
\gap \triangleq & \abs{\inprod{\MC}{\MX} - \inprod{\vb}{\vy} } / ( 1 +  \abs{ \inprod{\MC}{\MX} } + \abs{ \inprod{\vb}{\vy} } ),
\eeal
where $\norm{\MX} = \sum_{i=1}^l \norm{\MX_i}$ for any $\MX \in \bbX$. We define $\kkt \triangleq \max\{\pfeas,\dfeas,\gap \}$ as the \emph{maximum KKT residual}.

{\bf SDP solvers}. 
The most robust approach for solving SDP \eqref{eq:primalSDP} (and~\eqref{eq:dualSDP}) is based on \emph{primal-dual interior point methods} (IPM) \cite{Alizadeh98siam-ipmSDP,todd1998nesterov}, \eg~{\sdpt \cite{tutuncu03MP-SDPT3} and \mosek \cite{mosek}}. For problems of small to medium size (\eg $n_1 \leq 5000, m \leq 50,000$), IPMs can solve the SDP to arbitrary accuracy, \ie $\kkt < \varepsilon$ for $\varepsilon$ arbitrarily small, with a typical per-iteration complexity $\calO(n_1^3 + m^2 n_1^2 + m^3)$.\footnote{$\calO(n_1^3)$ for spectral decomposition of dense primal and dual iterates $(\MX,\MS)$, $\calO(m^2 n_1^2)$ for forming the Schur complement system, and $\calO(m^3)$ for factorizing and solving the Schur complement system.} If each linear constraint only involves a small number of blocks (\ie for each $j=1,\dots,m$, $\MA_{ij} = \zero$ for many blocks $i=1,\dots,l$), then IPMs can be made much more efficient using \emph{dualization} \cite{Zhang20MP-sparseSDP}.\footnote{\revise{``Dualization'' switches the primal-dual data structure in numerical solvers (\eg writing the dual \eqref{eq:dualSDP} with the structure of the primal \eqref{eq:primalSDP} such that $\vy$ is represented as an unconstrained cone, or difference of two nonnegative cones, with dimension $m$) \cite{lofberg09OMS-dualize}. When sparsity exists, dualization can lead to better numerical performance.}} Nevertheless, such sparsity is not always present and generally IPMs cannot solve large-scale problems on an ordinary workstation. 

First-order methods based on ADMM and Augmented Lagrangian, \eg~{\cdcs} \cite{Zheng20MP-CDCS}, and \sdpnal \cite{Yang2015mpc-sdpnalplus}, can handle large-scale problems but exhibit slow convergence, and hence can only obtain solutions of moderate accuracy. 

For single-block problems ($l=1$) with low-rank solutions (\ie $\rank{\MXstar} \ll n_1 $) and $m = \calO(n_1)$, the Burer-Monteiro (B-M) low-rank factorization method \cite{Burer03mp,Boumal16nips} is preferable. Section \ref{sec:introduction} mentioned the success of SDP relaxations in solving \emph{outlier-free} perception problems. 
 This success is attributed to the following facts: (a) most of the SDPs arising in outlier-free estimation have $n_1 < 100$ and $m < 1000$, and can be solved by IPMs in less than one second; (b) although some SDPs (\eg~\cite{Rosen19IJRR-sesync}) can have $n_1 > 10,000$, they can be efficiently solved by B-M because the optimal solution is low-rank and $m \approx n_1$ \cite{Rosen20wafr-scalableLowRankSDP}. 

{\bf Challenges}. Unfortunately, \emph{none} of the existing solvers can solve the SDPs presented in this paper to a desired accuracy. In particular, our SDPs have $n_1 < 5000$ but $m = \calO(n_1^2)$ as large as a few millions, rendering IPMs and B-M factorization inapplicable. Moreover, our SDPs admit rank-one optimal solutions and are necessarily degenerate~\cite{Alizadeh97mp-nondegenerateSDP} (loosely speaking, degeneracy is a property that often leads to slower convergence in SDP solvers and prevents the application of B-M). Our previous work \cite{Yang21arxiv-stride} shows that first-order methods perform poorly on degenerate problems. 

\maybeOmit{
{\bf New Frontiers}. Large-scale degenerate SDPs are an unsolved puzzle in the mathematical optimization community \cite{Yang2015mpc-sdpnalplus}.
{\strideplus}, originally proposed in~\cite{Yang21arxiv-stride}, not only achieves strong performance on solving degenerate SDPs in certifiable outlier-robust perception in this paper, but also enables solving degenerate SDP relaxations from mathematics and machine learning that were previously deemed too difficult to be solved~\cite{Yang21arxiv-stride,Yang21report-STRIDE}.} 

\subsection{Polynomial Optimization and Lasserre's Hierarchy}
\label{sec:pre-pop}
{\bf Polynomial optimization}. Given $\vxx = [x_1 \vcat x_2 \vcat \ldots \vcat x_d] \in \Real{d}$,
a \emph{monomial} in $\vxx$ is a product of $x_i$'s with \emph{nonnegative} integer exponents, \ie $\vxx^{\valpha} \triangleq x_1^{\alpha_1}\cdots x_d^{\alpha_d}$ for $\valpha \in \nnint^d$ 
(for instance $x_1^2 x_5 x_6^3$ is a monomial). The sum of the exponents, $\norm{\valpha}_1$, \revise{or $\inprod{\one}{\valpha}$,} is called the \emph{degree} of the monomial (\eg the monomial $x_1^2 x_5 x_6^3$ has degree $6$). 

A real \emph{polynomial} $p(\vxx)$ is a finite sum of monomials with real coefficients. 
\revise{We shorthand $p$ in place of $p(\vxx)$ when the variable $\vxx$ is clear.}
The degree of a polynomial $p$, denoted by $\deg{p}$, is the \emph{maximum} degree of its monomials. 
The ring of real polynomials is denoted by $\polyring{\vxx}$. A standard polynomial optimization problem (POP) reads
\begin{equation}\label{eq:pop}
\pstar  \triangleq \min_{\vxx \in \Real{d}} \cbrace{p(\vxx) \ \middle\vert\ \substack{ \displaystyle h_i(\vxx) = 0, i=1,\dots,l_h \\ \displaystyle g_j(\vxx) \geq 0, j = 1,\dots,l_g } }, \tag{POP}
\end{equation} 
where $p, h_i, g_j \in \polyring{\vxx}$. Problem \eqref{eq:pop} is easily seen to be NP-hard \cite{lasserre10book-momentsOpt}, \eg it can model combinatorial binary constraints $x_i \in \{+1,-1\}$ via $x_i^2 - 1 = 0,i=1,\dots,d$.

{\bf Lasserre's hierarchy}. We now give a simplified (and somewhat less conventional) introduction to Lasserre's hierarchy that is sufficient for understanding our paper. 
For a comprehensive treatment, we refer the reader to~\cite{lasserre10book-momentsOpt}.  

We define $[\vxx]_{\kappa} \triangleq \{ \vxx^{\valpha} \mid \norm{\valpha}_1 \!\leq \!\kappa, \valpha \!\in\! \nnint^d \}$ to be the \revise{vector} of monomials of degree up to $\kappa$. For example, if $\vxx = [x_1 \vcat x_2]$ and $\kappa=2$, then $[\vxx]_2 = [1\vcat x_1 \vcat x_2 \vcat x_1^2 \vcat x_1 x_2 \vcat x_2^2]$. The dimension of $[\vxx]_\kappa$ is $\binomial{d}{\kappa} \triangleq \nchoosek{d+\kappa}{\kappa}$. With $[\vxx]_\kappa$, we form the so-called \emph{moment matrix} $\MX_{\kappa} \triangleq [\vxx]_\kappa [\vxx]_{\kappa}\tran$. 
For instance, for $\vxx = [x_1 \vcat x_2]$ and $\kappa=2$ (\cf with $[\vxx]_2$ above):
\beq\label{eq:momentMatrix}
\MX_{\kappa} \triangleq  [\vxx]_2 [\vxx]_2\tran \!=\!
\small{  
\left[
\begin{array}{cccccc}
	1 &  x_1  &  x_2  & x_1^2 & x_1 x_2 & x_2^2 \\
	x_1 &  x_1^2  &  x_1 x_2  & x_1^3 & x_1^2 x_2 & x_1 x_2^2 \\
	x_2 &  x_1 x_2  &  x_2^2  & x_1^2 x_2 & x_1 x_2^2 & x_2^3 \\
	x_1^2 &  x_1^3  &  x_1^2 x_2  & x_1^4 & x_1^3 x_2 & x_1^2 x_2^2 \\
	 x_1 x_2 &   x_1^2 x_2  &   x_1 x_2^2  & x_1^3 x_2 & x_1^2 x_2^2 & x_1 x_2^3 \\
	 x_2^2 &  x_1 x_2^2 &  x_2^3  & x_1^2 x_2^2 & x_1 x_2^3 & x_2^4 
\end{array}
\right]
}.
\eeq

By construction, $\MX_{\kappa} \in \psd{\binomial{d}{\kappa}}$ is positive semidefinite and has $\rank{\MX_{\kappa}} = 1$. 
Moreover, the set of \emph{unique} entries in $\MX_{\kappa}$ is simply $[\vxx]_{2\kappa}$, \ie the set of monomials of degree up to $2\kappa$ (these monomials typically appear multiple times in $\MX_{\kappa}$, \eg see $x_1 x_2$ in eq.~\eqref{eq:momentMatrix}).
Therefore, a key fact is that \emph{---for a suitable matrix $\MA$--- 
the linear function $\inprod{\MA}{\MX_{\kappa}}$
can express any polynomial in $\vxx$ of degree up to $2\kappa$.}

The key idea of Lasserre's hierarchy is to (i) rewrite~\eqref{eq:pop} using the moment matrix $\MX_{\kappa}$, 
(ii) relax the (non-convex) rank-1 constraint on $\MX_{\kappa}$, and
(iii) add redundant constraints that are trivially satisfied in~\eqref{eq:pop}; 
as we show below, this leads to a \emph{convex} semidefinite program.

\emph{(i) Rewriting~\eqref{eq:pop} using $\MX_{\kappa}$}. 
We pick a positive integer $\kappa\!\in\!\pint$ (the \emph{order} of the relaxation) such that $2\kappa \geq \max \{\deg{p}\!,\deg{h_1}\!, \ldots, \deg{h_{l_h}}\!,
\deg{g_1}\!, \ldots, \deg{g_{l_g}}\!\}.$ 
(this way we can express both objective function and constraints using $\MX_{\kappa}$). 
For instance, we can rewrite the objective and the equality constraints as: 
\beal\label{eq:objective}
\!\!\!\!\!\!\text{\grayout{objective}}: & \inprod{\MC_1}{\MX_\kappa} \\
\eeal
\beal \label{eq:eqConstraints1}
\!\!\!\text{\grayout{equality constraints}}: & \!\!\! \inprod{\MA_{\meq,j}}{\MX_\kappa} = 0, \; j=1,\ldots,l_h \\
\eeal
for suitable matrices $\MC_1$ and $\MA_{\meq,j}$.
Note that using $\MX_{\kappa}$ is already a relaxation since we are no longer enforcing the entries of $\MX_{\kappa}$
to be monomials 
(\eg we do not enforce  
the entry $x_1 x_2$ in~\eqref{eq:momentMatrix} to be the product of the entries $x_1$ and $x_2$, which would be a non-convex constraint).


 \emph{(ii) Relaxing the (non-convex) rank-$1$ constraint on $\MX_{\kappa}$}. 
 At the previous point we noticed we can rewrite objective and constraints in~\eqref{eq:pop} as linear (hence convex) 
 functions of $\MX_\kappa$. However, $\MX_\kappa$ still belongs to the set of positive-semidefinite rank-1 matrices, which is a non-convex set
 due to the rank constraint. Therefore, we simply relax the rank constraint and only enforce:
\beal \label{eq:eqMomentIsPSD}
\!\!\!\text{\grayout{moment matrix}}: & \MX_\kappa \succeq 0. \\
\eeal

 \emph{(iii) Adding redundant constraints}. Since we have relaxed~\eqref{eq:pop} by re-parametrizing it in $\MX_\kappa$ 
 and dropping the rank constraint, the final step to obtain Lasserre's relaxation consists in adding extra constraints to make the relaxation tighter. 
 First of all, we observe that there are multiple repeated entries in the moment matrix (\eg in~\eqref{eq:momentMatrix}, the entry $x_1 x_2$ 
 appears 4 times in the matrix). Therefore, we can enforce these entries to be the same. In general, this leads to 
 $\mmom = \trinum(\binomial{d}{\kappa}) - \binomial{d}{2\kappa} + 1$ linear constraints,
where $\trinum(n) \triangleq \frac{n(n+1)}{2}$ is the dimension of $\sym{n}$. These constraints are typically called \emph{moment constraints}:
\beal\label{eq:momentConstraints}
\text{\grayout{moment constraints}}: & \revise{\inprod{\MA_{\mathmom,0}}{\MX_\kappa} = 1}, \\
& \inprod{\MA_{\mathmom,j}}{\MX_\kappa} = 0, \\
&  j = 1, \ldots, \trinum(\binomial{d}{\kappa}) - \binomial{d}{2\kappa},
\eeal
\revise{where $\MA_{\mathmom,0}$ is all-zero except $[\MA_{\mathmom,0}]_{11} =1$, and it defines the constraint $[\MX_\kappa]_{11} = 1$, following from the 
definition of the moment matrix (see eq.~\eqref{eq:momentMatrix}).}

Second, we can also add \emph{redundant} equality constraints. Simply put, if $h_i = 0$, then also $h_i \cdot x_1 = 0$, $h_i \cdot x_2 = 0$, and so on, for any monomial we multiply by $h_i$. Since via $\MX_\kappa$ we can represent any polynomial of degree up to $2\kappa$, we can write as linear constraints any polynomial equality in the form $h_i \cdot [\vxx]_{2\kappa - \deg{h_i}} = \zero$ (the order of the monomials is chosen such that the product does not exceed order $2\kappa$). These new equalities can again be written linearly as:
\beal\label{eq:redundantEqualityConstraints}  
\hspace{-3mm}\text{\grayout{(redundant) equality constraints}}: \inprod{\MA_{\mathreq,ij}}{\MX_\kappa} = 0, \\
\quad\quad i = 1, \ldots, l_h, \ \ 
j = 1, \ldots, \binomial{d}{2\kappa - \deg{h_i}}\!\!\!\!\!\!\!\!\!
\eeal
for suitable $\MA_{\mathreq,ij}$.
Since the first entry of $[\vxx]_{2\kappa - \deg{h_i}}$ is always 1 (\ie the monomial of order zero),~eq.~\eqref{eq:redundantEqualityConstraints}  already includes the original 
equality constraints in~\eqref{eq:eqConstraints1}.

Finally, we observe that if $g_j \geq 0$, then for any positive semidefinite matrix $\MM$, it holds $g_j \cdot \MM \succeq 0$.
Since we can represent any polynomial of order up to $2\kappa$ as a linear function of $\MX_\kappa$, 
we can add redundant constraints in the form $g_j \cdot \MX_{\kappa - \ceil{\deg{g_j}/2}} \succeq 0$ 
(by construction $g_j \cdot \MX_{\kappa - \ceil{\deg{g_j}/2}}$ only contains polynomials of degree up to $2\kappa$).
To phrase the resulting 
relaxation in the standard form~\eqref{eq:primalSDP}, it is common to add extra matrix variables $\MX_{g_j} = g_j \cdot \MX_{\kappa - \ceil{\deg{g_j}/2}}$ for $j=1,\ldots,l_g$ 
(the so called \emph{localizing matrices} \cite[\S 3.2.1]{lasserre10book-momentsOpt})
and then force these matrices to be a linear function of  $\MX_\kappa$:
\beal\label{eq:locMatrices}
\text{\grayout{localizing matrices}}: & \MX_{g_j} \succeq 0, \;\; j=1,\ldots,l_g
\eeal
\beal \label{eq:localizingConstraints}  
\hspace{-2mm} \text{\grayout{{localizing} constraints}}: \inprod{\MA_{\mathloc,jkh}}{\MX_\kappa} = [\MX_{g_j}]_{hk} \\
\quad \quad j = 1, \ldots, l_g,\ \  
1 \leq h\leq k \leq \binomial{d}{\kappa - \ceil{\deg{g_j}/2}}
\eeal
where the linear constraints (for some $\MA_{\mathloc,jkh}$) enforce each entry of $\MX_{g_j}$ to be a linear combination of entries in $\MX_\kappa$.

Following steps (i)-(iii) above, it is straightforward to obtain the following (convex) semidefinite program:
\begin{equation}\label{eq:lasserre}
\hspace{-4mm} \fstar_{\kappa} =\!\! \displaystyle\min_{\MX = (\MX_\kappa, \MX_1, \ldots, \MX_{l_g})} \cbrace{\inprod{\MC_1}{\MX_\kappa} \mid \calA(\MX)\!=\!\vb,\MX\! \succeq\! 0}\!,\!\!\! \tag{LAS}
\end{equation}
where the variable $\MX = (\MX_\kappa, \MX_1,\dots,\MX_{l_g})$ is a collection of positive-semidefinite matrices (\cf~\eqref{eq:eqMomentIsPSD} and~\eqref{eq:locMatrices}, and we shorthand $\MX_j = \MX_{g_j}$ for notation convenience), 
the objective is the one given in~\eqref{eq:objective}, and the linear constraints $\calA(\MX)=\vb$
collect all the constraints in~\eqref{eq:momentConstraints},~\eqref{eq:redundantEqualityConstraints}, and~\eqref{eq:localizingConstraints}.
Problem \eqref{eq:lasserre} can be readily formulated as a multi-block SDP in the primal form~\eqref{eq:primalSDP}, which matches
  the data format used by common SDP solvers. 
Problem \eqref{eq:lasserre} is commonly known as the \emph{dense} Lasserre's relaxation because a fully dense monomial basis $[\vxx]_\kappa$ is used to build the moment matrix \cite{Lasserre01siopt-lasserrehierarchy}. One can solve the relaxation for different choices of $\kappa$, leading to a \emph{hierarchy} of convex relaxations.

While we presented Lasserre's hierarchy in a somewhat procedural way, the importance of the hierarchy lies in its stunning theoretical properties, that we review below.
\begin{theorem}[Lasserre's Hierarchy \cite{Lasserre01siopt-lasserrehierarchy,lasserre10book-momentsOpt,Nie14mp-finiteConvergenceLassere}]
\label{thm:lasserre}
Let $-\infty < \pstar < \infty$ be the optimum of \eqref{eq:pop} and \revise{$\fstar_{\kappa}$ (resp. $\MXstar_\kappa$) be the optimum (resp. one optimizer) of \eqref{eq:lasserre},} assume \eqref{eq:pop} satisfies the Archimedeanness condition (a stronger form of compactness, \cf \cite[Definition 3.137]{Blekherman12Book-sdpandConvexAlgebraicGeometry}), then 
\begin{enumerate}[label=(\roman*)]
\item \revise{(lower bound and convergence)} $\fstar_\kappa$ converges to $\pstar$ from below as $\kappa \rightarrow \infty$, and convergence occurs at a finite $\kappa$ under suitable technical conditions \cite{Nie14mp-finiteConvergenceLassere};
\item \revise{(rank-one solutions)} if $\fstar_\kappa = \pstar$ at some finite $\kappa$, then for every global minimizer $\vxxstar$ of \eqref{eq:pop}, 
$\MXstar_\kappa \triangleq [\vxxstar]_{\kappa} [\vxxstar]_{\kappa}\tran$ is optimal for \eqref{eq:lasserre}, and every rank-one optimal solution $\MXstar_\kappa$ of \eqref{eq:lasserre} can be written as $[\vxxstar]_\kappa [\vxxstar]_{\kappa}\tran$ for some $\vxxstar$ that is optimal for \eqref{eq:pop};
\item \revise{(optimality certificate)} if $\rank{\MXstar_\kappa} = 1$ at some finite $\kappa$, then $\fstar_\kappa = \pstar$.
\end{enumerate}
\end{theorem}
Theorem \ref{thm:lasserre} states that~\eqref{eq:lasserre} provides a hierarchy of lower bounds for~\eqref{eq:pop}. 
When the relaxation is exact ($\pstar\!=\!\fstar_\kappa$), global minimizers of~\eqref{eq:pop} correspond to rank-one solutions of~\eqref{eq:lasserre}. \revise{Moreover, after solving the convex SDP \eqref{eq:lasserre}, one can check the rank of the optimal solution $\MXstar_\kappa$ to obtain a \emph{certificate} of global optimality. In practice, rank computation can be subject to numerical inaccuracies, and we introduce a continuous metric for evaluating the exactness of the relaxation in Section \ref{sec:sdprelax} (\cf Theorem \ref{thm:sparserelaxtls}).}

\maybeOmit{
{\bf Curse of Dimensionality}. As we will see in Section~\ref{sec:robustandpop}, for outlier-robust geometric perception problems, (i) $d$ 
---the size of the variable in the original~\eqref{eq:pop}--- increases \wrt the number of measurements and can be a few hundreds (contrarily, outlier-free problems have $d$ fixed and typically less than $20$), (ii) $\kappa=2$ is the minimum relaxation order because $\deg{p} > 2$, leading to $n_1 = \binomial{d}{2}$ and $m \geq \mmom = \trinum(\binomial{d}{2}) - \binomial{d}{4} + 1$, which both grow quickly \wrt $d$ (contrarily, outlier-free problems typically have $\deg{p}=2$, and one can use $\kappa=1$ in \eqref{eq:lasserre}, which  is much more scalable).
Therefore, Lasserre's hierarchy, at least in its dense form \eqref{eq:lasserre}, is impractical for outlier-robust perception. In Section \ref{sec:sdprelax}, we present a \emph{sparse} version of \eqref{eq:lasserre} for outlier-robust perception that significantly improves scalability.
}

\section{Outlier-Robust Estimation as POP}
\label{sec:robustandpop}

\newcommand{\myhspaceone}{\hspace{-2mm}}
\newcommand{\mpwseven}{2.7cm}
\begin{figure*}[t]
	\begin{center}
	\begin{minipage}{\textwidth}
	\begin{tabular}{ccccccc}%
		   \hspace{-3mm} \myhspaceone \hspace{-4mm}
			\begin{minipage}{\mpwseven}%
			\centering%
			\includegraphics[width=\columnwidth]{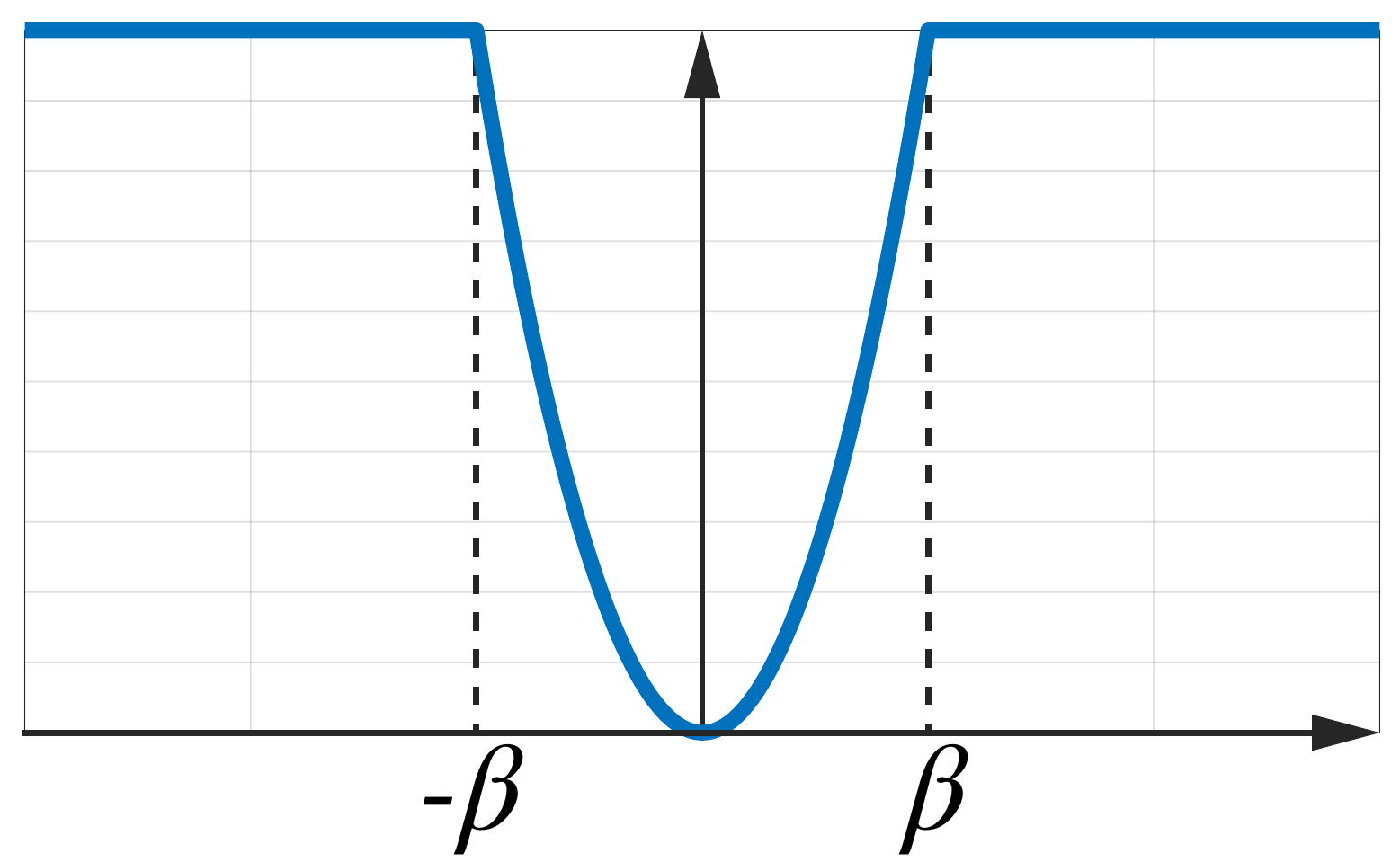}\\
			{(a) TLS}
			\end{minipage}
		&  \myhspaceone \hspace{-3mm}
			\begin{minipage}{\mpwseven}%
			\centering%
			\includegraphics[width=\columnwidth]{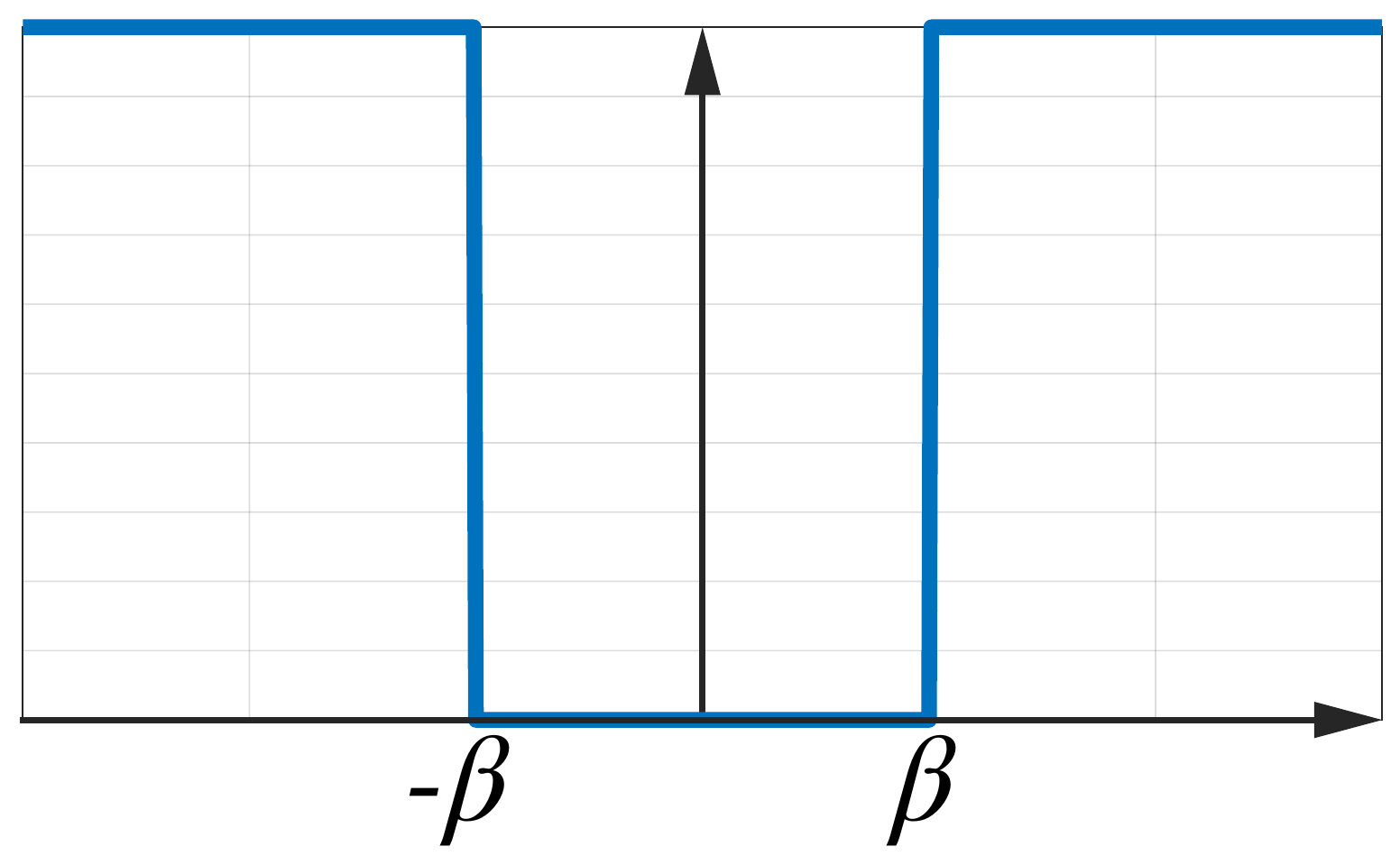}\\
			{(b) MC}
			\end{minipage}
		&  \myhspaceone \hspace{-3mm}
			\begin{minipage}{\mpwseven}%
			\centering%
			\includegraphics[width=\columnwidth]{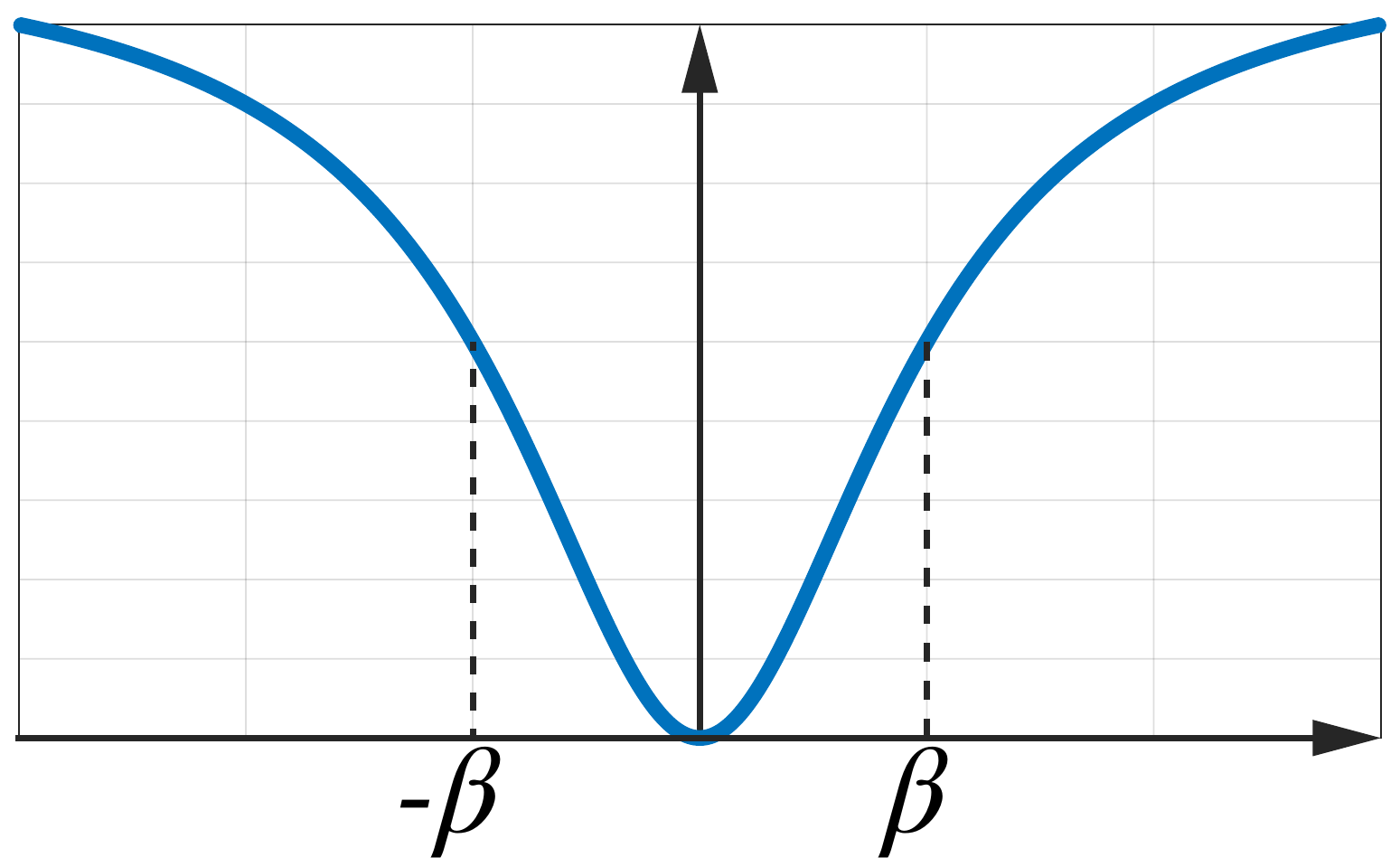}\\
			{(c) GM}
			\end{minipage}
		&  \myhspaceone \hspace{-3mm}
			\begin{minipage}{\mpwseven}%
			\centering%
			\includegraphics[width=\columnwidth]{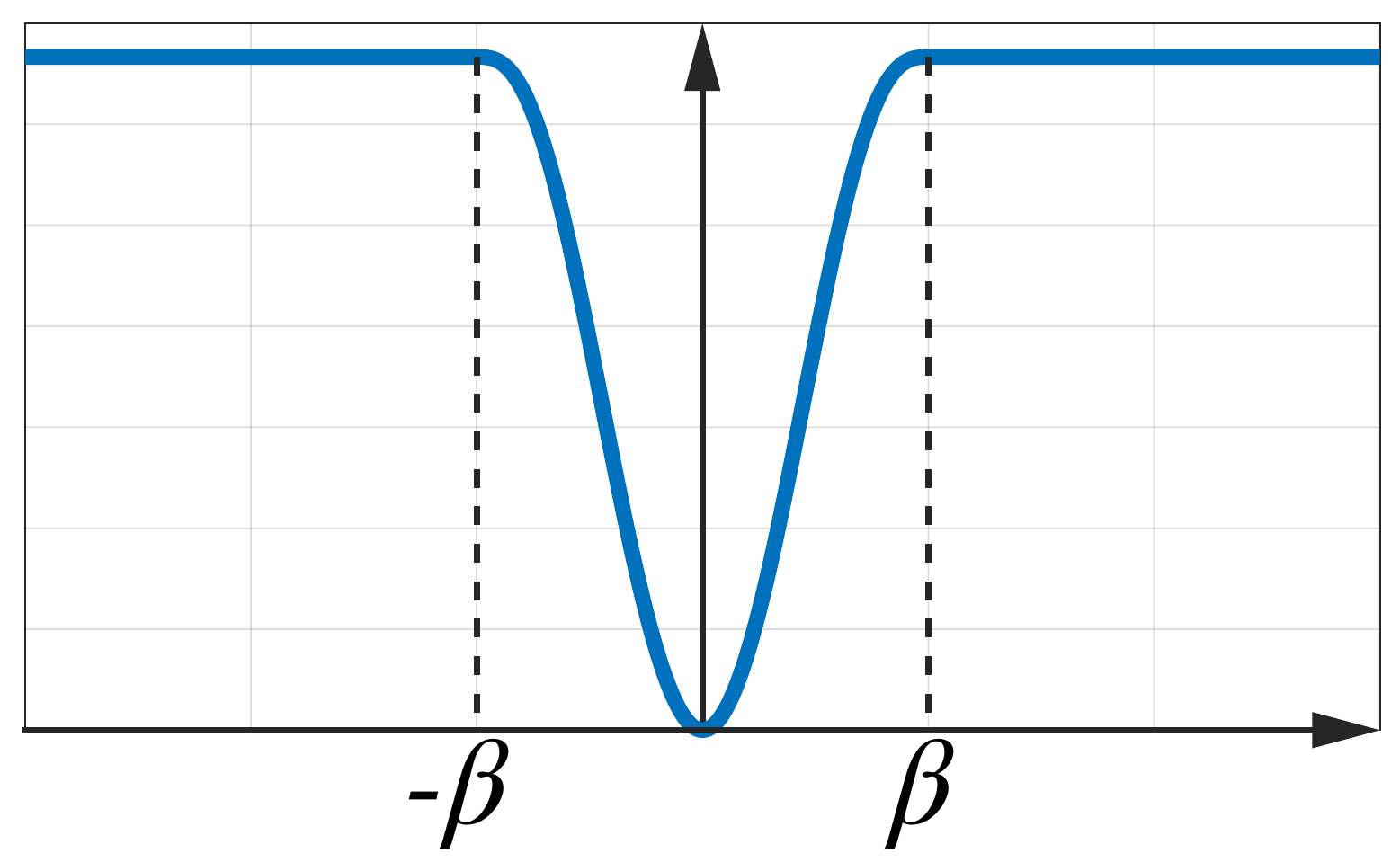}\\
			{(d) TB}
			\end{minipage} 
		&  \myhspaceone \hspace{-3mm}
			\begin{minipage}{\mpwseven}%
			\centering%
			\includegraphics[width=\columnwidth]{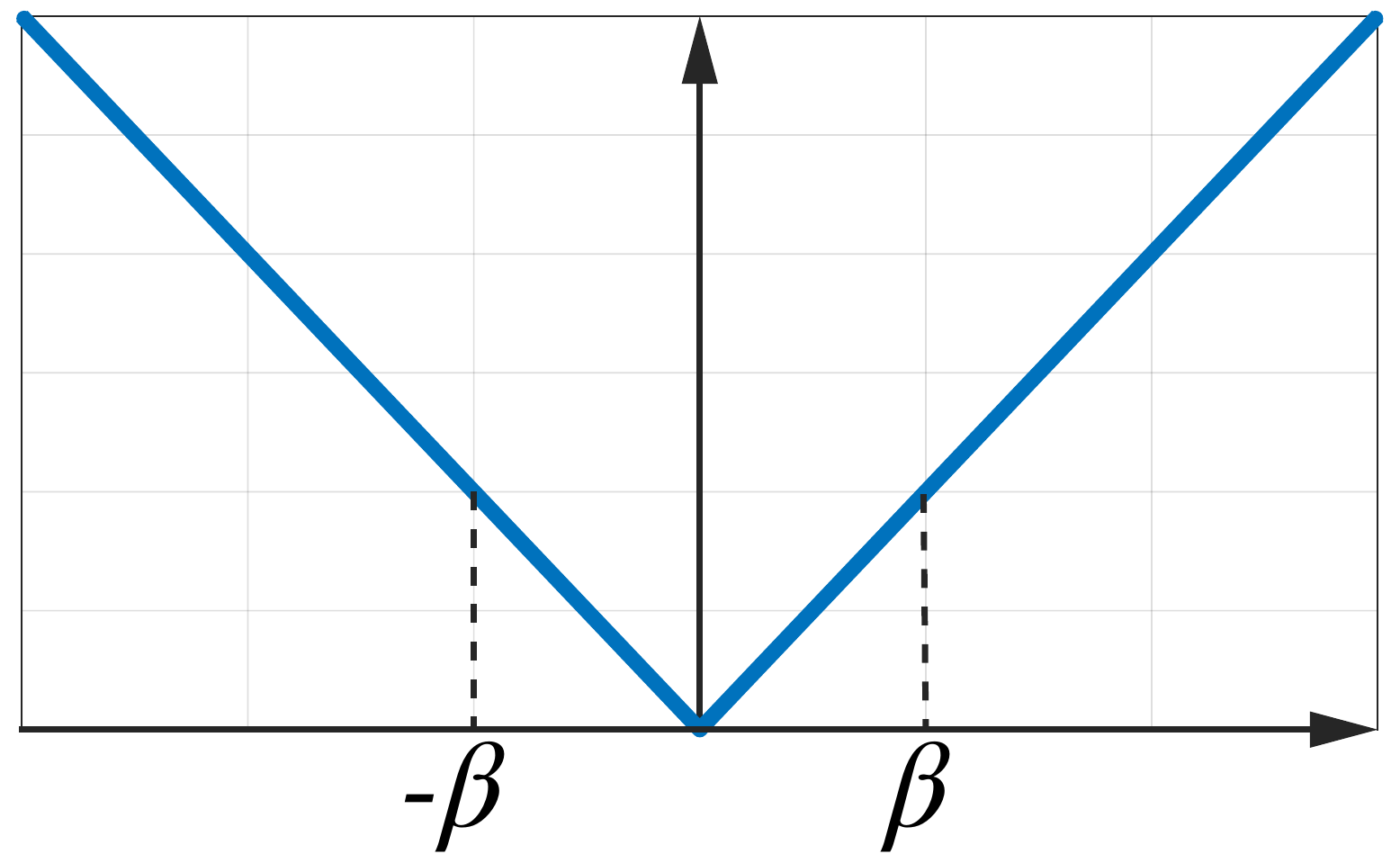}\\
			{(e) L1}
			\end{minipage}
		&  \myhspaceone \hspace{-3mm}
			\begin{minipage}{\mpwseven}%
			\centering%
			\includegraphics[width=\columnwidth]{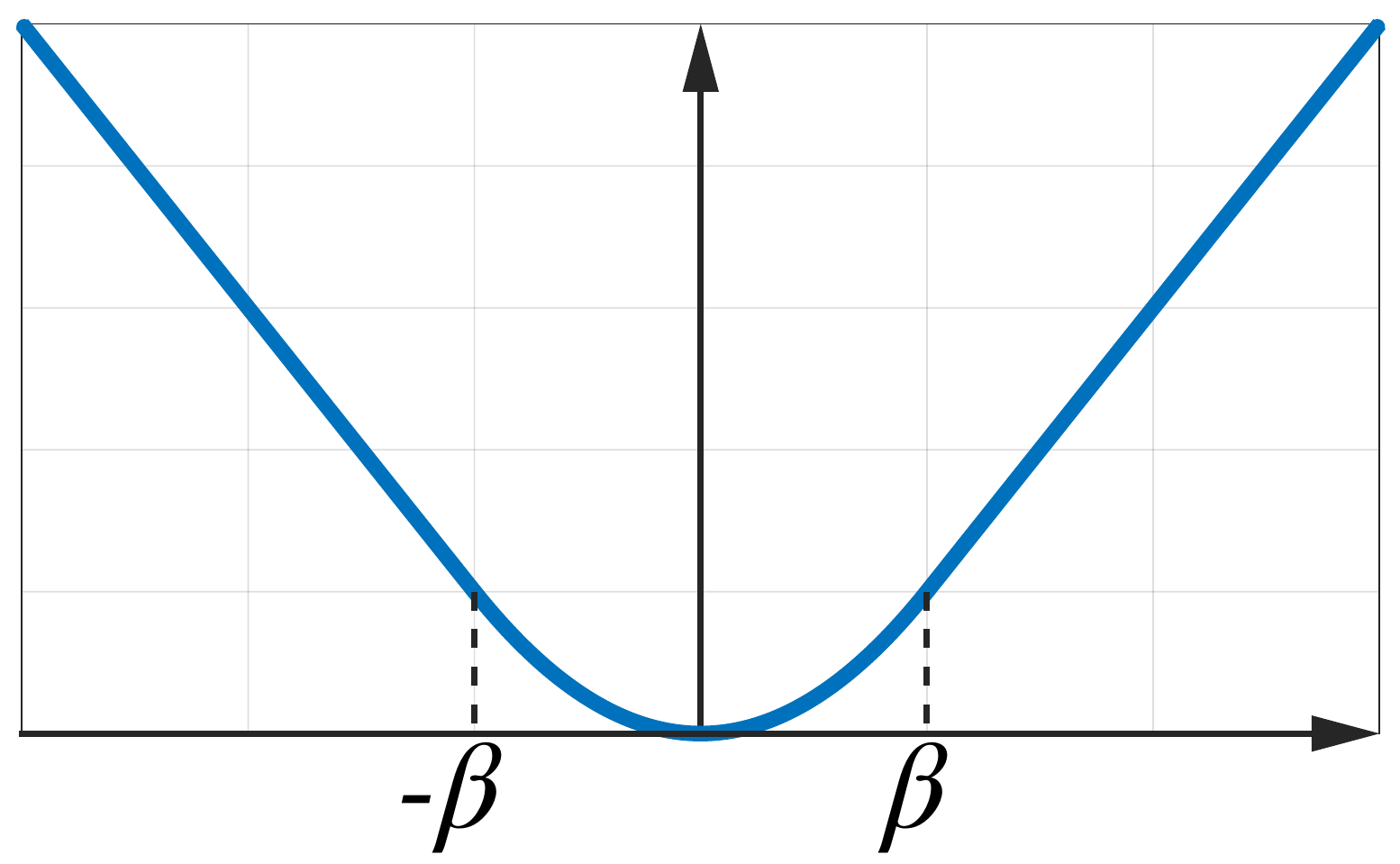}\\
			{(f) Huber}
			\end{minipage}
		&  \myhspaceone \hspace{-3mm}
			\begin{minipage}{\mpwseven}%
			\centering%
			\includegraphics[width=\columnwidth]{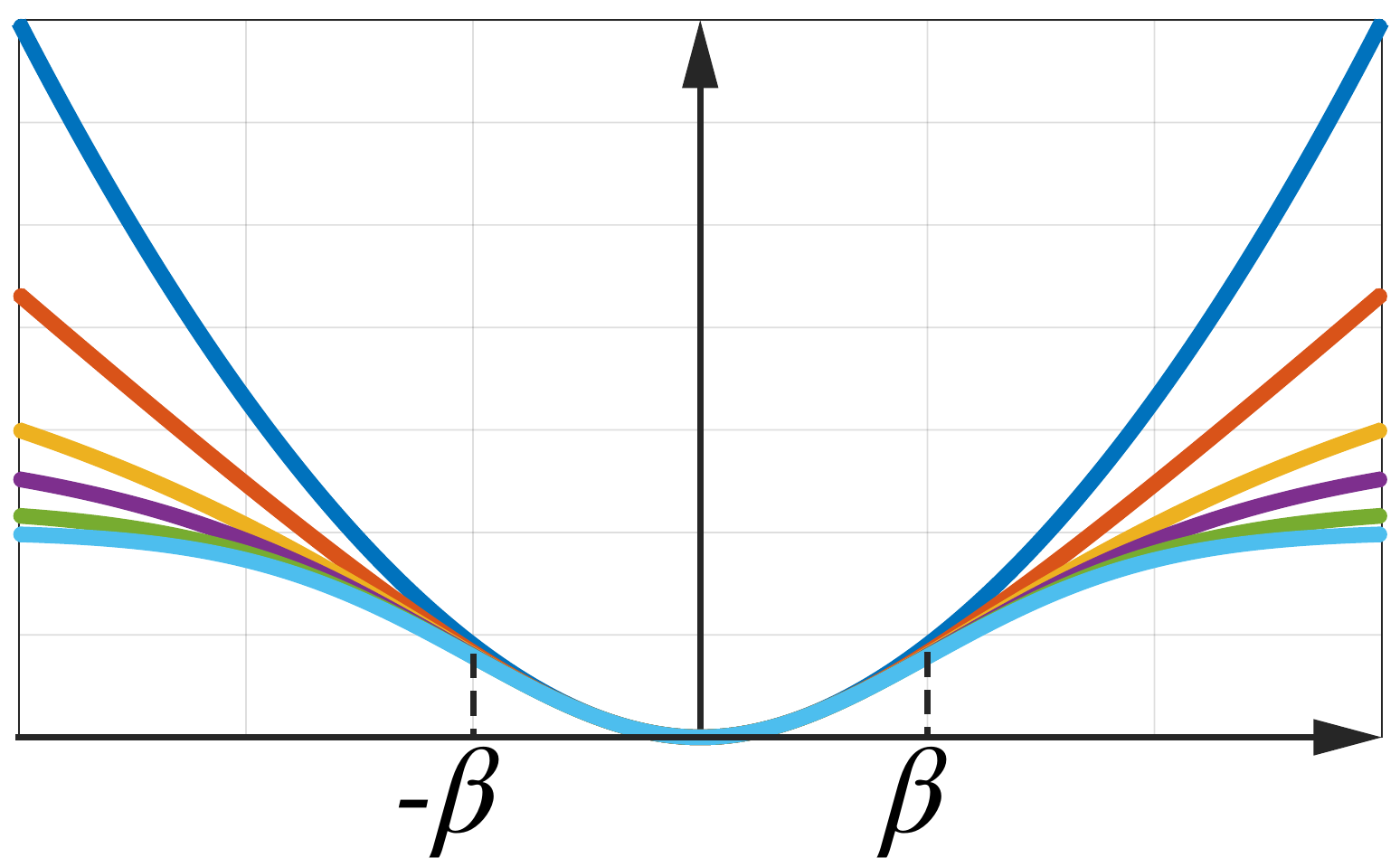}\\
			{(g) Adaptive}
			\end{minipage}
	\end{tabular}
	\end{minipage} 
	\caption{Estimation using common robust cost functions can be equivalently reformulated as polynomial optimization problems (\cf Proposition \ref{prop:robustaspop}).
	\label{fig:robust-costs}} 
	\vspace{-7mm} 
	\end{center}
\end{figure*}

In this section, we consider a general formulation of estimation with robust cost functions. We show that, for \nrCosts popular robust costs, this formulation can be recast as a~\eqref{eq:pop}. We conclude the section by showcasing the resulting formulation on six perception problems.

{\bf Outlier-robust estimation}. Given a set of $N$ measurements $\calZ = \{\vz_i\}_{i=1}^N$ (\eg 2D image keypoints, 3D point clouds, relative poses), we consider the problem of using $\calZ$ to estimate an unknown geometric model $\vxx \in \calX \subseteq \Real{d}$ (\eg camera poses, rigid transformations, 3D shapes, robot trajectory), despite the fact that the measurement set $\calZ$ may contain a large amount of \emph{outliers}. Building on standard M-estimation \cite{Maronna19book-robustStats,MacTavish15crv-robustEstimation}, we perform outlier-robust estimation by solving the following optimization problem
\begin{equation}\label{eq:robust}
\min_{\vxx \in \calX \subseteq \Real{d}}\ \ \sum_{i=1}^N \rho(r(\vxx,\vz_i), \beta_i) + \regularizer,  
\tag{Robust}
\end{equation}
where $r(\vxx,\vz_i)$ is a (scalar) \emph{residual} function that measures the mismatch between $\vxx$ and $\vz_i$ (\eg Euclidean distances, pose errors), $\beta_i > 0$ (set by the user) is the \emph{maximum admissible residual} for a measurement to be considered as an \emph{inlier} (or minimum residual to be an outlier), $\rho(r,\beta_i)$ is a \emph{robust} cost function that penalizes outliers much \emph{less} than inliers to prevent outliers from contaminating the estimate. We include a \emph{regularization} term $\regularizer$ in~\eqref{eq:robust}, 
to keep full generality: as we will see in the examples below, a regularizer is often added to high-dimensional estimation problems 
to ensure the solution is unique and well-behaved.
We make the following assumption on problem \eqref{eq:robust}.

\begin{assumption}[Polynomial Residual, Constraint, and Regularization]
\label{assumption:polynomialsrobust} In \eqref{eq:robust}, assume (i) $r^2, \psi$ are polynomials; (ii) the constraint $\vxx \in \calX$ can be described by finitely many polynomial equalities and inequalities, \ie $\calX = \{\vxx \in \Real{d}\mid h_i(\vxx)=0,i\in 1,\dots,l_h, g_j(\vxx) \geq 0,j=1,\dots,l_g \}$.
\end{assumption}

Assumption \ref{assumption:polynomialsrobust} is the prerequisite for applying the machinery of semidefinite relaxation for POP in Section \ref{sec:pre-pop}. These assumptions are often mild in geometric perception problems, a point that will become clearer when we introduce the six examples later in this section \mbox{(\cf Proposition \ref{prop:polynomialExpressibility})}.

Now the only component of \eqref{eq:robust} that may prevent it from being a POP is the robust cost $\rho(r,\beta_i)$. In \emph{outlier-free} estimation, $\rho = r^2/\beta_i^2$ is chosen as the \emph{least squares}  cost and \eqref{eq:robust} is immediately in the form of \eqref{eq:pop}. However, in outlier-robust estimation, $\rho$ is typically not a polynomial. For instance, let us consider the \emph{truncated least squares} (TLS) cost, which will be extensively used in this paper:
\begin{equation} \label{eq:tlsdef}
\rho_{\mtls}(r,\beta_i) \triangleq 
\min \cbrace{ \frac{r^2}{\beta_i^2}, 1 } = 
\begin{cases}
\frac{r^2}{\beta_i^2} & \abs{r} \leq \beta_i \\
1 & \text{otherwise}
\end{cases},
\end{equation}
The TLS cost~\eqref{eq:tlsdef} is apparently not a polynomial, and it is not even a \emph{smooth} function (\cf Fig. \ref{fig:robust-costs}(a)). 

{\bf Reformulation as POP}. To build intuition, we now show that \eqref{eq:robust} with the TLS cost \eqref{eq:tlsdef} can be reformulated as a POP; then we generalize this conclusion to other cost functions in Proposition~\ref{prop:robustaspop}. 
The key observation is that, for any $a,b \in \Real{}$, $\min \{a, b\} \equiv \min_{\theta \in \{ +1,-1\}} \frac{1+\theta}{2} a + \frac{1-\theta}{2} b$  which allows recasting \eqref{eq:robust} with $\rho = \rho_{\mtls}$ as 
\begin{equation}\label{eq:binaryTLS}
\min_{\substack{\vxx \in \calX \subseteq \Real{d}, \\ \vtheta \in \{\pm 1\}^N} }\  \sum_{i=1}^N  \frac{1+\theta_i}{2} \frac{r^2(\vxx,\vz_i)}{\beta_i^2} + \frac{1-\theta_i}{2} + \regularizer, \tag{TLS}
\end{equation}
where each binary variable $\theta_i \in \{+1,-1\}$ decides whether the $i$-th measurement $\vz_i$ is an inlier ($\theta_i = +1$) or an outlier ($\theta_i=-1$). By recalling that $\theta_i \in \{+1,-1\} \Leftrightarrow \theta_i^2 - 1 = 0$, we see that problem \eqref{eq:binaryTLS} is an instance of \eqref{eq:pop}, with the decision variables now being $(\vxx, \vtheta) \in \Real{d+N}$. The next proposition states that, the reformulation above can be generalized to a broader set of robust cost functions. 

\begin{proposition}[Robust Estimation as POP]
\label{prop:robustaspop}
Under Assumption \ref{assumption:polynomialsrobust}, if the cost function $\rho$ in \eqref{eq:robust} is one of the following:
\begin{enumerate}[label=(\roman*)]
\item truncated least squares (TLS): $\rho_{\mtls} \triangleq \min \cbrace{ \displaystyle \frac{r^2}{\beta_i^2}, 1 } $;
\item maximum consensus: $\rho_{\mmc} \triangleq \begin{cases} 0 & \abs{r} \leq \beta_i \\ 1 & \text{otherwise} \end{cases}$;

\item Geman-McClure: $\rho_{\mgm} \triangleq \frac{\displaystyle r^2/\beta_i^2}{\displaystyle 1+r^2/\beta_i^2}$;

\item Tukey's Biweight: $\rho_{\mtb} \triangleq \begin{cases} \frac{r^2}{\beta_i^2} - \frac{r^4}{\beta_i^4} + \frac{r^6}{3\beta_i^6} & \abs{r}\leq \beta_i \\ \frac{1}{3} & \text{otherwise} \end{cases}$,
\end{enumerate}
then \eqref{eq:robust} can be recast as a \eqref{eq:pop} with $d+N$ variables, where each of the additional $N$ variables indicates the confidence of the corresponding measurement being an inlier. Moreover, \eqref{eq:robust} with the following costs can also be written as a \eqref{eq:pop}
\begin{enumerate}[label=(\roman*)]
\setcounter{enumi}{4}
\item L1: $\rho_{\mlone} \triangleq \abs{r}/\beta_i $;
\item Huber: $\rho_{\mhuber} \triangleq \begin{cases} \frac{r^2}{2\beta_i^2} & \abs{r} \leq \beta_i \\ \frac{\abs{r}}{\beta_i} - \frac{1}{2} & \text{otherwise} \end{cases}$;
\item Adaptive \cite{Barron19cvpr-adaptRobustLoss}: $\rho_{\madaptive,s} \triangleq \frac{\abs{s-2}}{s} \parentheses{ \parentheses{ \frac{r^2/\beta_i^2}{\abs{s-2}} + 1 }^{\frac{s}{2}} - 1}  $, \\for a given scale parameter $s \in \bbQ \backslash \{0,2\}$,
\end{enumerate}
by adding slack variable(s) for each measurement.
\end{proposition}
Fig. \ref{fig:robust-costs} plots the seven robust costs (Fig. \ref{fig:robust-costs}(g) shows $\rho_{\madaptive,s}$ for six different values of $s$). While we postpone the proof to \supp, the key insight is that for common robust cost functions we can either (a)
use Black-Rangarajan duality~\cite{Black96ijcv-unification} to convert them into polynomials by introducing additional slack variables -- one for each measurement (we use this approach for (i)-(iv)), 
or (b) directly manipulate them into polynomials by change of variables (for (v)-(vii)).

\newcommand{\mpwsix}{3.1cm}
\begin{figure*}[t]
	\begin{center}
	\begin{minipage}{\textwidth}
	\begin{tabular}{cccccc}%
		   \hspace{-2mm} \myhspaceone \hspace{-4mm}
			\begin{minipage}{\mpwsix}%
			\centering%
			\includegraphics[width=\columnwidth]{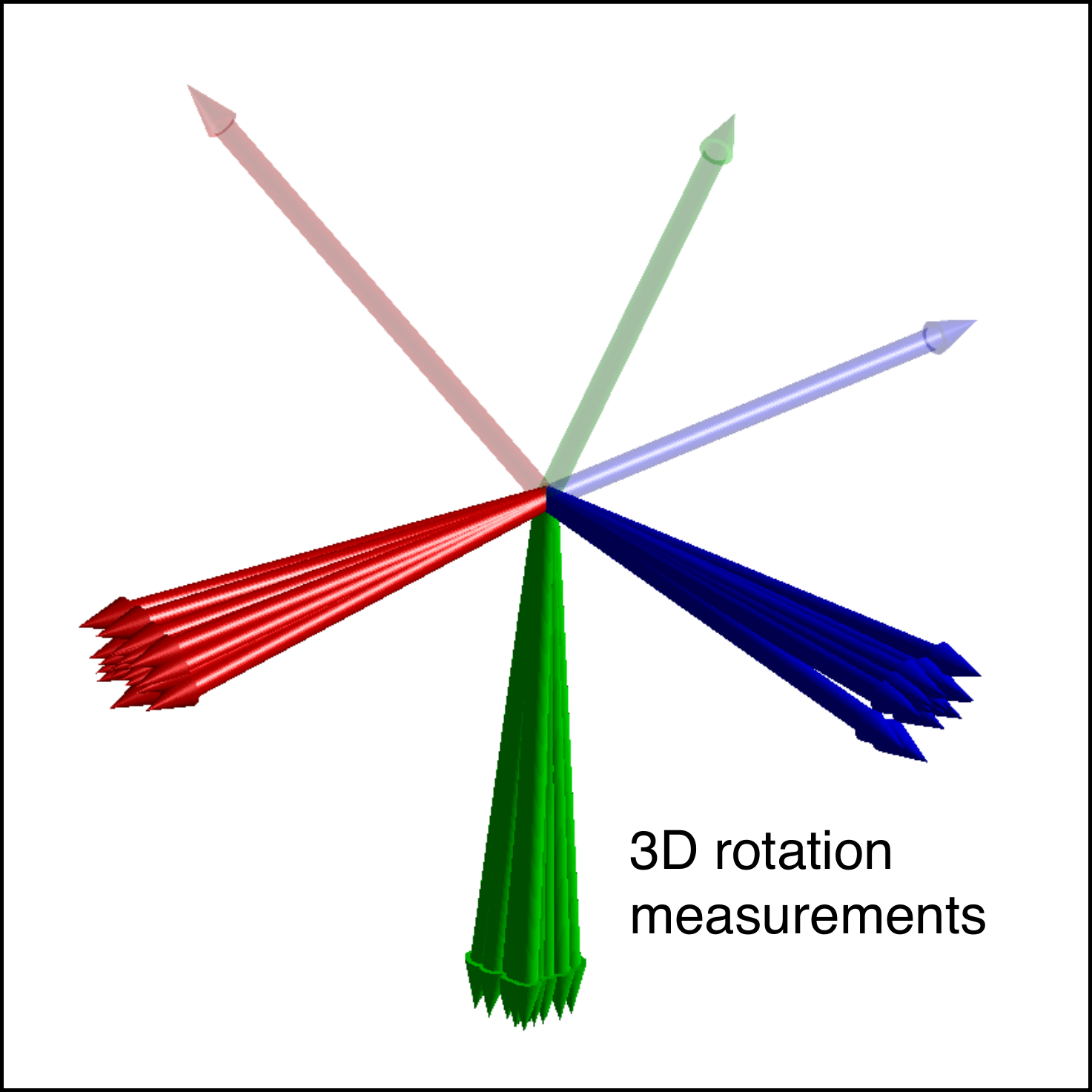}\\
			{\fontsize{6pt}{6pt}\selectfont (a) Single rotation averaging}
			\end{minipage}
		&  \myhspaceone \hspace{-3mm}
			\begin{minipage}{\mpwsix}%
			\centering%
			\includegraphics[width=\columnwidth]{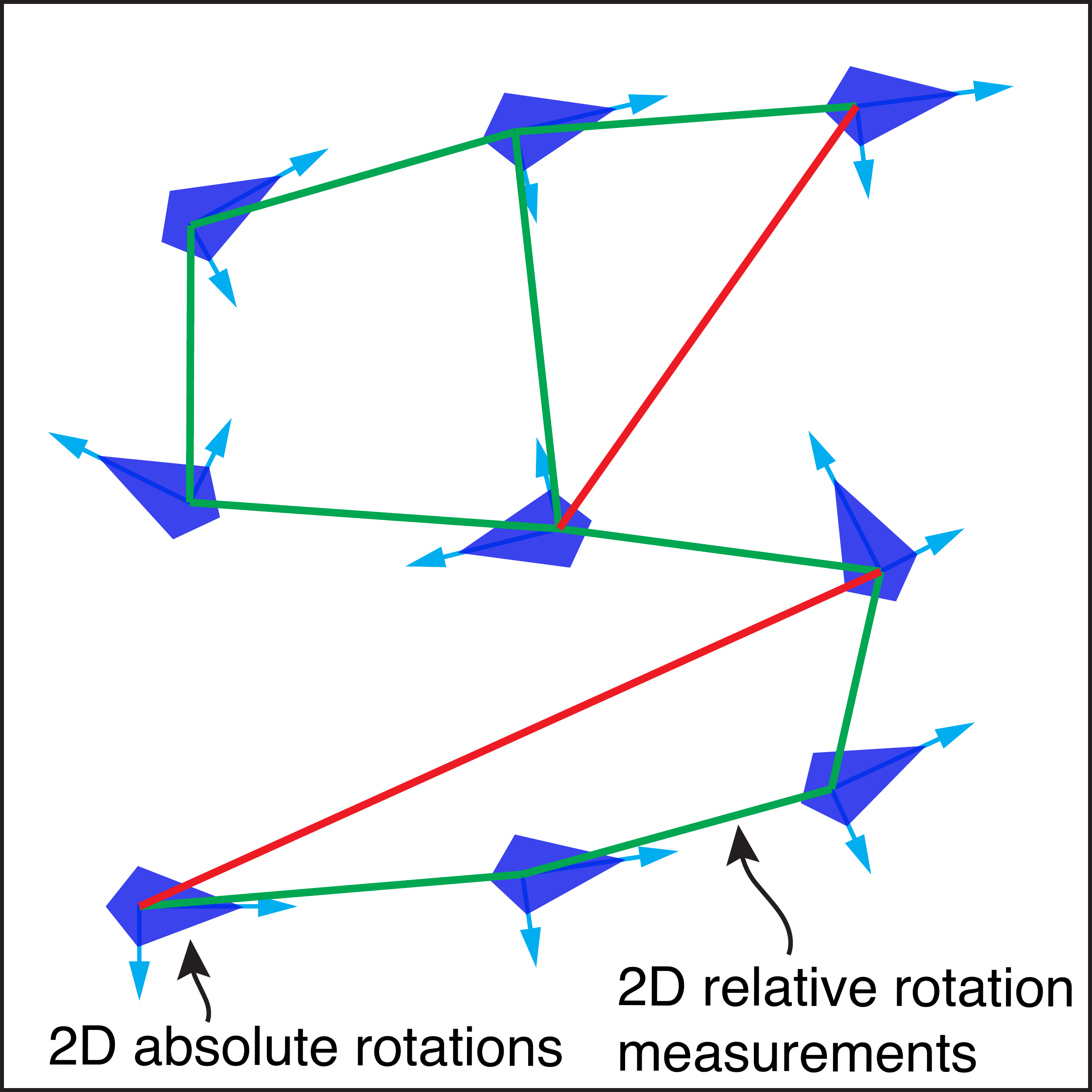}\\
			{\fontsize{6pt}{6pt}\selectfont (b) Multiple rotation averaging}
			\end{minipage}
		&  \myhspaceone \hspace{-3mm}
			\begin{minipage}{\mpwsix}%
			\centering%
			\includegraphics[width=\columnwidth]{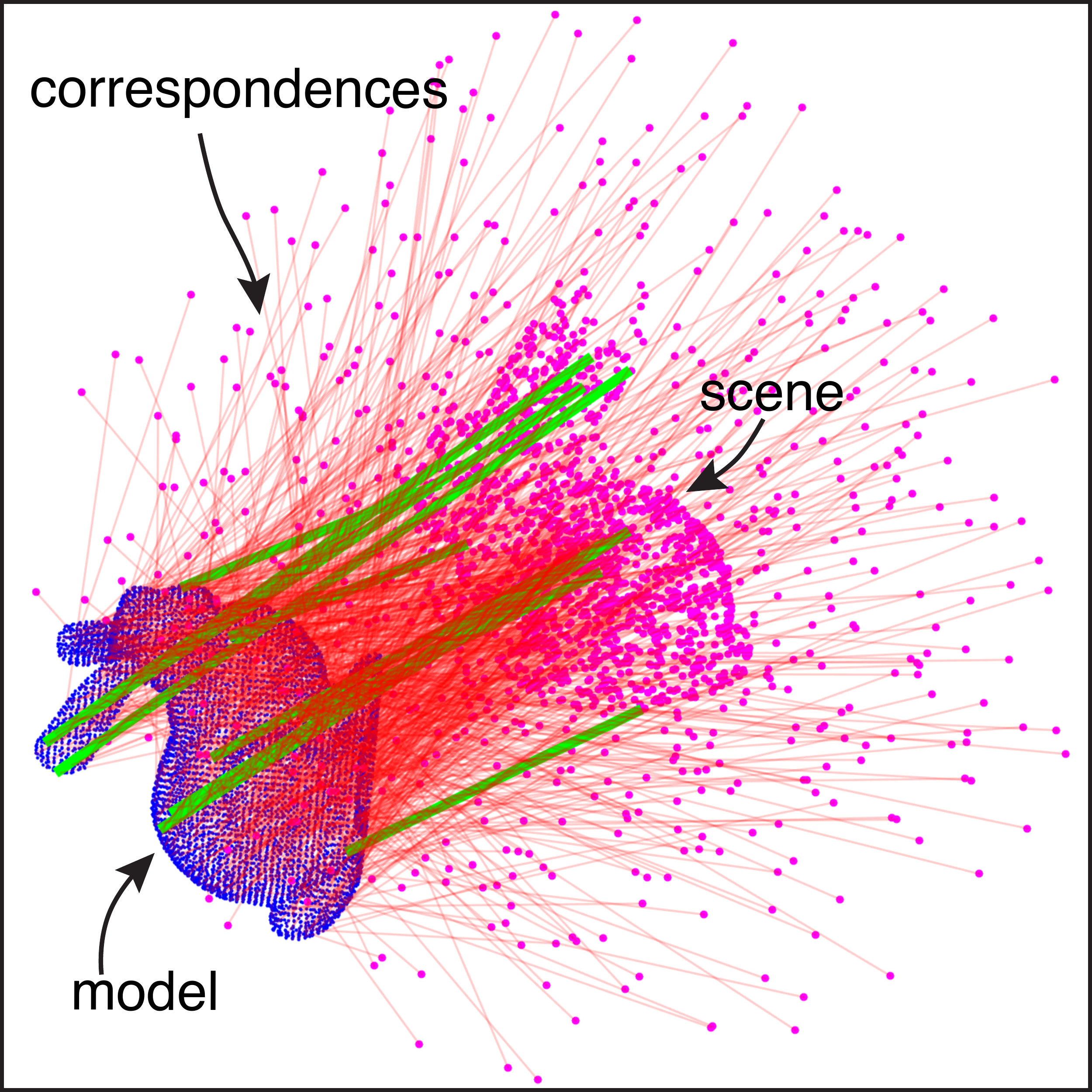}\\
			{\fontsize{6pt}{6pt}\selectfont (c) Point cloud registration}
			\end{minipage}
		&  \myhspaceone \hspace{-3mm}
			\begin{minipage}{\mpwsix}%
			\centering%
			\includegraphics[width=\columnwidth]{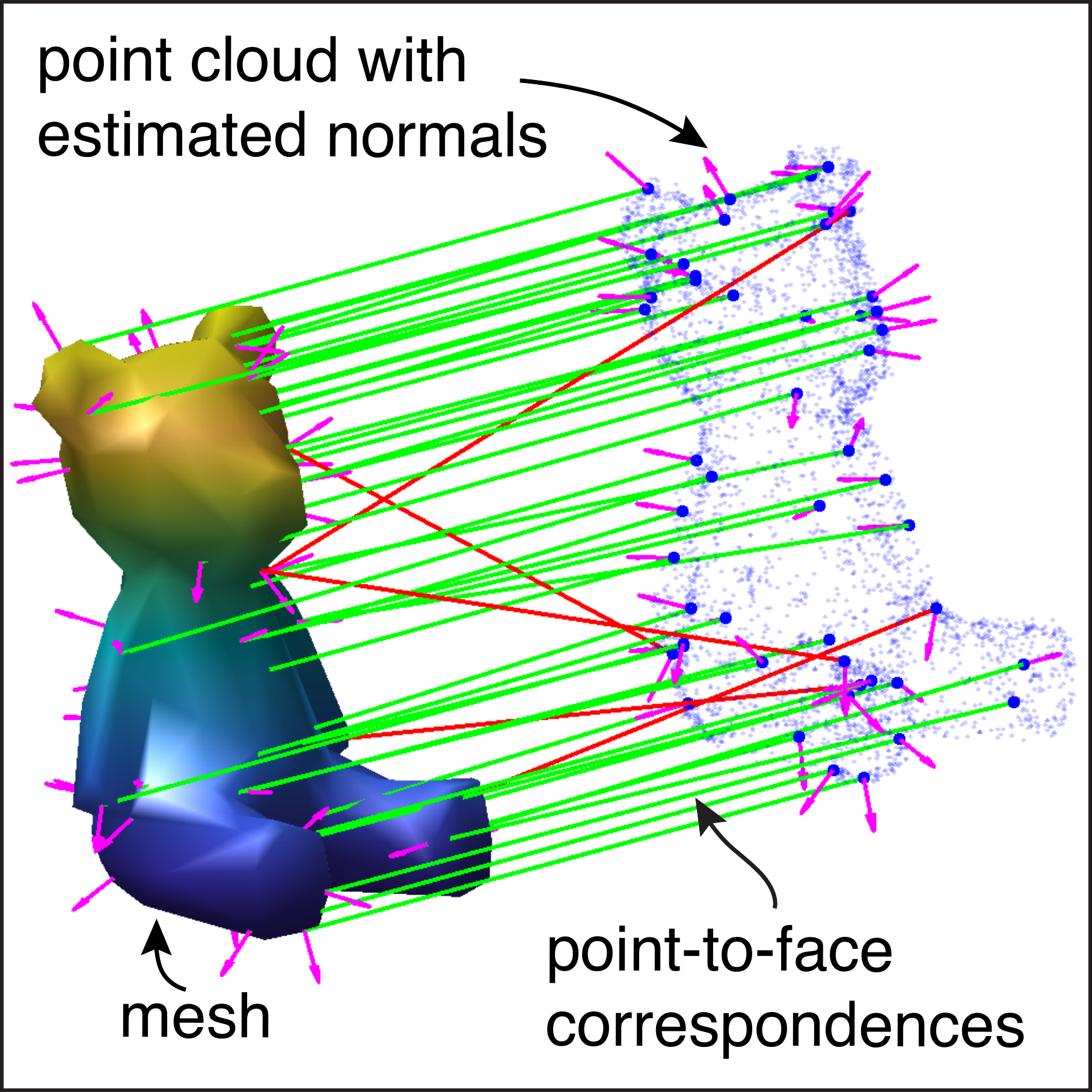}\\
			{\fontsize{6pt}{6pt}\selectfont (d) Mesh registration}
			\end{minipage} 
		&  \myhspaceone \hspace{-3mm}
			\begin{minipage}{\mpwsix}%
			\centering%
			\includegraphics[width=\columnwidth]{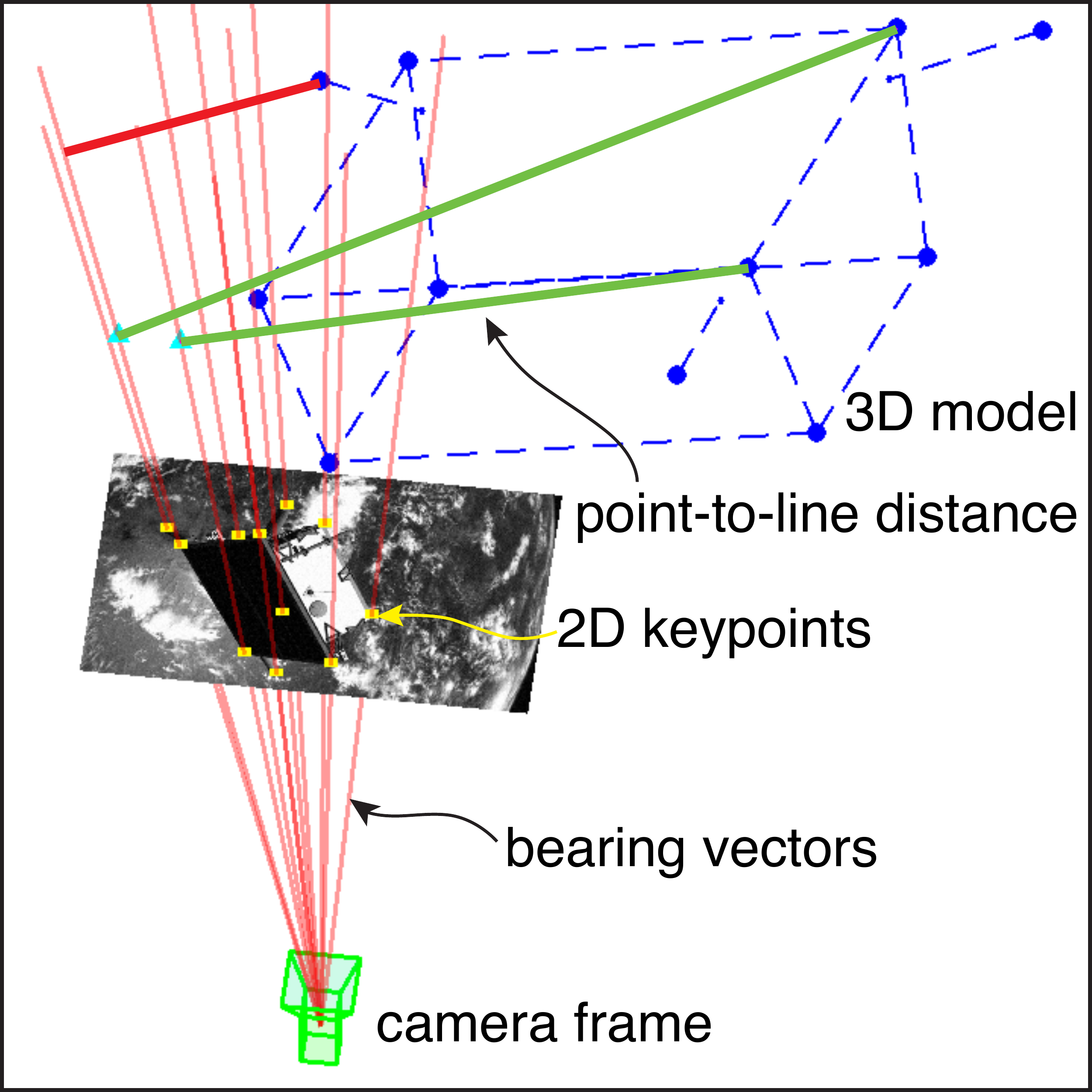}\\
			{\fontsize{6pt}{6pt}\selectfont (e) Absolute pose estimation}
			\end{minipage}
		&  \myhspaceone \hspace{-3mm}
			\begin{minipage}{\mpwsix}%
			\centering%
			\includegraphics[width=\columnwidth]{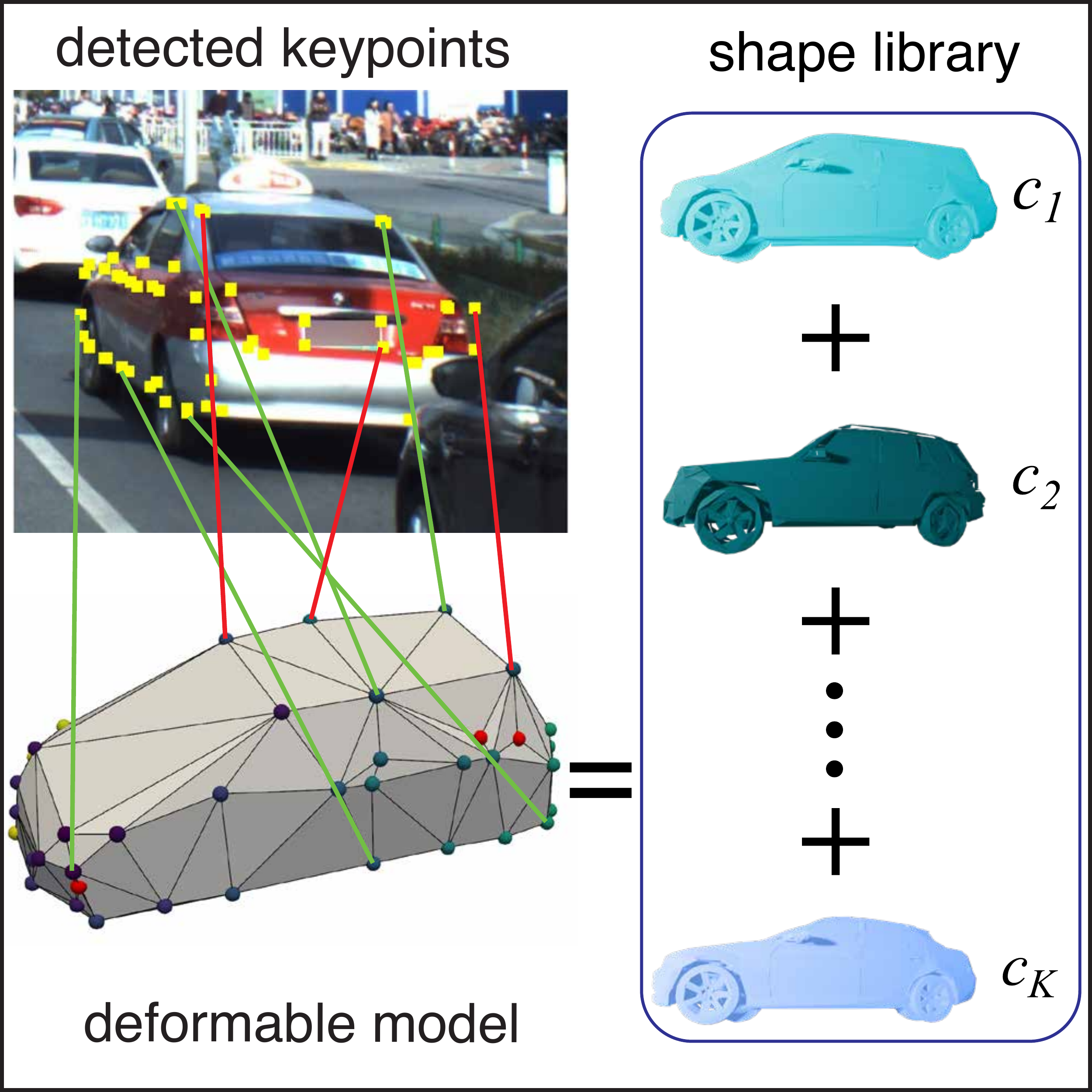}\\
			{\fontsize{6pt}{6pt}\selectfont (f) Category-level perception}
			\end{minipage}
	\end{tabular}
	\end{minipage} 
	\caption{Perception examples considered in this paper that can be modeled as polynomial optimization problems (\cf Examples \ref{ex:singlerotation}-\ref{ex:category}).
	\label{fig:applications}} 
	\vspace{-7mm} 
	\end{center}
\end{figure*}

{\bf Perception examples}. 
We now shed some light on the generality of the formulation~\eqref{eq:robust} and Assumption~\ref{assumption:polynomialsrobust} by considering 
six outlier-robust geometric perception problems. 
We first present the examples and then conclude they all satisfy Assumption~\ref{assumption:polynomialsrobust} in Proposition~\ref{prop:polynomialExpressibility}.
We assume $\regularizer = 0$ unless otherwise mentioned.



\setcounter{theorem}{0}
\begin{example}[Single Rotation Averaging \cite{Hartley13ijcv}] 
\label{ex:singlerotation}
Given $N$ measurements of an unknown $\dimrot$-dimensional rotation $\{ \vz_i = \tldMR_i \in \mySOd \}_{i=1}^N$, single rotation averaging seeks to find the best average rotation $\vxx = \MR \in \mySOd$. The residual function is chosen as the chordal distance between $\MR$ and $\tldMR_i$: $r(\vxx, \vz_i) = \Vert \MR - \tldMR_i \Vert$. Fig.~\ref{fig:applications}(a) plots an instance of 3D single rotation averaging with $20$ measurements (rotations are plotted as 3D coordinate frames), among which there is a single outlier (shown as transparent).
\end{example}

\begin{example}[Multiple Rotation Averaging \cite{Eriksson18cvpr-strongDuality,Carlone16TRO-planarPGO,Lajoie19ral-DCGM}] 
\label{ex:multirotation}
Let $\calG = (\calV,\calE)$ be an undirected graph with vertex set $\calV = [n]$ and edge set $\calE$. Each vertex $i \in \calV$ is associated with an unknown rotation $\MR_i \in \mySOd$ (typically $\dimrot=2$ or $\dimrot=3$), while each edge $(i,j) \in \calE$ gives a relative rotation measurement $\tldMR_{ij} \in \mySOd$ between the unknown rotations at vertex $i$ and $j$.
Multiple rotation averaging estimates the set of absolute rotations on the vertices $\vxx = \{\MR_i\}_{i \in \calV} \in \mySOd^{n}$ from relative measurements over $\calE$. The residual function is chosen as the chordal distance between $\MR_i \tldMR_{ij}$ and $\MR_j$ for $(i,j) \in \calE$: $r(\vxx,\vz_{ij}) = \Vert \MR_i \tldMR_{ij} - \MR_j \Vert$. 
\maybeOmit{Optionally, if a set $\calR$ of relative measurements, 
 is known to be free of outliers (\eg odometry measurements in robot navigation), then a regularization $\regularizer = \sum_{(i,j) \in \calR} \Vert \MR_i \tldMR_{ij} - \MR_j \Vert^2$ is added to \eqref{eq:robust}.} 
 Fig. \ref{fig:applications}(b) plots an instance of 2D multiple rotation averaging with $9$ (unknown) absolute rotations and $11$ (measured) relative measurements, two of which are outliers (shown in red).
\end{example}

\begin{example}[Point Cloud Registration \cite{Yang20tro-teaser}]
\label{ex:pointcloud} 
Given two sets of 3D points with putative correspondences $\{ \vz_i = (\vp_i, \vq_i) \}_{i=1}^{N}$ (\eg matched by deep-learned features \cite{Yang21cvpr-sgp}), point cloud registration seeks the best rigid transformation $\vxx = (\MR,\vt) \in \SOthree \times \Real{3}$ to align them.
The residual function is chosen as the Euclidean distance between pairs of points after applying the rigid transformation: $r(\vxx,\vz_i) = \norm{\vq_i - \MR \vp_i - \vt}$. For mathematical convenience (\ie to satisfy the Archimedeanness condition in Theorem \ref{thm:lasserre}), we assume the translation to be bounded: $\vt \in \ball^3_T$, where $ \ball^q_T \triangleq \{\vt \in \Real{q}\mid \norm{\vt} \leq T \}$ defines a $q$-dimensional ball centered at the origin with radius $T$. Fig. \ref{fig:applications}(c) plots an instance of point cloud registration using the {\bunny} dataset \cite{Curless96siggraph} (outlier correspondences are shown in red).

\end{example}

\begin{example}[Mesh Registration \cite{Briales17cvpr-registration,Shi21icra-robin}] 
\label{ex:mesh}
Given a set of $N$ putative correspondences from a 3D point cloud to a 3D mesh, where the point cloud $\{(\vp_i,\vu_i) \}_{i=1}^N$ is represented as a collection of points ($\vp_i \in \Real{3}$) with estimated normals ($\vu_i \in \usphere{2}$), and the mesh $\{(\vq_i,\vv_i )\}_{i=1}^N$ is represented as a collection of faces with unit normals $(\vv_i \in \usphere{2})$ and arbitrary points that belong to them $(\vq_i \in \Real{3})$, mesh registration seeks the best rigid transformation $\vxx = (\MR,\vt) \in \SOthree \times \Real{3}$ to align the point cloud with the mesh. The residual function is chosen as: $r(\vxx,\vz_i) =\sqrt{ \norm{\inprod{\vv_i}{\vq_i - \MR \vp_i - \vt}}^2 + \norm{\vv_i - \MR \vu_i}^2 }$, where $\norm{\inprod{\vv_i}{\vq_i - \MR \vp_i - \vt}}$ is the point-to-plane distance, and $\norm{\vv_i - \MR \vu_i}$ is the normal-to-normal distance. Similar to Example \ref{ex:pointcloud}, we enforce $\vt \in \ball^3_T$. Fig. \ref{fig:applications}(d) visualizes an instance of mesh registration using the {\teddybear} model from the {\homebrew} dataset \cite{Kaskman19-homebrewedDB} (outlier correspondences shown in red).
\end{example}

\begin{example}[Absolute Pose Estimation \cite{Kneip2014ECCV-UPnP,Schweighofer2008bmvc-SOSforPnP,Yang21arxiv-damp}]
\label{ex:absolutepose}
Consider a camera with field of view (FOV) $\alpha \in (0,\pi)$ picturing a 3D object {(conventionally centered at zero)}. 
Given a set of $N$ putative correspondences between 3D keypoints $\{\vp_i \in \Real{3} \}_{i=1}^N$ on the object and 2D image keypoint detections $\{\vu_i \in \usphere{2}\}_{i=1}^N$,
where $\vu_i$ denotes the unit bearing vector corresponding to the $i$-th 2D keypoint, absolute pose estimation (also known as \emph{Perspective-$n$-Points}) seeks to estimate the absolute camera pose $\vxx = (\MR,\vt)\in \SOthree \times \Real{3}$ from the 2D-3D correspondences. The residual function is chosen as: $r(\vxx,\vz_i)\!=\!\sqrt{\inprod{\MR \vp_i + \vt}{(\eye_3 - \vu_i\vu_i\tran)(\MR \vp_i + \vt)} }$, \ie the point-to-line distance from the transformed 3D keypoint $\MR\vp_i + \vt$ (in camera frame) to the bearing vector $\vu_i$.\footnote{Instead of using the geometric reprojection error as the residual (a rational function), we follow \cite{Yang21arxiv-damp,Schweighofer2008bmvc-SOSforPnP} and choose the point-to-line distance as the residual so that $r^2$ is a polynomial per Assumption \ref{assumption:polynomialsrobust}.} 
In this paper, we enforce $\vt \in \ball^3_T \cap \calC_{\alpha}$,
where $\calC_{\alpha}\triangleq \{ \vt \in \Real{3} \mid \tan(\frac{\alpha}{2}) t_3 \geq \sqrt{t_1^2 + t_2^2} \}$ defines the 3D cone corresponding to the camera FOV;
the constraint $\vt \in \calC_\alpha$ enforces 
the center of the 3D object (\ie~$\MR\cdot\zero + \vt = \vt$ in camera frame) to lie inside the FOV. 
Fig.~\ref{fig:applications}(e) shows an instance of absolute pose estimation using a satellite \mbox{image from the {\speed} dataset~\cite{Sharma19arxiv-SPEED} (outliers in red).}
\end{example}

\begin{example}[Category-Level Object Pose and Shape Estimation \cite{Shi21rss-pace}]
\label{ex:category}
Given $N$ 3D semantic keypoint observations $\{ \vp_i \}_{i=1}^N$ of an object of a certain category (\eg car, chair), 
category-level perception estimates the object pose and shape. 
We consider the standard \emph{active shape model}, where the unknown shape of the object is described as a nonnegative combination 
of $K$ shapes in a library $\{ \{\vq_{k,i}\}_{i=1}^N \}_{k=1}^K$ (the \emph{bases}, which intuitively correspond to examples of objects in that category).
Hence,  category-level perception estimates the pose $(\MR,\vt) \in \SOthree \times \Real{3}$ and shape coefficients $\vc \in \bbR^{K}_{+}$ describing the object.
The residual function is chosen as: $r(\vxx,\vz_i) = \Vert \MR \vp_i + \vt - \sum_{k=1}^K c_k \vq_{k,i} \Vert$, \ie the Euclidean distance between the transformed 3D keypoint detections and the nonnegative combination of the shape bases. We include $\psi(\vxx,\lambda) = \lambda \norm{\vc}^2$ as a regularization for the shape parameters $\vc$, as in~\cite{Shi21rss-pace}. Again, we enforce $\vt \in \ball^3_T, \vc \in \ball^K_T$ to be both bounded. Fig. \ref{fig:applications}(f) pictures an example of category-level perception from the {\apollo} dataset~\cite{Wang19pami-apolloscape}, where one estimates the pose and shape of a vehicle given 2D semantic keypoint detections with associated depth values (outliers shown in red).
\end{example}


\begin{proposition}[Polynomial Expressibility]
\label{prop:polynomialExpressibility}
Examples \ref{ex:singlerotation}-\ref{ex:category} satisfy Assumption~\ref{assumption:polynomialsrobust}. Precisely, (i) $r^2$ and $\psi$ (if $\psi \neq 0$) are quadratic polynomials (\ie $\deg{r^2}=\deg{\psi}=2$); (ii) the constraint set $\calX$ can be described by polynomial equalities $h_i$'s and inequalities $g_j$'s with degree up to 2 (\ie $\deg{h_i},\deg{g_j} \leq 2$).
\end{proposition}
While we postpone the proof to \supp, we observe that the key insights behind the proof are simple but powerful: (i) rigid body transformations can be expressed as linear functions (\eg $\MR \vp_i + \vt$ for a given point $\vp_i$), 
(ii) squared residuals $r^2$ (and our regularizer $\psi$) are commonly squared L2 norms, that can be written as quadratic functions, 
and (iii) the set of poses and rotations can be described by quadratic (in-)equality constraints, a fact already used in, \eg~\cite{Tron15RSSW-rotationdeterminant,Carlone15icra-verification,Briales18cvpr-global2view,Yang20cvpr-perfectshape}.

Proposition \ref{prop:robustaspop} and \ref{prop:polynomialExpressibility} together establish that outlier-robust geometric perception \eqref{eq:robust} with TLS, MC, GM, TB, L1, Huber and Adaptive costs (Fig. \ref{fig:robust-costs}), when applied to Examples \ref{ex:singlerotation}-\ref{ex:category} (Fig. \ref{fig:applications}), are instances of~\eqref{eq:pop}. The expert reader will also recognize other geometric perception problems that satisfy Assumption~\ref{assumption:polynomialsrobust}, including 2D-2D relative pose estimation \cite{Briales18cvpr-global2view}, triangulation \cite{Aholt12eccv-qcqptriangulation}, rotation search (Wahba problem) \cite{Yang19iccv-quasar}, 
pose graph optimization \cite{Rosen19IJRR-sesync}, among others. \revise{Although the bundle adjustment problem \cite{Agarwal10eccv} cannot be written as a POP using the geometric reprojection error, adopting the point-to-line error can put bundle adjustment in the form of a POP \cite{Schweighofer06bmvc}. Nevertheless, bundle adjustment typically involves too many variables (\eg hundreds of camera poses and hundreds of thousands of 3D points) to be practically solvable using existing semidefinite relaxations.}

For the rest of the paper, we will focus on designing certifiable algorithms and semidefinite relaxations for~\eqref{eq:robust} with the~\eqref{eq:binaryTLS} cost function. However, semidefinite relaxations proposed in Section \ref{sec:sdprelax} can be extended to the other costs in Proposition~\ref{prop:robustaspop}, and we leave that exercise to the interested reader. We end this section with a remark about why we prefer the TLS cost over the others in Proposition~\ref{prop:robustaspop}.

\begin{remark}[Preference for TLS]
\label{remark:TLSvsothers}
(i) Compared to GM, TB, L1 and Huber, which still penalize outliers, TLS completely discards outliers. Consequently, TLS can often achieve better robustness to outliers \cite{Yang20ral-gnc,MacTavish15crv-robustEstimation}. (ii) MC also completely discards outliers, but it does not select a model to minimize the inlier residuals. Therefore, there can be an infinite number of solutions to problem \eqref{eq:robust} with equal cost (number of outliers). (iii) The adaptive cost typically leads to POPs with high-degree polynomials, which requires a large $\kappa$ from the relaxation hierarchy and results in SDPs that are intractable.  (iv) TLS can be shown as a maximum likelihood estimator, when the inliers have a Gaussian distribution and the outliers are 
uniformly distributed, see \cite[Proposition 5]{Antonante20TRO-outlier}.
\end{remark}

\section{Sparse Semidefinite Relaxation}
\label{sec:sdprelax}

In the previous section, we showed how to rephrase the TLS cost as a nonconvex polynomial optimization in $\tldvxx \triangleq [\vxx \vcat \vtheta] \in \Real{d+N}$. The goal of this section is to design algorithms that can solve~\eqref{eq:binaryTLS} to certifiable global optimality. 

{\bf Can we just use Lasserre's hierarchy?} 
Before introducing our sparse semidefinite relaxation, let us attempt to apply the dense Lasserre's hierarchy \eqref{eq:lasserre}  
to~\eqref{eq:binaryTLS}. We know that the objective in \eqref{eq:binaryTLS} has degree $3$,\footnote{The residuals $r^2(\vxx,\vz_i)$ are quadratic from Proposition \ref{prop:polynomialExpressibility}, 
hence the terms $\theta_i r^2(\vxx,\vz_i)$ in the objective of \eqref{eq:binaryTLS} become cubic.} thus $\kappa\geq2$ is needed for \eqref{eq:lasserre}. In fact, as we have shown in \cite{Yang20neurips-onering}, \eqref{eq:lasserre} at $\kappa=2$ is empirically exact (on small problem instances). However, as we can see from Examples \ref{ex:singlerotation}-\ref{ex:category}, the problems we care about have minimum $d=9$ (a 3D rotation in Example \ref{ex:singlerotation}) and maximum $d=9n$ ($n$ 3D rotations in Example \ref{ex:multirotation}) with $n$ being as large as a few hundreds, and meanwhile, it is desirable to be able to handle $N=100$ measurements. Choosing $d=10,N=100,\kappa=2$, the SDP resulting from the dense relaxation \eqref{eq:lasserre} has $n_1 =6216,\mmom =12,649,561$; when $d=100,N=100,\kappa=2$, such SDP would have $n_1 \approx 2 \times 10^4, \mmom \approx 1.4\times 10^8$. In both cases, 
it is hopeless to solve the resulting SDPs using existing solvers.

{\bf Sparse semidefinite relaxation (SSR)}. Now we present a semidefinite relaxation that is much more scalable than \eqref{eq:lasserre}. Note that the fundamental reason why \eqref{eq:lasserre} leads to an intractable SDP is the use of the \emph{dense} monomial basis $[\tldvxx]_\kappa$ for building the moment matrix $\MX_\kappa$. 
Although the full set of monomials $[\tldvxx]_\kappa$ is necessary when the polynomials $p,h_i,g_j$ 
contain all monomials up to degree $2\kappa$, in practice $p,h_i,g_j$ are almost always \emph{sparse} (\ie include a small set of monomials). Therefore, the crux of our semidefinite relaxation is to construct a {sparse} set of monomials 
that result in a much smaller moment matrix. Towards this, we analyze the sparsity of the objective and constraint polynomials in~\eqref{eq:binaryTLS} and observe they only contain three types of monomials: 
\begin{enumerate}[label=(\roman*)]
\item \label{item:tlsmono1} $[\vxx]_2$, coming from $r^2$ and $\psi$ in the objective, and polynomials defining the feasible set $\calX$ (\cf Proposition \ref{prop:polynomialExpressibility});

\item \label{item:tlsmono2} $\theta_i \cdot [\vxx]_2,i=1\dots,N$, coming from $\theta_i r^2$ and $\theta_i$ in the objective for $i=1,\dots,N$; and

\item \label{item:tlsmono3} $\theta_i^2,i=1,\dots,N$, coming from the equality constraints $\theta_i^2-1=0$ for $i=1,\dots,N$.
\end{enumerate}
Therefore, it is easy to see that, with the Kronecker product denoted by ``$\kron$'', choosing the sparse basis
\bea\label{eq:sparsebasis}
\vv(\tldvxx) \triangleq [1 \vcat \vxx \vcat \vtheta \vcat \vtheta \kron \vxx] \in \Real{n_1},\ n_1 \triangleq (1+d)(1+N)
\eea
leads to the following moment matrix
\bea \label{eq:sparsemomentmat}
\MX_v \triangleq \vv \vv\tran\!\! =\!\! 
\bmat{cccc}
1 & \vxx\tran & \vtheta\tran & \vtheta\tran \kron \vxx\tran \\
\vxx & \vxx \vxx\tran & \vxx\vtheta\tran &\!\!\!\! \vxx (\vtheta\tran \kron \vxx\tran)\!\!\!\! \\
\vtheta & \vtheta \vxx\tran & \vtheta \vtheta\tran &\!\!\!\! \vtheta (\vtheta\tran \kron \vxx\tran)\!\!\!\! \\
\!\!\!\vtheta \kron \vxx\! &\!\!\! (\vtheta \kron \vxx)\vxx\tran\!\!\! &\!\!\! (\vtheta \kron \vxx) \vtheta\tran\!\! &\!\!\!\! \vtheta\vtheta\tran \kron \vxx\vxx\tran\!\!\!\!
\emat
\eea
that contains all the three types of monomials ($[\vxx]_2$, $\theta_i \cdot [\vxx]_2$, and $\theta_i^2$) in \ref{item:tlsmono1}-\ref{item:tlsmono3}. Therefore, \emph{we can write the objective and constraint polynomials in \eqref{eq:binaryTLS}  
as linear functions of the smaller moment matrix~\eqref{eq:sparsemomentmat}.} 
Clearly, the advantage is that the size of the moment matrix is now $(1+d)(1+N)$, which is much smaller than $\nchoosek{d+N+\kappa}{\kappa}$ (for $\kappa=2$) from Lasserre's hierarchy.

Now we can formulate our sparse relaxation using $\MX_v$ in~\eqref{eq:sparsemomentmat}, by following the same procedure as in Section \ref{sec:pre-pop}.

\emph{(i) Rewriting \eqref{eq:binaryTLS} using the sparse moment matrix $\MX_v$}. Because the sparse moment matrix $\MX_v$ contains all monomials in the objective and constraint polynomials of \eqref{eq:binaryTLS}, we can write them as linear functions of $\MX_v$. For example, the objective 
can be written as $\inprod{\MC_1}{\MX_v}$.

\emph{(ii) Relaxing the rank-$1$ constraint on $\MX_v$}. By construction, $\MX_v$ belongs to the set of rank-one positive semidefinite matrices. Since the rank constraint is non-convex, we drop it and only enforce $\MX_v$ to be positive semidefinite: $\MX_v \succeq 0$.

\emph{(iii) Adding redundant constraints}. First, similar to the dense relaxation case, some monomials can repeat themselves at multiple entries of $\MX_v$. For example, in \eqref{eq:sparsemomentmat}, the ``$\vtheta \kron \vxx$'' block is the same as the ``$\vtheta \vxx\tran$'' block up to rearrangement of entries. In fact, the number of \emph{unique} monomials in $\MX_v$ is $m_{2v} = \trinum(d+1)\trinum(N+1)$, while the dimension of $\MX_v$ (in terms of a symmetric matrix) is $\trinum((1+d)(1+N))$. Therefore, we can add a total number of $\mmom = \trinum((1+d)(1+N)) - m_{2v} + 1$ \emph{moment constraints}:
\beal\label{eq:momentConstraintssparse}
\text{\grayout{moment constraints}}:&  \revise{\inprod{\MA_{\mathmom,0}}{\MX_v} = 1, } \\
& \inprod{\MA_{\mathmom,j}}{\MX_v} = 0, \\
&  j = 1, \ldots, \mmom-1,
\eeal
to enforce the repeating monomials in $\MX_v$ to be equal to each other, as well as the leading entry $[\MX_v]_{11} = 1$ \revise{(similar to \eqref{eq:momentConstraints}, $\MA_{\mathmom,0}$ is all zero except $[\MA_{\mathmom,0}]_{11} =1$)}.

Second, we add redundant equality constraints. For each equality constraint $h_i$ in $\eqref{eq:binaryTLS}$, we denote $[\tldvxx]_{h_i}$ as the largest set of unique monomials such that $h_i \cdot [\tldvxx]_{h_i}$ only contains monomials in $\MX_v$. Formally,
\bea
[\tldvxx]_{h_i} \triangleq \{\tldvxx^{\valpha} \mid \mono{ h_i \cdot \tldvxx^{\valpha} } \subseteq \mono{\MX_v} \}, \label{eq:liftequalities}
\eea
where $\mono{\cdot}$ returns the set of unique monomials of a polynomial (or of a matrix of polynomials). Consequently, we can write $h_i \cdot [\tldvxx]_{h_i} = \zero$ as linear equalities in $\MX_v$:
\beal\label{eq:redundantEqualityConstraintssparse}  
\hspace{-3mm} \text{\grayout{(redundant) equality constraints}}: \inprod{\MA_{\mathreq,ij}}{\MX_v} = 0, \\
\quad\quad \quad i = 1, \ldots, l_h,\ \ 
j = 1, \ldots, \abs{[\tldvxx]_{h_i}}.
\eeal 
Note that since each $[\tldvxx]_{h_i}$ must include the monomial ``1'', eq.~\eqref{eq:redundantEqualityConstraintssparse} includes the original equality constraints $h_i$ in $\eqref{eq:binaryTLS}$.

Finally, for each inequality constraint $g_j$ (recall $\deg{g_j} \leq 2$ by Proposition \ref{prop:polynomialExpressibility}), we denote by $[\MX_1]_{\calI_j,\calI_j}$ the maximum principal submatrix of $\MX_1$ (\ie order-one full moment matrix) such that $g_j \cdot [\MX_1]_{\calI_j,\calI_j}$ only contains monomials in $\MX_v$. Formally, the indices $\calI_j$ are selected as:
\bea
&  \hspace{-6mm} \calI_j  =  \displaystyle \argmax_{\calJ} \{ \abs{\calJ} \mid \mono{ g_j\! \cdot\! [\MX_1]_{\calJ,\calJ} } \subseteq \mono{\MX_v} \}. \label{eq:liftPSDsubblks}
\eea
As a result, calling $\MX_{g_j} = g_j \cdot [\MX_1]_{\calI_j,\calI_j}$, which is positive semidefinite by construction, we can write down the following localizing matrices and constraints:
\beal\label{eq:locMatricessparse}
\text{\grayout{localizing matrices}}: & \MX_{g_j} \succeq 0, \;\; j=1,\ldots,l_g
\eeal 
\beal \label{eq:localizingConstraintssparse}  
\text{\grayout{{localizing} constraints}}: \inprod{\MA_{\mathloc,jkh}}{\MX_v} = [\MX_{g_j}]_{hk} \\
\quad\quad\quad j = 1, \ldots, l_g, \ \ 
1 \leq h\leq k \leq \abs{\calI_j},
\eeal
where the linear constraints in \eqref{eq:localizingConstraintssparse} simply enforce each entry of $\MX_{g_j}$ to be a linear combination of entries in $\MX_v$.

Steps (i)-(iii) above lead to the following SDP:
\begin{equation}\label{eq:sparserelax}
\hspace{-3mm} \fstar =\!\! \min_{\MX = (\MX_v, \MX_1,\dots,\MX_{l_g})} \cbrace{\inprod{\MC_1}{\MX_v} \mid \calA(\MX)\! =\! \vb, \MX \succeq 0}\!,\!\!\! \tag{SSR}
\end{equation}
where we have shorthanded $\MX_j = \MX_{g_j}$ for notation convenience, and $\calA(\MX)=\vb$ collects all the linear equality constraints in \eqref{eq:momentConstraintssparse}, \eqref{eq:redundantEqualityConstraintssparse} and \eqref{eq:localizingConstraintssparse}.

Similar to Theorem \ref{thm:lasserre} for \eqref{eq:lasserre}, we have the following result for \eqref{eq:sparserelax} about certifiable global optimality.

\begin{theorem}[Sparse Semidefinite Relaxation for \eqref{eq:binaryTLS}]
\label{thm:sparserelaxtls}
Denote by $p(\vxx,\vtheta)$ the objective function of \eqref{eq:binaryTLS}, by $\pstar$ the optimum of \eqref{eq:binaryTLS}, and by $\fstar$ \revise{(resp. $\MXstar_v$)} the optimum \revise{(resp. one optimizer)} of \eqref{eq:sparserelax}, we have
\begin{enumerate}[label=(\roman*)]
\item (lower bound) $\fstar \leq \pstar$;
\item (rank-one solutions) if $\fstar = \pstar$, then for each global minimizer $\tldvxxstar = (\vxxstar,\vthetastar)$ of \eqref{eq:binaryTLS}, its rank-one lifting $\MX_v = \vv (\tldvxxstar) \vv (\tldvxxstar)\tran$ is optimal for \eqref{eq:sparserelax}, \revise{and every rank-one optimal solution $\MXstar_v$ of \eqref{eq:sparserelax} can be written as $\vv (\tldvxxstar) \vv (\tldvxxstar)\tran$ for some $\tldvxxstar$ that is optimal for \eqref{eq:binaryTLS}};
\item \revise{(optimality certificate) if $\rank{\MXstar_v} = 1$, then $\fstar = \pstar$.}
\end{enumerate}
\end{theorem}
Theorem \ref{thm:sparserelaxtls} states that \eqref{eq:sparserelax} is a relaxation for \eqref{eq:binaryTLS} \revise{and solving the convex SDP \eqref{eq:sparserelax} can provide a certificate for the exactness of the relaxation if the rank of the optimal solution $\MXstar_v$ equals one. In practice, rank computation can be subject to numerical inaccuracies (\eg it can be difficult to decide if the relaxation is exact when the second largest eigenvalue is, say $10^{-3}$). Therefore, we now introduce a continuous metric for evaluating the exactness of the relaxation (that also applies to the dense relaxation \eqref{eq:lasserre}).
}

{\bf Relative suboptimality}. 
\revise{Assume $\MXstar_v$ is an optimal solution of \eqref{eq:sparserelax} and let $\vv$ be the eigenvector corresponding to the maximum eigenvalue of $\MXstar_v$. If the maximum eigenvalue has multiplicity larger than one, then choose $\vv$ as any of the eigenvectors corresponding to the maximum eigenvalue.
Define the rounding function $(\hatvxx,\hatvtheta) = {\rounding}(\vv)$ that returns from $\vv$ a \emph{feasible} solution to \eqref{eq:binaryTLS} as}
\begin{equation}\label{eq:rounding}
\vv \leftarrow \vv / \vv_1,\ \hatvxx = \Pi_{\calX} (\vv_{x}),\ \hatvtheta = \sign\parentheses{\vv_{\theta}},
\end{equation}
where $\vv_{x}$ (resp. $\vv_{1},\vv_{\theta}$) takes the entries of $\vv$ corresponding to monomials $\vxx$ (resp. $1,\vtheta$) in \eqref{eq:sparsebasis}, $\sign(a)$ returns the sign of a scalar ``$a$'', and $\Pi_{\calX}$ denotes the projection onto set $\calX$.\footnote{For our Examples \ref{ex:singlerotation}-\ref{ex:category}, the feasible set $\calX$ includes $\mySOd$, whose projections can be performed in closed form, and $\ball^3_T$, $\calC_\alpha$, $\ball^3_T \cap \calC_\alpha$, $\ball^K_T \cap \bbR^K_{+}$, all of which are \emph{low-dimensional convex} sets whose projections can be computed to arbitrary accuracy using standard convex solvers. Therefore, the {\rounding} procedure~\eqref{eq:rounding} can be done efficiently.} Denoting $\hatp \triangleq p(\hatvxx,\hatvtheta)$ as the cost attained by the rounded solution, we have $\fstar \leq \pstar \leq \hatp$. Moreover, we can compute a \emph{relative suboptimality} of the rounded solution $(\hatvxx,\hatvtheta)$
\bea \label{eq:subopt}
\subopt \triangleq \abs{\fstar - \hatp}/ \parentheses{ 1 + \abs{\fstar} + \abs{\hatp} }
\eea
as a measure of suboptimality. \revise{Intuitively, the relative suboptimality certifies that a solution $(\hatvxx,\hatvtheta)$ with objective value \emph{at most} $\subopt$ (\eg $0.1\%$) away from the unknown global optimizer has been found.} Evidently, $\subopt = 0$ implies $(\hatvxx,\hatvtheta)$ is optimal and \eqref{eq:sparserelax} is exact. \revise{In fact, for \emph{any feasible solution} $(\hatvxx,\hatvtheta)$, not necessarily obtained from the SDP solution $\MXstar_v$, we can evaluate $\hatp = p(\hatvxx,\hatvtheta)$ to compute the relative suboptimality at the given feasible solution using \eqref{eq:subopt}. Similarly, if $\subopt = 0$ is attained at any feasible solution, we can certify the exactness of the relaxation and the global optimality of the feasible solution.}
As an advanced reading, in \supp, we discuss how to compute a relative suboptimality measure that is not sensitive to potential numerical inaccuracies in 
the computation of $\fstar$ \revise{(as mentioned in Section \ref{sec:pre-sdp}, it can be challenging to compute $\fstar$ to high accuracy for large-scale SDPs)}. 

\begin{figure}
\centering
\includegraphics[width=0.8\columnwidth]{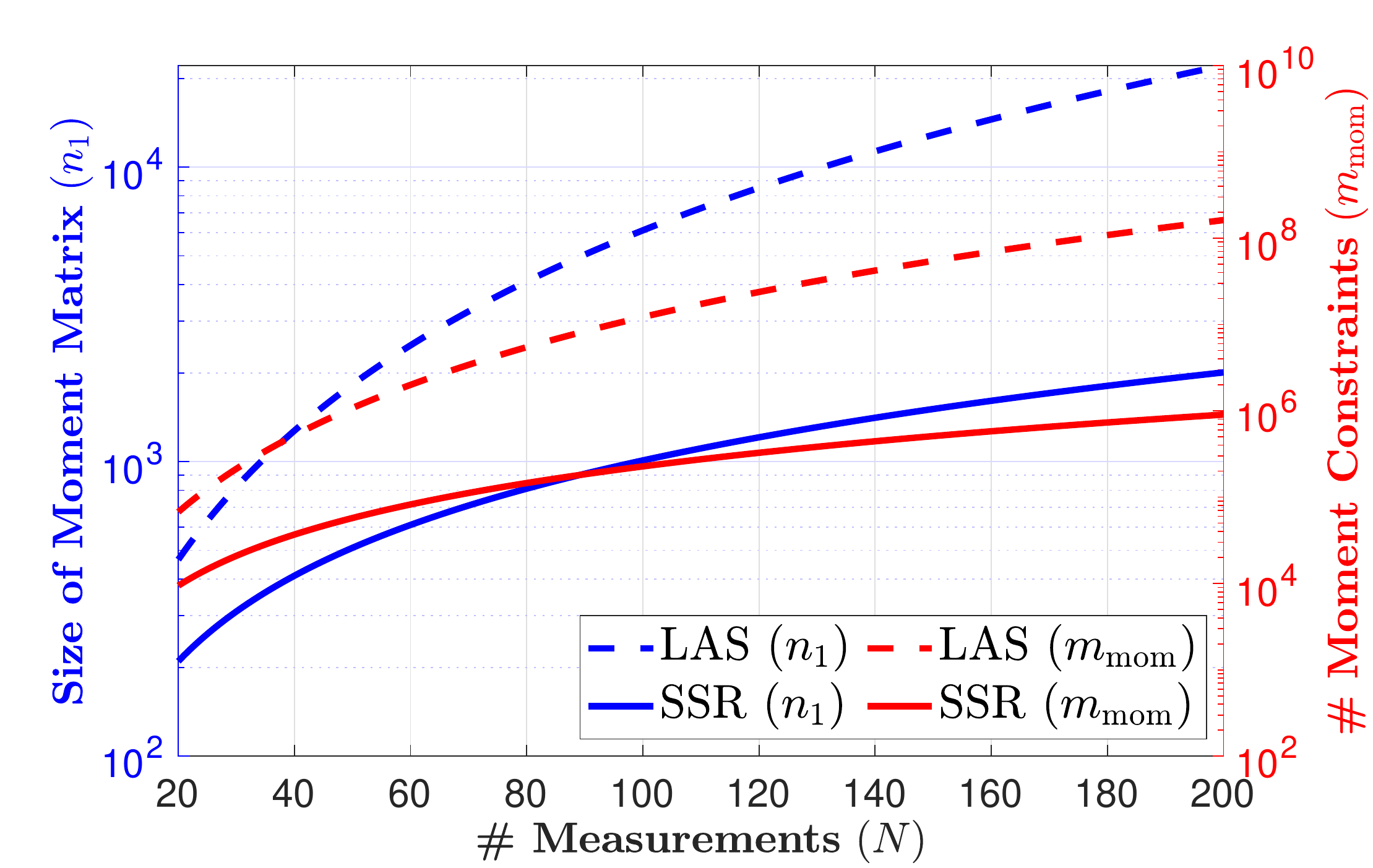}
\vspace{-4mm}
\caption{Comparison of SDP sizes when applying \eqref{eq:lasserre} and \eqref{eq:sparserelax} to Example \ref{ex:singlerotation} ($d=9$) with $N$ increased from $20$ to $200$.\label{fig:LASvsSSR} }
\vspace{-6mm}
\end{figure}

{\bf Scalability improvement}. 
Table \ref{table:LASvsSSR} compares the size of the SDP from our sparse relaxation \eqref{eq:sparserelax} with that from the standard Lasserre's hierarchy \eqref{eq:lasserre}, in terms of the size of the maximum positive semidefinite block $n_1$ and the number of moment constraints $\mmom$ (in our problems, over $60\%$ of the equality constraints are moment constraints, hence $\mmom$ is representative of the size of the SDP). For illustration purpose, Fig. \ref{fig:LASvsSSR} plots $n_1$ and $\mmom$ as $N$ increases from $20$ to $200$, when applying \eqref{eq:lasserre} and \eqref{eq:sparserelax} to Example \ref{ex:singlerotation} ($d=9$). We can observe a drastic reduction in both $n_1$ and $\mmom$ when using \eqref{eq:sparserelax}. Notably, when $N=200$, $n_1 > 20,000$ and $\mmom > 100,000,000$ if using \eqref{eq:lasserre}, while $n_1 \approx 2,000$ and $\mmom \approx 1,000,000$ if using \eqref{eq:sparserelax}. This is about $10$ times reduction in $n_1$ and $100$ times reduction in $\mmom$. Certainly, such scalability improvement would be meaningless if \eqref{eq:sparserelax} is \emph{inexact} and fails to solve the original \eqref{eq:binaryTLS} problem to global optimality. However, as we will show in Section \ref{sec:experiments}, \eqref{eq:sparserelax} is {empirically exact across all Examples \ref{ex:singlerotation}-\ref{ex:category}, even in the presence of many outliers}.

\begin{table}[t]
\caption{Comparison of the sizes of two semidefinite relaxations applied to problem \eqref{eq:binaryTLS}. $n_1$: size of the largest positive semidefinite block (\ie the moment matrix), $\mmom$: number of moment constraints. \label{table:LASvsSSR}}
\vspace{-3mm}
\renewcommand{\arraystretch}{1.2}
\adjustbox{max width=\linewidth}{%
\begin{tabular}{c|c|c}
\hline
Relaxation & $n_1$ & $\mmom$ \\
\hline
\eqref{eq:lasserre}, $\kappa=2$ & $\binomialc{d+N}{2}$ & $\trinum( \binomialc{d+N}{2} ) - \binomialc{d+N}{4} + 1$  \\
\eqref{eq:sparserelax} & $(1+d)(1+N)$ & $\trinum((1+d)(1+N)) - \trinum(1+d)\trinum(1+N) + 1$ \\
\hline
\end{tabular}
}%
\vspace{-4mm}
\end{table}

{\bf Further reduction on Example \ref{ex:multirotation}}. For multiple rotation averaging, the dimension of the geometric model is $d=2n$ (2D) or $d=9n$ (3D), where $n$ is the number of nodes of a graph. Practical rotation averaging problems in structure from motion and SLAM can have $n$ and $N$ being a few hundreds to a few thousands \cite{Rosen19IJRR-sesync,Eriksson18cvpr-strongDuality}. Taking $d=400, N=20$ leads to $\mmom=16,842,001$ that is still too large. In \supp, we present a method to further reduce the size of the sparse monomial basis in \eqref{eq:sparsebasis}.

We end this section with a remark about how to exploit sparsity while preserving exactness of the relaxation.

\begin{remark}[Exploiting Sparsity]
\label{remark:sparsity}
(i) A sparse relaxation can be exact only when the dense relaxation \eqref{eq:lasserre} is \revise{exact}. Therefore, we believe it is good practice to first obtain an \revise{empirically exact} relaxation using \revise{the dense hierarchy} \eqref{eq:lasserre} at certain $\kappa$ (as we have done in \cite{Yang20neurips-onering} \revise{with extensive experimental validation}), and then try to find a sparse monomial basis at that $\kappa$. (ii) When the dense relaxation is exact, it is nontrivial to decide if a sparse relaxation will be \revise{exact} without empirical evaluation. For example, replacing \eqref{eq:sparsebasis} with $\vv(\tldvxx) = [[\vxx]_2 \vcat \vtheta]$ is also a sparse relaxation ---the corresponding moment matrix includes all monomials in \ref{item:tlsmono1}-\ref{item:tlsmono3}--- but it is far from being exact. (iii) Parallel to our work \cite{Yang20neurips-onering}, \cite{Wang21SIOPT-tssos} has presented a methodology, {\tssos}, to systematically exploit term sparsity for general POPs.
However, {\tssos} tends to find a larger monomial basis when compared to problem-specific techniques such as \eqref{eq:sparserelax} in this paper. For Example \ref{ex:absolutepose} with $N=10$ measurements, the dense monomial basis has dimension $276$, our sparse basis \eqref{eq:sparsebasis} has dimension $143$, but {\tssos} with ``maximal chordal extension'' finds a sparse basis that has dimension $246$ and is a strict superset of \eqref{eq:sparsebasis}.
\end{remark}

\section{{\sf STRIDE}: Scalable SDP Solver}
\label{sec:scalableopt}
The sparse relaxation \eqref{eq:sparserelax} leads to an SDP that can still have $m$ as large as hundreds of thousands when $N$ is large (\cf Fig.~\ref{fig:LASvsSSR}). Therefore, with IPMs such as \mosek, the scale at which \eqref{eq:sparserelax} can be solved is still quite limited (recall IPMs can typically handle $m$ up to $50,000$). This section presents \strideplus (\emph{\namelong}), 
an SDP solver that goes far beyond IPMs and enables solving~\eqref{eq:sparserelax} on problems of 
moderate but realistic size.

{\bf Intuition}. The key insight behind {\strideplus} comes from Theorem~\ref{thm:sparserelaxtls}(ii): assuming the relaxation \eqref{eq:sparserelax} is exact, then the SDP \eqref{eq:primalSDP} admits \emph{rank-one} optimal solutions $\MXstar_v = \vv(\tldvxxstar)\vv(\tldvxxstar)\tran$, where $\tldvxxstar = (\vxxstar,\vthetastar)$ corresponds to the global minimizer of \eqref{eq:binaryTLS}. 
Therefore, {\strideplus} tries to move between rank-one matrices in the feasible set of the SDP (these are the \emph{vertices} of the spectrahedron \cite{Blekherman12Book-sdpandConvexAlgebraicGeometry}), searching for a globally optimal solution. 
More in detail, 
{\strideplus} employs a globally convergent \revise{\emph{projected gradient descent} (PGD)} method as the backbone for solving the convex SDP \eqref{eq:sparserelax}, but blends \emph{short} \revise{\pgm} steps with \emph{long} rank-one steps generated by fast NLP algorithms on the POP~\eqref{eq:binaryTLS}.
Intuitively, the long rank-one steps circumvent the slow convergence of \revise{\pgm}, while the \revise{\pgm} backbone allows escaping local minima where the NLP algorithm can be stuck in.

With this insight, we now develop the details of \strideplus. 

{\bf Short {\pgm} step}.
The backbone of {\strideplus} implements a {\pgm} for solving the primal SDP \eqref{eq:primalSDP}. Given an initial point $\MX^0 \in \bbX$, the $k$-th ($k \geq 0$) iteration of {\pgm} performs
\begin{equation}\label{eq:pgd}
\MX^{k+1} = \Pi_{\setsdpp} \parentheses{\MX^k - \sigma_k \MC}, \tag{\pgm} 
\end{equation}
for a given constant $\sigma_k > 0$, where $\Pi_{\setsdpp}$ denotes the metric projection onto the spectrahedron $\setsdpp \triangleq \{\MX \in \bbX \mid \calA(\MX)=\vb, \MX \succeq 0 \}$ (\ie the feasible set of \eqref{eq:primalSDP}). In words, the \eqref{eq:pgd} step first moves along the direction of the negative gradient for some step size $\sigma_k$ (recall the objective of \eqref{eq:primalSDP} is $\inprod{\MC}{\MX}$ with a constant gradient $\MC$), and then projects the new point $\MX^k - \sigma_k \MC$ onto the feasible set $\setsdpp$. 
It is well known that \eqref{eq:pgd} guarantees to converge to an optimal solution of \eqref{eq:primalSDP}, provided that $\sigma_{k+1} \geq \sigma_{k}, \forall k \geq 0$ (see \cite{Jiang12siopt-PGMSDP,Beck09SIIS-FISTA,Bertsekas99book-nlp}). In \supp, we show the Lagrangian dual of the projection subproblem in \eqref{eq:pgd} 
can be reformulated as a \emph{smooth unconstrained optimization}, which allows solving~\eqref{eq:pgd} for large-scale problems using a limited-memory BFGS (\lbfgs) algorithm. \revise{For this reason, in \eqref{eq:strideprojection} we also output the dual optimal solution.}

{\bf Long rank-one step}.
The issue with \eqref{eq:pgd} is that the convergence can be slow, particularly when the optimal $\MXstar$ is rank-one and degenerate (as in \eqref{eq:sparserelax}). Here we propose to exploit the low-rankness of $\MXstar$ and accelerate the convergence by generating long rank-one steps. Towards this goal, calling $\barMX^{k+1} := \Pi_{\setsdpp}(\MX^k - \sigma_k \MC)$, and $\barMX_v^{k+1} \in \psd{n_1}$ as the first block in $\barMX^{k+1}$ (\ie the moment matrix), we compute a potentially better rank-one iterate via three steps:
\begin{enumerate}[label=(\roman*)]
\item\label{item:rounding} {\bf (Rounding)}. Let $\barMX^{k+1}_v = \sum_{i=1}^{n_1} \lambda_i \vv_i \vv_i\tran$ be the spectral decomposition of $\barMX^{k+1}_v$ with $\lambda_1 \geq \dots \geq \lambda_{n_1}$ in nonincreasing order. Compute $r$ hypotheses from the leading $r$ eigenvectors  
	\bea \label{eq:roundingrestate}
		(\barvxx^{k+1}_{i},\barvtheta^{k+1}_i) = \rounding(\vv_i), \quad i = 1,\dots,r,
	\eea
	where the function~\rounding~is defined as in \eqref{eq:rounding}.
	
\item {\bf (Local search)}. Apply a local search method for \eqref{eq:binaryTLS} using NLP with initial point chosen as $ (\barvxx^{k+1}_{i},\barvtheta^{k+1}_i) $ for each $i=1,\dots,r$. Denoting the solution of each local search as $ (\hatvxx^{k+1}_i,\hatvtheta_i^{k+1}) $, with associated objective value $p(\hatvxx^{k+1}_i,\hatvtheta_i^{k+1})$, choose the best local solution with \emph{minimum} objective value. Formally,
 	\begin{subequations}
	\bea
	\hspace{-3mm} (\hatvxx_i^{k+1},\hatvtheta_i^{k+1}) =&\!\!\!\! \nlp(\barvxx^{k+1}_i,\barvtheta^{k+1}_i),\ \ i=1,\dots,r, \label{eq:nlpinlocalsearch}\\
	\hspace{-3mm} (\hatvxx^{k+1},\hatvtheta^{k+1}) =&\!\!\!\! \displaystyle \argmin_{(\hatvxx^{k+1}_i,\hatvtheta^{k+1}_i), i=1\dots,r} p(\hatvxx^{k+1}_i,\hatvtheta^{k+1}_i).
	\eea
	\end{subequations}

\item\label{item:lifting} {\bf (Lifting)}. Perform a rank-one lifting of the best local solution $\tldvxx^{k+1} \triangleq (\hatvxx^{k+1},\hatvtheta^{k+1}) $
    \begin{subequations}\label{eq:lifting}
	\bea 
	\hspace{-3mm} \hatMX^{k+1}_v =&\!\!\!  \vv(\tldvxx^{k+1}) \vv(\tldvxx^{k+1})\tran, \ \  (\cf\ \eqref{eq:sparsebasis}) \\
	\hspace{-3mm} \hatMX^{k+1}_{g_j} =&\!\!\! \MX^{k+1}_{g_j} (\tldvxx^{k+1}), j = 1,\dots,l_g, \ \ (\cf\ \eqref{eq:liftPSDsubblks}) \\
	\hspace{-3mm} \hatMX^{k+1} = &\!\!\! (\hatMX^{k+1}_v,\dots,\hatMX^{k+1}_{g_j},\dots)_{j=1}^{l_g},
	\eea
	\end{subequations}
	where $\hatMX^{k+1},\hatMX^{k+1}_{g_j},j=1,\dots,l_g$ are computed by \emph{evaluating} the moment and localizing matrices at $\tldvxx^{k+1}$.
\end{enumerate}

{\bf Taking the right step}.
Now we are given two candidates for the next iteration, namely the short {\pgm} step $\barMX^{k+1}$ (generated by computing the projection of $\MX^k - \sigma_k \MC$ onto $\setsdpp$) and the long rank-one step $ \hatMX^{k+1} $ (obtained by rounding, local search, and lifting). Which one should we choose to be the next iterate $\MX^{k+1}$ such that the entire sequence $\{\MX^k\}$ is globally convergent? The answer to this question is quite natural --we accept $\hatMX^{k+1}$ if and only if it attains a strictly lower cost than $\barMX^{k+1}$ (\cf eq. \eqref{eq:accept-reject}). \maybeOmit{-- \red{and guarantees that the algorithm visits a sequence of rank-one vertices (\ie local minima via NLP) with descending costs}.}

\let\oldnl\nl
\newcommand{\nonl}{\renewcommand{\nl}{\let\nl\oldnl}}
\setlength{\intextsep}{0pt}
\begin{algorithm}[ht!]
\nonl 
Given $ (\MX^{0}, \MS^{0}, \vy^0)\in \calK\times\calK\times \Real{m} $, a tolerance $ \tol > 0 $, an integer $r \in [1,n]$, a constant $\epsilon>0$, a nondecreasing sequence $\{ \sigma_k >0 \}$. Perform the following steps for $ k = 0,1,\dots $.
\\
{\bf (Short \revise{\pgm} step)}. 
Solve the projection problem with initial point $(\MX^{k}, \MS^{k}, \vy^k)$
\bea \label{eq:strideprojection}
	 (\barMX^{k+1},\MS^{k+1},\vy^{k+1}) = \Pi_{\setsdpp}\left( \MX^{k} - \sigma_k \MC \right),
\eea
using the L-BFGS algorithm in \cite[Section 4]{Yang21arxiv-stride}. \label{line:shortpgm}
\\
{\bf (Long rank-one step)}. Compute a candidate rank-one iterate $ \hatMX^{k+1} $ from \eqref{eq:roundingrestate}-\eqref{eq:lifting}.
\\
{\bf (Update primal variable)}. Update $\MX^{k+1}$ as 
\begin{equation}
	\label{eq:accept-reject}
	\hspace{-4mm}
	\MX^{k+1} = 
		\begin{cases}
			\hatMX^{k+1} & {\rm if }\; \substack{\displaystyle f(\hatMX^{k+1}) \leq  f(\barMX^{k+1}) - \epsilon \\ \displaystyle \hatMX^{k+1} \in \setsdpp }  \\
			\barMX^{k+1} & {\rm otherwise }
		\end{cases}.
\end{equation}
\\
{\bf (Check convergence)}. 
Compute KKT residuals at $(\MX^{k+1}, \MS^{k+1}, \vy^{k+1})$ from \eqref{eq:KKTresiduals}, if $\kkt < \tol$, then output $(\MX^{k+1}, \MS^{k+1}, \vy^{k+1})$. Otherwise, go to Step \ref{line:shortpgm}. 
\caption{{\strideplus} \label{alg-iPGMnlp}}
\end{algorithm}

The full {\stride} algorithm is presented in Algorithm~\ref{alg-iPGMnlp}.
\begin{theorem}[Global Convergence]\label{thm:strideconverge}
Suppose the Slater condition for \eqref{eq:primalSDP} holds and $\{ (\MX^k,\vy^k,\MS^k) \}$ is generated by {\strideplus}, then $\{f(\MX^k) \}$ converges to $\fstar$, where $\fstar$ is the optimum of \eqref{eq:primalSDP}.
\end{theorem}

While we provide the proof in \supp, the intuition is that 
eq. \eqref{eq:accept-reject} ensures the rank-one ``strides'' are accepted only if they strictly decrease the objective value.
Therefore, either the last rank-one point is already optimal, or ---if it is suboptimal--- it still provides an improved reinitialization for \eqref{eq:pgd} to globally converge to the optimal $\MXstar$.
Note that the \revise{\pgm} backbone allows {\strideplus} to converge even when the optimal solution has rank higher than one. \revise{In \cite{Yang21arxiv-stride}, we show it is also possible to \emph{accelerate} and \emph{generalize} the \eqref{eq:pgd} backbone using \emph{proximal gradient methods}.}

Although {\stride} is a globally convergent algorithm for solving the primal SDP \eqref{eq:primalSDP}, the initial guess $(\MX^0,\MS^0,\vy^0)$ can have a significant impact on its convergence speed. The next remark states that existing fast heuristics for robust perception can be readily incorporated into {\stride}.

\begin{remark}[Fast Heuristics and Certification]
\label{remark:fastheuristics}
Existing fast heuristics for robust estimation, such as graduated non-convexity (\gnc) \cite{Yang20ral-gnc,Black96ijcv-unification} and {\ransac} \cite{Fischler81}, can typically return the \emph{globally optimal} solution to \eqref{eq:binaryTLS} when the measurement set $\calZ$ contains a low or medium portion of outliers (\eg below $70\%$). Therefore, we use {\gnc} or {\ransac} to generate an initial guess for the SDP relaxation \eqref{eq:sparserelax}. Formally, calling $(\hatvxx,\hatvtheta)$ the candidate solution obtained by solving \eqref{eq:binaryTLS} using {\gnc} or {\ransac}, we generate $\MX^0$ (for {\stride}) by applying the lifting procedure in \eqref{eq:lifting} to $(\hatvxx,\hatvtheta)$. Notably, when $(\hatvxx,\hatvtheta)$ is already globally optimal to \eqref{eq:binaryTLS} (hence $\MX^0$ is an optimizer of \eqref{eq:sparserelax} as long as the relaxation is exact), {\stride} only finds a \emph{certificate of optimality} for $(\hatvxx,\hatvtheta)$ by performing one step of \eqref{eq:pgd} (\cf \eqref{eq:strideprojection} in Algorithm \ref{alg-iPGMnlp}). 
\end{remark}

Fast heuristics provide a good \emph{primal} initialization for {\stride}. However, little information is known about how to obtain a good \emph{dual} initialization. In {\supp}, we describe a dual initialization procedure that exploits \emph{correlative sparsity} \cite{Wang21siopt-chordaltssos} and leverages a fast first-order algorithm called \emph{semi-proximal ADMM} (also known as {\admmplus}) \cite{Sun15siopt-admmplus}. We also give more implementation details about how to use Riemannian optimization to perform local search.
\section{Experiments}
\label{sec:experiments}
In this section, we test the sparse relaxation \eqref{eq:sparserelax} and the SDP solver {\stride} on Examples \ref{ex:singlerotation}-\ref{ex:category} using both synthetic and real data \revise{(we defer the results for Example \ref{ex:mesh} mesh registration to {\supp} due to space constraints)}. The goal of our experiments is not to claim state-of-the-art efficiency or robustness (\eg against problem-specific implementations), but rather to show that \eqref{eq:sparserelax} and {\stride}, for the first time, provide a general framework to solve large-scale nonconvex outlier-robust perception problems to certifiable global optimality within reasonable computation time. We believe with the advancement of SDP solvers, our framework will eventually run in real time.


{\bf Baselines}. We use two state-of-the-art SDP solvers, {\mosek} \cite{mosek} and {\sdpnal} \cite{Yang2015mpc-sdpnalplus}, as baseline solvers to compare against {\stride}. We omit {\mosek} whenever the SDP becomes too large to be solved by {\mosek} (\ie when $m > 50,000$). We use default settings for both {\mosek} and {\sdpnal}.

{\bf {\stride}'s settings}. In Algorithm \ref{alg-iPGMnlp}, we choose $\tol\!=\!1\ee{-6}$, $r\!=\!3$, $\epsilon\!=\!1\ee{-12}$, $\sigma_k\!=\!10,\forall k$, and run it for a maximum of $5$ iterations. As described in Remark \ref{remark:fastheuristics}, we use {\gnc} or {\ransac} to initialize the primal variable, and {\admmplus} to initialize the dual variable. The local search is performed using {\manopt} with a trust region solver. Details about local search and  {\admmplus} can be found in \supp.

{\bf Evaluation metrics}. Let $(\hatMX,\hatvy,\hatMS)$ be the solution for \eqref{eq:sparserelax} returned by an SDP solver 
and $(\hatvxx,\hatvtheta)$ be the corresponding rounded solution for \eqref{eq:binaryTLS}.
 We evaluate the performance of the solver using four metrics: 
(i)~the estimation errors  of $\hatvxx$ compared to the groundtruth,  
whenever the groundtruth is available;
(ii)~SDP solution quality, using the maximum KKT residual $\kkt$ from \eqref{eq:KKTresiduals}; 
(iii)~certified suboptimality, using the {\rounding} procedure in \eqref{eq:rounding} and the relative suboptimality measure $\subopt$ in \eqref{eq:subopt} (we deem a rounded solution as globally optimal if $\subopt < 1\ee{-3}$); 
(iv) solver CPU time in seconds. {For simulation experiments, statistics are computed over $20$ Monte Carlo runs per setup.}

{\bf Hardware}. Experiments are performed on a Linux PC with 12-core Intel i9-7920X CPU@2.90GHz and 128GB RAM.


\subsection{Single Rotation Averaging}


\newcommand{\mpwfour}{5.0cm}
\newcommand{\myhspace}{\hspace{-4mm}}
\begin{figure*}[t]
	\begin{center}
	\begin{minipage}{\textwidth}
	\begin{tabular}{cccc}%
		   \myhspace
			\begin{minipage}{\mpwfour}%
			\centering%
			\includegraphics[width=\columnwidth]{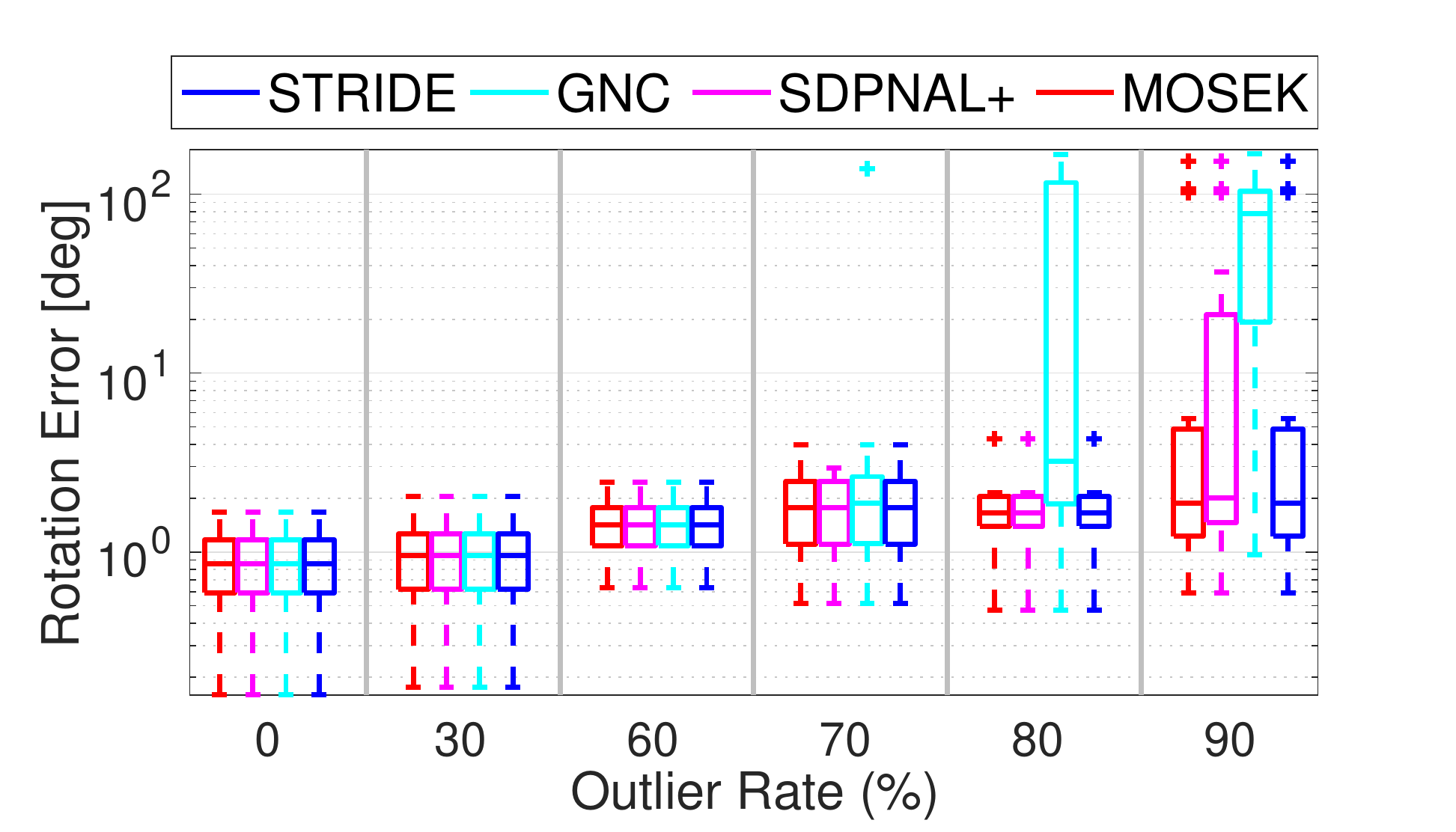}
			\end{minipage}
		&  \myhspace \hspace{-4mm}
			\begin{minipage}{\mpwfour}%
			\centering%
			\includegraphics[width=\columnwidth]{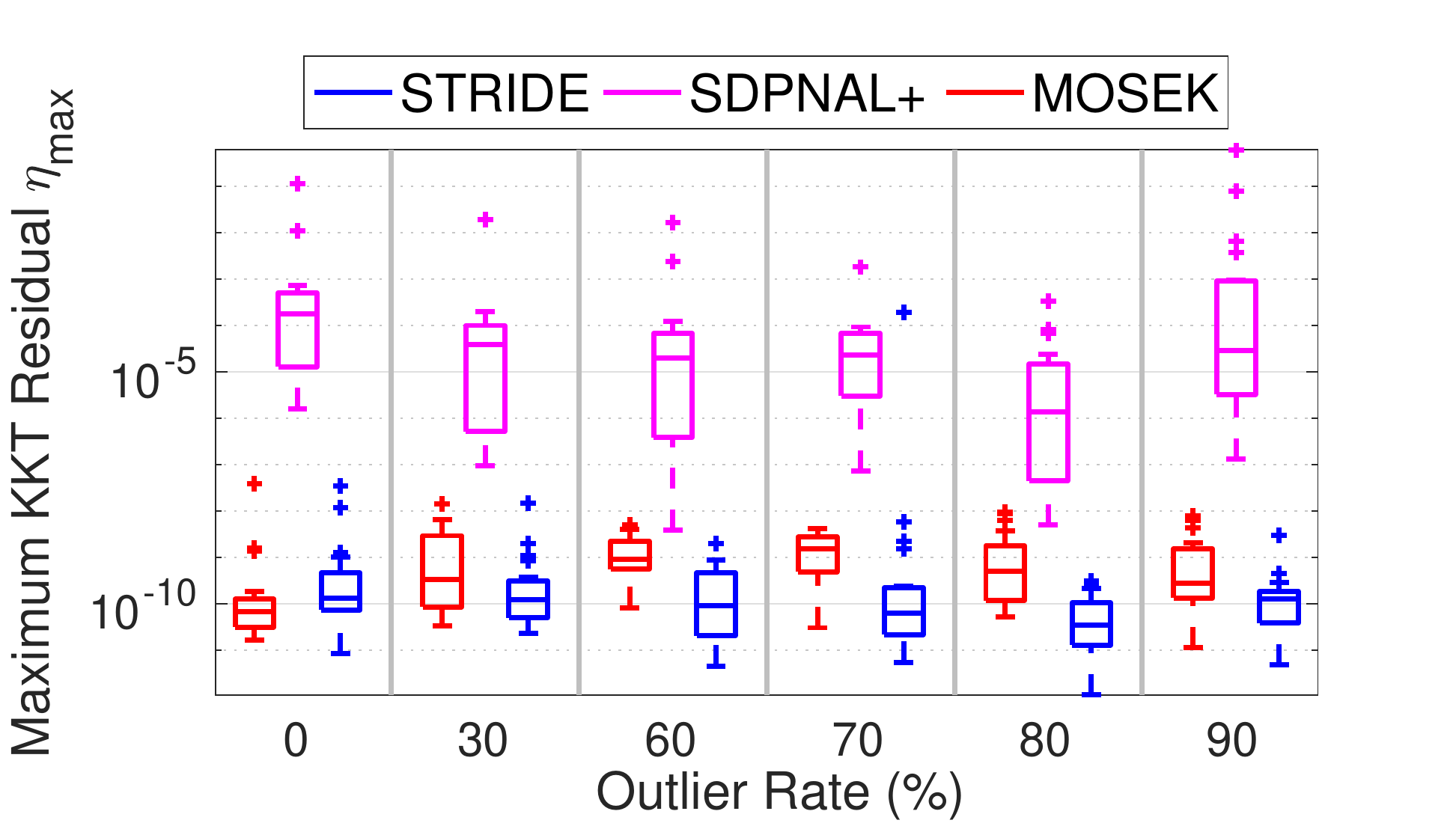}
			\end{minipage}
		&  \myhspace \hspace{-5mm}
			\begin{minipage}{\mpwfour}%
			\centering%
			\includegraphics[width=\columnwidth]{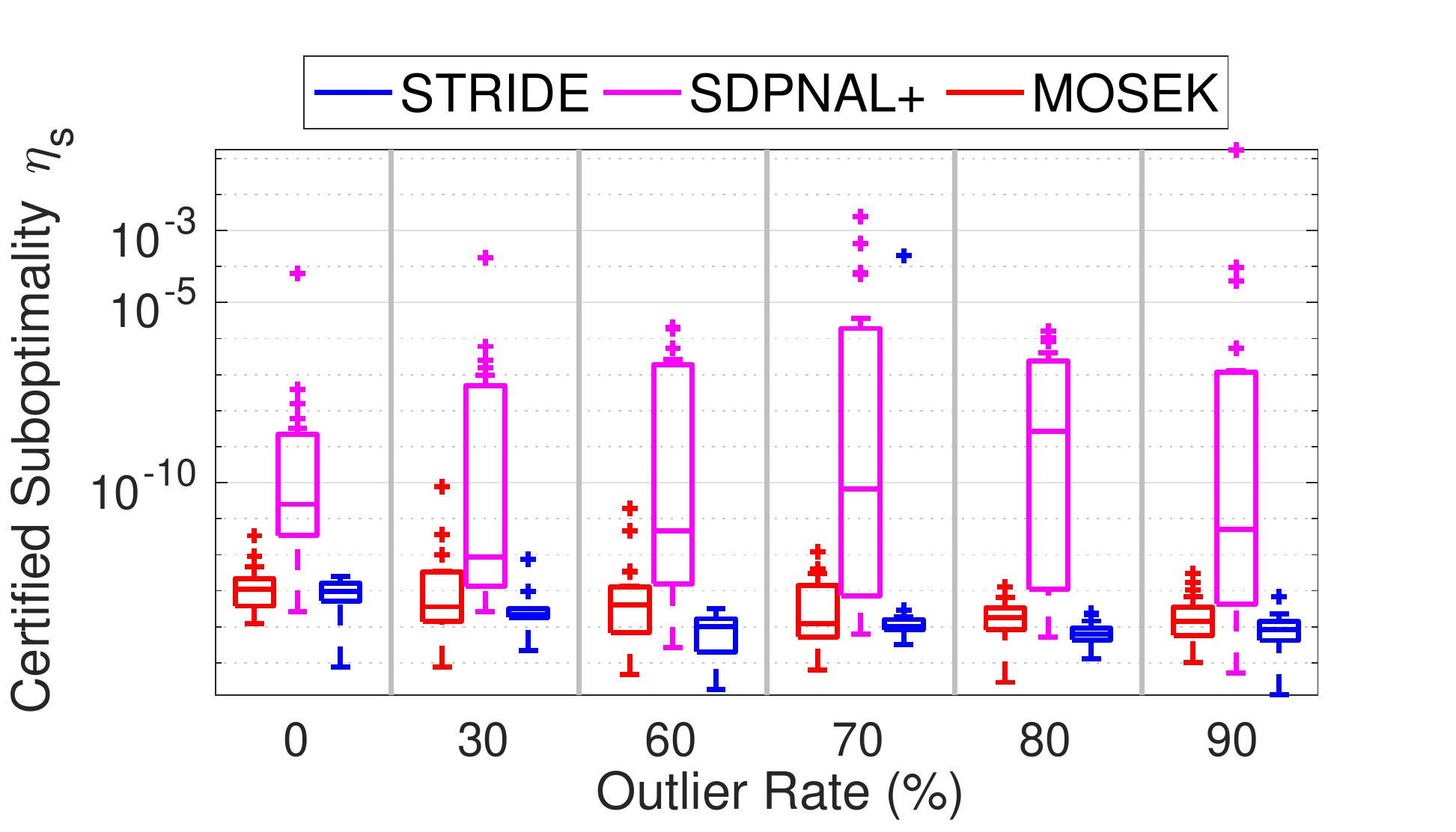}
			\end{minipage}
		&  \myhspace \hspace{-5mm}
			\begin{minipage}{\mpwfour}%
			\centering%
			\includegraphics[width=\columnwidth]{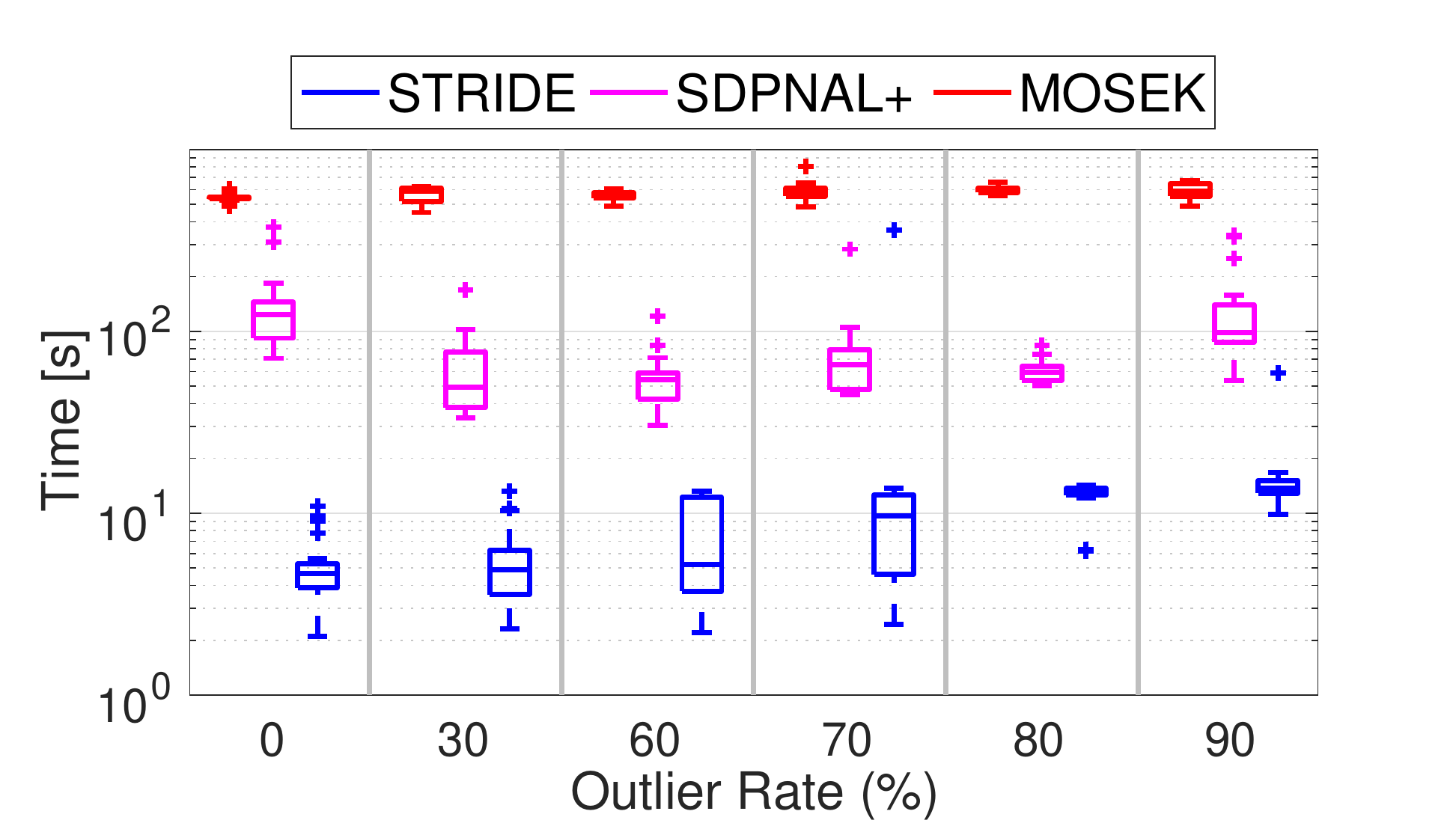}
			\end{minipage} \\
		 \multicolumn{4}{c}{\subcapsize (a) $N=30$, $n_1=310$, $m=30,016$}  \\
		   \myhspace
			\begin{minipage}{\mpwfour}%
			\centering%
			\includegraphics[width=\columnwidth]{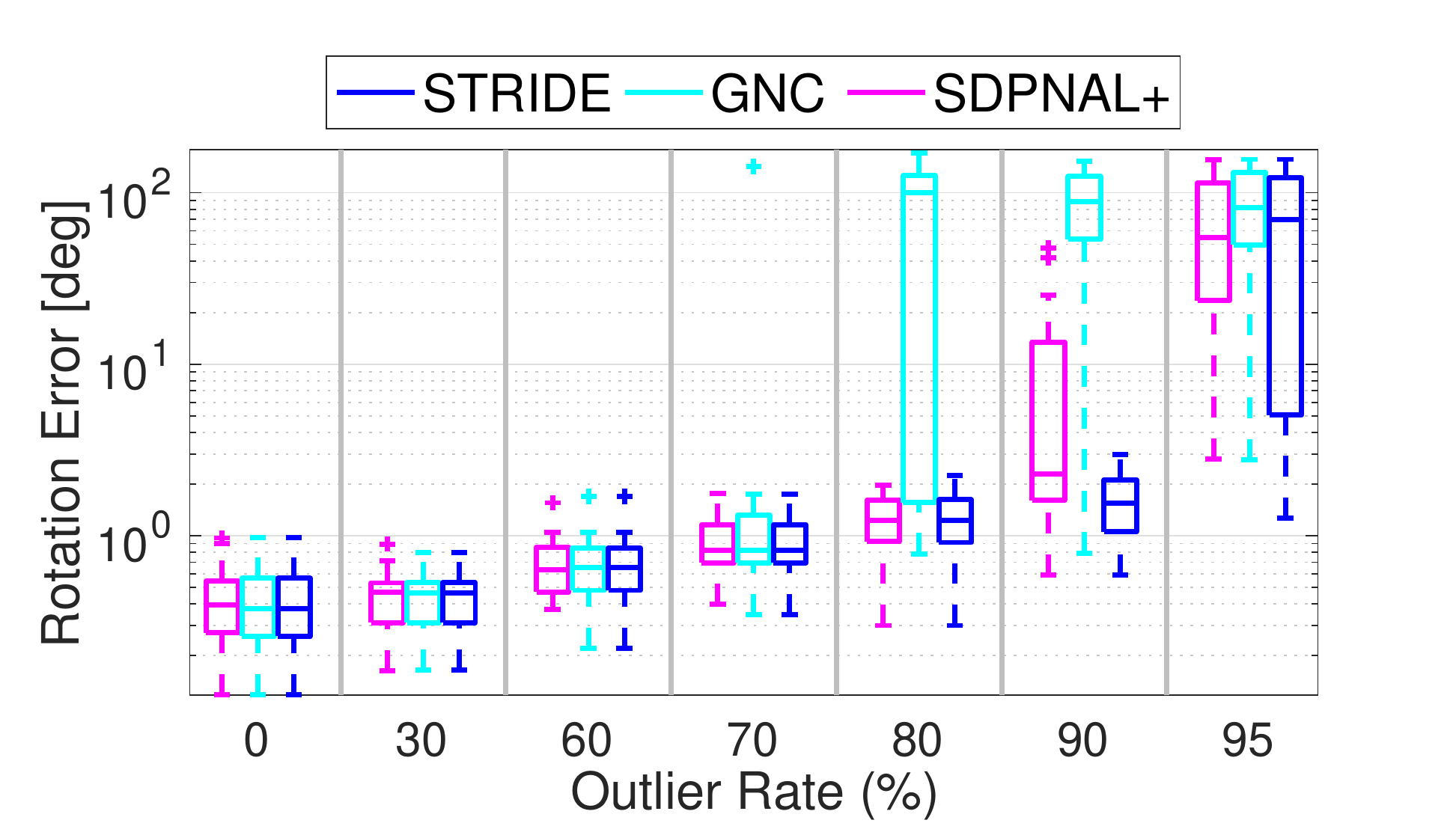}
			\end{minipage}
		&  \myhspace \hspace{-4mm}
			\begin{minipage}{\mpwfour}%
			\centering%
			\includegraphics[width=\columnwidth]{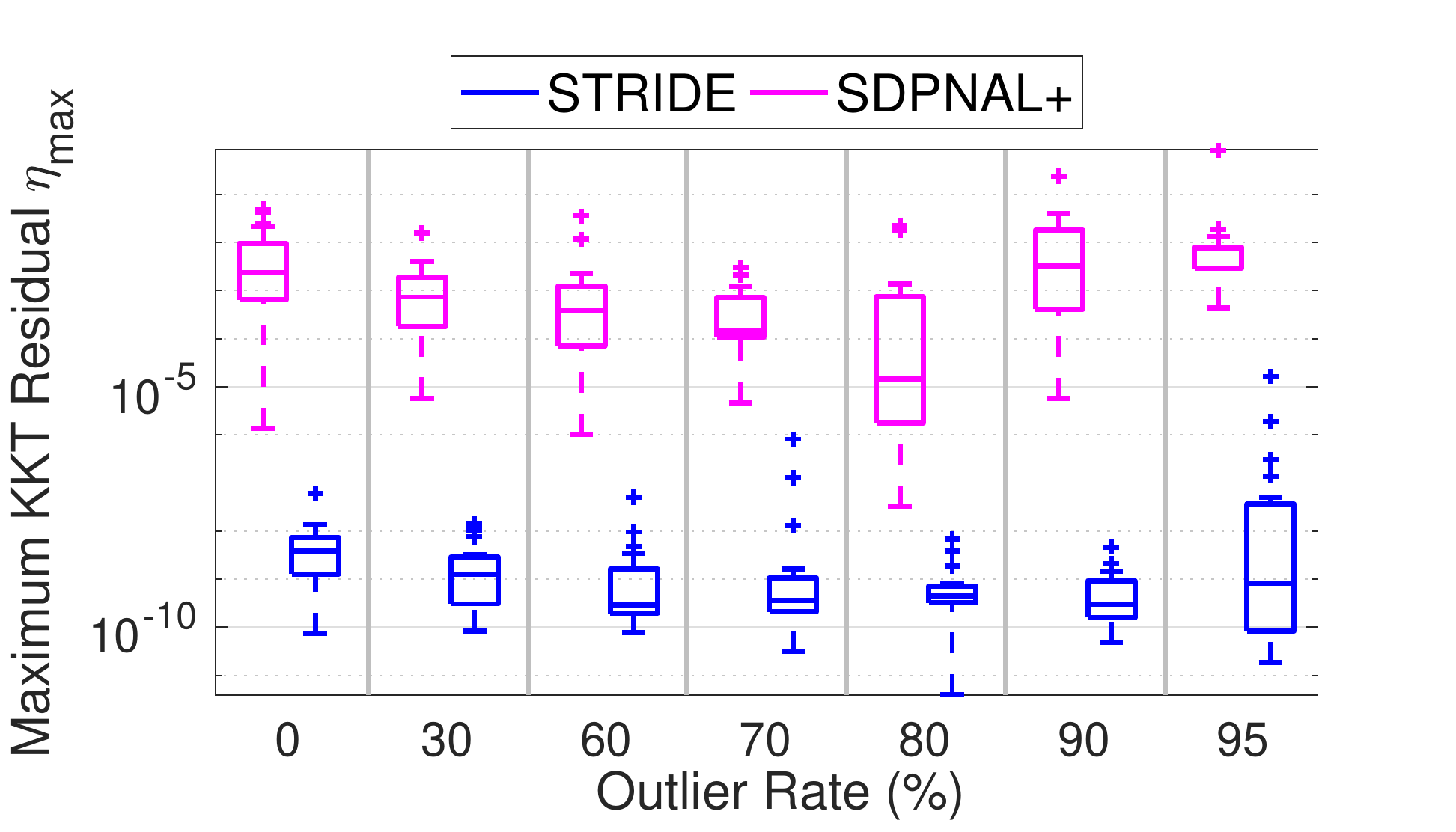}
			\end{minipage}
		&  \myhspace \hspace{-5mm}
			\begin{minipage}{\mpwfour}%
			\centering%
			\includegraphics[width=\columnwidth]{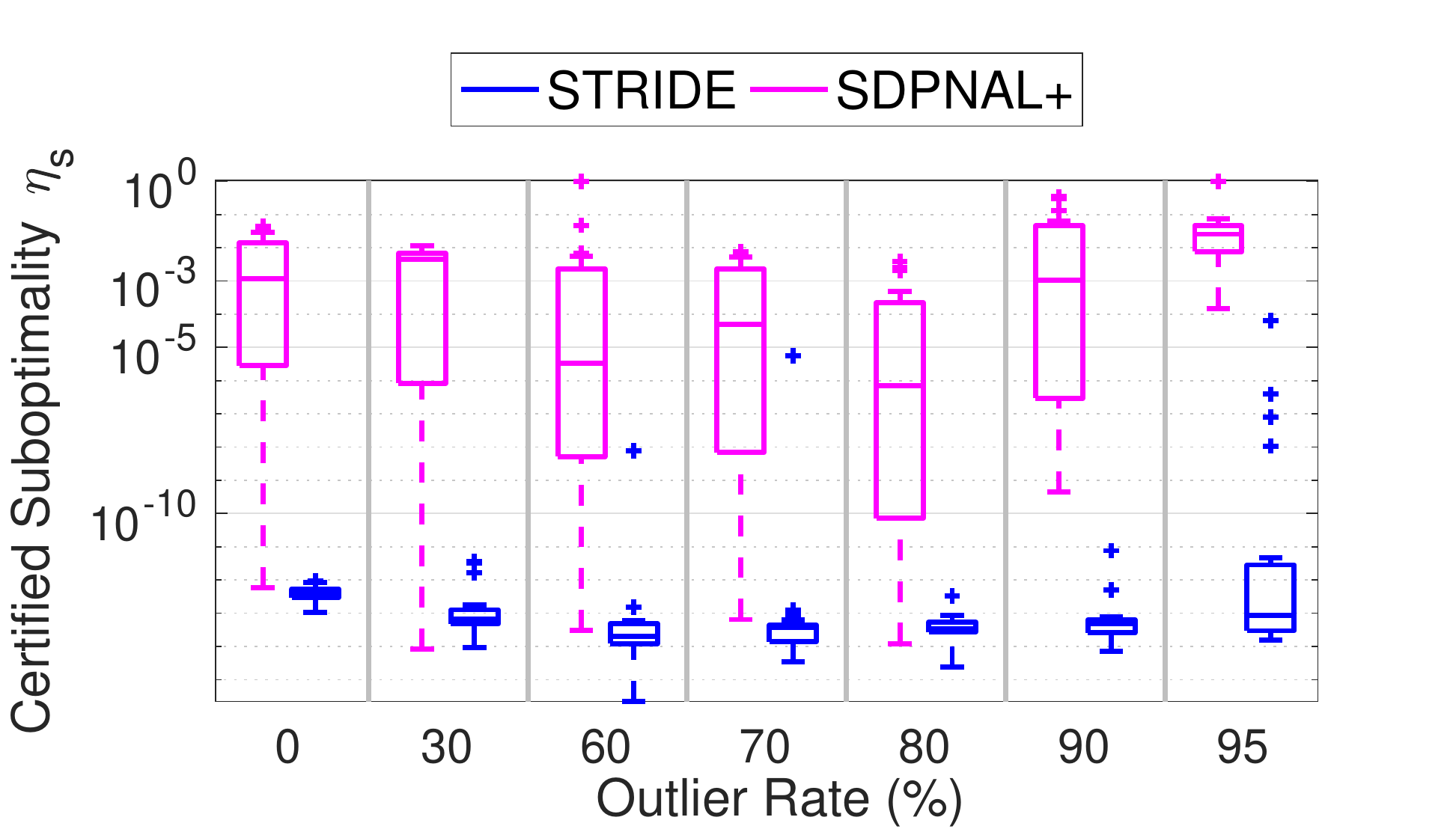}
			\end{minipage}
		&  \myhspace \hspace{-5mm}
			\begin{minipage}{\mpwfour}%
			\centering%
			\includegraphics[width=\columnwidth]{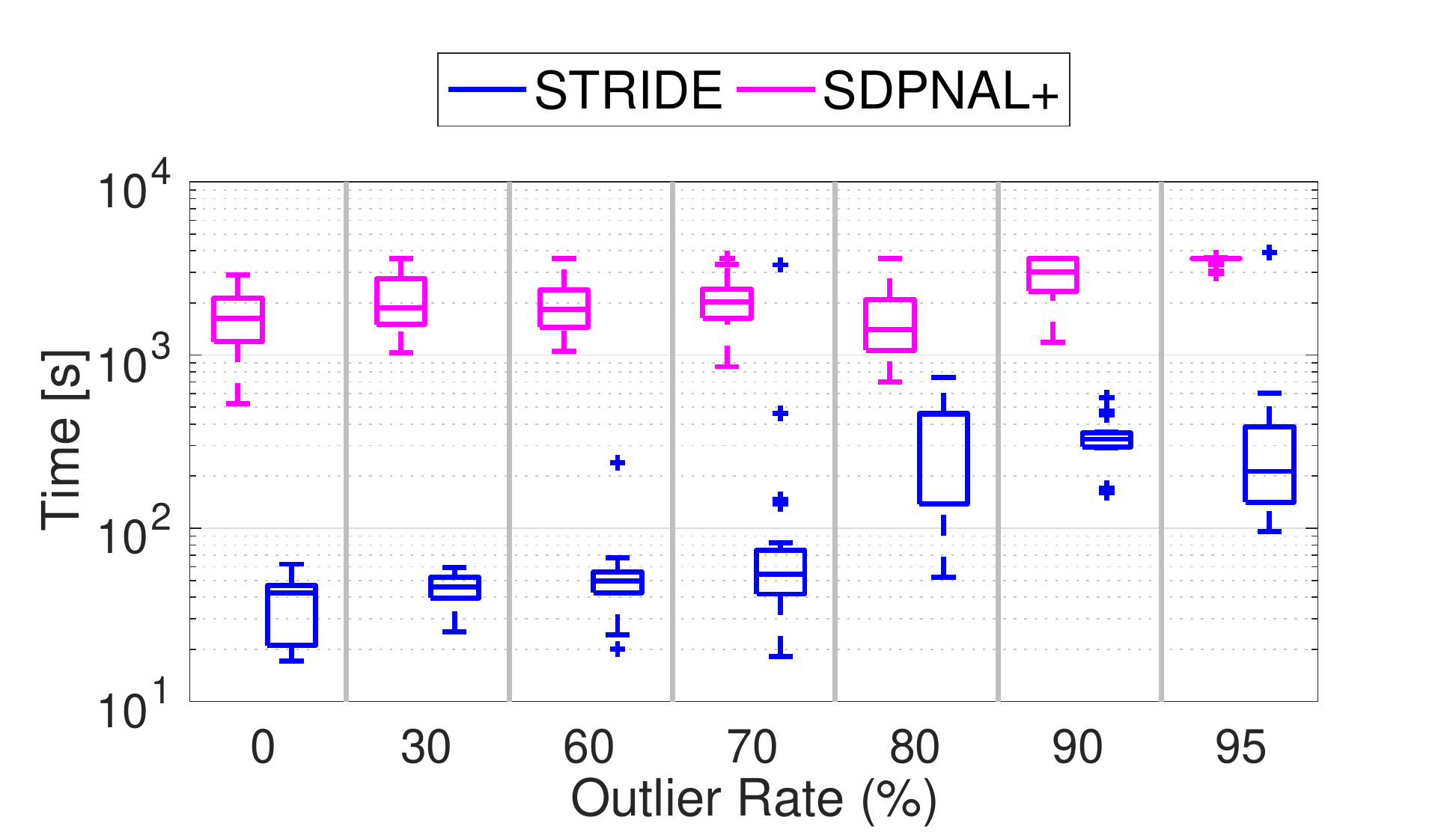}
			\end{minipage} \\
		\multicolumn{4}{c}{\subcapsize (b) $N=100$, $n_1=1010$, $m=310,016$ \vspace{-2mm}}
	\end{tabular}
	\end{minipage} 
	\caption{Single Rotation Averaging (Example \ref{ex:singlerotation}).
	\label{fig:exp-sra-results}} 
	\vspace{-6mm} 
	\end{center}
\end{figure*}

{\bf Setup}. At each Monte Carlo run, we first randomly generate a groundtruth 3D rotation $\MRgt$; then inliers are generated by $\MRin = \MRgt \MRnoise$,
where the inlier noise $\MRnoise$ is generated by randomly sampling a rotation axis and a rotation angle $\varepsilon \sim \calN(0,\sigma^2)$ with $\sigma = 5^{\circ}$; outliers are arbitrary random rotations. We test two setups with $N=30$ and $N=100$. At $N=30$, we sweep the outlier ratio from $0\%$ to $90\%$, while at $N=100$, we sweep the outlier ratio up to $95\%$.

{\bf Results}. Fig. \ref{fig:exp-sra-results}(a)-(b) plot the evaluation metrics for $N=30$ and $N=100$, respectively. We make the following observations. (i) Our sparse relaxation \eqref{eq:sparserelax} is exact with up to $90\%$ outliers when $N=30$ and up to $95\%$ outliers when $N=100$ (the suboptimality $\subopt$ is below $1\ee{-3}$ for all test runs). (ii) For $N=30$, {\stride} solves the SDP to an accuracy that is comparable to {\mosek} (\cf the $\kkt$ plot), but is about $100$ (and up to $270$) times faster (\cf the time plot). (iii) For $N=100$, {\mosek} cannot run. While {\sdpnal} still runs, its accuracy is at least five orders of magnitude worse than {\stride} (\cf the $\kkt$ plot, where {\stride} attains $1\ee{-8}$ accuracy, but {\sdpnal} only attains $1\ee{-3}$ accuracy), and its runtime is about $40$ times slower than {\stride}. (iv) {\stride} safeguards {\gnc}. While {\gnc} is used to initialize {\stride}, {\stride} can \emph{certify} the global optimality of {\gnc} and escapes the local minima of {\gnc} (\eg at $80\%$ outlier rate in the rotation error plot, while {\gnc} fails many times, the solution of {\stride} is always correct and optimal). (v) When the outlier rate is too high, global optimality not necessarily implies a correct estimation (in the sense of being close to the groundtruth). For example, at $90\%$ outlier rate when $N=30$, {\stride} and {\mosek} both obtain certifiable optimality ($\subopt = 0$), but the rotation error can be quite large (about $100^{\circ}$). Similarly, at $95\%$ outlier rate when $N=100$, the optimal estimate obtained by {\stride} also has large rotation errors. 
For further discussion about this point, 
 we refer the reader to the notion of \emph{estimation contract} in~\cite{Yang20tro-teaser}, which 
 ties the number of inliers to the accuracy of the optimal solution of \eqref{eq:binaryTLS} \wrt the groundtruth, and 
 reports estimation contracts for a 3D registration problem. 


\subsection{Multiple Rotation Averaging}

\begin{figure*}[t]
	\begin{center}
	\begin{minipage}{\textwidth}
	\begin{tabular}{cccc}%
		   \myhspace
			\begin{minipage}{\mpwfour}%
			\centering%
			\includegraphics[width=\columnwidth]{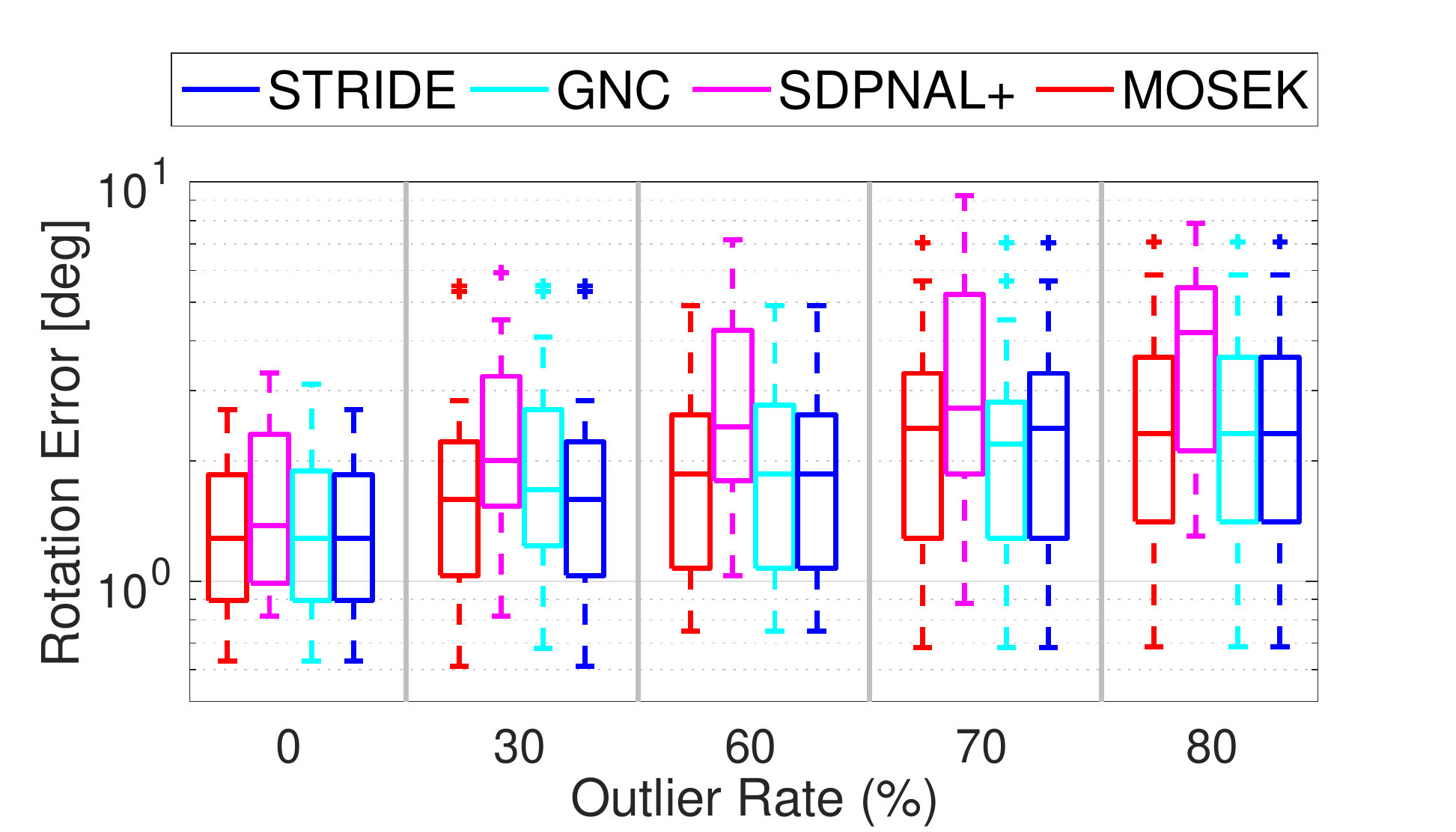}
			\end{minipage}
		&  \myhspace \hspace{-4mm}
			\begin{minipage}{\mpwfour}%
			\centering%
			\includegraphics[width=\columnwidth]{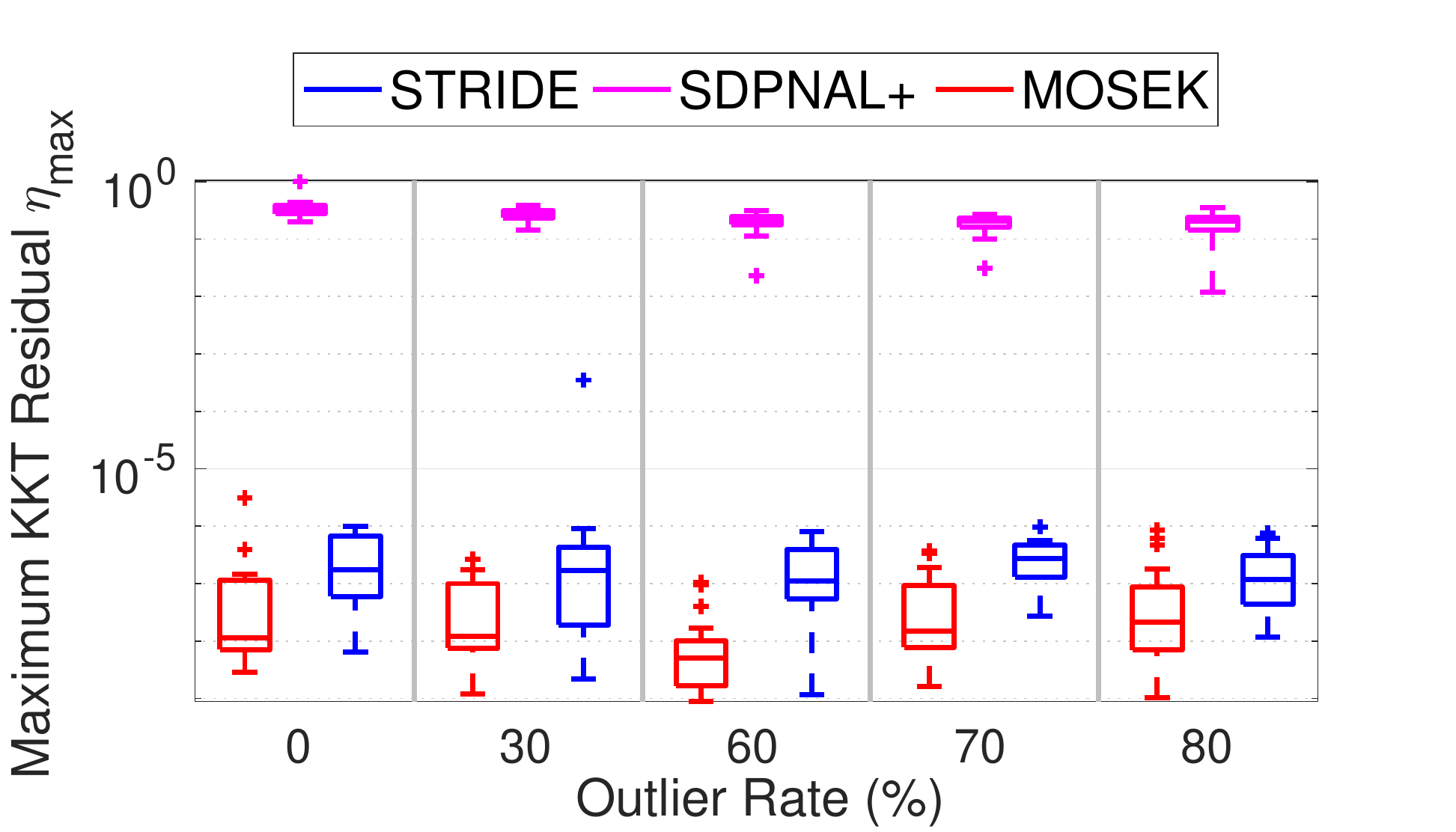}
			\end{minipage}
		&  \myhspace \hspace{-4mm}
			\begin{minipage}{\mpwfour}%
			\centering%
			\includegraphics[width=\columnwidth]{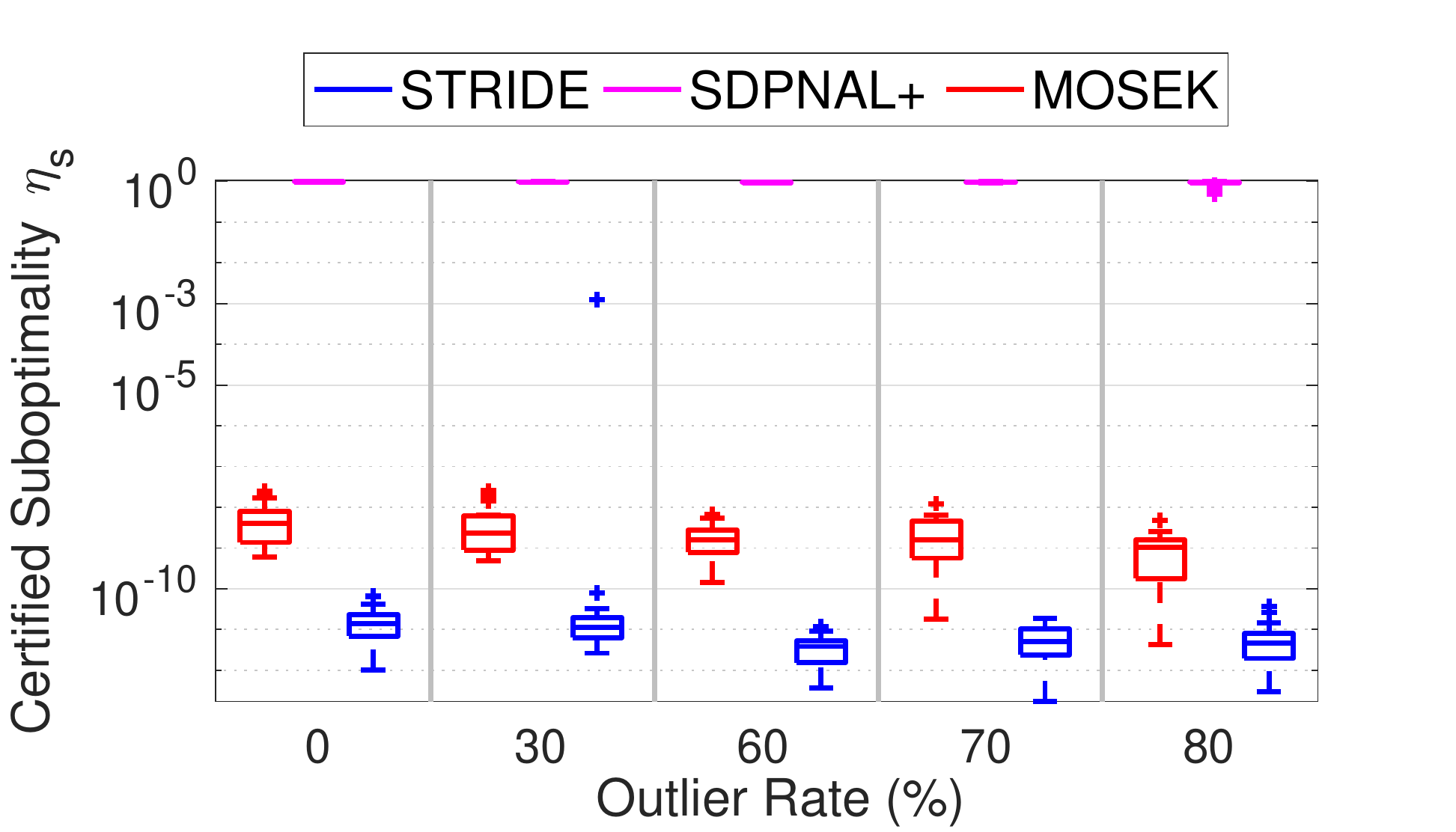}
			\end{minipage}
		&  \myhspace \hspace{-4mm}
			\begin{minipage}{\mpwfour}%
			\centering%
			\includegraphics[width=\columnwidth]{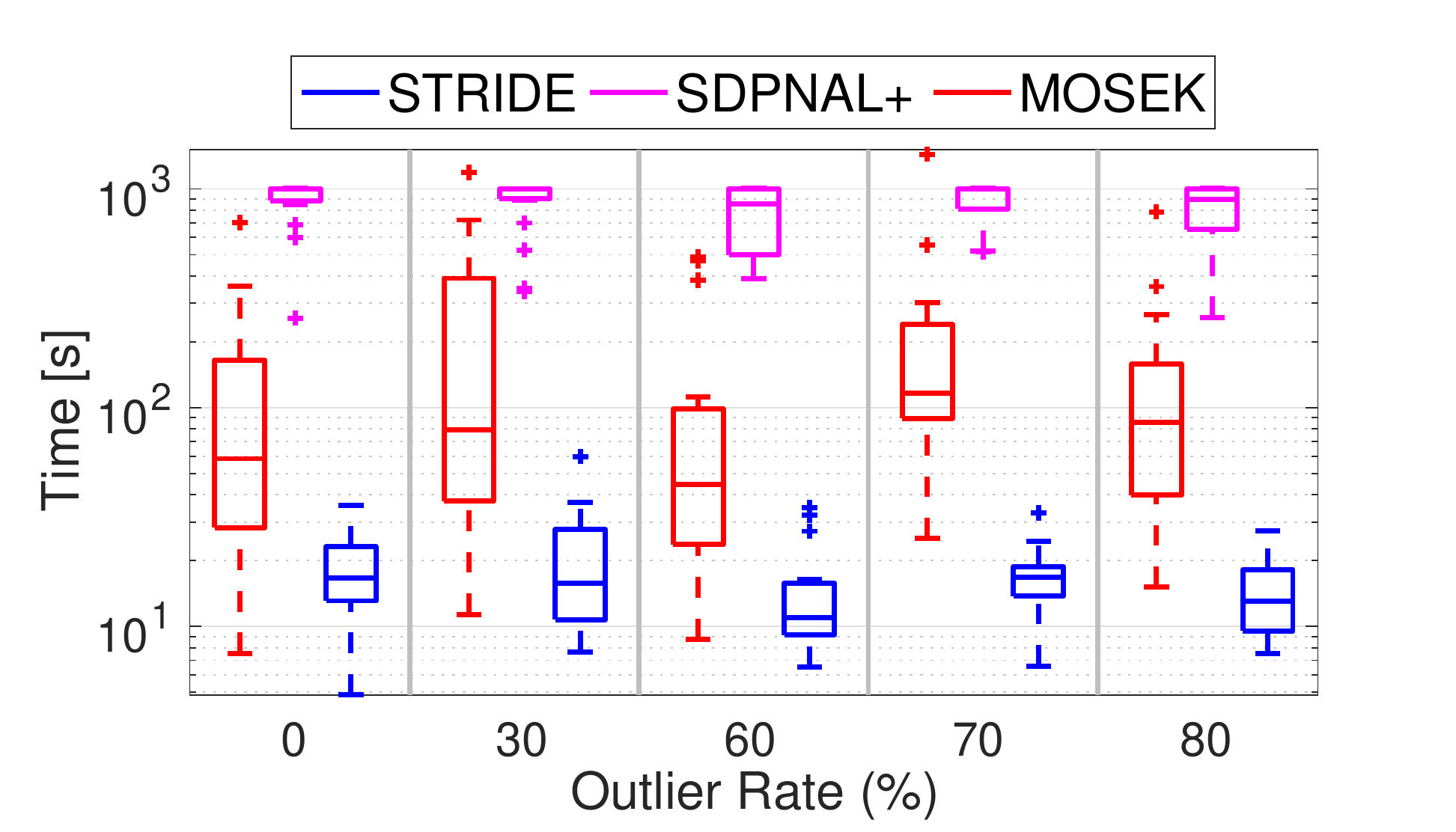}
			\end{minipage} \\
		\multicolumn{4}{c}{\subcapsize (a) $10\times 10$ 2D grid, $N=10$ loop closures, $343 \leq n_1 \leq 549$, $3452 \leq m \leq 35,246$}
		\\
		   \myhspace
			\begin{minipage}{\mpwfour}%
			\centering%
			\includegraphics[width=\columnwidth]{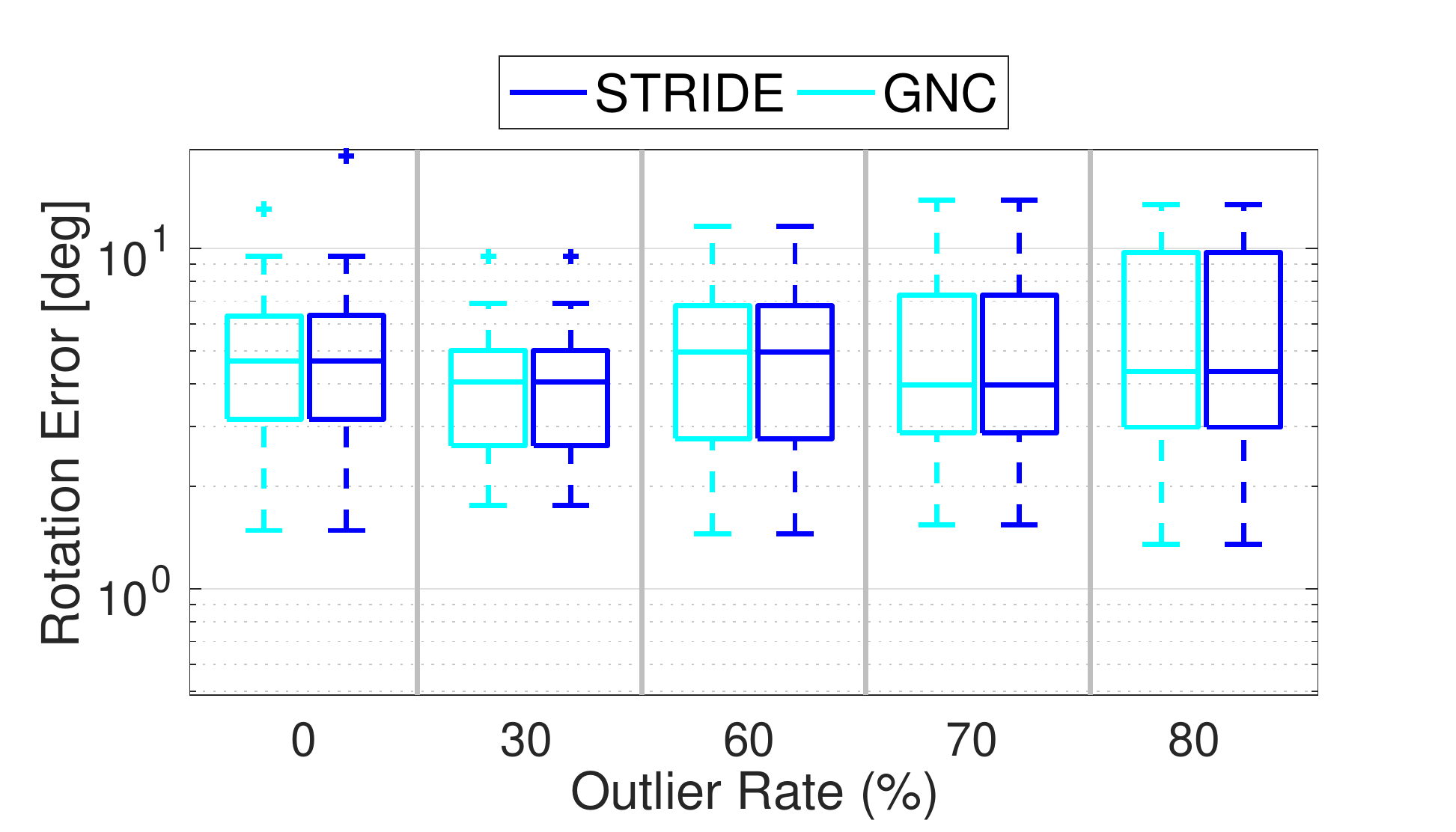}
			\end{minipage}
		&  \myhspace \hspace{-4mm}
			\begin{minipage}{\mpwfour}%
			\centering%
			\includegraphics[width=\columnwidth]{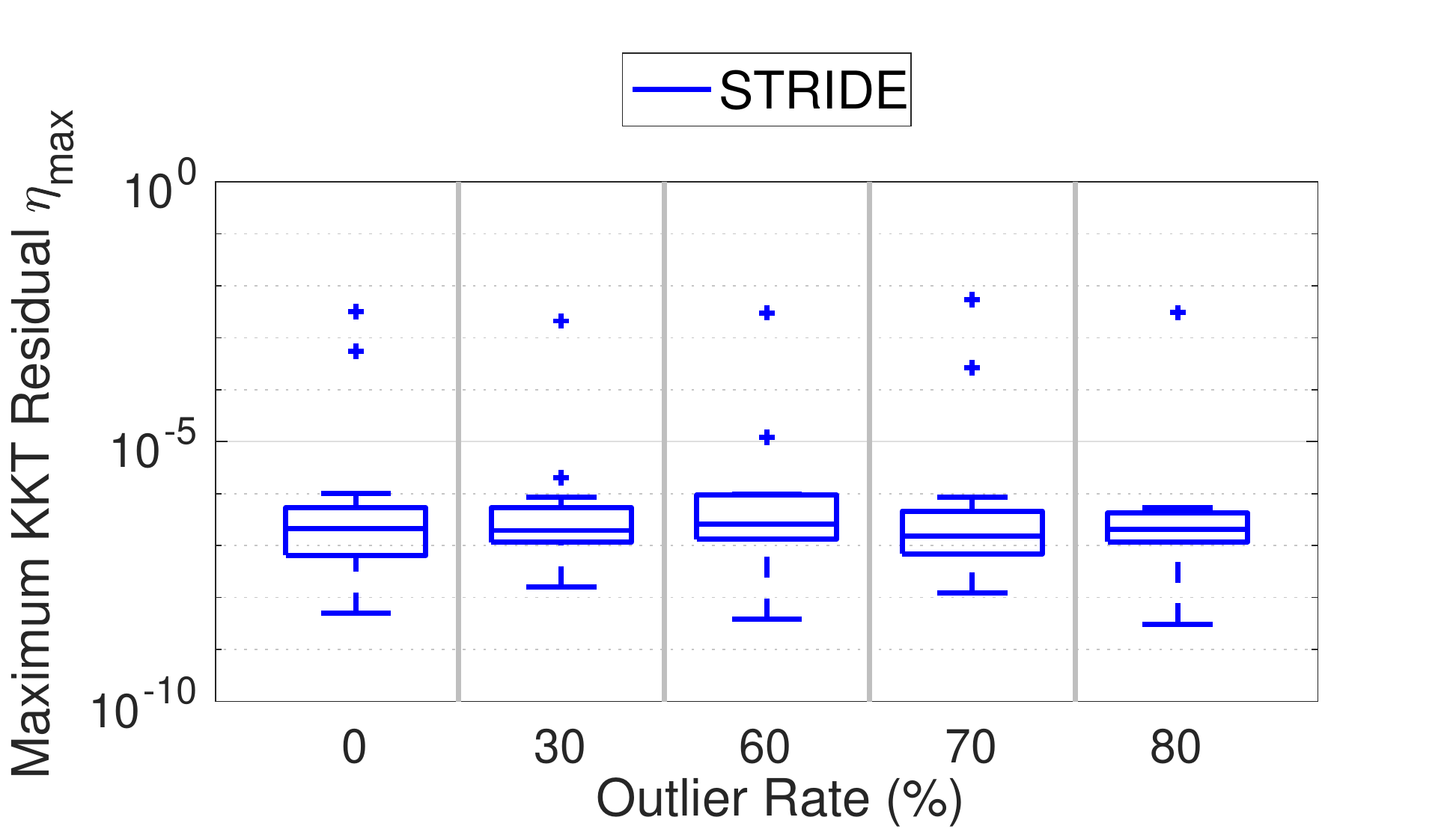}
			\end{minipage}
		&  \myhspace \hspace{-4mm}
			\begin{minipage}{\mpwfour}%
			\centering%
			\includegraphics[width=\columnwidth]{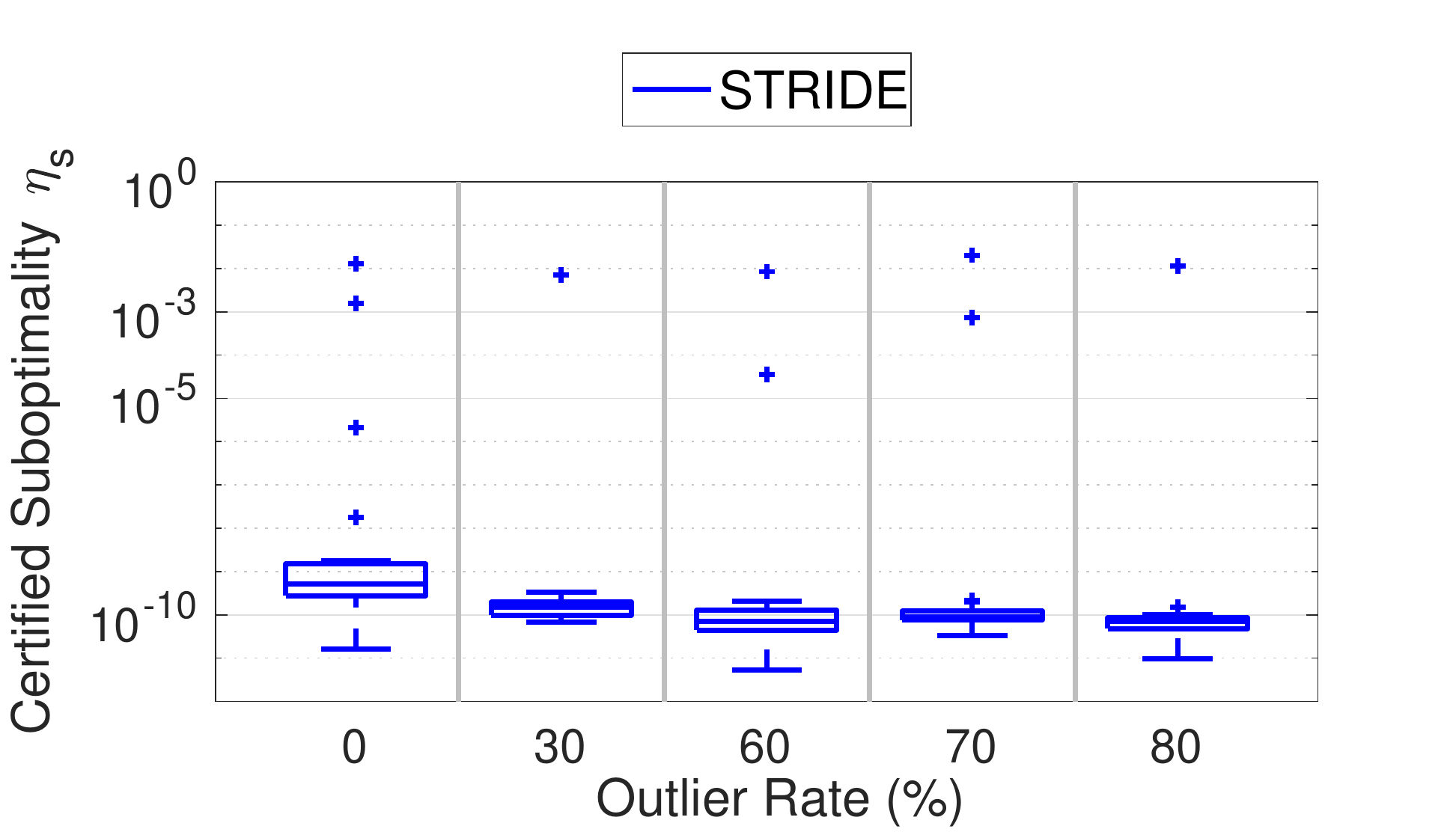}
			\end{minipage}
		&  \myhspace \hspace{-4mm}
			\begin{minipage}{\mpwfour}%
			\centering%
			\includegraphics[width=\columnwidth]{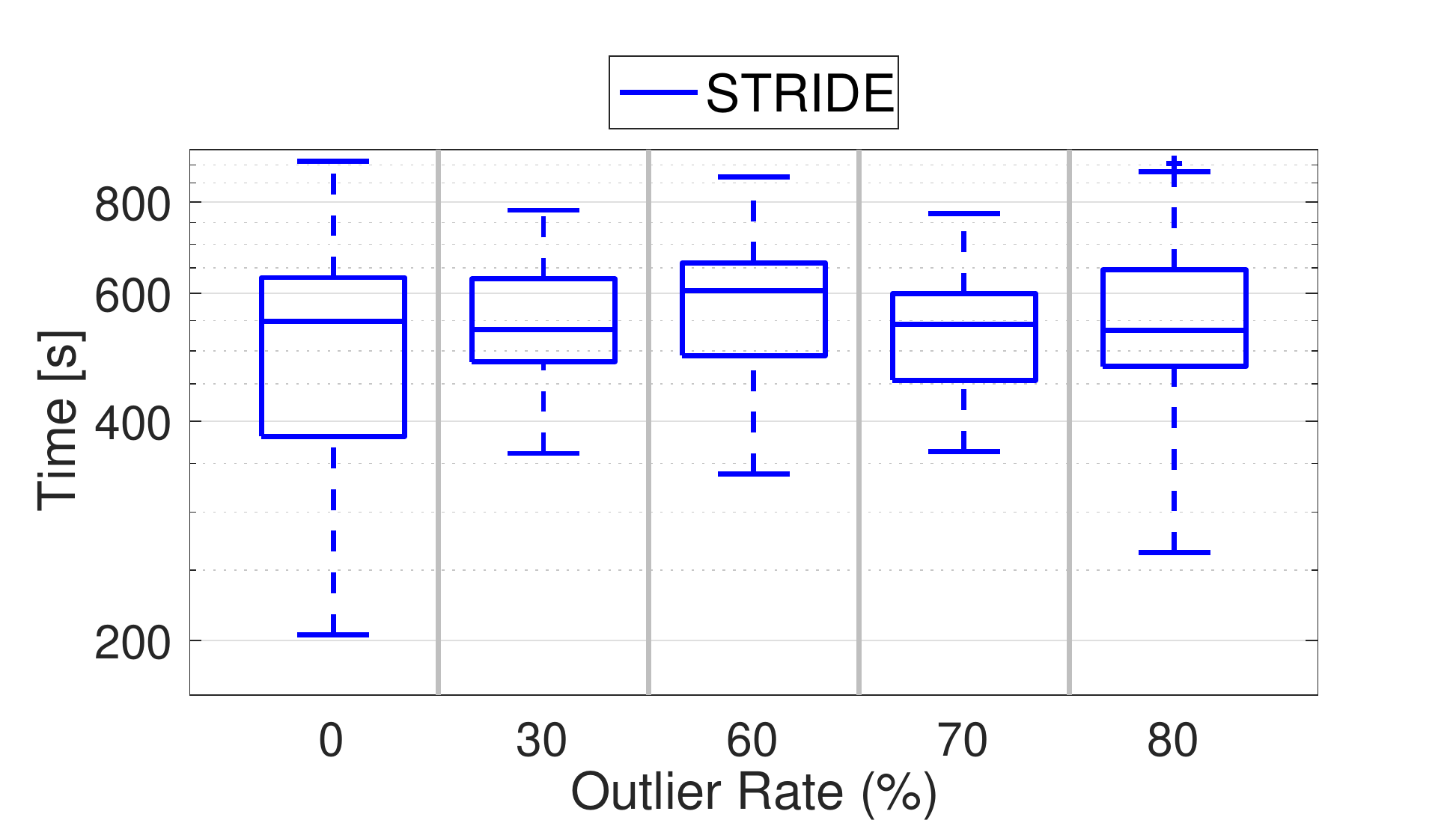}
			\end{minipage} \\
		\multicolumn{4}{c}{\subcapsize (b) $20\times 20$ 2D grid, $N=20$ loop closures, $1483 \leq n_1 \leq 1761$, $64,384 \leq m \leq 189,198$ \vspace{-2mm}}
	\end{tabular}
	\end{minipage} 
	\caption{2D Multiple Rotation Averaging (Example \ref{ex:multirotation}).
	\label{fig:exp-mra-results}} 
	\vspace{-6mm} 
	\end{center}
\end{figure*}

{\bf Setup}. We test 2D multiple rotation averaging in a SLAM setting, where a robot traverses a trajectory following a 2D grid pattern (\eg Fig. \ref{fig:applications}(b) shows a $3\times 3$ grid) with both odometry measurements (between consecutive nodes) and loop closures. We assume the odometry measurements are outlier-free (\ie we include them in the function $\regularizer$ in Example \ref{ex:multirotation}) and only the loop closures could be corrupted by outliers as in~\cite{Yang20ral-gnc}. Inlier relative rotations are generated by $\MRin = \MRgt \MRnoise$ where $\MRgt = \MR_i\tran \MR_j$ is the groundtruth relative rotation between nodes $(i,j)$ and $\MRnoise$ is a random 2D rotation with angle $\varepsilon \sim \calN(0,\sigma^2)$ ($\sigma=0.6^{\circ}$). Outlier relative rotations are arbitrary 2D rotations. We test two cases with increasing outlier rates: a $10\times 10$ grid with $N=10$ loop closures, and a $20 \times 20$ grid with $N=20$ loop closures.

{\bf Results}. Fig. \ref{fig:exp-mra-results}(a)-(b) plot the evaluation metrics for both cases. We make the following observations. (i) For $10 \times 10$ grid with $N=10$, our relaxation is always exact, with up to $80\%$ outliers. In this case, {\stride} can solve the SDP to an accuracy that is comparable to {\mosek}, while being about $20$ times faster (up to $40$ times faster). {\sdpnal}, unfortunately, completely fails in this problem. 
Therefore, we did not run {\sdpnal} for the more challenging $20 \times 20$ grid. (ii) For $20\times 20$ grid with $N=20$, our relaxation is also almost always exact, with up to $80\%$ outliers. However, there exist 1-2 runs per outlier rate where {\stride} fails to obtain an $\subopt < 1\ee{-3}$. In such cases, we suspect the relaxation is inexact. 


\subsection{Point Cloud Registration}


\newcommand{\mpwtwo}{9.0cm}
\begin{figure*}[t]
	\begin{center}
	\begin{minipage}{\textwidth}
	\begin{tabular}{cccc}%
		   \myhspace
			\begin{minipage}{\mpwfour}%
			\centering%
			\includegraphics[width=\columnwidth]{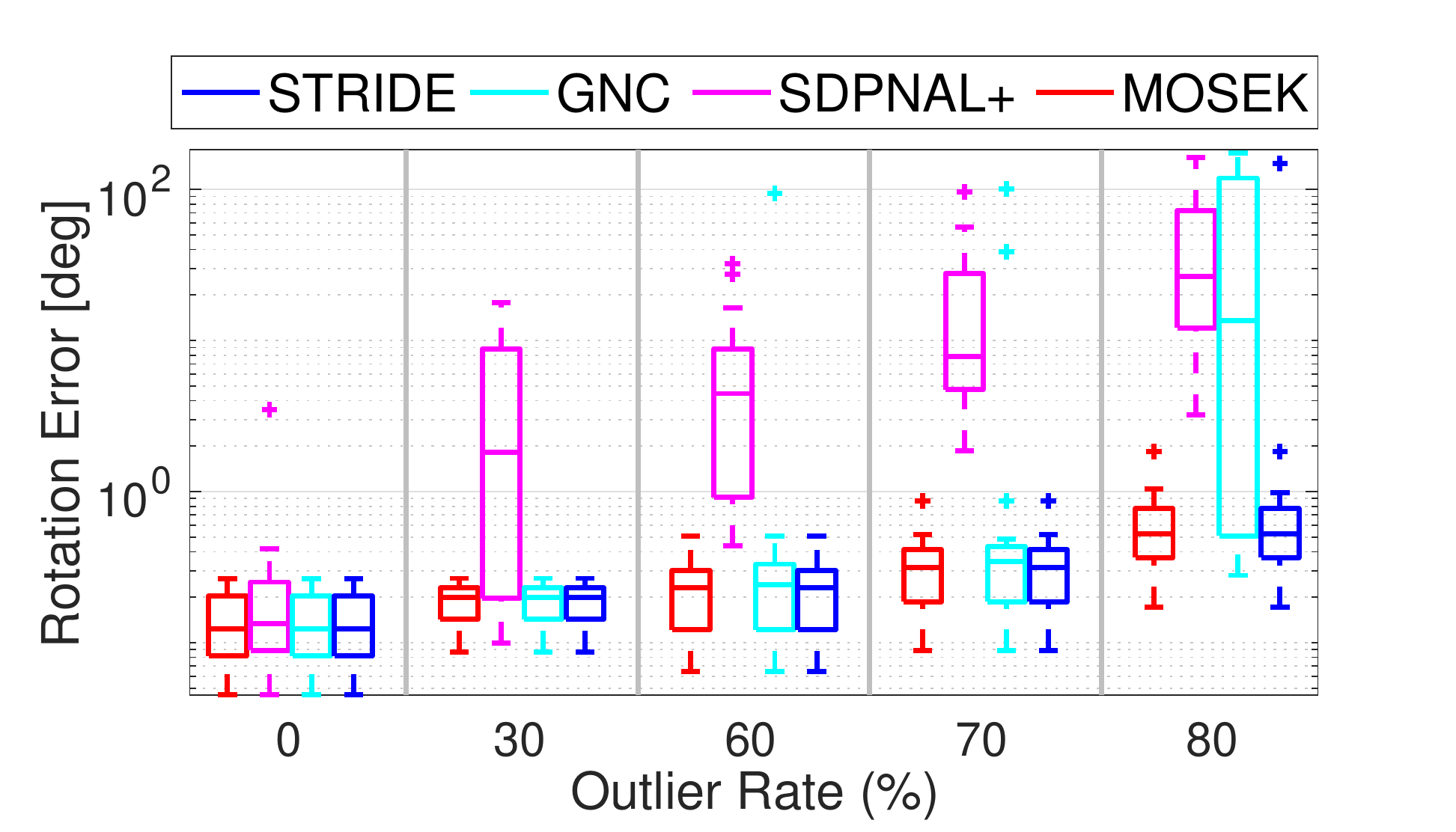}
			\end{minipage}
		&  \myhspace \hspace{-4mm}
			\begin{minipage}{\mpwfour}%
			\centering%
			\includegraphics[width=\columnwidth]{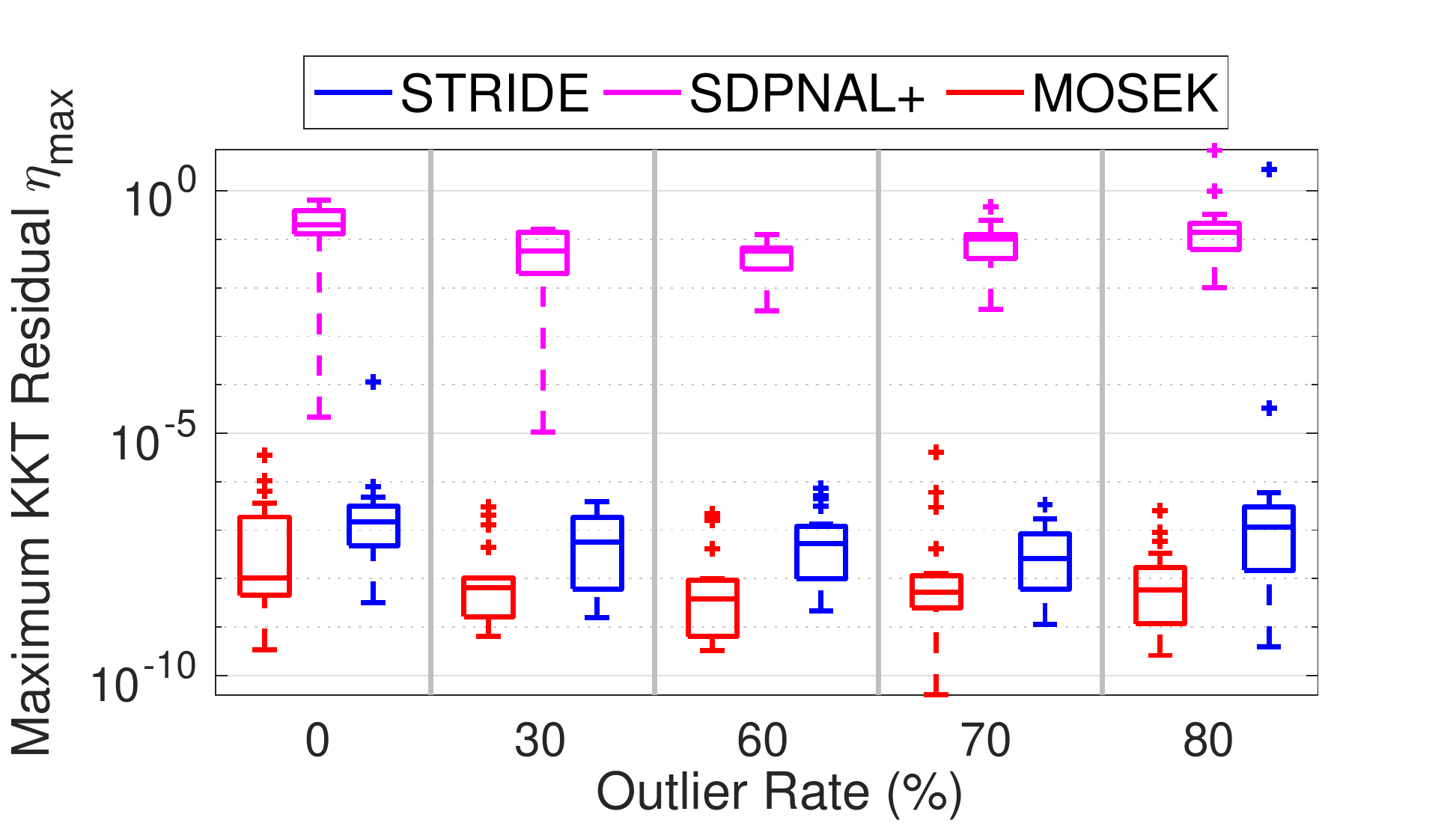}
			\end{minipage}
		&  \myhspace \hspace{-5mm}
			\begin{minipage}{\mpwfour}%
			\centering%
			\includegraphics[width=\columnwidth]{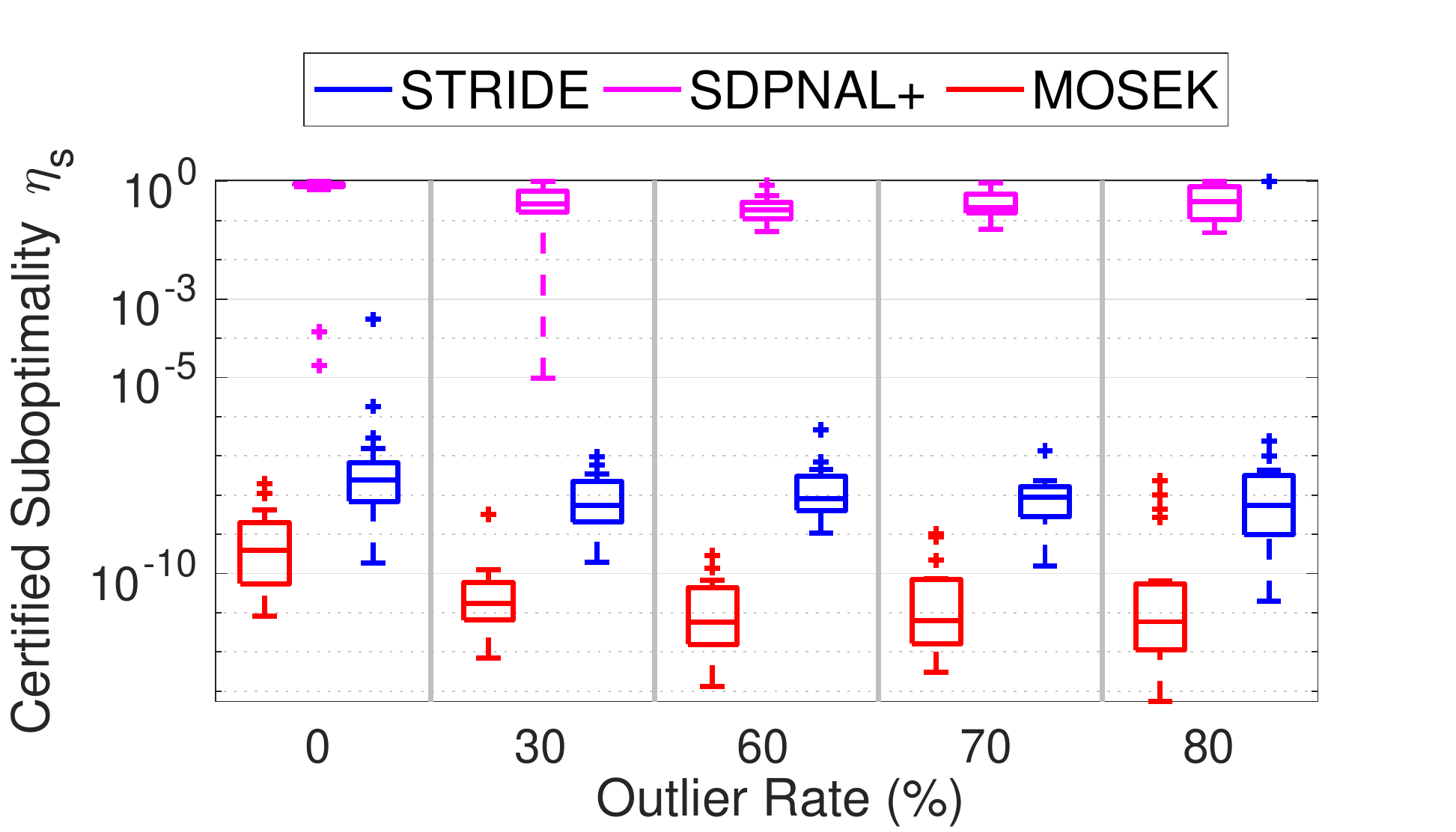}
			\end{minipage}
		&  \myhspace \hspace{-5mm}
			\begin{minipage}{\mpwfour}%
			\centering%
			\includegraphics[width=\columnwidth]{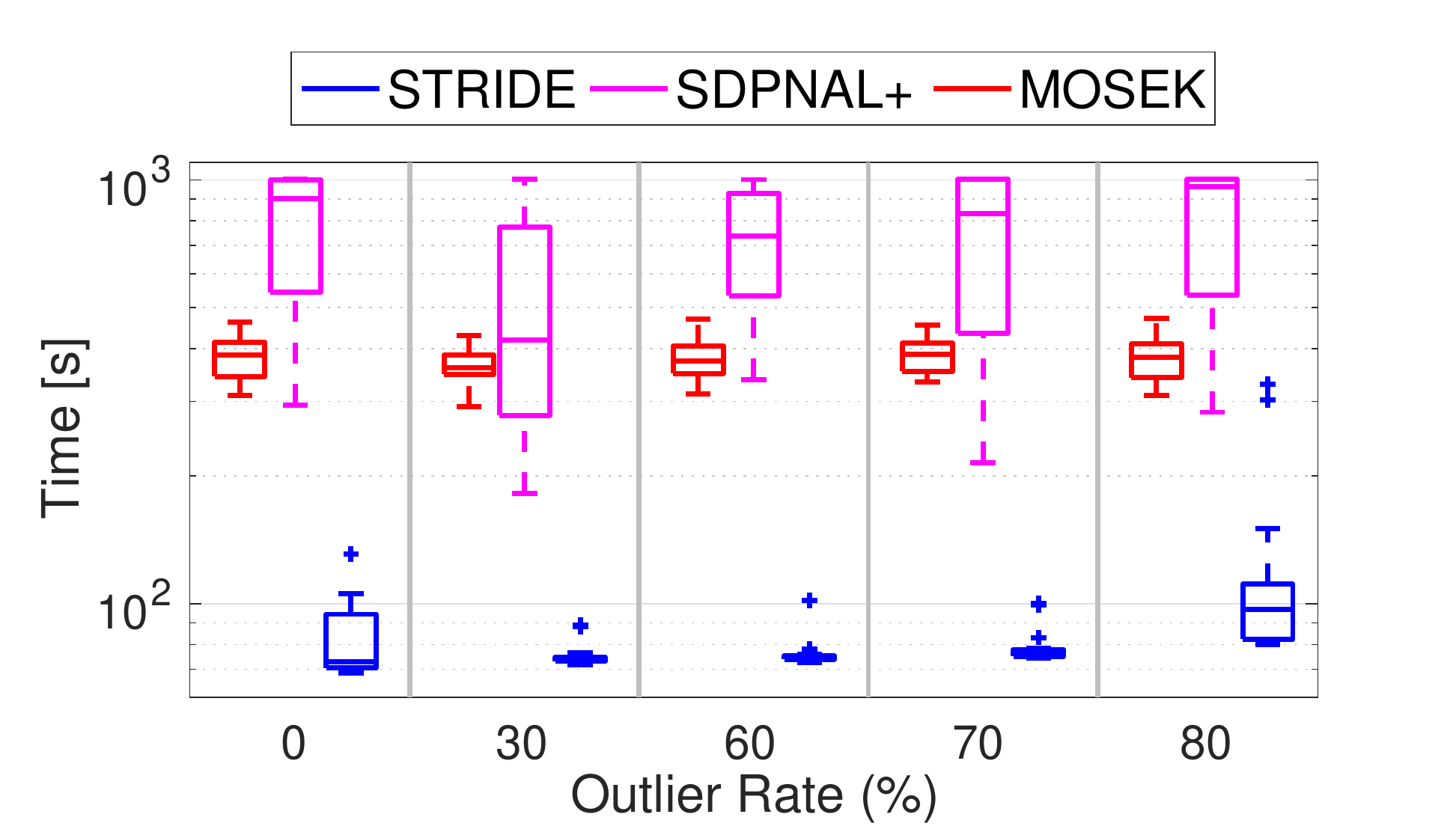}
			\end{minipage} \\
		\multicolumn{4}{c}{\subcapsize (a) $N=20$, $n_1=273$, $m=21,897$}
		\\
		   \myhspace
			\begin{minipage}{\mpwfour}%
			\centering%
			\includegraphics[width=\columnwidth]{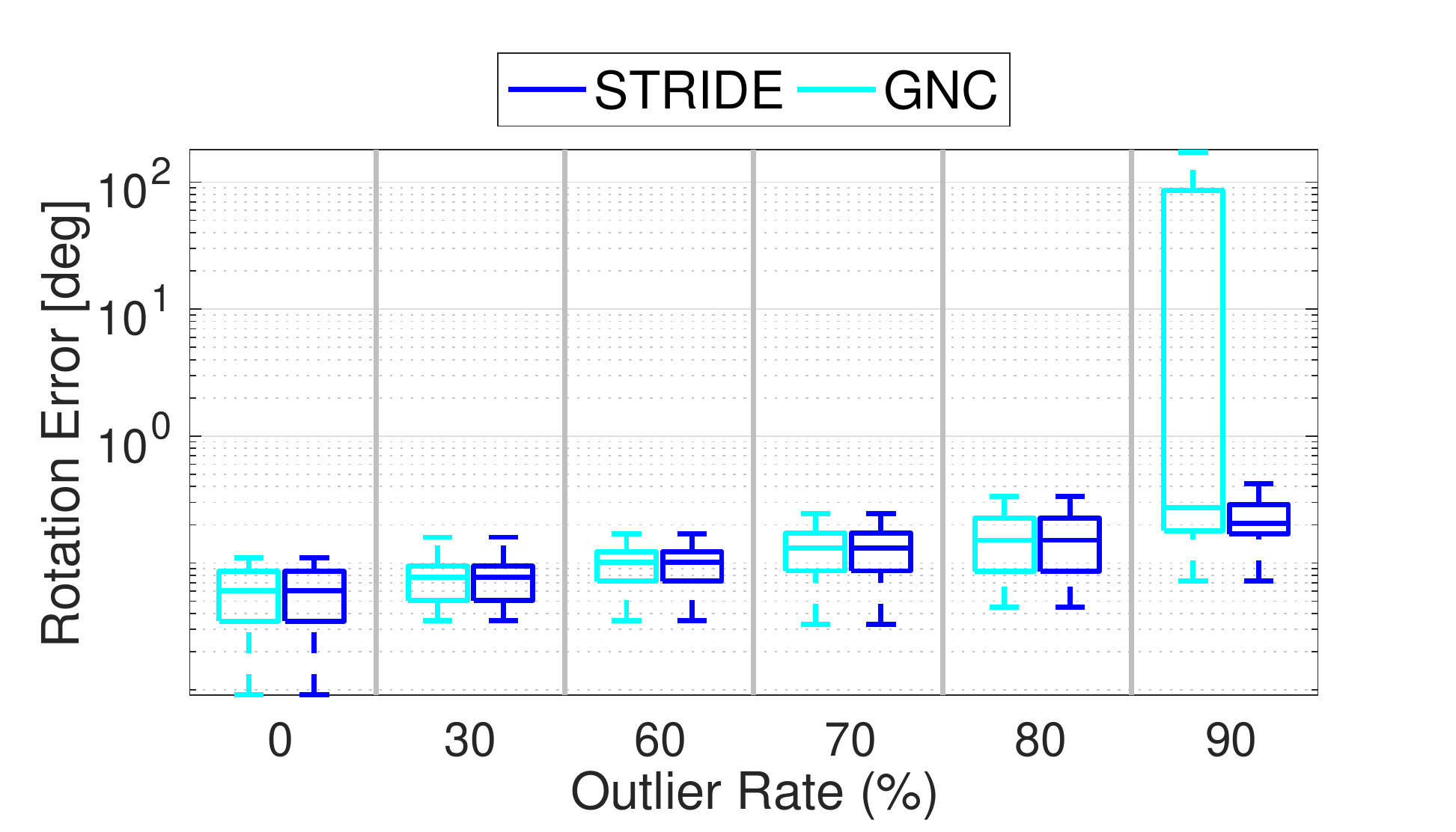}
			\end{minipage}
		&  \myhspace \hspace{-4mm}
			\begin{minipage}{\mpwfour}%
			\centering%
			\includegraphics[width=\columnwidth]{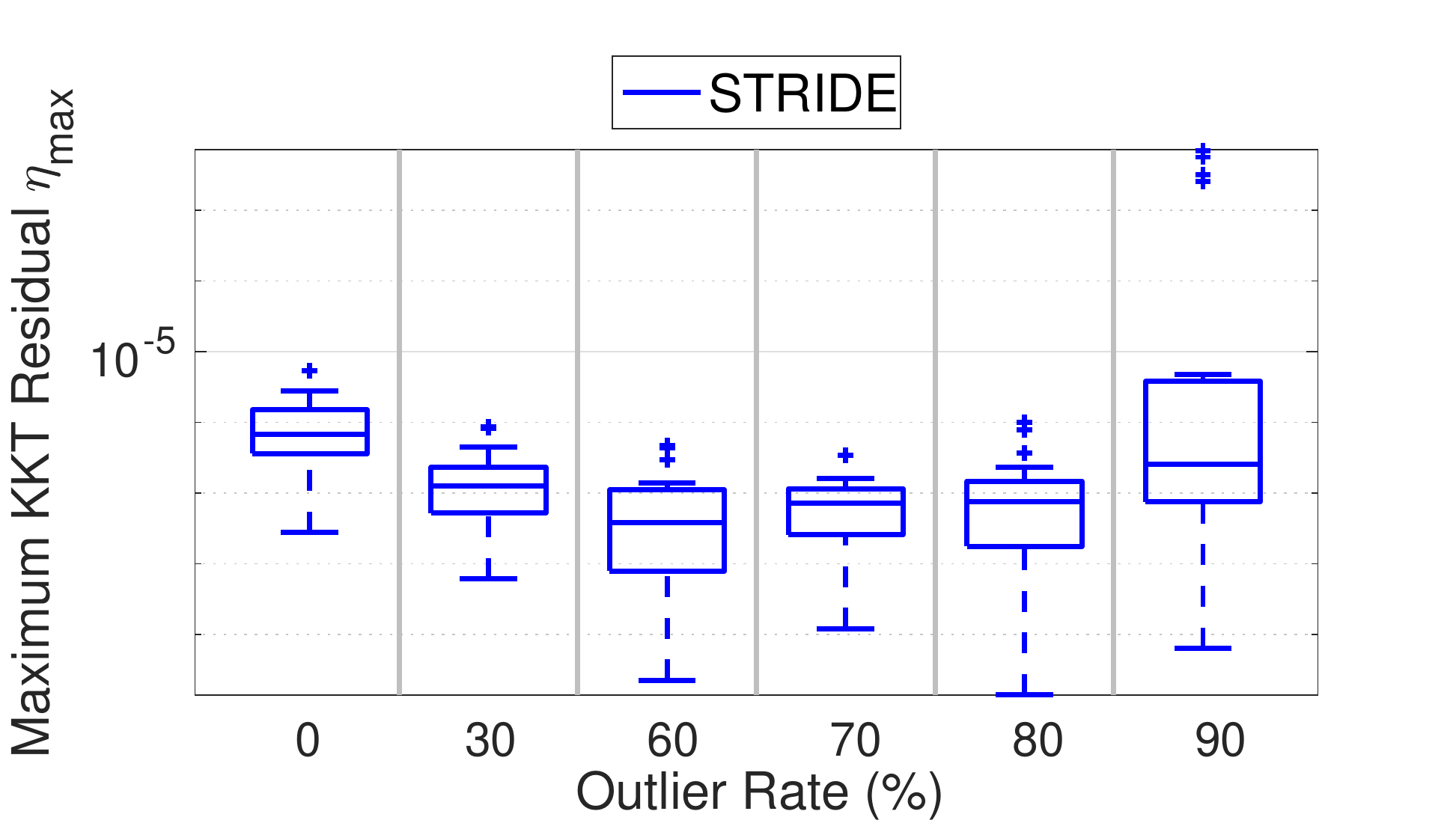}
			\end{minipage}
		&  \myhspace \hspace{-5mm}
			\begin{minipage}{\mpwfour}%
			\centering%
			\includegraphics[width=\columnwidth]{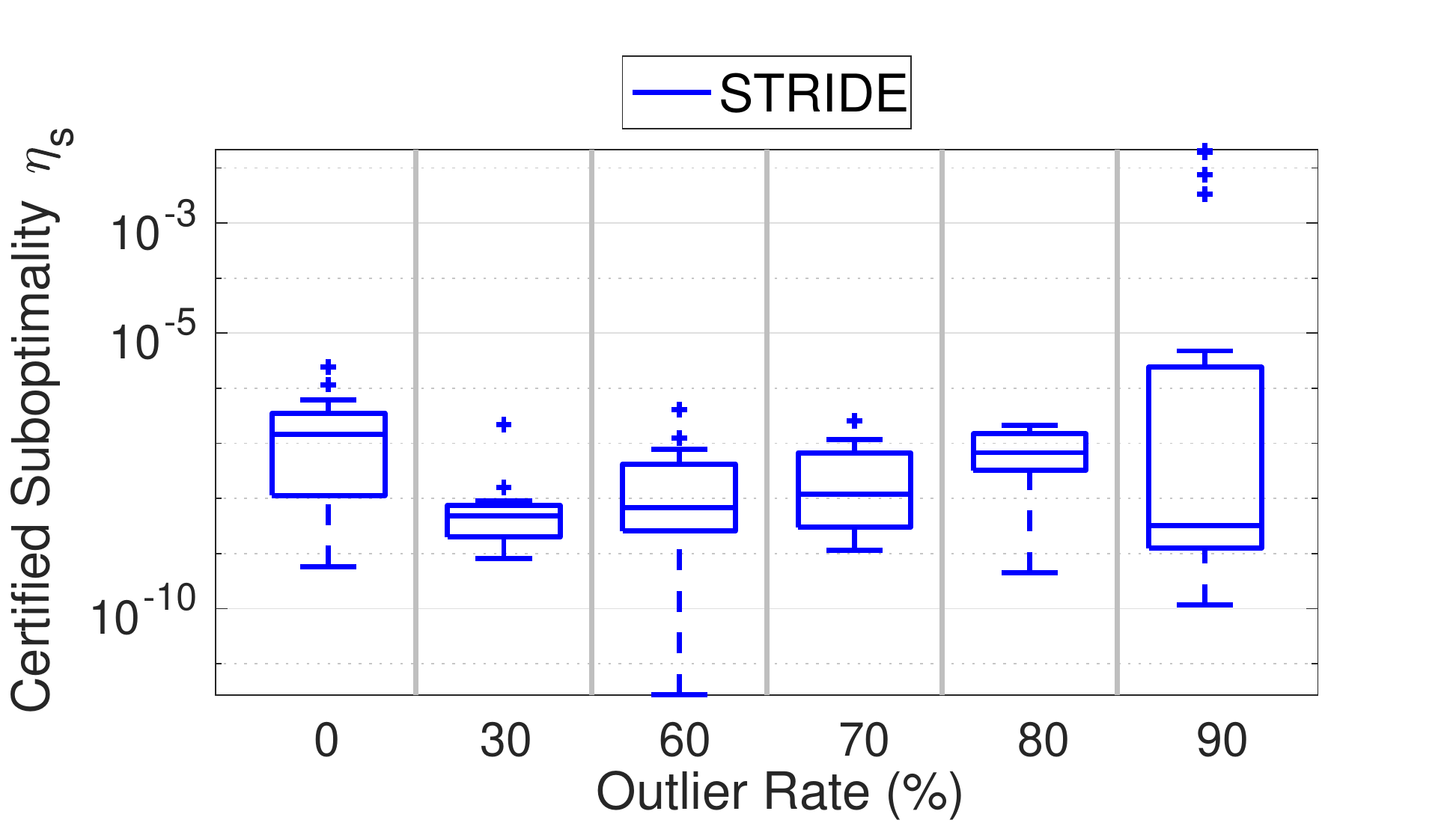}
			\end{minipage}
		&  \myhspace \hspace{-5mm}
			\begin{minipage}{\mpwfour}%
			\centering%
			\includegraphics[width=\columnwidth]{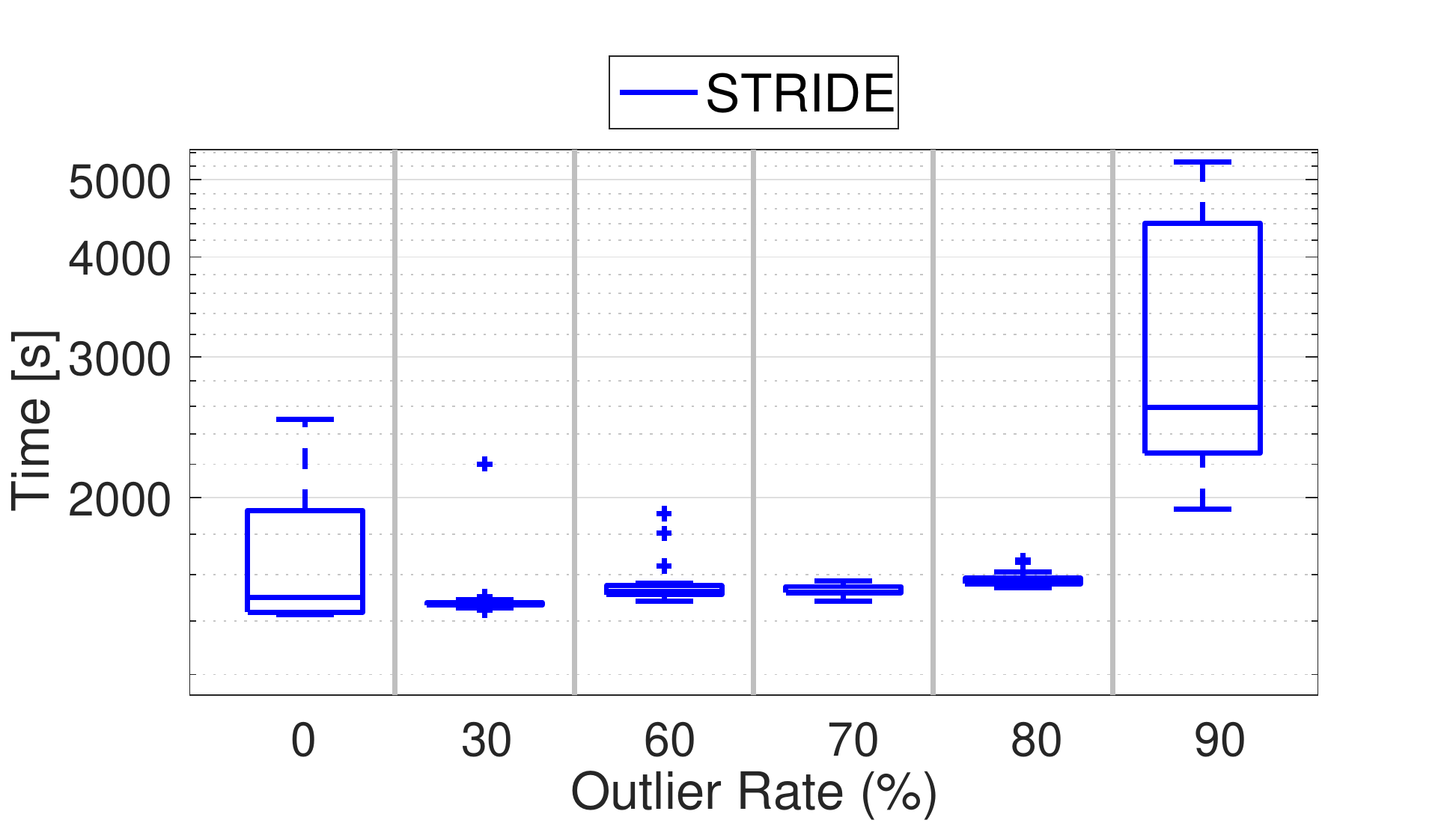}
			\end{minipage} \\
		\multicolumn{4}{c}{\subcapsize (b) $N=100$, $n_1=1313$, $m=485,417$}
		\\
		\multicolumn{2}{c}{
		\myhspace \hspace{-5mm}
		\begin{minipage}{\mpwtwo}%
		\centering%
		\includegraphics[width=0.9\columnwidth]{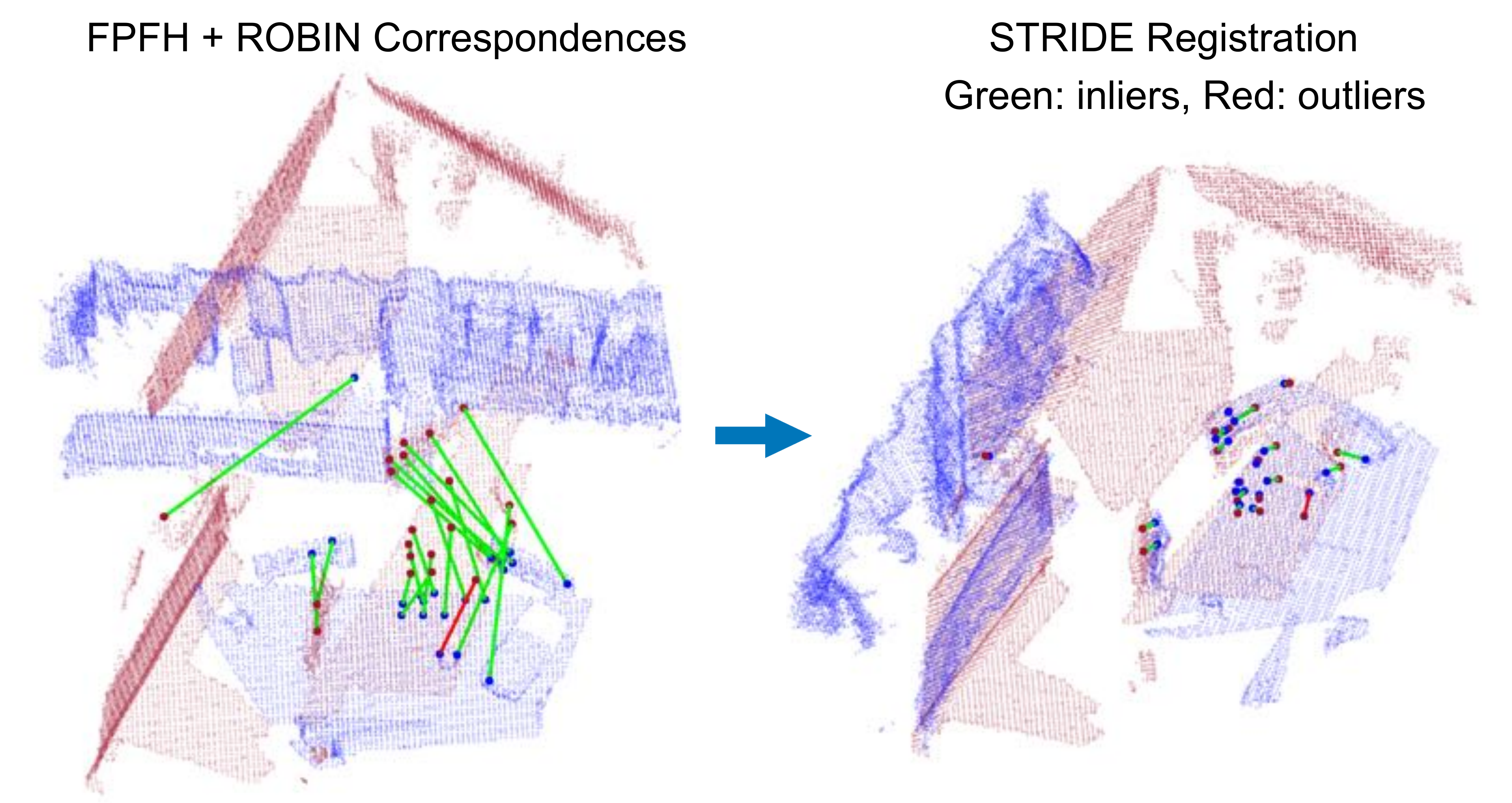}
		\\
		{\subcapsize (c) $\MR$ error: $3.5^{\circ}$, $\vt$ error: $9.5\ee{-2}$,  $\subopt = 1.1\ee{-10}$, time: $75$ [s]}
		\end{minipage} 
		}
		& 
		\multicolumn{2}{c}{
		\myhspace \hspace{-8mm}
		\begin{minipage}{\mpwtwo}%
		\centering%
		\includegraphics[width=0.9\columnwidth]{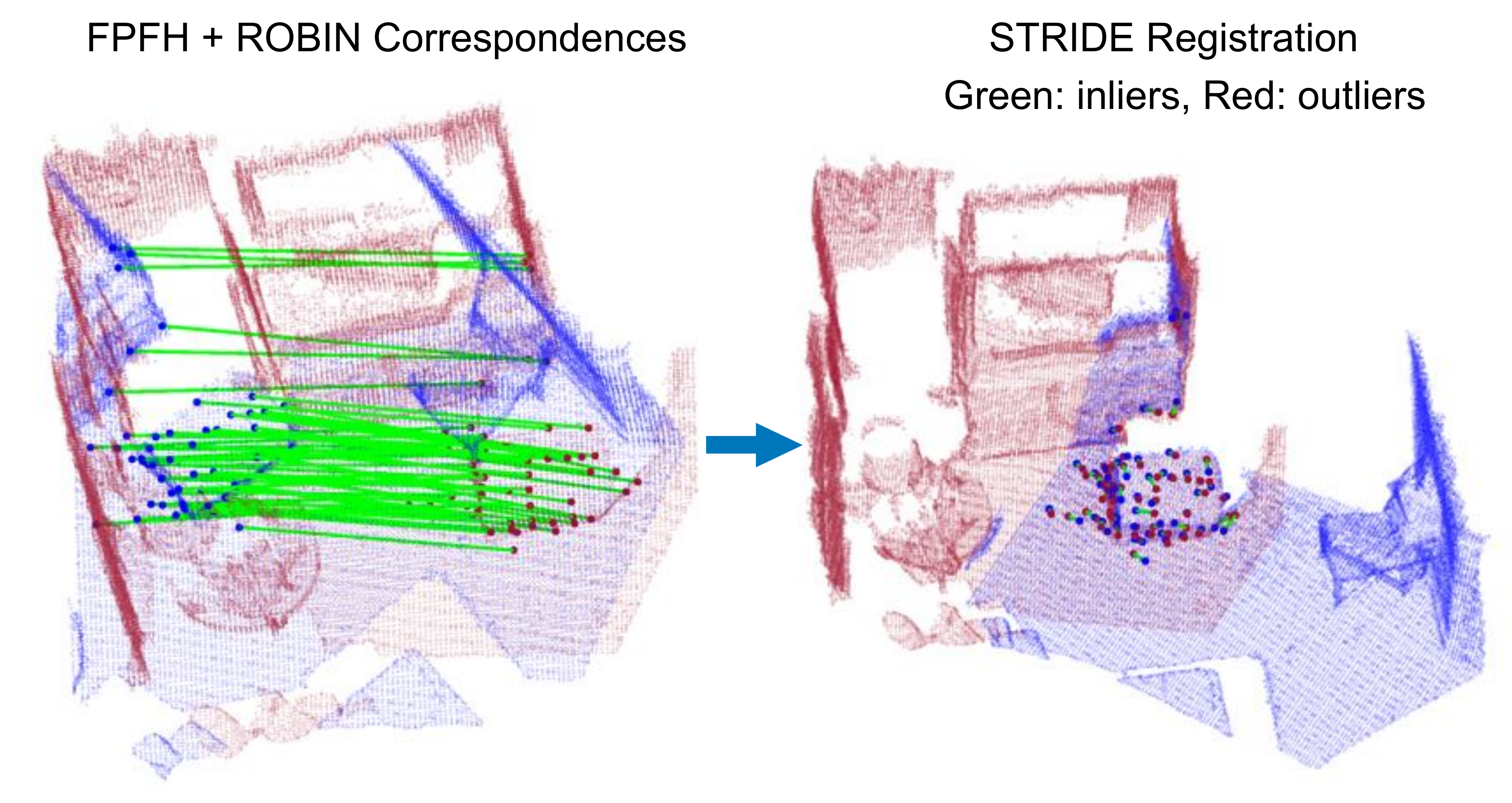}
		\\
		{\subcapsize (d) $\MR$ error: $3.3^{\circ}$, $\vt$ error: $2.0\ee{-1}$,  $\subopt = 1.1\ee{-7}$, time: $558$ [s]}
		\end{minipage}
		}
	\end{tabular}
	\end{minipage} 
	\caption{Point Cloud Registration (Example \ref{ex:pointcloud}).
	\label{fig:exp-pcr-results}} 
	\vspace{-7mm} 
	\end{center}
\end{figure*}

{\bf Setup}. We first sample a random set of 3D points $\{ \vp_i\}_{i=1}^N$, where each $\vp_i \sim \calN(\zero,\eye_3)$. Then we generate a random rotation and translation $(\MRgt,\vtgt)$ such that $\norm{\vtgt} \leq T = 10$. Using $(\MRgt,\vtgt)$, we generate $\{\vq_i\}_{i=1}^N$ by $\vq_i = \MRgt \vp_i + \vtgt + \vvarepsilon_i$ ($\vvarepsilon_i \sim \calN(\zero,0.01^2 \eye_3)$) if $\vq_i$ is an inlier, or by $\vq_i \sim \calN(\zero,\eye_3)$ if $\vq_i$ is an outlier. We test $N = 20$ and $N=100$. 

{\bf Results}. Fig. \ref{fig:exp-pcr-results}(a)-(b) plot the evaluation metrics for $N=20$ and $N=100$, respectively. We make the following observations. (i) When $N=20$, our relaxation is tight with up to $80\%$ outlier correspondences. Both {\mosek} and {\stride} can obtain a certifiably optimal solution, except that {\stride} failed once to attain sufficient accuracy (within $5$ iterations) at $80\%$ outlier rate.
\footnote{Consistent with \cite{Yang20neurips-onering}, we empirically noticed that the relaxation breaks earlier when fewer measurements are available. We remark that the formulation considered in this paper is more challenging that the rotation-only version in~\cite{Yang19iccv-quasar}, which remains tight at $90\%$ outliers.}
However, {\stride} is about $5$ times faster than {\mosek}. {\sdpnal} completely fails in this problem. (ii) When $N=100$, our relaxation is exact with up to $90\%$ outliers and {\stride} is the only solver that can certify exactness. At $90\%$ outlier rate, {\stride} certified the global optimality for $17$ runs, while failed to do so for $3$ runs. (iii) {\stride} can certify the success of {\gnc} and escape local minima when {\gnc} fails (\eg at $60-80\%$ when $N=20$ and at $90\%$ when $N=100$).

{\bf Scan matching on {\threedmatch}}. To showcase the practical value of {\stride}, we perform scan matching using the {\threedmatch} test data \cite{Zeng17cvpr-3dmatch}. We use {\fpfh} \cite{Rusu09icra-fast3Dkeypoints} to generate putative feature matches, followed by using {\robin} \cite{Shi21icra-robin} to filter out gross outliers. The result of {\fpfh} and {\robin} is typically a set of sparse keypoint matches with only a few outliers. We then use {\stride} to \emph{certifiably} estimate the rigid transformation. Fig. \ref{fig:exp-pcr-results}(c)-(d) visualize two examples where {\stride} returns certified globally optimal estimates ($\subopt < 1\ee{-6}$). More examples are provided in {\supp}.

\subsection{Absolute Pose Estimation}
\begin{figure*}[t]
	\begin{center}
	\begin{minipage}{\textwidth}
	\begin{tabular}{cccc}%
		   \myhspace \hspace{-4mm}
			\begin{minipage}{\mpwfour}%
			\centering%
			\includegraphics[width=\columnwidth]{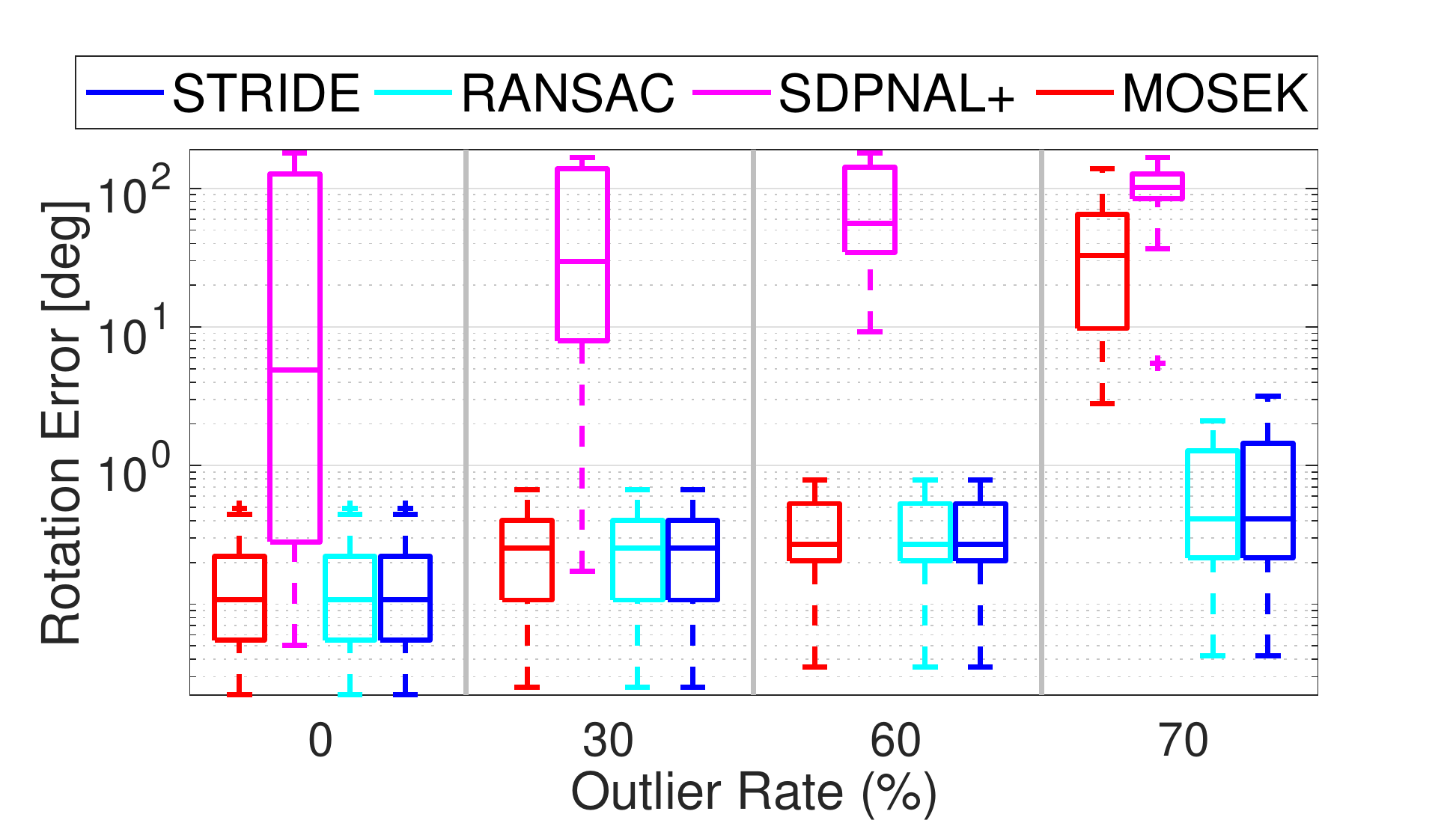}
			\end{minipage}
		&  \myhspace \hspace{-3mm}
			\begin{minipage}{\mpwfour}%
			\centering%
			\includegraphics[width=\columnwidth]{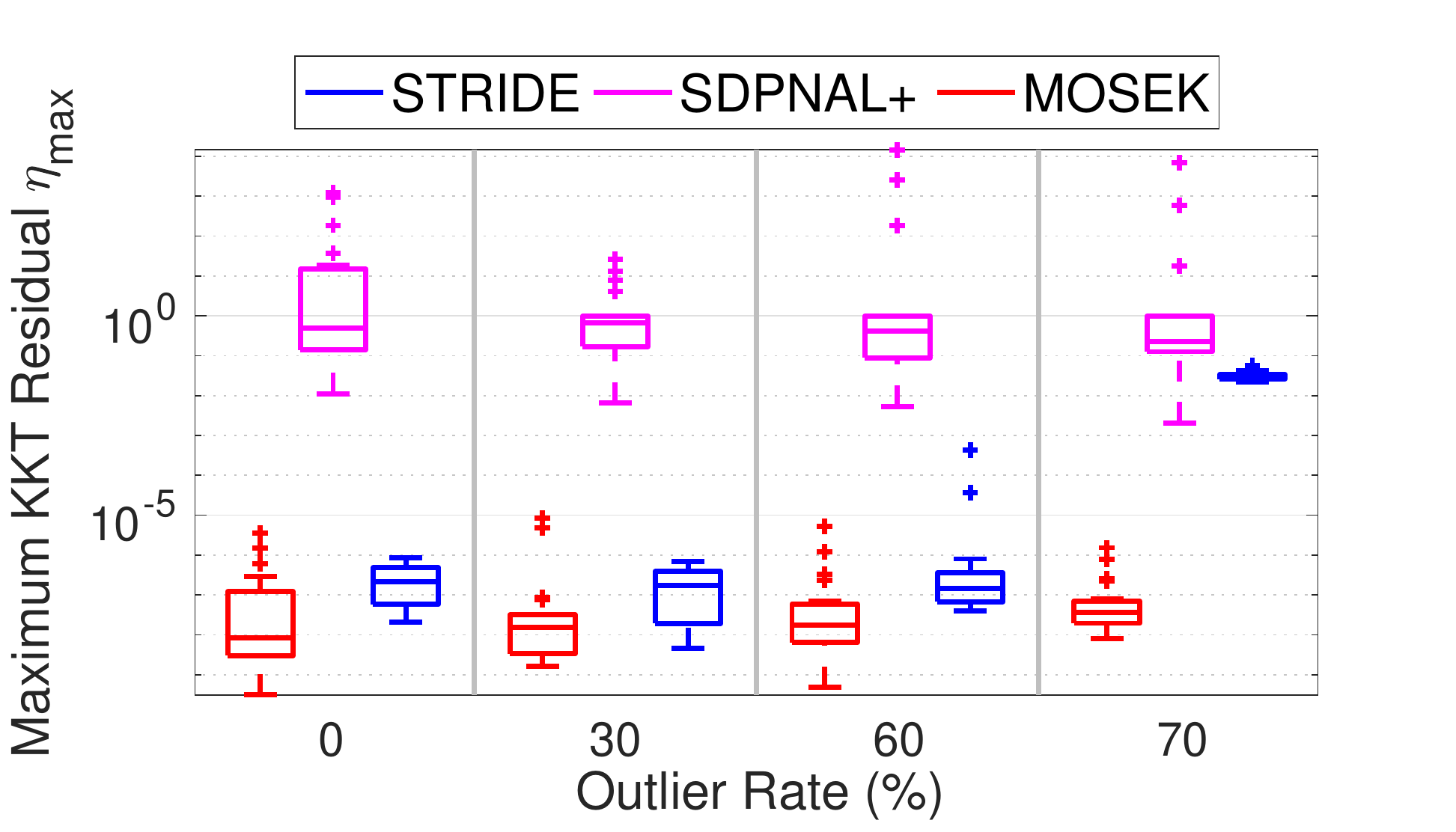}
			\end{minipage}
		&  \myhspace \hspace{-3mm}
			\begin{minipage}{\mpwfour}%
			\centering%
			\includegraphics[width=\columnwidth]{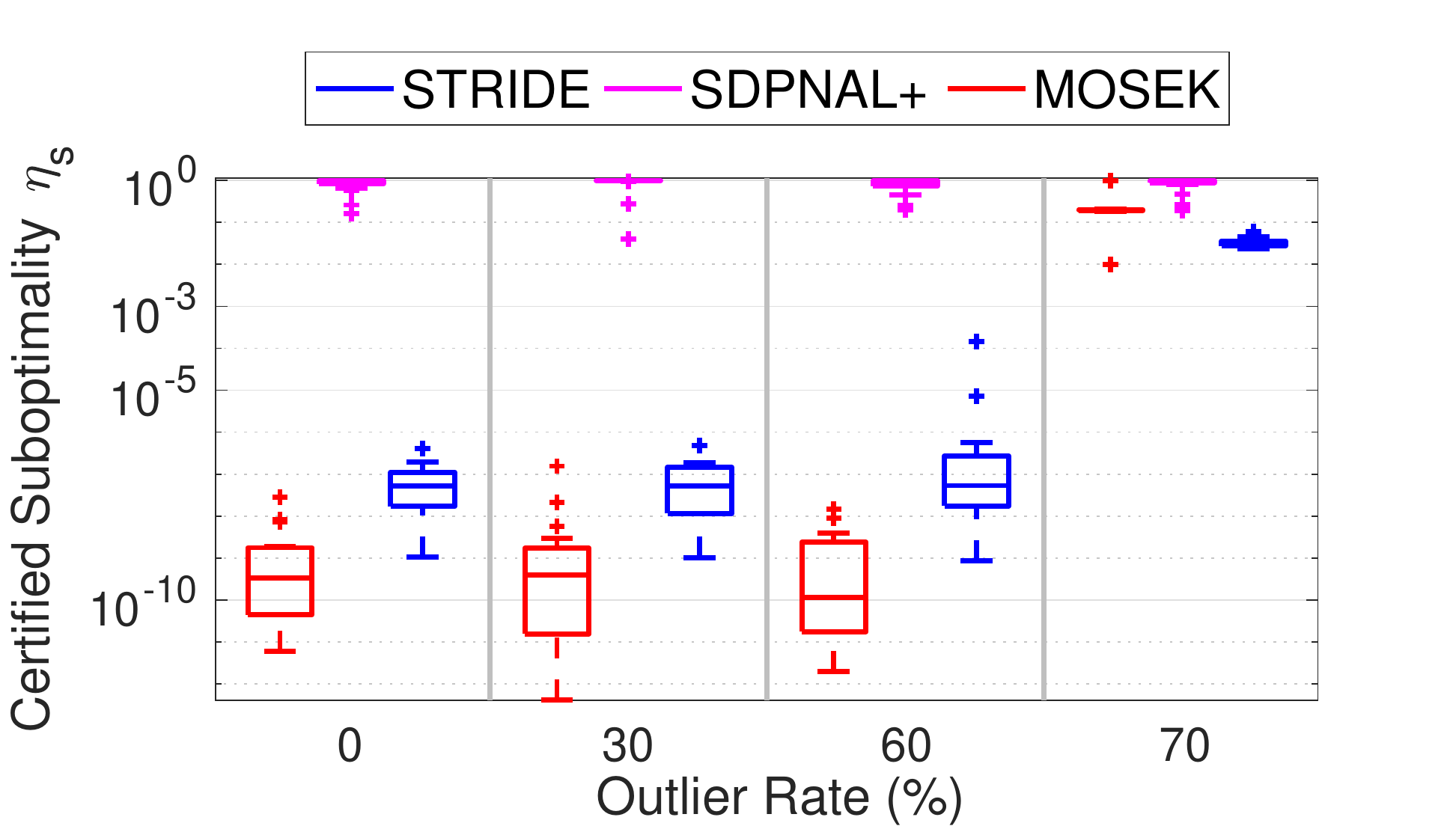}
			\end{minipage}
		&  \myhspace \hspace{-3mm}
			\begin{minipage}{\mpwfour}%
			\centering%
			\includegraphics[width=\columnwidth]{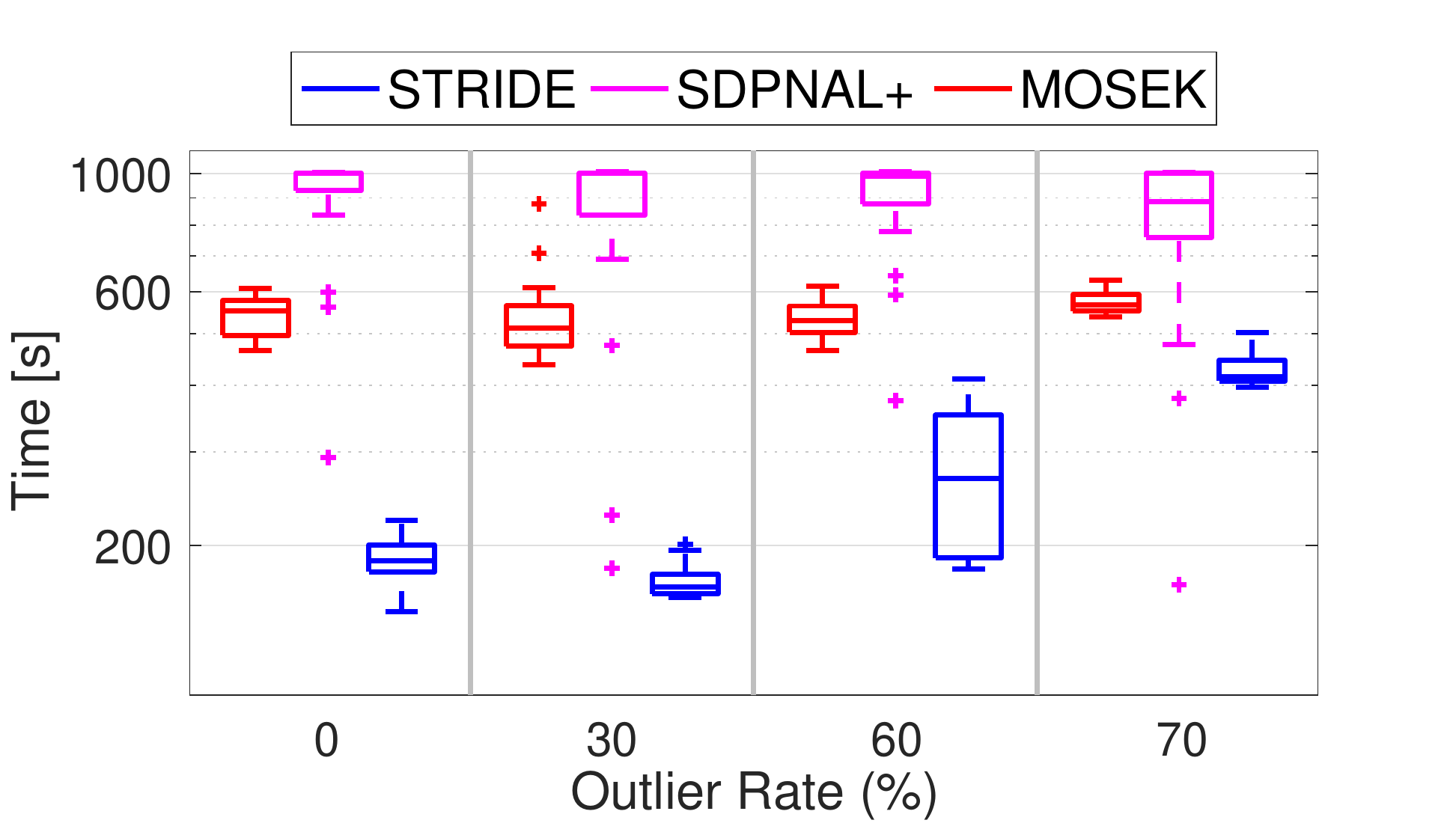}
			\end{minipage} \\
		\multicolumn{4}{c}{\subcapsize (a) $N=20$, $n_1=273$, $m=25,824$}
		\\
		   \myhspace \hspace{-4mm}
			\begin{minipage}{\mpwfour}%
			\centering%
			\includegraphics[width=\columnwidth]{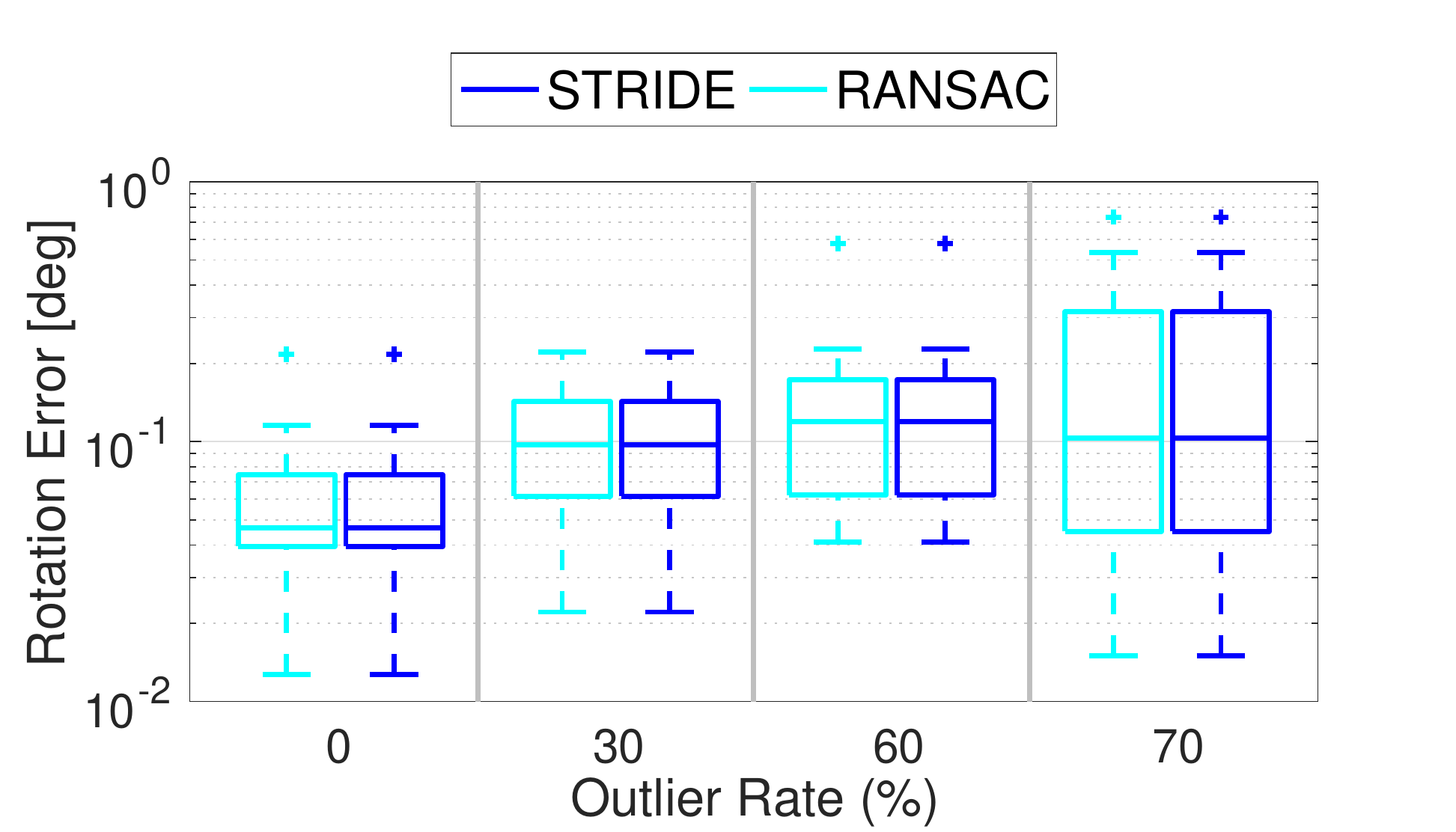}
			\end{minipage}
		&  \myhspace \hspace{-3mm}
			\begin{minipage}{\mpwfour}%
			\centering%
			\includegraphics[width=\columnwidth]{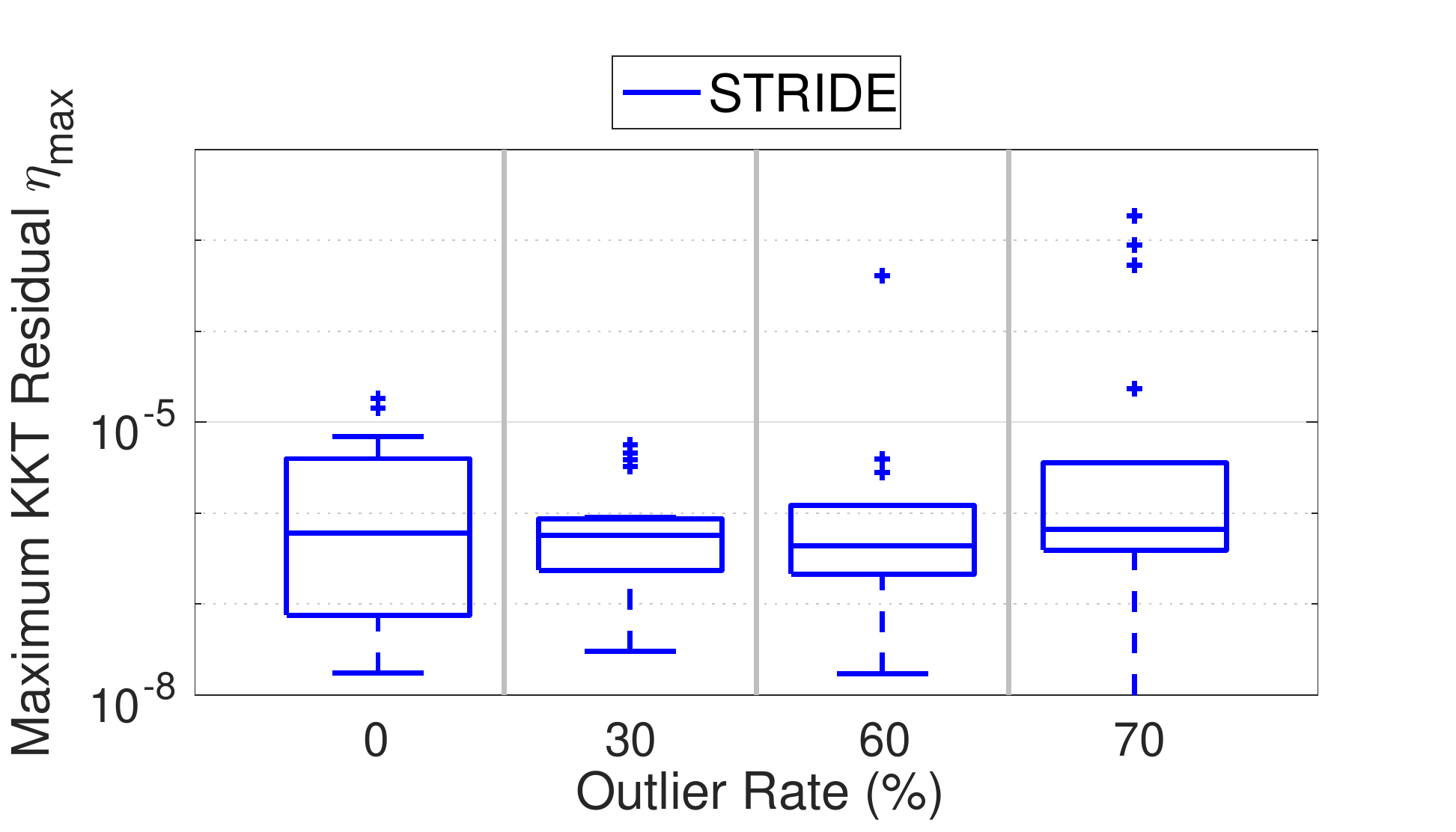}
			\end{minipage}
		&  \myhspace \hspace{-3mm}
			\begin{minipage}{\mpwfour}%
			\centering%
			\includegraphics[width=\columnwidth]{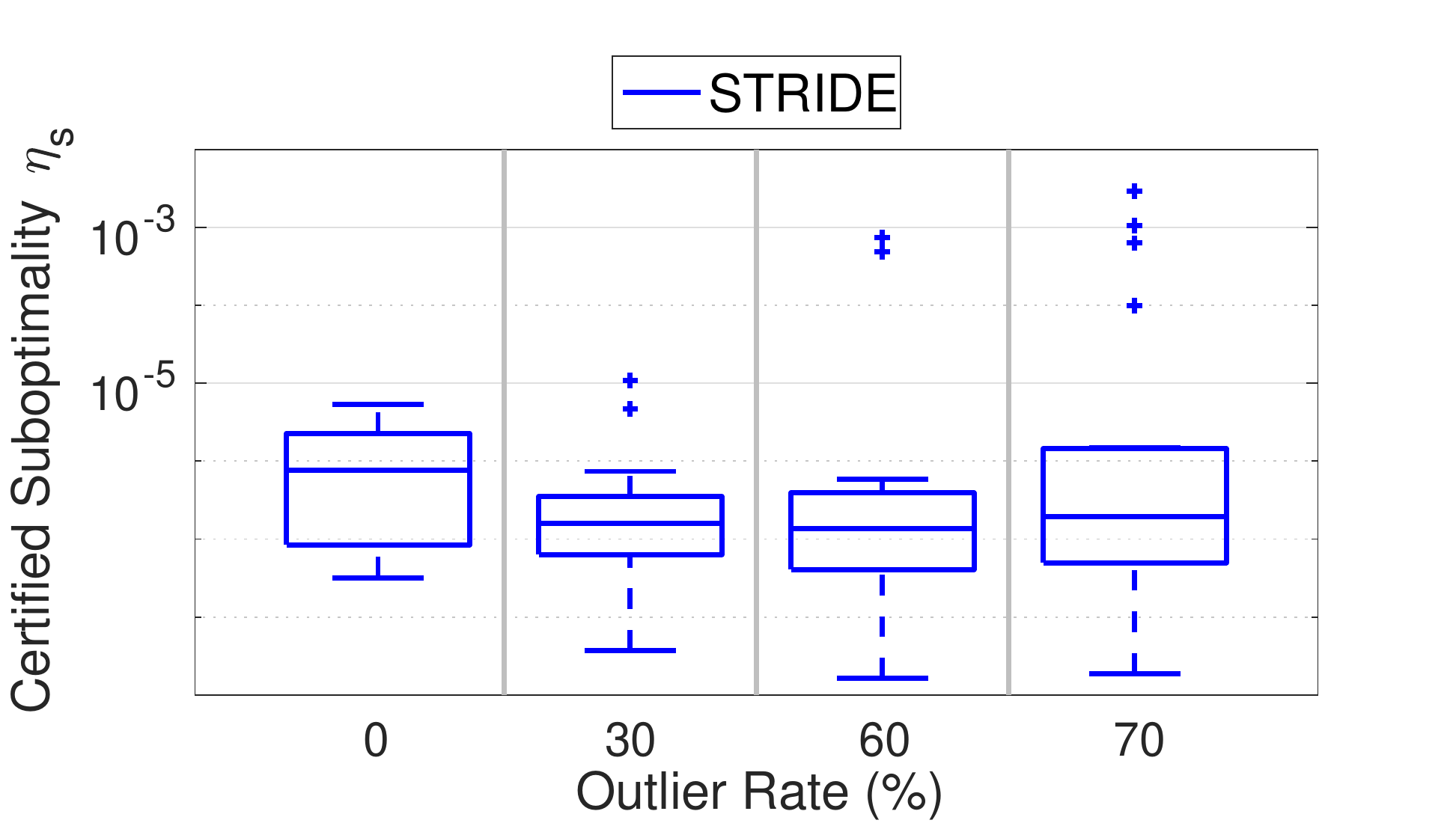}
			\end{minipage}
		&  \myhspace \hspace{-3mm}
			\begin{minipage}{\mpwfour}%
			\centering%
			\includegraphics[width=\columnwidth]{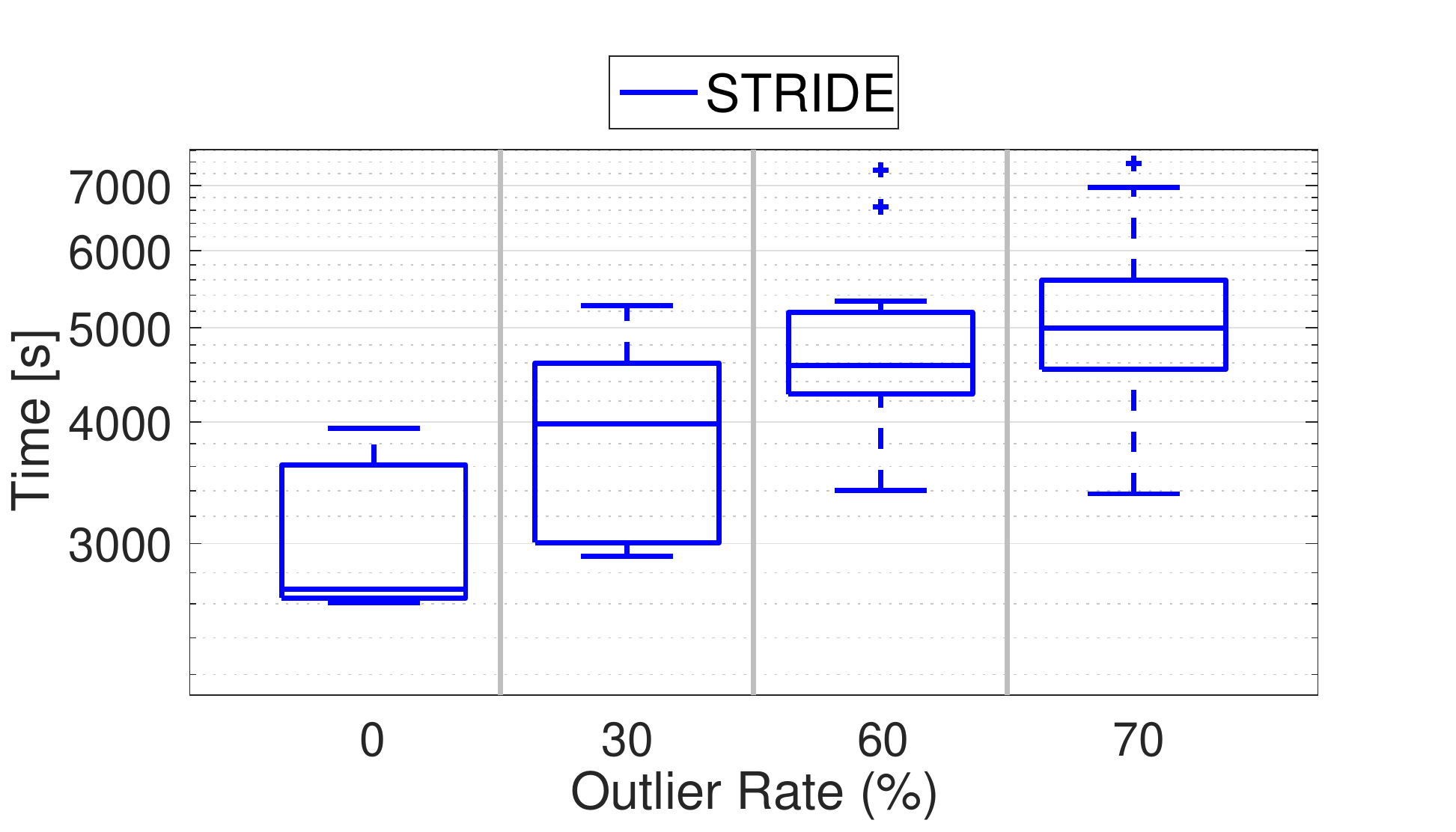}
			\end{minipage} \\
		\multicolumn{4}{c}{\subcapsize (b) $N=100$, $n_1=1313$, $m=572,984$}
		\\
		   \myhspace \hspace{-4mm}
			\begin{minipage}{\mpwfour}%
			\centering%
			\includegraphics[width=0.95\columnwidth]{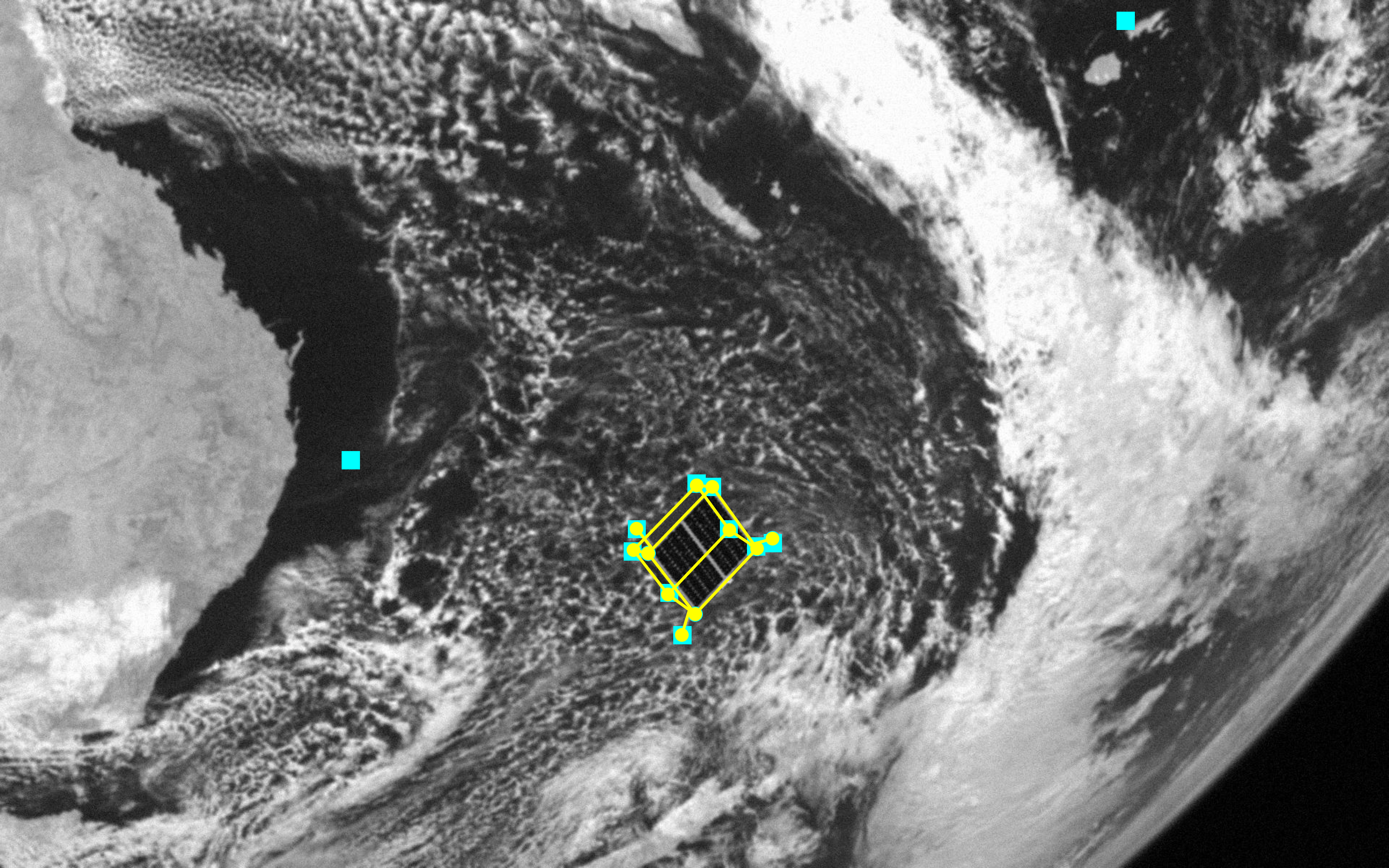}\\
			\subcapsize $2$ outliers, 
			 $\MR$ error: $1.0^{\circ}$, $\vt$ error: $0.25$ \\ 
			 $\subopt=8.8\ee{-8}$, time: $29$ [s]
			\end{minipage}
		&  \myhspace \hspace{-3mm}
			\begin{minipage}{\mpwfour}%
			\centering%
			\includegraphics[width=0.95\columnwidth]{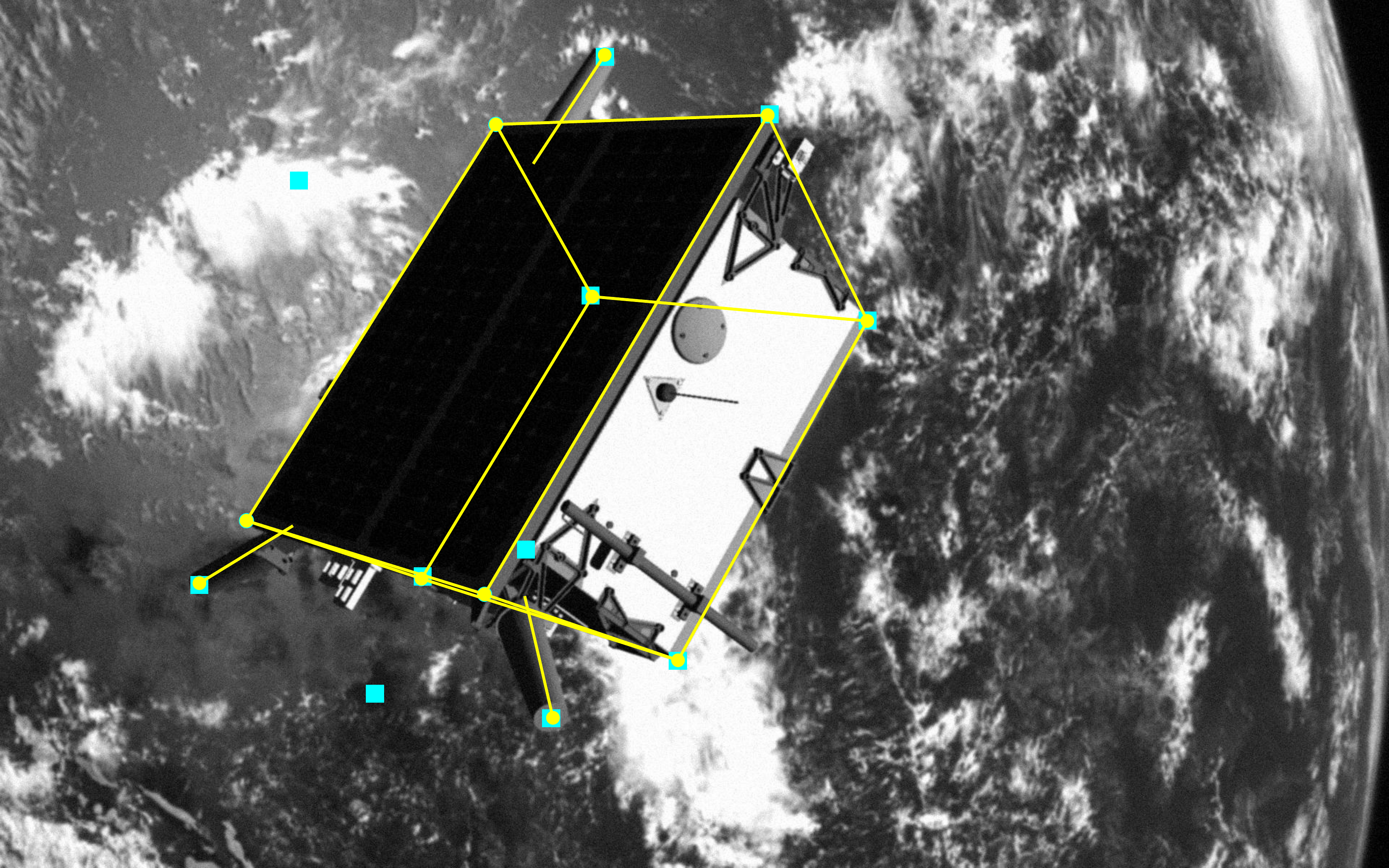}\\
			\subcapsize $3$ outliers, 
			 $\MR$ error: $0.2^{\circ}$, $\vt$ error: $0.01$ \\ 
			 $\subopt=1.0\ee{-9}$, time: $33$ [s]
			\end{minipage}
		&  \myhspace \hspace{-3mm}
			\begin{minipage}{\mpwfour}%
			\centering%
			\includegraphics[width=0.95\columnwidth]{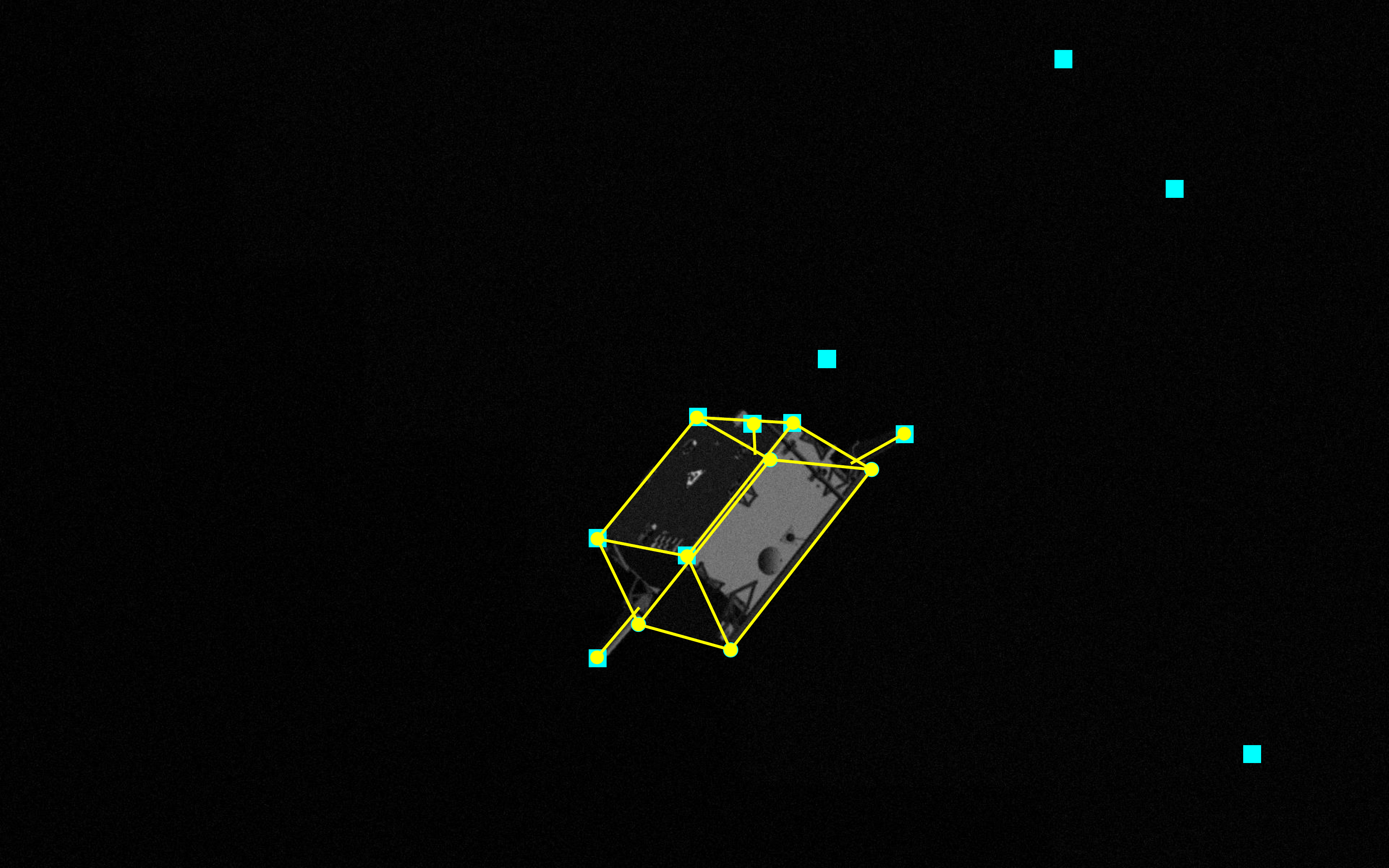}\\
			\subcapsize $4$ outliers, 
			 $\MR$ error: $0.5^{\circ}$, $\vt$ error: $0.03$ \\ 
			 $\subopt=9.6\ee{-8}$, time: $39$ [s]
			\end{minipage}
		&  \myhspace \hspace{-3mm}
			\begin{minipage}{\mpwfour}%
			\centering%
			\includegraphics[width=0.95\columnwidth]{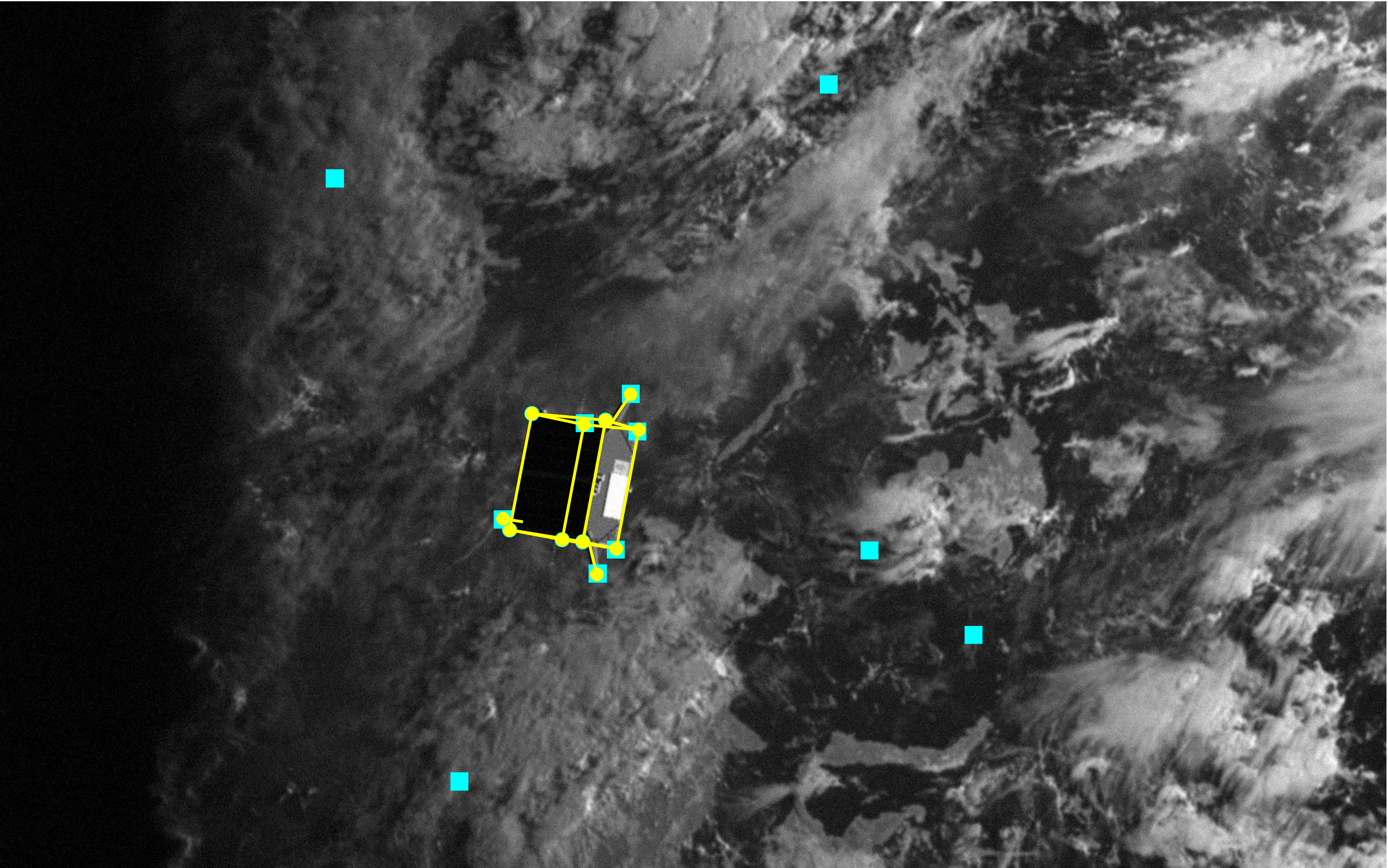}\\
			\subcapsize $5$ outliers, 
			 $\MR$ error: $0.3^{\circ}$, $\vt$ error: $0.07$ \\ 
			 $\subopt=2.5\ee{-8}$, time: $39$ [s]
			\end{minipage}
			\vspace{1mm}
		\\
		\multicolumn{4}{c}{\hspace{-3mm} \subcapsize (c) Satellite pose estimation on {\speed} \cite{Sharma19arxiv-SPEED}. Cyan square: detected 2D keypoints; Yellow circle: reprojected 3D keypoints. \vspace{-2mm}}
	\end{tabular}
	\end{minipage}
	\caption{Absolute Pose Estimation (Example \ref{ex:absolutepose}).
	\label{fig:exp-ape-results}} 
	\vspace{-8mm} 
	\end{center}
\end{figure*}

{\bf Setup}. We first generate a set of random 3D points $\{ \vp_i \}_{i=1}^N$ that are centered at zero. We then generate a random pose $(\MRgt,\vtgt)$ such that $\norm{\vtgt} \leq T=10$ and $\vtgt$ lies inside the camera FOV cone $\calC_\alpha$ with $\alpha = \frac{\pi}{2}$. Using $(\MRgt,\vtgt)$ and $\{ \vp_i \}_{i=1}^N$, we generate 2D keypoints by projecting the transformed 3D points onto the imaging plane, \ie $\vv_i = \calP(\MRgt \vp_i + \vtgt)$, where $\calP: \Real{3} \rightarrow \Real{2}$ is defined as $\calP(\va) = [a_1/a_3\vcat a_2/a_3]$. We then generate the inlier bearing vectors from the 2D keypoints by $\vu_i = \normalize([\vv_i + \vvarepsilon_i \vcat 1])$, where $\vvarepsilon_i \sim \calN(\zero,0.001^2 \eye_2)$ is a random 2D Gaussian noise. For outliers, we generate $\vu_i$ as random unit vectors inside the FOV cone. We test $N=20$ and $N=100$ with increasing outlier rates. We use {\ransac} implemented in the Matlab function $\texttt{estimateWorldCameraPose}$ to initialize {\stride}.

{\bf Results}. Fig. \ref{fig:exp-ape-results}(a)-(b) plot the evaluation metrics. We make the following observations. (i) When $N=20$, our relaxation is exact with up to $60\%$ outliers. At $70\%$ outlier rate, even if {\mosek} solves the SDP to high accuracy, since the solution in not rank one, the rounding procedure obtains a pose estimation that is far from the groundtruth. (ii) When $N=100$, our relaxation becomes mostly tight at $70\%$ outlier rate, which suggests that increasing the total number of matches could lead to a tighter relaxation.

{\bf Satellite pose estimation on {\speed}}. We showcase {\stride} on satellite pose estimation using the {\speed} dataset \cite{Sharma19arxiv-SPEED}. We use the 3D satellite model provided in \cite{Chen19ICCVW-satellitePoseEstimation} (with $N=11$ keypoints) and spoil the groundtruth 2D keypoints with outliers. Fig. \ref{fig:exp-ape-results}(c) shows four examples with 2-5 outliers, where {\stride} obtains accurate pose estimates with certified global optimality in less than one minute. More examples are provided in {\supp}.

\subsection{Category-Level Object Perception}
\begin{figure*}[t]
	\begin{center}
	\begin{minipage}{\textwidth}
	\begin{tabular}{cccc}%
		   \myhspace \hspace{-4mm}
			\begin{minipage}{\mpwfour}%
			\centering%
			\includegraphics[width=\columnwidth]{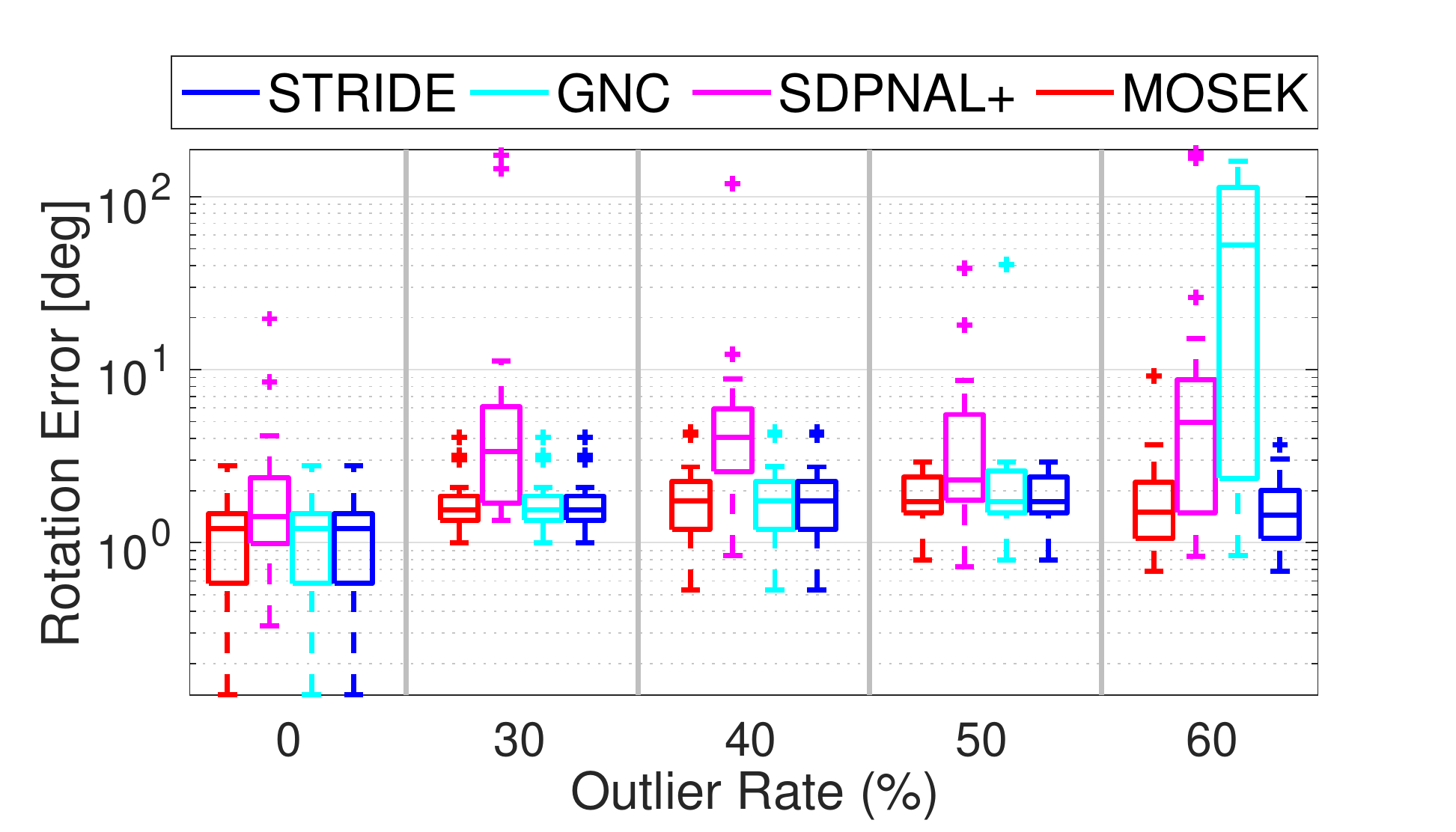}
			\end{minipage}
		&  \myhspace \hspace{-3mm}
			\begin{minipage}{\mpwfour}%
			\centering%
			\includegraphics[width=\columnwidth]{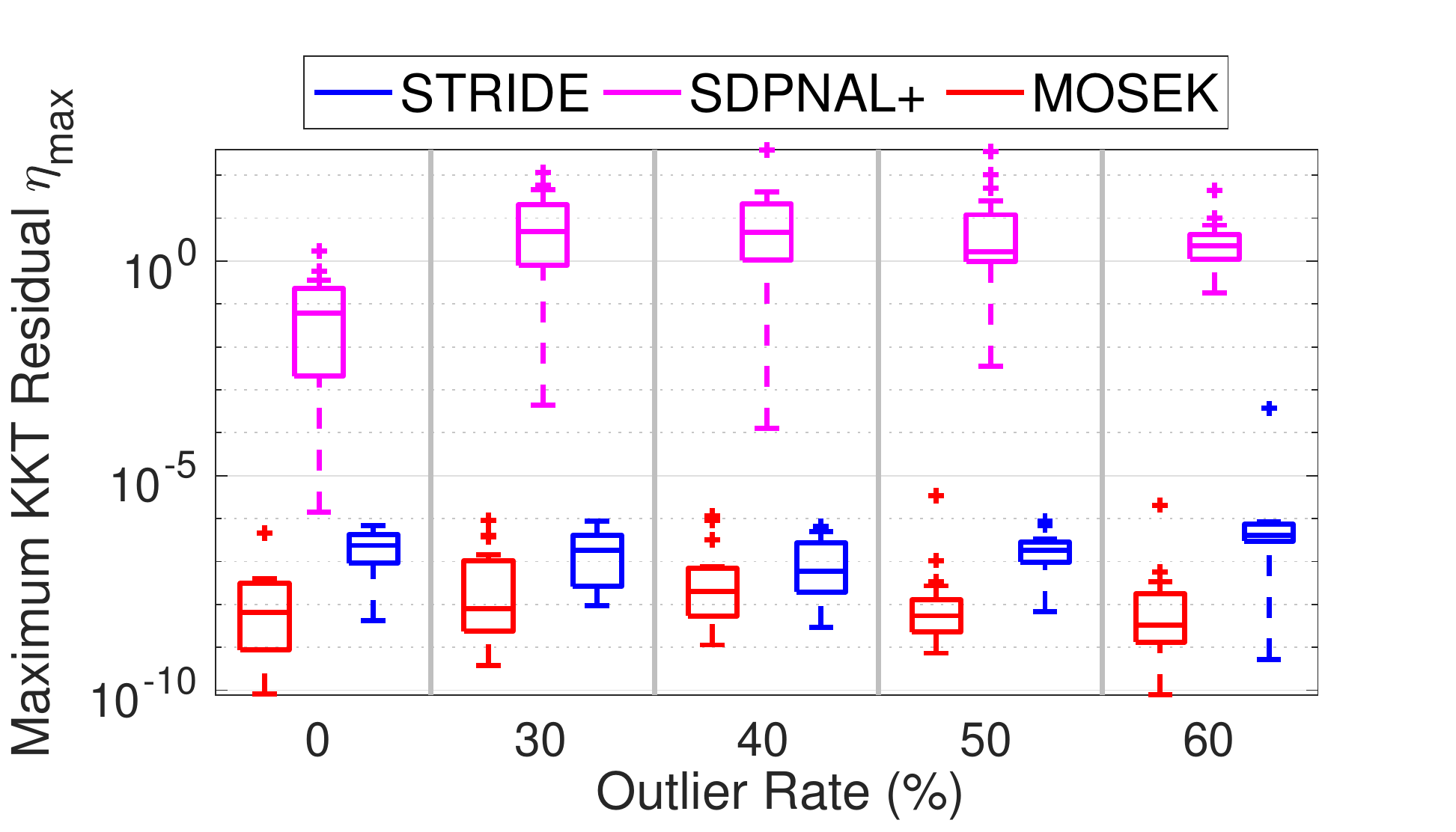}
			\end{minipage}
		&  \myhspace \hspace{-3mm}
			\begin{minipage}{\mpwfour}%
			\centering%
			\includegraphics[width=\columnwidth]{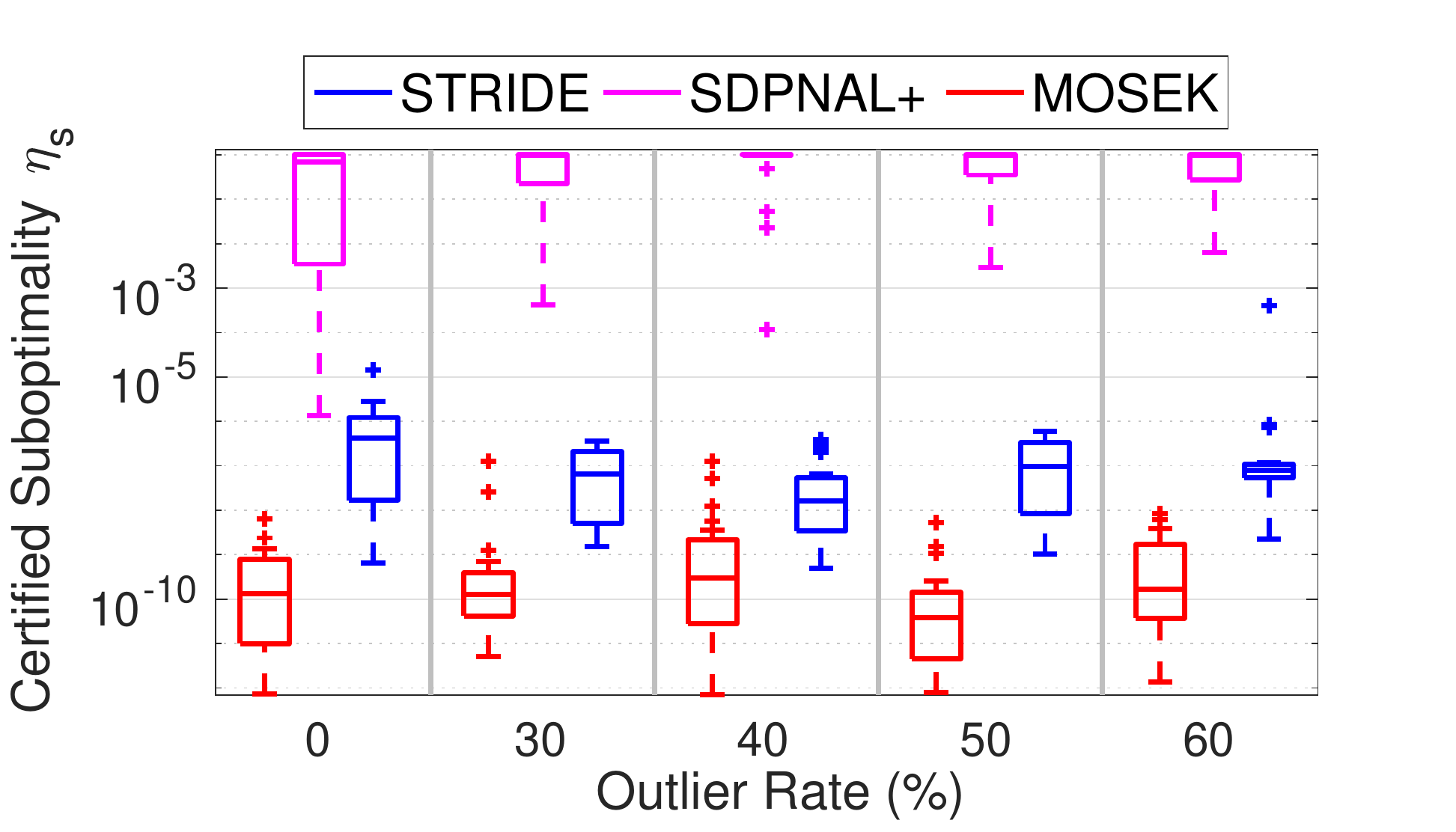}
			\end{minipage}
		&  \myhspace \hspace{-3mm}
			\begin{minipage}{\mpwfour}%
			\centering%
			\includegraphics[width=\columnwidth]{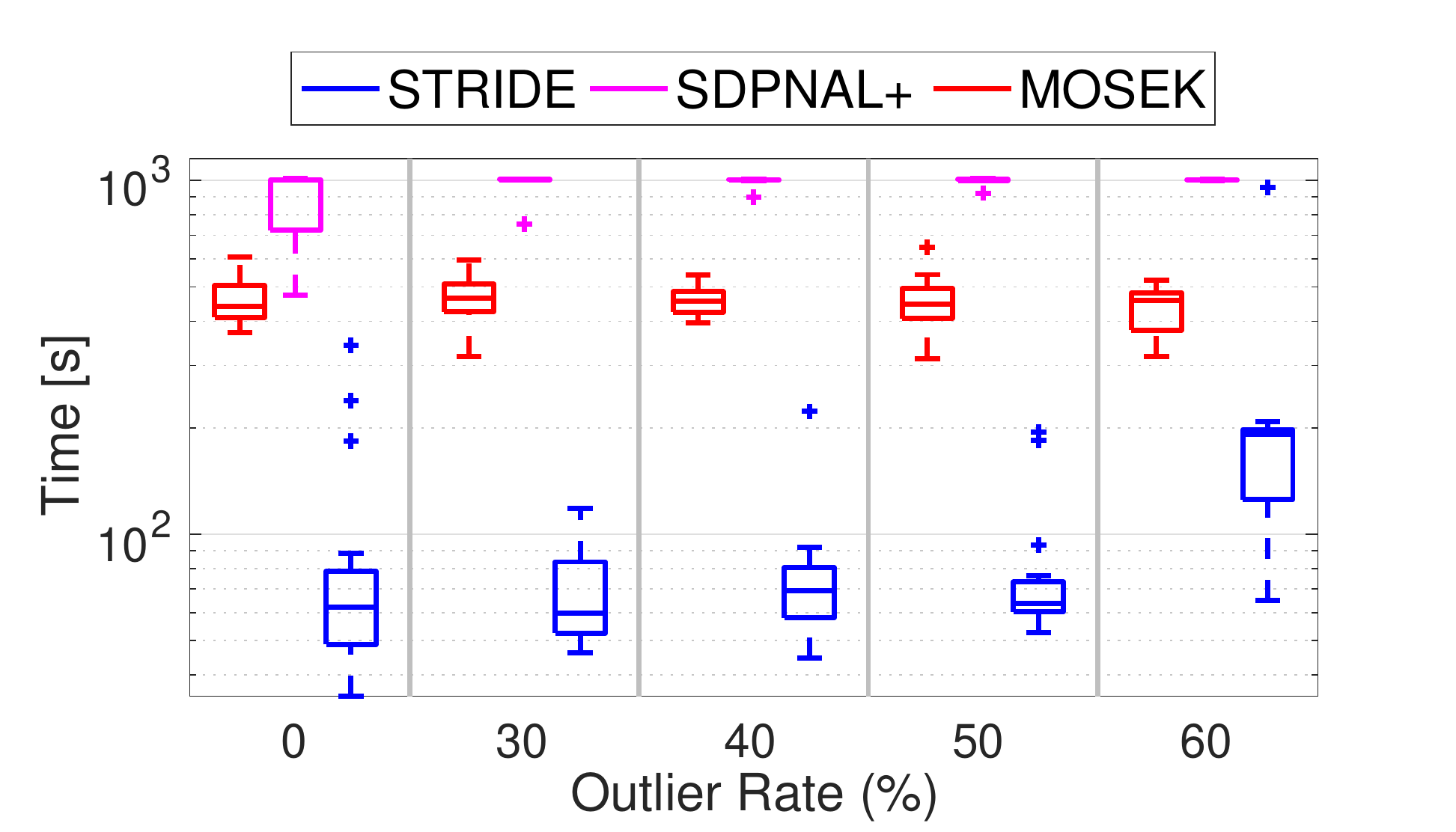}
			\end{minipage} \\
		\multicolumn{4}{c}{\subcapsize (a) \emph{car} category in \pascal \cite{Xiang14-Pascal3DPlus}: $N=12$, $K=9$, $n_1=286$, $m=23,241$ \vspace{1mm}} \\
		   \myhspace \hspace{-4mm}
			\begin{minipage}{\mpwfour}%
			\centering%
			\includegraphics[width=0.95\columnwidth]{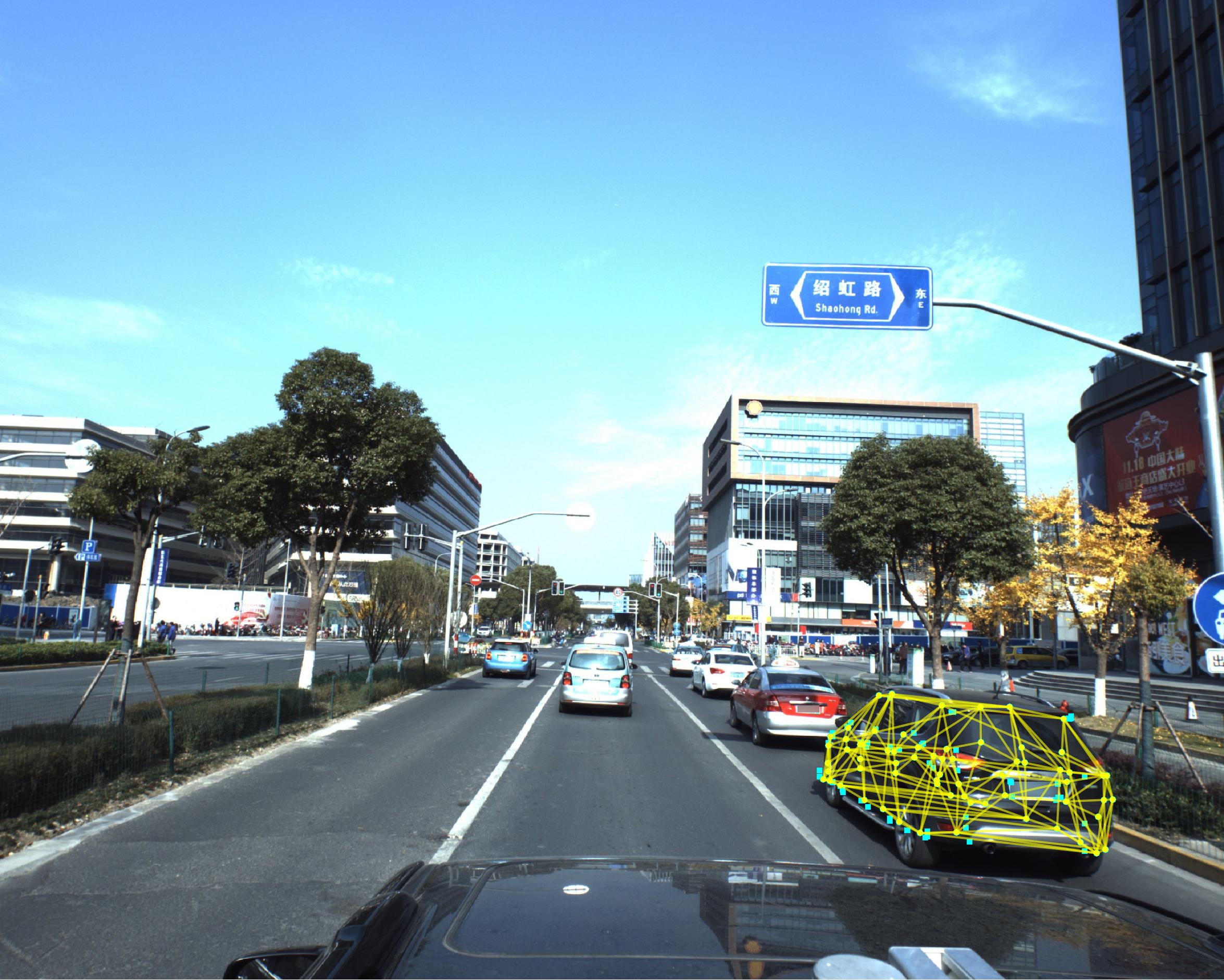}
			\end{minipage}
		&  \myhspace \hspace{-3mm}
			\begin{minipage}{\mpwfour}%
			\centering%
			\includegraphics[width=0.95\columnwidth]{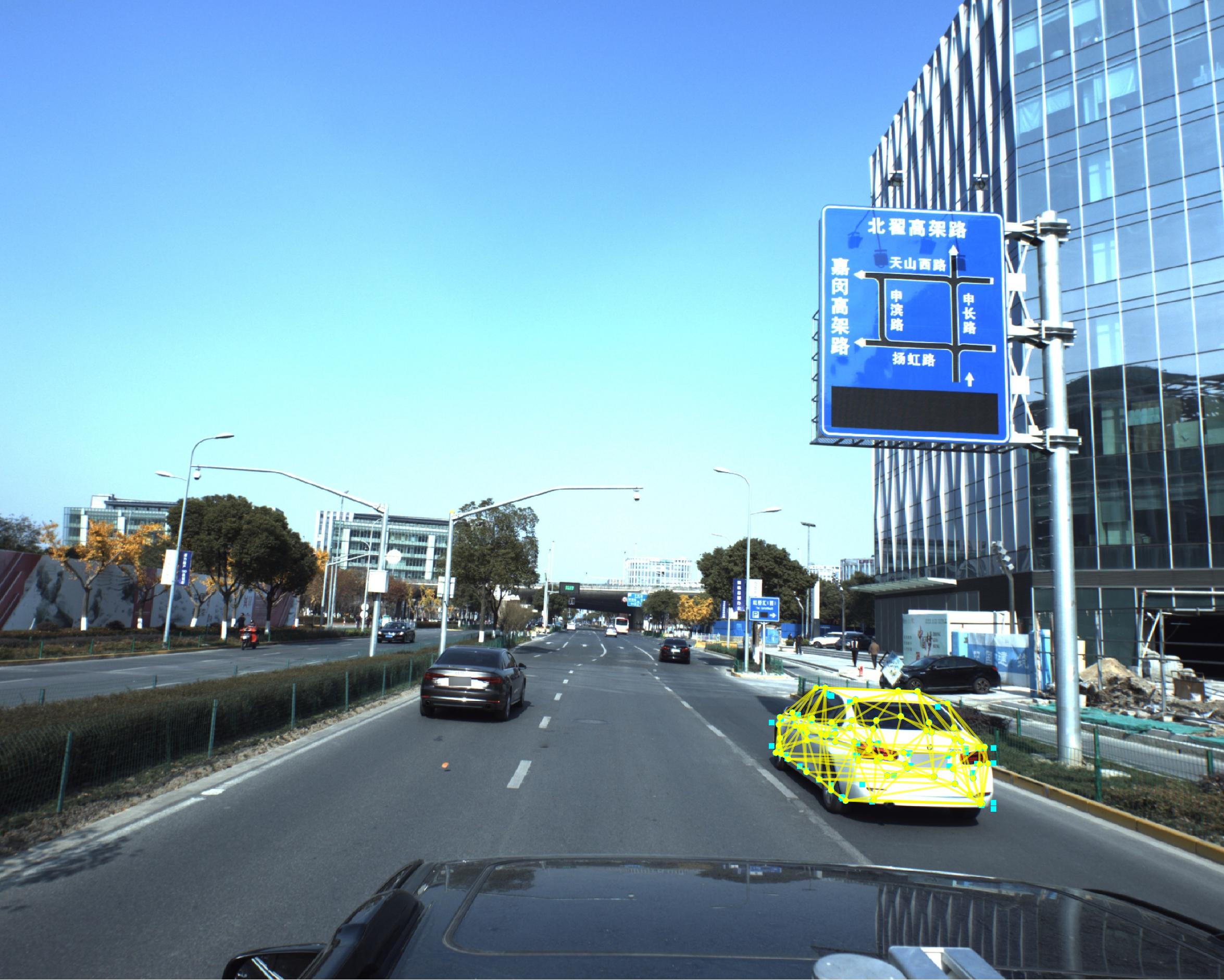}
			\end{minipage}
		&  \myhspace \hspace{-3mm}
			\begin{minipage}{\mpwfour}%
			\centering%
			\includegraphics[width=0.95\columnwidth]{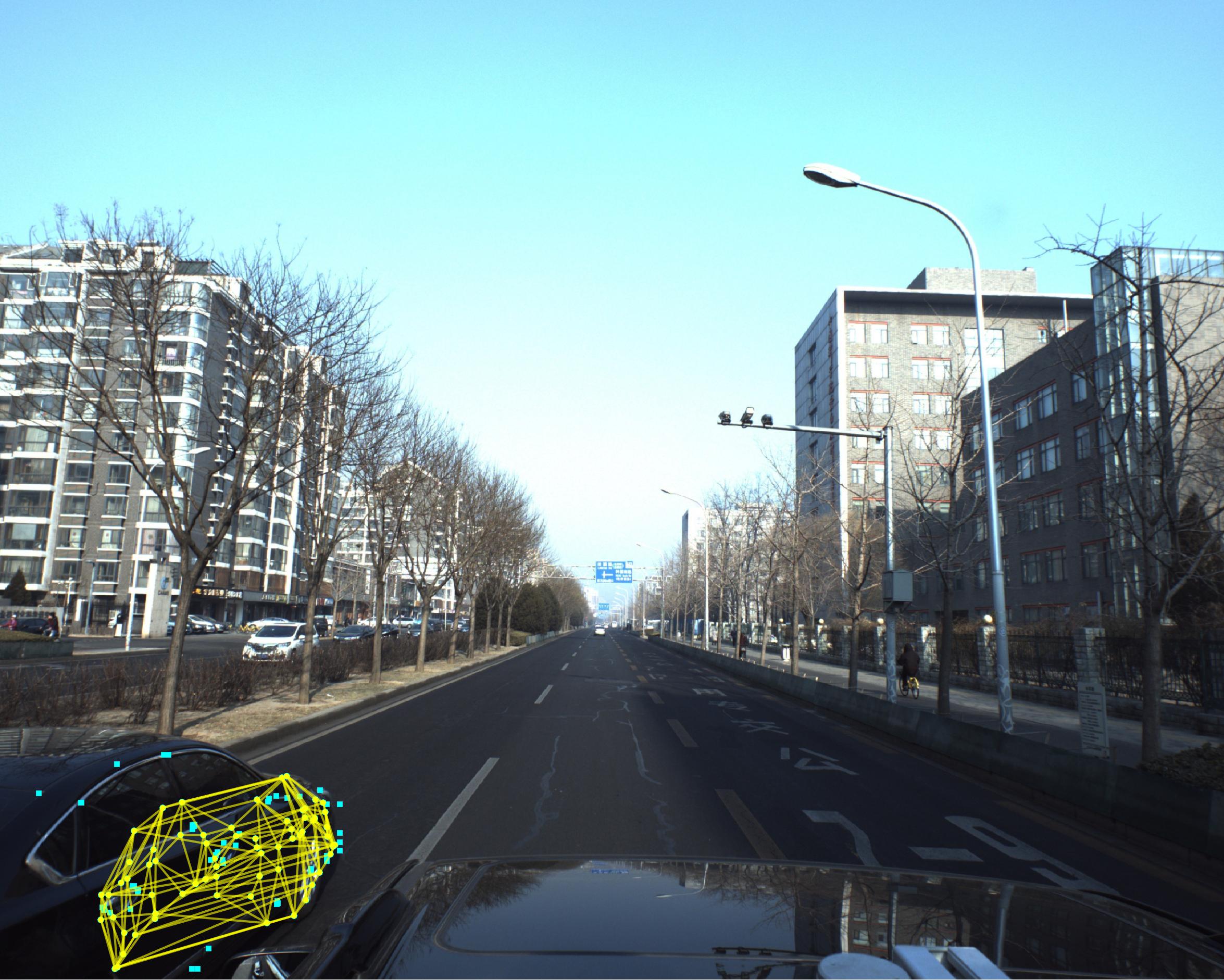}
			\end{minipage}
		&  \myhspace \hspace{-4mm}
			\begin{minipage}{\mpwfour}%
			\centering%
			\includegraphics[width=0.95\columnwidth]{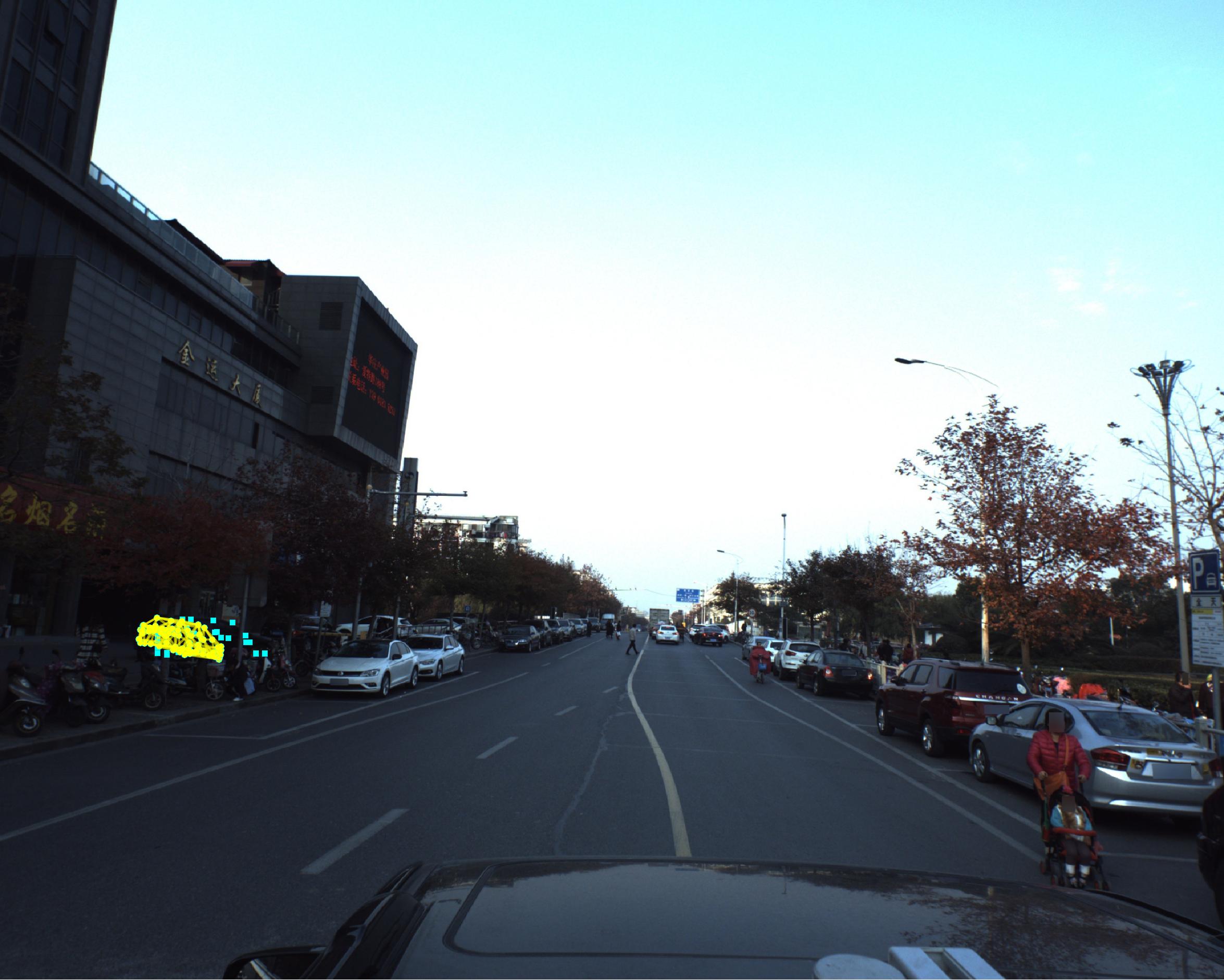}
			\end{minipage}
		\\
		   \myhspace \hspace{-4mm}
			\begin{minipage}{\mpwfour}%
			\centering%
			\includegraphics[width=0.95\columnwidth]{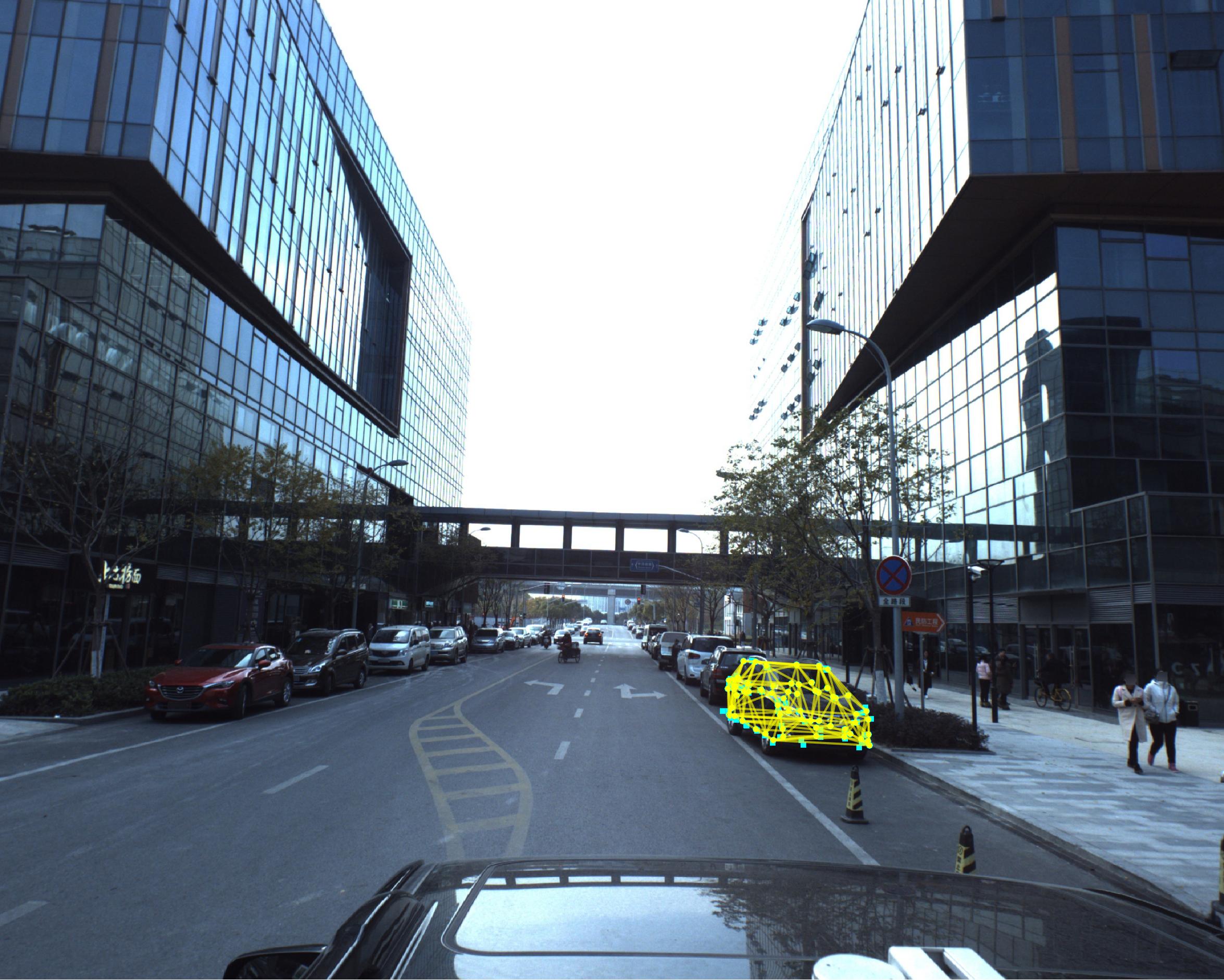}
			\end{minipage}
		&  \myhspace \hspace{-3mm}
			\begin{minipage}{\mpwfour}%
			\centering%
			\includegraphics[width=0.95\columnwidth]{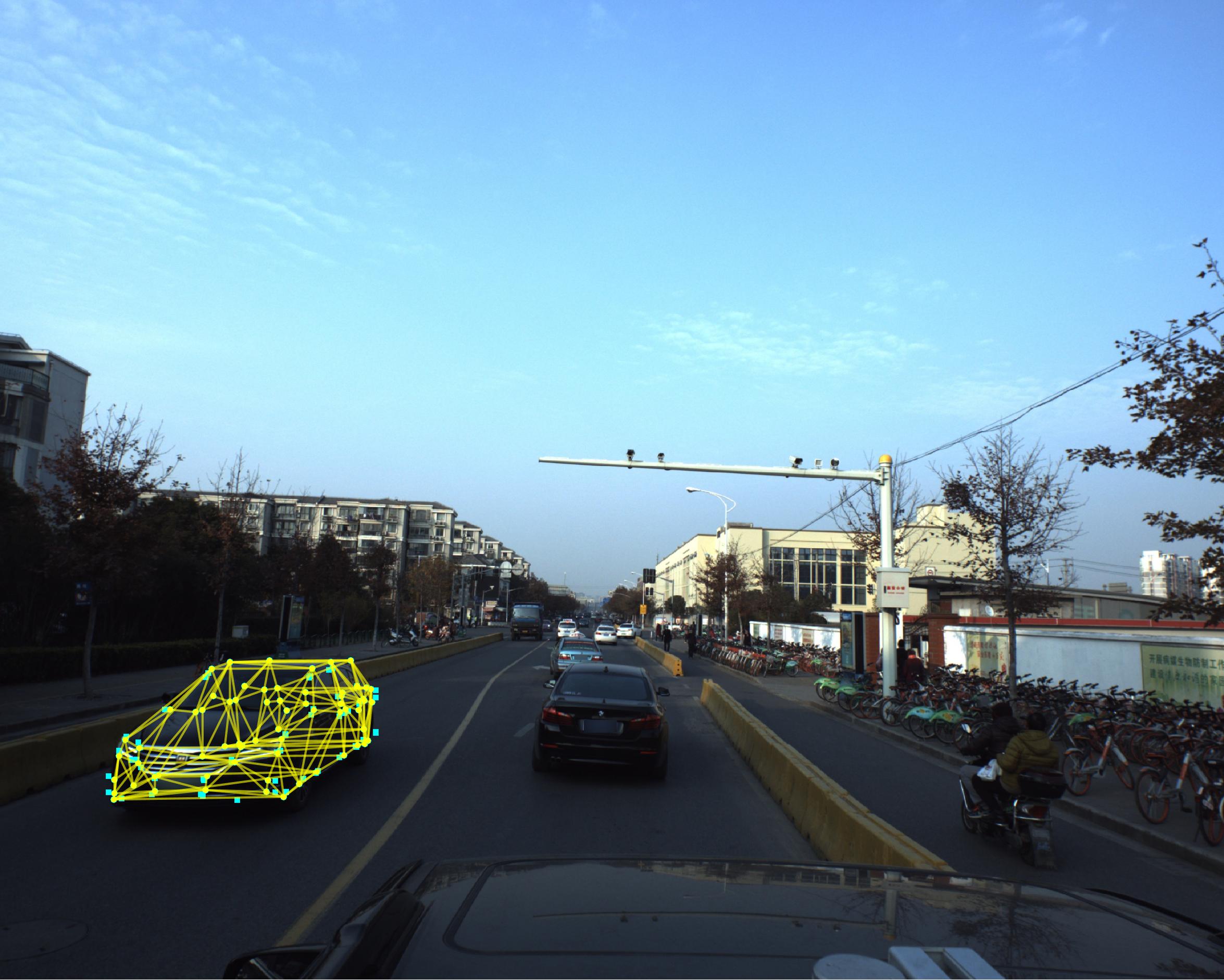}
			\end{minipage}
		&  \myhspace \hspace{-3mm}
			\begin{minipage}{\mpwfour}%
			\centering%
			\includegraphics[width=0.95\columnwidth]{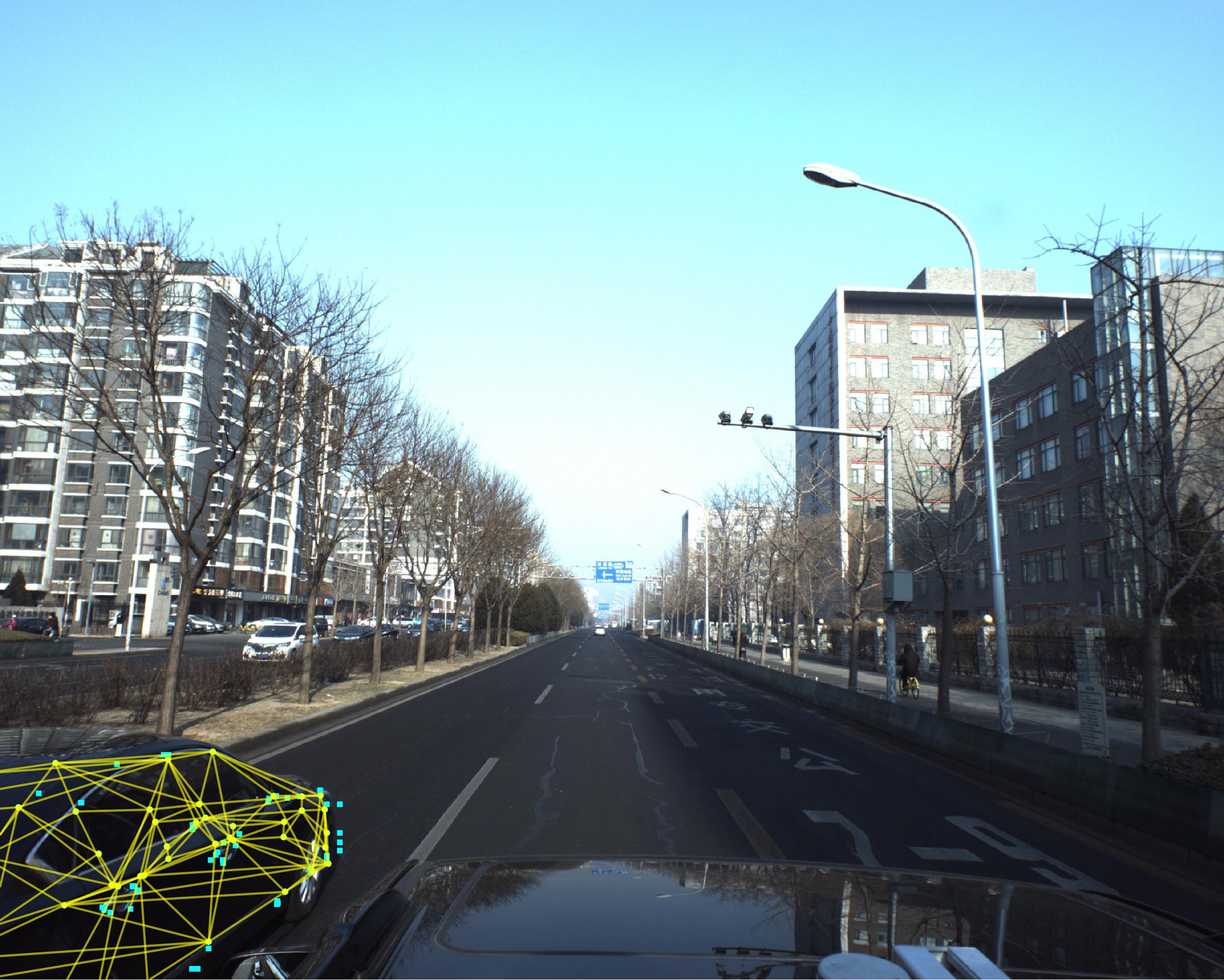}
			\end{minipage}
		&  \myhspace \hspace{-3mm}
			\begin{minipage}{\mpwfour}%
			\centering%
			\includegraphics[width=0.95\columnwidth]{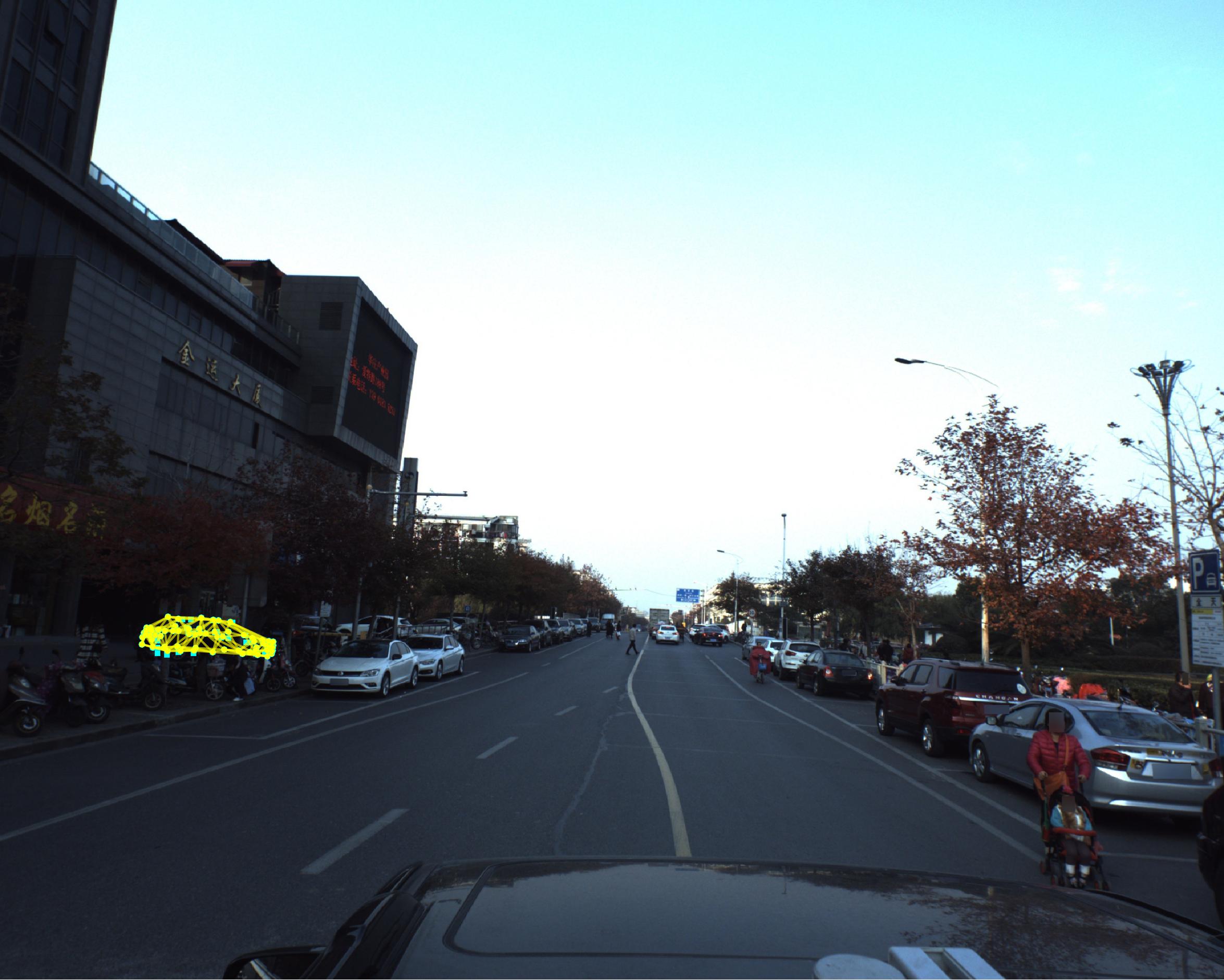}
			\end{minipage}
			\vspace{1mm}
		\\
		\multicolumn{2}{c}{\hspace{-8mm}\begin{minipage}{0.48\textwidth} \subcapsize (b-1) Vehicle pose and shape estimation on {\apollo} \cite{Wang19pami-apolloscape}. Four examples where globally optimal estimates returned by {\gnc} are \emph{certified} by {\stride}. \end{minipage} \vspace{-2mm}}
		&
		\multicolumn{2}{c}{\hspace{-8mm}\begin{minipage}{0.48\textwidth} \subcapsize (b-2) Two examples where {\gnc} converges to suboptimal estimates (top row), but {\stride} escapes the local minima and obtains \emph{certifiably optimal} estimates. \end{minipage}}
	\end{tabular}
	\end{minipage} 
	\caption{Category-level Object Perception (Example \ref{ex:category}). 
	\label{fig:exp-catreg-results}} 
	\vspace{-8mm} 
	\end{center}
\end{figure*}

{\bf Setup}. We use the ``\emph{car}'' category from {\pascal} dataset \cite{Xiang2014WACV-PASCAL+} for simulation experiments, which contains $N=12$ keypoints with $K=9$ basis shapes. We generate an unknown instance of the category by sampling a random vector of shape coefficients $\vcgt \in \bbR^K_{+}$ such that $\sum_{k=1}^K c_i^{\circ} = 1$ and using $\vcgt$ to linearly combine the $K$ basis shapes. We then add random Gaussian noise (with standard deviation $0.01$) to the new instance and transform it with a random rigid transformation $(\MRgt,\vtgt)$ with $\norm{\vtgt} \leq T = 10$. We test increasing outlier rates up to $60\%$ with 20 runs per outlier rate. We use a regularization parameter $\lambda = 1$.

{\bf Results}. Fig. \ref{fig:exp-catreg-results}(a) plots the evaluation metrics: (i) our relaxation is exact with up to $60\%$ outliers; (ii) {\stride} can can certify the global optimality of {\gnc} and escapes its local minima; (iii) {\stride} is about $10$ times faster than {\mosek}.

{\bf Vehicle pose and shape estimation on {\apollo}}. We use {\stride} to jointly estimate the pose and shape of an unknown vehicle from the {\apollo} self-driving dataset \cite{Wang19pami-apolloscape}. We use a set of $K=5$ basis shapes, each with $N=66$ annotated 3D keypoints. Given a 2D image depicting an unknown vehicle, we use the pretrained {\gsnet} \cite{Ke20-gsnet} to detect 2D keypoints of the unknown vehicle with groundtruth depth (same setup as one of the tests in \cite{Shi21rss-pace}). Fig. \ref{fig:exp-catreg-results}(b-1) shows four examples where {\stride} certified the global optimality of solutions returned by {\gnc} 
($\subopt = 1.5\ee{-7},1.3\ee{-9},1.4\ee{-10},1.6\ee{-9}$)
, and Fig. \ref{fig:exp-catreg-results}(b-2) shows two example where {\stride} escapes the suboptimal solutions returned by {\gnc} and finds the certified globally optimal solutions ($\subopt = 3.2\ee{-4},4.6\ee{-4}$). More examples are provided in {\supp}.

\subsection{Summary and Discussion}

Table \ref{table:overalltiming} summarizes the timing results of {\stride}, compared with {\mosek}, for all six problems. We make a few comments. (i) {\stride} is able to solve problems far beyond the reach of {\mosek} (in fact, the SDPs solved in this paper are among the largest in all semidefinite programming literature). (ii) When fast heuristics converge to the globally optimal solution, {\stride} just needs to perform optimality certification and can be $2$-$5$ times faster (\cf~{\stride} (Certify) vs. {\stride} (Escape)). (iii) For problems of similar sizes (in terms of $n_1$ and $m$), the speed of {\stride} can be \emph{application-dependent} (\eg~{\stride} is much faster in single rotation averaging than in other applications). This suggests that relaxations of different applications lead to SDP problems of \emph{drastically different geometry}. Understanding the geometry and leveraging new tools to further speedup the computation is an exciting research venue. For example, it could be promising to use data-driven methods to ``learn'' the geometry of different problems to generate high-quality initializations.


\begin{table*}[t]
\caption{Average timing of {\stride} and {\mosek} (in seconds). We omit {\sdpnal} because it failed to solve most of the problems. ``$**$'' indicates {\mosek} out of memory. ``N/A'' indicates {\gnc} or {\ransac} succeeded for all instances and there was no data for {\stride} escaping local minima. \label{table:overalltiming}}
\vspace{-3mm} 
\adjustbox{max width=\linewidth}{%
\begin{tabular}{c|cc|cc|cc|cc|cc|c}
\hline
 & \multicolumn{2}{c|}{Single Rotation Averaging} & \multicolumn{2}{c|}{Multiple Rotation Averaging} & \multicolumn{2}{c|}{Point Cloud Registration} & \multicolumn{2}{c|}{Mesh Registration} & \multicolumn{2}{c|}{Absolute Pose Estimation} & Category-level Perception \\
  & $N=30$ & $N=100$ & $10\times 10,N=10$ & $20 \times 20,N=20$ & $N=20$ & $N=100$ & $N=20$ & $N=100$ & $N=20$ & $N=100$ & $N=12,K=9$ \\
\hline
\hline
\eqref{eq:sparserelax} size ($n_1$) & $310$  & $1010$  & $343$ -- $549$ & $1483$ -- $1761$ & $273$ & $1313$ & $273$ & $1313$ & $273$ & $1313$ & $286$\\
\eqref{eq:sparserelax} size ($m$) & $30,016$ & $310,016$ & $3452$ -- $35,246$ & $64,384$ -- $189,198$ & $21,897$ & $485,417$ & $21,897$ & $485,417$ & $25,824$ & $572,984$ & $23,241$\\
\hline
\hline
  {\mosek} & $573$ & $**$ & $174$ & $**$ & $377$ & $**$ & $372$ & $**$ & $547$ & $**$ & $454$ \\
  {\stride} (Certify) & $11$ & $99$ & $17$ & $564$ & $78$ & $1675$ & $66$ & $1381$ & $211$ & $4168$ & $88$ \\
  {\stride} (Escape) & $15$ & $506$ & N/A  & N/A & $127$ & $4274$ & $163$ & $3136$ & N/A & N/A & $196$\\
\hline
\end{tabular}
}%
\vspace{-5mm}
\end{table*}

\section{Related Work}
\label{sec:relatedwork}
We review related works on outlier-free and outlier-robust geometric perception, while we refer the interested reader to \cite{Yang2015mpc-sdpnalplus,Wang21SIOPT-tssos} for recent progress in semidefinite programming and semidefinite relaxations.

{\bf Outlier-free geometric perception} algorithms can be divided into \emph{minimal solvers} and \emph{non-minimal solvers}. Minimal solvers assume \emph{noiseless} measurements (\ie~$r(\vxx,\vz_i)=0,\forall \; i$ in~\eqref{eq:robust}) and use the minimum number of measurements necessary to estimate $\vxx$, which typically
leads to solving a system of polynomial equations~\cite{Kukelova2008ECCV-automaticGeneratorofMinimalProblemSolvers}. 
Non-minimal solvers account for measurement noise and estimate $\vxx$ via nonlinear least squares (NLS), \ie~$\rho(r) = r^2/\beta_i^2$ in~\eqref{eq:robust}.
While in rare cases NLS can be solved in closed form~\cite{Horn87josa} or by solving the polynomial equations arising from the first-order optimality conditions~\cite{Kneip2014ECCV-UPnP}, in general 
they lead to
nonconvex problems and are attacked using local solvers~\cite{Schonberger16cvpr-SfMRevisited} 
or exponential-time methods (\eg \emph{Branch and Bound}~\cite{Olsson09pami-bnbRegistration}).

\emph{Certifiable algorithms} for outlier-free perception have recently emerged as an approach to compute globally optimal 
NLS
solutions in polynomial time.
These algorithms
relax 
the NLS minimization
into a convex optimization, using Lasserre's hierarchy of semidefinite relaxations for \emph{polynomial optimizations}~\cite{lasserre10book-momentsOpt,Kahl07IJCV-GlobalOptGeometricReconstruction}. By solving the SDP resulting from the convex relaxations, 
certifiable algorithms compute global solutions to NLS problems and provide a certificate of optimality, 
which usually depends on the rank of the SDP solution or the duality gap.
Empirically tight convex relaxations have been discovered in pose graph optimization~\cite{Carlone16TRO-planarPGO,Rosen19IJRR-sesync}, rotation averaging~\cite{Eriksson18cvpr-strongDuality,Fredriksson12accv}, triangulation~\cite{Aholt12eccv-qcqptriangulation}, 3D registration~\cite{Briales17cvpr-registration,Maron16tog-PMSDP,Chaudhury15Jopt-multiplePointCloudRegistration}, absolute pose estimation~\cite{Agostinho2019arXiv-cvxpnpl}, relative pose estimation~\cite{Briales18cvpr-global2view,Zhao20pami-relativepose}, hand-eye calibration~\cite{Heller14icra-handeyePOP} and shape and pose estimation from 2D or 3D landmarks~\cite{Yang20cvpr-perfectshape,Shi21rss-pace}. More recently, theoretical analysis of when and why the relaxations are tight is also emerging~\cite{Carlone15icra-verification,Aholt12eccv-qcqptriangulation,Eriksson18cvpr-strongDuality,Rosen19IJRR-sesync,Cifuentes17arxiv,Zhao20pami-relativepose,Chaudhury15Jopt-multiplePointCloudRegistration,Dym17Jopt-exactPMSDP,Iglesias20cvpr-PSRGlobalOptimality,Eriksson19pami-rotavgstrongduality}.
Tight relaxations also enable 
optimality certification (\ie checking if a given solution is optimal), which ---in outlier-free perception--- can sometimes be performed in closed form~\cite{Carlone16TRO-planarPGO,Eriksson18cvpr-strongDuality,Garcia21IVC-certifiablerelativepose,Boumal16nips,Burer03mp,Rosen20wafr-scalableLowRankSDP,Iglesias20cvpr-PSRGlobalOptimality}. {Despite being certifiably optimal, these solvers assume all measurements are inliers (\ie~have small noise), which rarely occurs in practice, and hence give poor estimates even in the presence of a single outlier. 

{\bf Outlier-robust geometric perception} algorithms can be divided into \emph{fast heuristics} and \emph{globally optimal solvers}. Two general frameworks for designing fast heuristics are \ransac~\cite{Fischler81} and \emph{graduated non-convexity} (\gnc)~\cite{Black96ijcv-unification,Yang20ral-gnc,Antonante20TRO-outlier}. {\ransac} robustifies minimal solvers and acts as a fast heuristics to solve \emph{consensus maximization}~\cite{Chin17slcv-maximumConsensusAdvances}, while {\gnc} robustifies non-minimal solvers and acts as a fast heuristics to solve \emph{M-estimation} (\ie~using a robust cost function $\rho$ in~\eqref{eq:robust}). Local optimization 
is also a popular fast heuristics~\cite{Schonberger16cvpr-SfMRevisited,Agarwal13icra} for the case where an initial guess is available. Approximate but deterministic algorithms have also been designed to solve consensus maximization \cite{Le19pami-deterministicApproximateMC}. 
On the other hand, globally optimal solvers are typically designed using Branch and Bound~\cite{Bazin12accv-globalRotSearch,Bustos18pami-GORE,Izatt17isrr-MIPregistration,Yang2014ECCV-optimalEssentialEstimationBnBConsensusMax,Paudel15iccv-robustSOS,Li09cvpr-robustFitting,Chin15cvpr-CMTreeAstar,Li07iccv-3DRegistration}. 

\emph{Certifiable outlier-robust algorithms} relax problem~\eqref{eq:robust} with a robust cost into a tight convex optimization. 
While certain robust costs, such as L1~\cite{Wang13ima} and Huber~\cite{Carlone18ral-robustPGO2D}, are already convex, they have low breakdown points (\ie 
they can be compromised by a single outlier~\cite{Maronna19book-robustStats}). Problem-specific certifiably robust algorithms have been proposed 
to deal with high-breakdown-point formulations, such as the TLS cost~\cite{Yang19rss-teaser,Yang19iccv-quasar,Lajoie19ral-DCGM}. 
\maybeOmit{Even optimality certification becomes harder and problem-specific in the presence of outliers, due to the lack of a closed-form characterization of the 
dual variables~\cite{Yang20tro-teaser}.}

\section{Conclusions}
\label{sec:conclusion}

We presented the first general {and \practical} framework to design certifiable algorithms for outlier-robust geometric perception. We first showed that estimation with several common robust cost functions can be reformulated as polynomial optimization problems. We then designed a  semidefinite relaxation scheme that exploits the sparsity of outlier-robust estimation to generate SDPs of much smaller sizes {while maintaining empirical exactness}. Finally, we proposed a robust and scalable SDP solver, {\stride}, that can solve the sparse relaxations to unprecedented scale and accuracy. We tested our framework on six geometric perception applications using both synthetic and real data, demonstrating its robustness, scalability, and capability to safeguard existing fast heuristics for robust estimation.
\section*{Acknowledgments}
The authors would like to thank Jie Wang, Victor Magron, and Jean B. Lasserre for the discussion about Lasserre's hierarchy and {\tssos}; Ling Liang and Kim-Chuan Toh for the discussion about SDP solvers; Bo Chen and Tat-Jun Chin for the {\speed} data; and Jingnan Shi for the {\apollo} data.

This work was funded by ARL DCIST CRA W911NF-17-2-0181,
ONR RAIDER N00014-18-1-2828, MathWorks, NSF CAREER award ``Certifiable Perception for Autonomous Cyber-Physical Systems'', and Lincoln Laboratory's Resilient Perception in Degraded Environments program.

\clearpage
\begin{center}
{\bf \Large Supplementary Material}
\end{center}

\renewcommand{\thesection}{A\arabic{section}}
\renewcommand{\theequation}{A\arabic{equation}}
\renewcommand{\thealgocf}{A\arabic{algocf}}
\renewcommand{\thetheorem}{A\arabic{theorem}}
\renewcommand{\thefigure}{A\arabic{figure}}
\renewcommand{\thetable}{A\arabic{table}}
\setcounter{equation}{0}
\setcounter{section}{0}
\setcounter{theorem}{0}
\setcounter{figure}{0}


\section{Proof of Proposition~\ref{prop:robustaspop}} 
\label{sec:app-proof-robust-pop}
\begin{proof}
We first prove (i)-(iv) using Black-Rangarajan duality \cite{Black96ijcv-unification}, and then (v)-(vii) by manipulating the cost functions.

{\bf Proof of (i)-(iv)}.
The TLS proof has been given in \eqref{eq:binaryTLS} of the main text. We start with (ii) MC. With a similar strategy of introducing a binary variable as in \eqref{eq:binaryTLS}, we can write the MC cost function as
\bea\label{eq:mcidentity}
\rho_{\mmc} \equiv \min_{\theta \in \{+1,-1\}} \cbrace{ \frac{1-\theta}{2} \ \middle\vert\ -\theta(r^2 - \beta_i^2) \geq 0 },
\eea
where the constraint $-\theta(r^2 - \beta_i^2) \geq 0$ enforces $\theta = -1$ if $r^2 > \beta_i^2$ (hence $\rho_{\mmc} = 1$), and $\theta = +1$ if $r^2 < \beta_i^2$ (hence $\rho_{\mmc} = 0$). If $r^2 = \beta_i^2$, then the minimization selects $\theta = +1$ as it minimizes the objective to be zero. Using the identity in \eqref{eq:mcidentity}, problem \eqref{eq:robust} with $\rho = \rho_{\mmc}$ is equivalent to
\begin{equation}\label{eq:dualMC}
\hspace{-4mm}
\min_{\substack{\vxx \in \calX \subseteq \Real{d}, \\ \vtheta \in \{\pm 1 \}^N}} \!\! \cbrace{ \sum_{i=1}^N\! \frac{1\!-\!\theta_i}{2}\! +\! \regularizer \middle\vert \substack{ \displaystyle -\theta_i (r^2(\vxx,\vz_i)\! -\! \beta_i^2)\! \geq\! 0,\! \\ \displaystyle \forall i=1,\dots,N} }\!,\!\!\! \tag{MC}
\end{equation}
which is an instance of \eqref{eq:pop} in $(\vxx,\vtheta)\in \Real{d+N}$.

To prove (iii), we leverage Black-Rangarajan duality \cite[Fig. 28]{Black96ijcv-unification} and write $\rho_{\mgm}$ as a minimization problem by introducing a confidence variable $w \in [0,1]$
\bea\label{eq:GMidentity}
\rho_{\mgm} \equiv \min_{w \in [0,1]} w \frac{r^2}{\beta_i^2} + (\sqrt{w}-1)^2.
\eea
One can check the correctness of \eqref{eq:GMidentity} by setting the derivative of the objective in \eqref{eq:GMidentity} \wrt $w$ to zero and obtain $w$ as a function of $r$ in closed form. Eq.~\eqref{eq:GMidentity}, however, does not directly lead to a POP due to the existence of $\sqrt{w}$. Nevertheless, with a change of variable $\theta := \sqrt{w} \in [0,1]$, we can write problem \eqref{eq:robust} with $\rho = \rho_{\mgm}$ as the following POP
\begin{equation}\label{eq:dualGM}
\min_{\substack{\vxx \in \calX \subseteq \Real{d}, \\ \vtheta \in [0,1]^N}} \sum_{i=1}^N \frac{\theta_i^2 r^2(\vxx,\vz_i)}{\beta_i^2} + (\theta_i - 1)^2 + \regularizer. \tag{GM}
\end{equation} 

Similarly, we can use Black-Rangarajan duality \cite[Fig. 25]{Black96ijcv-unification} to prove (iv) by introducing a confidence variable $w$ and writing $\rho_{\mtb}$ as the solution of the following minimization
\bea
\rho_{\mtb} \equiv \min_{w \in [0,1]} w \frac{r^2}{\beta_i^2} + \frac{1}{3} - w + \frac{2}{3} w^{\frac{3}{2}}.
\eea
Then, with a change of variable $\theta := \sqrt{w}$, we conclude that \eqref{eq:robust} with $\rho = \rho_{\mtb}$ can be written as the following POP
\begin{equation}\label{eq:dualTB}
\min_{\substack{\vxx \in \calX \subseteq \Real{d}, \\ \vtheta \in [0,1]^N}} \sum_{i=1}^N \frac{\theta_i^2 r^2(\vxx,\vz_i)}{\beta_i^2} + \frac{1}{3} - \theta_i^2 + \frac{2}{3} \theta_i^3 + \regularizer. \tag{TB}
\end{equation}
In \eqref{eq:binaryTLS} and \eqref{eq:dualMC}, $\theta_i$ is binary and discrete, with $\theta_i= +1$ (resp. $\theta_i = -1$) indicating the $i$-the measurement $\vz_i$ is an inlier (resp. outlier). While in \eqref{eq:dualGM} and \eqref{eq:dualTB}, $\theta_i$ is continuous, with $\theta_i \uparrow 1$ (resp. $\theta_i \downarrow 0$) indicating the $i$-the measurement $\vz_i$ is an inlier (resp. outlier). 

{\bf Proof of (v)-(vii)}. The L1 cost function can be simply rewritten as
\bea
\rho_{\mlone} \equiv \cbrace{ \frac{\gamma}{\beta_i}\ \middle\vert\ \gamma \geq 0, \gamma^2 = r^2 },
\eea
where $\gamma \geq 0$ and $\gamma^2 = r^2$ implies that $\gamma = \abs{r}$. Therefore, \eqref{eq:robust} with $\rho = \rho_{\mlone}$ is equivalent to the following POP:
\begin{equation}
\hspace{-5mm}\min_{\substack{\vxx \in \calX \subseteq \Real{d},\\ \vgamma \in \Real{N}} } \cbrace{ \sum_{i=1}^N \frac{\gamma_i}{\beta_i} \ \middle\vert\ \gamma_i \geq 0, \gamma_i^2 = r^2(\vxx,\vz_i),i=1,\dots,N}\!\!.\!\!\!\tag{L1}
\end{equation}

Now we prove \eqref{eq:robust} with the Huber loss \cite{Huber81} can also be written as a POP. We first perform a change of variable and let $\gamma = \abs{r}$ (which is equal to $\gamma \geq 0, \gamma^2 = r^2$):
\bea \label{eq:huberafterabs}
\rho_{\mhuber} = \begin{cases}
\frac{\gamma^2}{2\beta_i^2} & \gamma \leq \beta_i \\
\frac{\gamma}{\beta_i} - \half & \text{otherwise}
\end{cases}.
\eea
Then we introduce a binary variable $\theta \in \{ +1, -1\}$, and equivalently write \eqref{eq:huberafterabs} as
\bea
\rho_{\mhuber} = \cbrace{ \frac{1+\theta}{2} \frac{\gamma^2}{2\beta_i^2} + \frac{1-\theta}{2} \parentheses{\frac{\gamma}{\beta_i} - \half} \middle\vert \theta (\gamma - \beta_i) \leq 0 },
\eea
where the constraint $\theta (\gamma - \beta_i) \leq 0$ enforces $\theta = -1$ when $\gamma > \beta_i$ (hence $\rho_{\mhuber} = \frac{\gamma}{\beta_i} - \half$), $\theta = +1$ when $\gamma < \beta_i$ (hence $\rho_{\mhuber} = \frac{\gamma^2}{2\beta_i^2}$). Therefore, we can write \eqref{eq:robust} with $\rho = \rho_{\mhuber}$ as the following POP:
\begin{equation}
\hspace{-6mm}\min_{\substack{\vxx \in \calX, \vgamma \in \Real{N}, \\ \vtheta \in \{ \pm 1\}^N} } \cbrace{ \sum_{i=1}^N \frac{1+\theta_i}{2} \frac{\gamma_i^2}{2\beta_i^2} + \frac{1-\theta_i}{2} \parentheses{\frac{\gamma_i}{\beta_i} - \half}  \middle\vert \substack{ \gamma_i \geq 0, \\ \gamma_i^2 = r^2(\vxx,\vz_i^2), \\ \theta_i (\gamma_i - \beta_i) \leq 0,\\ i=1,\dots,N  }}\!\!. \tag{HB}
\end{equation}

Finally, we prove \eqref{eq:robust} with the adaptive cost function $\rho_{\madaptive,s}$, proposed by Barron \cite{Barron19cvpr-adaptRobustLoss}, can also be written as a POP, when we restrict the scale parameter $s$ to be rational numbers (we avoid $s = 0$ and $s=2$ because the cost function is not defined at those two parameters, and \cite{Barron19cvpr-adaptRobustLoss} augments the cost by taking its limits at $s = 0$ and $s=2$). Note that restricting $s$ to rational numbers preserves the expressiveness of the original adaptive cost in \cite{Barron19cvpr-adaptRobustLoss}, because the set of rational numbers is \emph{dense} in the set of real numbers. Because $s$ is a rational number, we can let $s = \frac{p}{q}$ with $p,q$ as integers, and write the adaptive cost as
\bea
\rho_{\madaptive,s} = \frac{\abs{s-2}}{s} \parentheses{ \parentheses{ \frac{r^2/\beta_i^2}{\abs{s-2}} + 1  }^{\frac{p}{2q}} - 1  }.
\eea
Now we perform a change of variable and let $\gamma = \parentheses{ \frac{r^2/\beta_i^2}{\abs{s-2}} + 1  }^{\frac{p}{2q}}$. This change of variable is equivalent to the following polynomial equality constraint:
\bea
0 = h(\gamma,r^2) := 
\begin{cases}
\gamma^{2q} - \parentheses{ \frac{r^2/\beta_i^2}{\abs{s-2}} + 1  }^p & p > 0\\
\gamma^{2q} \parentheses{ \frac{r^2/\beta_i^2}{\abs{s-2}} + 1  }^{\abs{p}} - 1 & p < 0
\end{cases}.
\eea
Therefore, we conclude that \eqref{eq:robust} with $\rho = \rho_{\madaptive,s}$ can be written as the following POP:
\begin{equation}
\hspace{-3mm} \min_{\substack{\vxx \in \calX, \\ \vgamma \in \Real{N}}} \cbrace{ \sum_{i=1}^N \frac{\abs{s-2}}{s} (\gamma_i - 1)\ \middle\vert\ h(\gamma_i,r^2(\vxx,\vz_i)) = 0,i=1,\dots,N}. \tag{ADT}
\end{equation}
This concludes the proof for all seven cost functions.
\end{proof}

\section{Proof of Proposition~\ref{prop:polynomialExpressibility}} 
\label{sec:app-proof-polynomial-express}

\begin{proof} (i) can be proved by inspection: all the $r^2$ and $\psi$ in Examples \ref{ex:singlerotation}-\ref{ex:category} are squared norms of \emph{affine} (degree-one) polynomials in $\vxx$, and are naturally quadratic.

To show (ii), we note that the $q$-dimensional ball $\ball^q_T$ can be described by a single quadratic inequality $\ball^q_T = \{\vt \in \Real{q} \mid T^2 - \inprod{\vt}{\vt} \geq 0 \}$, the 3D FOV cone $\calC_\alpha$ can be described by two inequalities $\calC_{\alpha} = \{ \vt \in \Real{3} \mid \tan^2(\frac{\alpha}{2}) t_3^2 - t_1^2 - t_2^2 \geq 0, t_3 \geq 0 \}$, where the first inequality is quadratic and the second is affine. Now it remains to show that 2D and 3D rotations can be described by polynomial equalities. First, any 2D rotation $\MR \in \SOtwo$ can be equivalently parametrized by
\bea\label{eq:2Drotationparam}
\MR = \cbrace{ \bmat{cc} r_1 & -r_2 \\ r_2 & r_1 \emat \ \middle\vert\ \vr \in \Real{2}, \inprod{\vr}{\vr} = 1 },
\eea
and hence described by a single quadratic equality. For a 3D rotation $\MR \in \SOthree$, we shorthand $\vr_i$ as its $i$-th column, and $\vr = [\vr_1 \vcat \vr_2 \vcat \vr_3] \in \Real{9}$ as its vectorization. Using the results from \cite{Briales17cvpr-registration,Yang20cvpr-perfectshape,Tron15RSSW-rotationdeterminant}, we know that $\MR \in \SOthree$ can be equivalently described by the following set of $15$ quadratic equality constraints
\begin{subequations}
\bea
\hspace{-8mm} \text{\grayout{Unit norm}} : &  h_{i} = 1 - \inprod{\vr_i}{\vr_i}, i=1,2,3, \label{eq:3Drotationunit} \\
\hspace{-8mm} \text{\grayout{Orthogonal}} : & h_{i,j} = \inprod{\vr_i}{\vr_j}, \parentheses {\substack{i \\j } } \in \{ \parentheses {\substack{1 \\2 } },\parentheses {\substack{2 \\3 } },\parentheses {\substack{3 \\1 } } \}, \label{eq:3Drotationorthogonal} \\
\hspace{-8mm} \text{\grayout{Right-hand}}: & \hspace{-3mm} h_{i,j,k}\! =\! \vr_i\! \times\! \vr_j\! -\! \vr_k,\! \parentheses {\substack{i \\j \\k } }\! \in\! \cbrace{\! \parentheses {\substack{1 \\2 \\3 } }\! ,\! \parentheses {\substack{2 \\3 \\1 } }\! ,\! \parentheses {\substack{3 \\1 \\2 } }\!}\!,\label{eq:3Drotationrighthand}
\eea
\end{subequations}
where ``$\times$'' denotes the vector cross product, and each $h_{i,j,k}$ defines a vector of $3$ equality constraints. Though the set of $15$ equalities is redundant (\eg \eqref{eq:3Drotationunit} and \eqref{eq:3Drotationrighthand} are sufficient for $\MR \in \SOthree$), we use all of them to enhance robustness and tightness of the relaxation in Section \ref{sec:sdprelax}.
\end{proof}
\section{Proof of Theorem~\ref{thm:sparserelaxtls}}
\label{sec:app-proof-theorem-sparserelax}

\begin{proof}
(i): Every $\tldvxx=(\vxx,\vtheta) \in \calX \times \{\pm 1\}^N$ of \eqref{eq:binaryTLS} leads to a rank-one lifting $\vv(\tldvxx)\vv(\tldvxx)\tran$ that is feasible for \eqref{eq:sparserelax}. Therefore, the feasible set of \eqref{eq:binaryTLS} is a subset of the feasible set of \eqref{eq:sparserelax}, and hence $\fstar \leq \pstar$.

\revise{
(ii) \& (iii) Since $\tldvxxstar$ is optimal for \eqref{eq:binaryTLS}, we have $p(\tldvxxstar) = \pstar$. Now because $\MX_v = \vv(\tldvxxstar)\vv(\tldvxxstar)\tran$ is a rank-one lifting of $\tldvxxstar$, we have that $\MX_v$ is feasible for the SDP \eqref{eq:sparserelax} and it attains $f(\MX_v) = \pstar = \fstar$. Therefore $\MX_v$ (and its corresponding localizing matrices $\MX_1,\dots,\MX_{l_g}$) are optimal for \eqref{eq:sparserelax}.

Now we prove that if an optimal SDP solution $\MXstar_v$ has rank one, then the relaxation is exact and a global optimizer can be extracted from $\MXstar_v$. Towards this goal, first observe that since $\rank{\MXstar_v} = 1$, $\MXstar_v \succeq 0$, and $[\MXstar_v]_{11} = 1$ (because $\MXstar_v$ is a feasible point of \eqref{eq:sparserelax}, which requires the leading entry to be one. \cf \eqref{eq:momentConstraintssparse}), we can perform a rank-one factorization of $\MXstar_v$ as
\bea
\MXstar_v = \bmat{c}
1 \\
\lowvxx \\
\lowvtheta \\
\lowxtheta
\emat
\bmat{cccc}
1 & \lowvxx\tran & \lowvtheta\tran & \lowxtheta\tran\emat \\
= \bmat{cccc}
1 & \lowvxx\tran & \lowvtheta\tran & \lowxtheta\tran \\
\lowvxx & \lowvxx\lowvxx\tran & \lowvxx \lowvtheta\tran & \lowvxx \lowxtheta\tran \\
\lowvtheta & \lowvtheta \lowvxx\tran & \lowvtheta \lowvtheta\tran & \lowvtheta \lowxtheta\tran \\
\lowxtheta & \lowxtheta \lowvxx\tran & \lowxtheta \lowvtheta\tran & \lowxtheta \lowxtheta\tran
\emat, \label{eq:expandrankone}
\eea
where $\lowvxx \in \Real{d}$, $\lowvtheta \in \Real{N}$, $\lowxtheta \in \Real{dN}$ and they correspond to the partition in the ``$\vv$'' monomial basis in \eqref{eq:sparsebasis} (note that here we overload the ``$\widehat{\cdot}$'' symbol only in the context of this proof). 

Now we first show that $\lowxtheta = \lowvxx \kron \lowvtheta$, \ie $\lowxtheta$ are second-order monomials in $\lowvxx$ and $\lowvtheta$ of the form $[\lowvxx]_i [\lowvtheta]_j$ for $1\leq i \leq d$ and $1\leq j \leq N$. This is evident when we observe that, asking $\MXstar_v$ to be a moment matrix (\ie enforcing the moment constraints in \eqref{eq:momentConstraintssparse}) requires that the block $\lowxtheta$ in \eqref{eq:expandrankone} is just a rearrangement of the entries of the block $\lowvxx\lowvtheta\tran$ in \eqref{eq:expandrankone}, where the latter block contains all the second-order monomials of the form $[\lowvxx]_i [\lowvtheta]_j$. 

Then we show that $\lowvxx \in \calX$ and $\lowvtheta \in \{+1,-1\}^N$, \ie $\lowvxx$ and $\lowvtheta$ are indeed \emph{feasible} points to the original \eqref{eq:binaryTLS} problem. This is equivalent to showing that all equality constraints hold: $h_i(\lowvxx,\lowvtheta) = 0,\forall i = 1,\dots,l_h$, and all inequality constraints hold: $g_j(\lowvxx,\lowvtheta) \geq 0,j=1,\dots,l_g$. This follows from the fact that (a) each $h_i(\lowvxx,\lowvtheta) = 0$ is enforced by one of the redundant constraints in \eqref{eq:redundantEqualityConstraintssparse}, and (b) each $g_j(\lowvxx,\lowvtheta) \geq 0$ is enforced by the one of the localizing constraints in \eqref{eq:localizingConstraintssparse}. 

At this point, we have shown that $(\lowvxx,\lowvtheta)$ is a feasible point of \eqref{eq:binaryTLS} that attains $p(\lowvxx,\lowvtheta) = f(\MXstar_v) = \fstar$. However, we know that $p(\lowvxx,\lowvtheta) \geq \pstar$ by the nature of the minimization problem \eqref{eq:binaryTLS}. Therefore, we have
\bea
\pstar \leq p(\lowvxx,\lowvtheta) = f(\MXstar_v) = \fstar, \nonumber
\eea
but $\fstar \leq \pstar$ by construction of the semidefinite relaxation. Hence $\pstar = \fstar$ and $(\lowvxx,\lowvtheta)$ is globally optimal for \eqref{eq:binaryTLS}.
} 
\end{proof}
\section{Relative Suboptimality}
\label{sec:app-compute-subopt}

This section is concerned with the computation of a formally correct suboptimality gap $\subopt$
for a given estimate (which we use as a performance metric in our experiments), whose validity is not hindered by potential numerical inaccuracies in the solution of the SDP relaxation~\eqref{eq:sparserelax}.

In Theorem \ref{thm:sparserelaxtls} and \eqref{eq:subopt}, we stated that, by solving the sparse SDP relaxation \eqref{eq:sparserelax} to global optimality with optimizer $\MXstar$ and associated optimum $\fstar$, one can round from $\MXstar$ a feasible solution $(\hatvxx_1,\hatvtheta_1)$ to the original \eqref{eq:binaryTLS} problem with associated cost $\hatp = p(\hatvxx_1,\hatvtheta_1)$. Then, a measure of suboptimality for the rounded solution $(\hatvxx_1,\hatvtheta_1)$ can be computed as follows (also in \eqref{eq:subopt}):
\bea \label{eq:app-subopt}
\subopt \triangleq \frac{\abs{\fstar - \hatp}}{1 + \abs{\fstar} + \abs{\hatp}}.
\eea
It is apparent that $\subopt = 0$ implies the relaxation \eqref{eq:sparserelax} is exact and the rounded solution $(\hatvxx_1,\hatvtheta_1)$ is indeed globally optimal for \eqref{eq:binaryTLS} (recall $\fstar \leq \pstar \leq \hatp$ by construction, where $\pstar$ is the unknown optimum of the nonconvex \eqref{eq:binaryTLS} problem).

However, the caveat in computing the relative suboptimality as in \eqref{eq:app-subopt} is that, although it is almost always possible to compute a rounded solution $(\hatvxx_1,\hatvtheta_1)$ with cost $\hatp$ (provided that the feasible set $\calX$ of \eqref{eq:binaryTLS} is simple to project, as in our examples), it can be quite challenging to obtain $\fstar$ (which acts as a valid lower bound for $\pstar$) \emph{to machine precision}, since $\fstar$ is computed by numerically solving the SDP~\eqref{eq:sparserelax}, which may still lead to small inaccuracies. Moreover, as shown in the experimental section of the main text, first-order solvers such as {\sdpnal} typically cannot solve the SDP to even moderate accuracy (with reasonable amount of iterations), in which case $\fstar$ in not attained.

Here we describe a procedure to compute a valid lower bound for $\pstar$, from any approximate solution $(\hatMX,\hatvy,\hatMS) \in \bbX \times \Real{m} \times \bbX$ of the SDP \eqref{eq:sparserelax}, without requiring it to be an optimal solution satisfying the KKT conditions \eqref{eq:sdpKKT}. In fact, as we will show soon, only $\hatvy \in \Real{m}$ is needed to compute a valid lower bound.

{\bf Bounded trace}. Let us first shows that, each block of the primal variable $\MX$ in \eqref{eq:sparserelax} has a bounded trace, when \eqref{eq:sparserelax} is applied to all six Examples \ref{ex:singlerotation}-\ref{ex:category}. Towards this goal, let us first observe that the variable $\vxx \in \calX$ has a bounded norm, \ie there exists $M_0>0$ such that $\norm{\vxx} \leq M_0, \forall \vxx \in \calX$. For example, in Example \ref{ex:singlerotation}, $\vxx = \MR \in \SOthree$ has $\norm{\vxx} = \sqrt{3} = M_0$; in Example \ref{ex:category}, $\vxx = (\MR,\vt,\vc)$ with $\MR \in \SOthree$, $\vt \in \ball^3_{T_t}$, and $\vc \in \ball^K_{T_c}$ has $\norm{\vxx} \leq \sqrt{3+T_t^2 + T_c^2} = M_0$, where $T_t$ and $T_c$ are the upper bounds for the norm of the translation and the norm of the shape parameters, respectively.

Now recall that the primal variable $\MX$ has $1+l_g$ blocks, with the first block being the moment matrix and the other $l_g$ blocks being localizing matrices. With the observation that $\norm{\vxx} \leq M_0, \forall \vxx \in \calX$, we can bound the trace of the moment matrix $\MX_v$ (\cf \eqref{eq:sparsemomentmat}) as
\bea \label{eq:traceboundmomentmat}
\trace{\MX_v} = & \trace{\vv(\tldvxx) \vv(\tldvxx)\tran} = \vv(\tldvxx)\tran \vv(\tldvxx) \nonumber \\
= & 1 + \norm{\vxx}^2 + \sum_{i=1}^N \theta_i^2 + \sum_{i=1}^N \theta_i^2 \norm{\vxx}^2 \nonumber \\
= & (1+N)(1+\norm{\vxx}^2) \nonumber \\
\leq &  (1+N)(1+M_0^2) =: M_1 , \nonumber \\
&  \forall \vxx \in \calX, \vtheta \in \{\pm 1\}^N.
\eea

Regarding the localizing matrices $\MX_{g_j} = g_j \cdot [\MX_1]_{\calI_j},j=1,\dots,l_g$ (where $\MX_1$ is the order-one moment matrix), we have that (recall $g_j \geq 0$ by definition) 
\bea \label{eq:traceboundlocalizemat}
\trace{\MX_{g_j}} = & g_j \cdot \trace{[\MX_1]_{\calI_j}} \nonumber \\
\leq &   g_j \cdot \trace{\MX_1} \nonumber \\
= & g_j \cdot (1+ \norm{\vxx}^2 + \sum_{i=1}^N \theta_i^2 ) \nonumber \\
\leq & g_j \cdot (1+N+M_0^2), \nonumber \\
& \forall \vxx \in \calX, \vtheta \in \{ \pm 1 \}^N.
\eea
Therefore, it suffices to show that $g_j$ is upper bounded for any $\vxx \in \calX$. This is obvious for all the examples in this paper. Particularly, there are only two types of inequality constraints among Examples \ref{ex:singlerotation}-\ref{ex:category}. (i) The ball constraint $\vt \in \ball^K_T$ (bounded translation and bounded shape parameters), which reads $g = T^2 - \norm{\vt}^2 \geq 0$, which certainly satisfies $g \leq T^2$ and is upper bounded. (ii) The camera FOV cone constraint $\vt \in \calC_\alpha$ that induces two inequality constraints $g_1 = t_3 \geq 0$ and $g_2 = \tan^2(\alpha/2) t_3^2 - t_1^2 - t_2^2 \geq 0$. However, since the translation also lies in the bounded ball $\ball^3_T$, we have $g_1 = t_3 \leq \norm{\vt} \leq T$, and $g_2 = \tan^2(\alpha/2) t_3^2 - t_1^2 - t_2^2 \leq \tan^2(\alpha/2) t_3^2 \leq \tan^2(\alpha/2) \norm{\vt}^2 \leq \tan^2(\alpha/2) T^2$ are both upper bounded. Therefore, we have shown that each localizing matrix also has bounded trace.

{\bf A valid lower bound}. Now suppose we are given a $\hatvy \in \Real{m}$, then for any $\vxx \in \calX, \vtheta \in \{ \pm 1\}^N$, we have
\bea
& p(\vxx,\vtheta) \nonumber\\
= & \inprod{\MC}{\MX} \nonumber \\
= & \inprod{\MC - \calAadj (\hatvy) }{\MX} + \inprod{\calAadj (\hatvy)}{\MX} \nonumber \\
= & \inprod{\MC - \calAadj (\hatvy) }{\MX} + \inprod{\calA (\MX)}{\hatvy} \nonumber\\
= & \inprod{\MC - \calAadj (\hatvy) }{\MX} + \inprod{\vb}{\hatvy} \label{eq:app-lower-bound-Xfeasible} \\
\geq &\!\!\!\!\! \underbrace{ \inprod{\vb}{\hatvy} + \displaystyle \sum_{i=1}^{1+l_g} M_i \cdot \min\{ \lambda_{\min}\parentheses{ [\MC - \calAadj (\hatvy)]_i },0\} }_{\plb(\hatvy)}, \label{eq:app-lower-bound-tracemineig}
\eea
where $M_i,i=1,\dots,1+l_g$ are the upper bounds for the traces of the moment matrix and the localizing matrices (shown in previous paragraphs and \eqref{eq:traceboundmomentmat}-\eqref{eq:traceboundlocalizemat}), $[\MC - \calAadj (\hatvy)]_i$ denotes the $i$-th block of $[\MC - \calAadj (\hatvy)]$ (recall that both $\MC$ and $\calAadj (\hatvy)$ are multi-block symmetric matrices, \cf \eqref{eq:adjointAmultiblk}), and $\lambda_{\min}(\cdot)$ denotes the minimum eigenvalue of a symmetric matrix. In \eqref{eq:app-lower-bound-Xfeasible}, we used that any $\MX$ that comes from moment matrix and localizing matrices must be primal feasible and hence $\calA(\MX) = \vb$. In \eqref{eq:app-lower-bound-tracemineig}, we used that $\inprod{\MA}{\MB} \geq \lambda_{\min}(\MA) \trace{\MB}$ for any $\MA\in \sym{n}$ and $\MB \in \psd{n}$. With this lower bound $\plb(\hatvy)$, we can compute the relative suboptimality from any $\hatvy \in \Real{m}$:
\bea
\subopt \triangleq \frac{\abs{\plb(\hatvy) - \hatp}}{1 + \abs{\plb(\hatvy)} + \abs{\hatp}}.
\eea

\section{Further Reduction on Multiple Rotation Averaging (Example \ref{ex:multirotation})} 
\label{sec:app-reduce-mra}
Recall that in multiple rotation averaging we are given a graph $\calG = (\calV,\calE)$ with vertex set $\calV = [n]$ and edge set $\calE$. Each vertex $i \in \calV$ is associated with an unknown rotation $\MR_i \in \mySOd$, and each edge $(i,j) \in \calE$ provides a relative measurement $\tldMR_{ij}$ between the unknown rotations $\MR_i$ and $\MR_j$ at vertices $i$ and $j$. Let $\calR$ be the set of edges whose relative measurements are known to be free of outliers (\eg odometry measurements in SLAM), and let $\calZ = \calE / \calR$ be the set of edges whose measurements are corrupted by outliers (\eg loop closures in SLAM). If no edge set is known to be free of outliers, then we set $\calR = \emptyset$.

We now present a further reduction for multiple rotation averaging. Let us denote by $\calV_{\calZ} \triangleq \{ i \in \calV \mid \exists j \in \calV, (i,j) \in \calZ  \} \subseteq \calV$ the subset of nodes that are attached to at least one edge in $\calZ$. Note that typically $\abs{\calV_\calZ} \ll n$ for SLAM applications (\ie these are the nodes at which loop closures occur). For each edge $(i,j) \in \calZ$, we define its \emph{depth-$\zeta$ neighbor set} for $\zeta \in \nnint$, in the following recursive manner:
\bea
\calV_{(i,j)}^0 \triangleq \{ i,j \}, \calV_{(i,j)}^{\zeta} \triangleq \{i \in \calV \mid \exists j \in \calV_{(i,j)}^{\zeta-1}, (i,j) \in \calE \},
\eea
where one can see that $\calV_{(i,j)}^{\zeta}$ (for $\zeta \geq 1$) is essentially the union of the $\zeta$-hop neighbor set of node $i$ with the $\zeta$-hop neighbor set of node $j$.
It is easy to see that $\calV_{(i,j)}^\zeta = \calV, \forall (i,j) \in \calZ$ when $\zeta$ is sufficiently large, as long as the graph $\calG$ is connected. With $\calV_\calZ$ and $\calV_{(i,j)}^\zeta$, for each edge $(i,j) \in \calZ$, we define 
\bea
\vxx_{(i,j)}^\zeta \triangleq \{\MR_k \mid k \in \calV_{(i,j)}^\zeta \cap \calV_\calZ \} \supseteq \{\MR_i, \MR_j \},
\eea
as the set of node-wise rotations in $\calV_\calZ$ that are attached to $(i,j)$ within depth $\zeta$. By definition, $\calV_{(i,j)}^\zeta \cap \calV_\calZ$ must contain node $i$ and node $j$, and hence $\vxx_{(i,j)}^\zeta$ contains at least two rotations (attached to the edge $(i,j)$). We now replace the sparse basis in \eqref{eq:sparsebasis} as
\bea
& \vv(\tldvxx) = [1 \vcat \vxx \vcat \vtheta \vcat \dots \theta_{ij} \vxx_{(i,j)}^\zeta \dots  ]_{(i,j) \in \calZ} \in \Real{n_1}, \nonumber \\
& 1+2n+5N \leq n_1 \leq 1+2n+ N(1+ 2\abs{\calV_\calZ}),
\eea
and use it to generate the semidefinite relaxation \eqref{eq:sparserelax}. It is worth noting that our relaxation recovers the hand-crafted SDP relaxation in \cite{Lajoie19ral-DCGM} with the choice of $\zeta = 0$, which is shown to be \emph{inexact} when the outlier rate is around $50\%$. In Section \ref{sec:experiments}, we show that, with a larger $\zeta$, we can achieve exact relaxation in the presence of over $70\%$ outliers.
\section{Solving the Projection Subproblem}
\label{sec:supp-projsubproblem}
In this section, we describe how to solve the projection subproblem in {\stride} (\cf \eqref{eq:pgd} and \eqref{eq:strideprojection} in Algorithm \ref{alg-iPGMnlp}). In particular, we show that the dual problem of \eqref{eq:pgd} admits an unconstrained formulation, which allows developing a scalable algorithm based on limited-memory BFGS.

Recall that the projection step \eqref{eq:strideprojection} of {\strideplus} seeks to compute the projection of a given point onto the spectrahedron $\setsdpp = \{ \MX \in \bbX \mid \calA(\MX)=\vb, \MX \in \calK \}$.
Formally, given a point $\MZ \in \bbX$, the projection problem seeks the closest point in $\setsdpp$ \wrt $\MZ$
\begin{equation}
	\label{prob:projection}
	\min_{\MX \in \bbX}\left\{ \half \left\lVert \MX - \MZ \right\rVert^2\ \middle\vert\ \MX\in \setsdpp \right\}.
\end{equation} 
Since $\setsdpp$ is the intersection of two convex sets, namely the hyperplane defined by $\calA(\MX)=\vb$ and the (product of) positive semidefinite cone $\calK$, a natural idea is to apply Dykstra’s projection algorithm (see \eg~\cite{Combettes11book-proximalSplitting}) to generate an approximate solution by alternating the projection onto the hyperplane and the projection onto the semidefinite cone, both of which are easy to compute. However, Dykstra’s projection is known to have slow convergence and it may take too many iterations until a satisfactory projection is found. As a result,
instead of solving \eqref{prob:projection} directly, we consider its dual problem
\begin{equation}
	\label{prob:projection-dual}
	\min_{\vy \in \Real{m}, \MS \in \bbX}\left\{ \half\left\lVert \MS + \calAadj (\vy) + \MZ \right\rVert^2 - \inprod{\vb}{\vy}\ \middle\vert\ \MS\in \calK \right\},
\end{equation}
where we have ignored the constant term $ -\half\left\lVert \MZ \right\rVert^2 $ and converted ``$\max$'' to ``$\min$'' by changing the sign of the objective. 
The KKT conditions for the pair \eqref{prob:projection} and \eqref{prob:projection-dual} are:
\begin{equation}
	\label{eq:KKT-proj}
	\hspace{-3mm}
	\calA(\MX) = \vb,\ \calAadj (\vy) + \MS = \MX - \MZ,\  \MX,\MS\in \calK, \  \inprod{\MX}{\MS} = 0.\!\!
\end{equation}

{\bf An unconstrained formulation}. Now we introduce a key observation that allows us to further simplify the dual \eqref{prob:projection-dual}. Fixing the unconstrained $\vy$, problem \eqref{prob:projection-dual} can be seen as finding the closest $\MS \in \calK$ \wrt the matrix $-\calAadj (\vy) - \MZ$, and hence admits a closed-form solution
\bea\label{eq:Sofy}
\MS 	= \Pi_{\calK} \parentheses{-\calAadj (\vy) - \MZ}.
\eea 
As a result, after inserting~\eqref{eq:Sofy}, problem~\eqref{prob:projection-dual} is equivalent to
\begin{equation}
	\label{prob:projetion-dual-y}
	\min_{\vy \in \Real{m}} \ \ \phi(\vy):= \frac{1}{2}\left\lVert \Pi_{\calK}(\calAadj (\vy) + \MZ) \right\rVert^2 - \left\langle \vb, \vy \right \rangle,
\end{equation}
with the gradient of $ \phi(\vy) $ given as 
\bea
\nabla \phi(\vy) = \calA \Pi_{\calK}(\calAadj (\vy) + \MZ) - \vb.
\eea
Thus, if $ \vystar $ is an optimal solution for problem~\eqref{prob:projetion-dual-y}, we can recover $\MSstar$ from~\eqref{eq:Sofy}, and $\MXstar$ from the KKT conditions \eqref{eq:KKT-proj}: 
\bea
\MXstar = \calAadj (\vystar) + \MSstar + \MZ. 
\eea
Formulating the dual problem as the unconstrained problem \eqref{prob:projetion-dual-y} has appeared multiple times in~\cite{Zhao10siopt-sdpnal,Malick09siopt-regularizationSDP}.

Now that~\eqref{prob:projetion-dual-y} is a \emph{smooth unconstrained convex} problem in $ \vy\in \Real{m} $, plenty of efficient algorithms are available, such as (accelerated) gradient descend~\cite{Nesterov18book-convexOptimization}, nonlinear conjugate gradient~\cite{Dai99siopt-ncg}, quasi-Newton methods~\cite{Nocedal06book-numericaloptimization} and the semismooth Newton method~\cite{Zhao10siopt-sdpnal}. In this paper, we apply the celebrated limited-memory BFGS (\lbfgs) method, see for example \cite[Algorithm 7.5]{Nocedal06book-numericaloptimization}. {\lbfgs} is easy to implement, can handle very large unconstrained optimization problems due to its low memory consumption, and is typically ``the algorithm of choice'' for large-scale problems~\cite[Chapter 7]{Nocedal06book-numericaloptimization}. Empirically, we observed that {\lbfgs} is efficient and robust for various applications. To the best of our knowledge, this is the first work that demonstrates the effectiveness of \lbfgs, or in general quasi-Newton methods, in solving large-scale and degenerate SDPs.

\section{Proof of Theorem~\ref{thm:strideconverge}}
\begin{proof}
Let $\calV = \{ \hatMX^{k(i)} \}$ be the sequence of all the $\hatMX$ that have been accepted due to \eqref{eq:accept-reject}, where $k(i)$ returns the iteration index of the $i$-th element in $\calV$. If $\calV = \emptyset$, then {\strideplus} reduces to \eqref{eq:pgd} and is globally convergent. If $\calV \neq \emptyset$, then we claim that $\calV$ must be finite. Note that, for any two consecutive elements $\hatMX^{k(i)}$ and $\hatMX^{k(i+1)}$ in $\calV$, we have 
\bea \label{eq:strictdescent}
f(\hatMX^{k(i+1)}) \leq& f(\barMX^{k(i+1)}) - \epsilon \nonumber \\
<&  f(\MX^{k(i+1) - 1}) - \epsilon  \leq f(\hatMX^{k(i)}) - \epsilon,
\eea
where the first inequality is due to \eqref{eq:accept-reject}, the second inequality is due to \eqref{eq:strideprojection} and the fact that projected gradient descent must strictly decrease the objective value when optimality has not been achieved \cite[Proposition 3.4.1]{Bertsekas99book-nlp}, and the last inequality holds because $k(i+1) - 1 \geq k(i)$. Eq.~\eqref{eq:strictdescent} states that the objective value must decrease by at least $\epsilon$ along each element of $\calV$. Therefore, we have $f_\min(\calV) \leq f_{\max}(\calV) - (|\calV|-1) \epsilon$, where $f_{\min}$ and $f_{\max}$ are the minimum and maximum objective values along $\calV$. Hence $|\calV|$ must be finite, otherwise $\fstar$ is unbounded below, contradicting Slater's condition and strong duality. Let $\hatMX^{k(|\calV|)}$ be the last element of $\calV$, then {\strideplus} reduces to \eqref{eq:pgd} with a new initial point at $\hatMX^{k(|\calV|)}$ and is globally convergent.
\end{proof}
\section{Implementation Details for {\stride}}
In Section \ref{sec:scalableopt}, we presented the {\strideplus} algorithm and proved its global convergence. We noted that the initial point $(\MX^0,\vy^0,\MS^0)$ could have a significant impact on the convergence speed of {\strideplus}. Therefore, in {\stride} we use existing fast heuristics ({\gnc}, {\ransac}) to generate a \emph{primal} initial guess (\cf Remark \ref{remark:fastheuristics}). In this section, we describe how to generate a \emph{dual} initial guess (Section \ref{sec:app-dual-warmstart}), and how to use Riemannian optimization for local search (Section \ref{sec:app-local-search}).

\subsection{Dual Warmstart} 
\label{sec:app-dual-warmstart}

We propose to use a combination of two techniques to generate a good dual initial point $(\vy^0,\MS^0)$. Section \ref{sec:app-cssr} describes a method to relax the \eqref{eq:binaryTLS} problem by exploiting correlative sparsity. Although such a relaxation is not tight, we show that its solution can be used to warmstart {\stride}. In Section \ref{sec:app-admmplus}, we present a fast first-order algorithm to refine both the primal and the dual initializations.

\subsubsection{Bootstrapping via Correlative Sparsity}
\label{sec:app-cssr}
The \eqref{eq:binaryTLS} problem has another special property called \emph{correlative sparsity} \cite{Wang21siopt-chordaltssos,Waki06jopt-SOSSparsity,Lasserre06siopt-correlativesparsity}, which, loosely speaking, refers to the property that there exists a partition of the variables $(\vxx,\vtheta)$ into a union of smaller groups, such that (i) each constraint of \eqref{eq:binaryTLS} involves only one group of the variables, and (ii) the objective of \eqref{eq:binaryTLS} can be decomposed into terms that each involves only one group of the variables (\cf \cite[Assumption 2]{Lasserre06siopt-correlativesparsity}).
Particularly, we observe that the objective polynomial, denoted by $p(\vxx,\vtheta)$, can be expressed as a sum of $N$ polynomials:
\bea
p(\vxx,\vtheta) = \sum_{i=1}^N \underbrace{ \parentheses{  \frac{1+\theta_i}{2} \frac{r^2(\vxx,\vz_i)}{\beta_i^2} + \frac{1-\theta_i}{2} + \frac{1}{N} \regularizer} }_{p_i(\vxx,\theta_i)} ,
\eea
where each $p_i$ is a polynomial that only involves $\tldvxx_i \triangleq [\vxx\vcat \theta_i] \in \Real{d+1}$. The constraint polynomials can also be partitioned into $N$ groups where the $i$-th group of constraints only involves $\tldvxx_i$. To see this, note that there are two types of constraints in \eqref{eq:binaryTLS}, the ones that constrain $\vxx$ (to be proper rotations and translations), denoted by $\calH[\vxx]$, and the ones that constrain each $\theta_i$ to be a binary variable, denoted by $\calH[\theta_i] = \{ \theta_i^2 -1 = 0\},i=1,\dots,N$. Therefore, defining $\calH_i \triangleq \{ \calH[\vxx],\calH[\theta_i] \}$, then each $\calH_i$ only contains polynomials in $\tldvxx_i$, and the union of $\calH_i$ for $i=1,\dots,N$ is the full constraint set of \eqref{eq:binaryTLS}. 
This correlative sparsity allows us to design an SDP relaxation for \eqref{eq:binaryTLS} using $N$ moment matrices $\MX_{v_i}, i=1,\dots,N$, where each $\MX_{v_i}$ is defined as
\bea
& \hspace{-3mm} \vv_i(\tldvxx_i) \triangleq [1\vcat \vxx \vcat \theta_i \vcat \theta_i \vxx] \in \Real{2d + 2}, \label{eq:csliftingmonomial}\\
& \hspace{-10mm}\MX_{v_i}\! \triangleq\! \vv_i(\tldvxx_i) \vv_i (\tldvxx_i)\tran\! =\! 
\bmat{cccc}
1 & \vxx\tran & \theta_i & \theta_i \vxx\tran \\
\vxx & \vxx \vxx\tran & \theta_i \vxx & \theta_i \vxx\vxx\tran \\
\theta_i & \theta_i \vxx\tran & \theta_i^2 & \theta_i^2 \vxx\tran \\
\theta_i \vxx & \theta_i \vxx\vxx\tran & \theta_i^2 \vxx & \theta_i^2 \vxx\vxx\tran
\emat \label{eq:csmomentmat}
\eea
and has a \emph{constant} size $2d+2$. It is easy to verify that $\MX_{v_i}$ contains all the monomials in $p_i(\tldvxx_i)$ and $\calH_i$. Therefore, by following similar steps as in the main text, we can derive an SDP relaxation that exploits correlative sparsity.

\emph{(i) Rewriting \eqref{eq:binaryTLS} using the moment matrices $\{ \MX_{v_i} \}_{i=1}^N$}. Because the sparse moment matrix $\MX_{v_i}$ contains all monomials in $p_i$, and the \eqref{eq:binaryTLS} cost is a sum of $p_i$'s, we can write the objective of \eqref{eq:binaryTLS} as a linear functions of $\{ \MX_{v_i} \}_{i=1}^N$:
\beal\label{eq:objectivesparsecs}
\!\!\!\!\!\!\text{\grayout{objective}}: & \sum_{i=1}^N \inprod{\MC_i}{\MX_{v_i}}.
\eeal

\emph{(ii) Relaxing the rank-$1$ constraint on $\{\MX_{v_i}\}_{i=1}^N$}. By construction, $\MX_{v_i}$ belongs to the set of rank-one positive semidefinite matrices. Since the rank constraint is non-convex, we drop it and only enforce each $\MX_{v_i}$ to be positive semidefinite:
\beal \label{eq:eqMomentIsPSDsparsecs}
\!\!\!\text{\grayout{moment matrices}}: & \MX_{v_i} \succeq 0, i=1,\dots,N. \\
\eeal

\emph{(iii) Adding redundant constraints}. Now we add moment constraints to each moment matrix $\MX_{v_i}$ and use the set of constraints $\calH_i$ to add redundant equality and localizing constraints for $\MX_{v_i}$. Because this procedure is the same for each moment matrix $\MX_{v_i}$, we will only describe it once for a fixed $i$. First, some monomials can repeat themselves at multiple entries of $\MX_{v_i}$. For example, in \eqref{eq:csmomentmat} the ``$\theta_i \vxx$'' block is the same as the ``$\theta_i \vxx\tran$'' block up to rearrangement of entries. In fact, the number of \emph{unique} monomials in $\MX_{v_i}$ is $m_{2v_i} = 3\trinum(d+1)$, while the dimension of $\MX_{v_i}$ (in terms of a symmetric matrix) is $\trinum(2d+2)$. Therefore, we can add a total number of $\mmomi = \trinum(2d+2) - m_{2v_i} + 1$ \emph{moment constraints}:
\beal\label{eq:momentConstraintssparsecs}
\text{\grayout{moment constraints}}:
& \inprod{\MA_{\mathmom,0}}{\MX_{v_i}} = 1, \\
 & \inprod{\MA_{\mathmom,j}}{\MX_{v_i}} = 0, \\
&  j = 1, \ldots, \mmomi-1,
\eeal
to enforce the repeating monomials in $\MX_{v_i}$ to be equal to each other, as well as the leading entry $[\MX_{v_i}]_{11} = 1$.

Second, we add redundant equality constraints. For each equality constraint $h_k$ in $\calH_i$, we denote $[\tldvxx_i]_{h_k}$ as the maximum set of unique monomials such that $h_k \cdot [\tldvxx_i]_{h_k}$ only contains monomials in $\MX_{v_i}$. Formally,
\bea
[\tldvxx_i]_{h_k} \triangleq \{\tldvxx_i^{\valpha} \mid \mono{ h_k \cdot \tldvxx_i^{\valpha} } \subseteq \mono{\MX_{v_i}} \}. \label{eq:csliftequalities}
\eea
Consequently, we can write $h_k \cdot [\tldvxx_i]_{h_k} = \zero$ as linear equalities in $\MX_{v_i}$:
\beal\label{eq:redundantEqualityConstraintssparsecs}  
\hspace{-3mm}\!\!\!\text{\grayout{(redundant) equality constraints}}: & \!\!\! \inprod{\MA_{\mathreq,kj}}{\MX_{v_i}} = 0, \\
\!\!\!&\!\!\!  k = 1, \ldots, l_{h_i}\\
\!\!\!&\!\!\!  j = 1, \ldots, \abs{[\tldvxx_i]_{h_k}},\!\!\!\!\!\!\!\!\!
\eeal 
where $l_{h_i}$ is the number of equality constraints in $\calH_i$.

Finally, for each inequality constraint $g_j$ in $\calH_i$ ($\deg{g_j} \leq 2$ by Proposition \ref{prop:polynomialExpressibility}), we denote by $[\MX_1]_{\calI_j}$ the maximum principal submatrix of $\MX_1$ (\ie order-one full moment matrix) such that $g_j \cdot [\MX_1]_{\calI_j}$ only contains monomials in $\MX_{v_i}$. Formally,
\bea
& [\MX_1]_{\calI_j} \triangleq  [\MX_1]_{\calI_j,\calI_j}, \text{ with } \nonumber \\
&  \hspace{-9mm} \calI_j \! =\!  \displaystyle \argmax_{\calJ} \{ \abs{\calJ} \mid \mono{ g_j\! \cdot\! [\MX_1]_{\calJ,\calJ} } \subseteq \mono{\MX_{v_i}} \}.
\eea
As a result, calling $\MX_{g_j} = g_j \cdot [\MX_1]_{\calI_j}$, which is positive semidefinite by construction, we can write down the following localizing matrices and constraints:
\beal\label{eq:cslocalizemat}
\text{\grayout{localizing matrices}}: & \MX_{g_j} \succeq 0, \;\; j=1,\ldots,l_{g_i}
\eeal 
\beal\label{eq:cslocalizecons}
\!\!\!\!\!\!\text{\grayout{{localizing} constraints}}: \!\!\!& \!\!\!\inprod{\MA_{\mathloc,jkh}}{\MX_{v_i}} = [\MX_{g_j}]_{hk} \\
\!\!\!\!\!\!&\!\!\!   j = 1, \ldots, l_{g_i}, \\
\!\!\!\!\!\!&\!\!\! 1 \leq h\leq k \leq \abs{\calI_j},
\eeal
where the linear constraints simply enforce each entry of $\MX_{g_j}$ to be a linear combination of entries in $\MX_{v_i}$, and $l_{g_i}$ is the number of inequality constraints in $\calH_i$.

\emph{(iv) Adding overlapping constraints}. The extra step that needs to be performed when there are multiple moment matrices is to add constraints that enforce \emph{overlapping entries} to be the same. Clearly, from \eqref{eq:csmomentmat}, one can see that the top left $2 \times 2$ blocks, \ie $[1\vcat \vxx] [1,\vxx\tran]$ is shared among $\MX_{v_i}$ for all $i=1,\dots,N$. Therefore, we add the following overlapping constraints
\beal\label{eq:csoverlapcons}
\hspace{-3mm}\!\!\!\text{\grayout{overlapping constraints}}: & \!\!\! [\MX_{v_i}]_{\mathovlp} = [\MX_{v_1}]_{\mathovlp}, \\
\!\!\!&\!\!\!  i = 2, \dots, N,
\eeal 
where $[\MX_{v_i}]_{\mathovlp}$ refers to the top-left $2\times 2$ blocks of $\MX_{v_i}$.

Steps (i)-(iv) above lead to the following SDP:
\begin{equation}\label{eq:correlativerelax}
\begin{split}
\hspace{-3mm} \min_{\MX} \cbrace{ \sum_{i=1}^N \inprod{\MC_i}{\MX_{v_i}}\ \middle\vert\ \calA(\MX)\! =\! \vb, \MX \succeq 0} \\
\text{with }\MX = \parentheses{ 
\begin{array}{c} \MX_{v_1},\MX_{1,1},\dots,\MX_{1, l_{g_1}} \\
\MX_{v_2},\MX_{2,1},\dots,\MX_{2, l_{g_2}} \\
\vdots \\
\MX_{v_N},\MX_{N,1},\dots,\MX_{N, l_{g_N}}
\end{array}},
\end{split} 
\tag{CSSR}
\end{equation}
where we have shorthanded $\MX_{i,j}$ as the $j$-th localizing matrix for the $i$-th moment matrix for notation convenience (\cf \eqref{eq:cslocalizemat}), and $\calA(\MX)=\vb$ collects all the linear equality constraints from \eqref{eq:momentConstraintssparsecs}, \eqref{eq:redundantEqualityConstraintssparsecs}, \eqref{eq:cslocalizecons}, and \eqref{eq:csoverlapcons}.

Comparing \eqref{eq:correlativerelax} with \eqref{eq:sparserelax}, we see that, although \eqref{eq:correlativerelax} has more positive semidefinite blocks than \eqref{eq:sparserelax}, the size of the blocks become much smaller, especially when $N$ is large (\eqref{eq:correlativerelax} has $n_1 = 2d+2$, while \eqref{eq:sparserelax} has $n_1 = (1+d)(1+N)$). Therefore, \eqref{eq:correlativerelax} can be solved much more efficiently using off-the-shelf interior point methods such as \mosek \cite{mosek}. However, the caveat is that \eqref{eq:correlativerelax} is not tight 
and cannot provide a certifiably optimal solution to the original \eqref{eq:binaryTLS} problem. 

{\bf Assembling a dual initialization for {\stride}}. Although the \eqref{eq:correlativerelax} relaxation is inexact, it is still useful to solve it because we can use its solution to warmstart {\strideplus}. To do this, let us recall the block structure of \eqref{eq:sparserelax} for the primal variable:
\bea
\MX = (\MX_v, \MX_1,\dots,\MX_{l_g}).
\eea
The dual variable $\MS$ has the same block structure:
\bea
\MS = (\MS_v, \MS_1,\dots,\MS_{l_g}),
\eea 
where each block of $\MS$ has the same size as the corresponding block of $\MX$. With a slight change of notation, let us rewrite the block structure of \eqref{eq:correlativerelax} as:
\bea
\MX_c = \parentheses{ 
\begin{array}{c} \MX_{v_1},\MX_{1,1},\dots,\MX_{1, l_g} \\
\MX_{v_2},\MX_{2,1},\dots,\MX_{2, l_g} \\
\vdots \\
\MX_{v_N},\MX_{N,1},\dots,\MX_{N, l_g}
\end{array}}, \\
\MS_c = \parentheses{ 
\begin{array}{c} \MS_{v_1},\MS_{1,1},\dots,\MS_{1, l_g} \\
\MS_{v_2},\MS_{2,1},\dots,\MS_{2, l_g} \\
\vdots \\
\MS_{v_N},\MS_{N,1},\dots,\MS_{N, l_g}
\end{array}},
\eea
where the subscript ``$c$'' indicates correlative, and we have used the fact that $l_{g_i} = l_g$ for all $i=1,\dots,N$ because the only inequality constraints in \eqref{eq:binaryTLS} come from $\vxx \in \calX$ and each $\calH_i$ has an equal number of $l_g$ inequality constraints. Our goal is to generate $\MS$, given $\MS_c$, for {\stride}. Note that the matrices $\MS_v$ ($\MX_v$) and $\MS_{v_i}$ ($\MX_{v_i}$) have different dimensions, so that it is inappropriate to just sum up all $\{\MS_{v_i} \}_{i=1}^N$ to get $\MS_v$. The correct way to ``assemble'' $\{\MS_{v_i} \}_{i=1}^N$ is as follows. For each $\MS_{v_i}$, we define $\barMS_{v_i}$ so that it satisfies the following polynomial equality
\bea \label{eq:definebarS}
\inprod{\barMS_{v_i}}{\MX_v} \equiv \inprod{\MS_{v_i}}{\MX_{v_i}}
\eea
for any $\MX_v$ and $\MX_{v_i}$ that are \emph{proper} moment matrices (note that both sides of \eqref{eq:definebarS} are polynomials and the equality implies that the coefficients of both polynomials must be equal). This is essentially creating $\barMS_{v_i}$ to be an all-zero matrix except that the principal submatrix of $\barMS_{v_i}$ indexed by the monomials $\vv_i(\tldvxx_i)$ is equal to $\MS_{v_i}$. Now that $\barMS_{v_i}$ has the same size as $\MX_v$ and $\MS_v$, we can assemble $\MS_v$ as
\bea \label{eq:assembleSv}
\MS_v = \sum_{i=1}^N \barMS_{v_i},
\eea
where the rationale for the sum can be partially understood from the complementarity condition of \eqref{eq:sdpKKT}.
By the same token, for each $\MS_{i,j}$, we create $\barMS_{i,j}$ such that
\bea\label{eq:definebarS2}
\inprod{\barMS_{i,j}}{\MX_{j}} \equiv \inprod{\MS_{i,j}}{\MX_{i,j}}, i = 1,\dots,N, j = 1,\dots,l_g,
\eea
for any $\MX_j$ and $\MX_{i,j}$ that are proper localizing matrices.
Then we assemble $\MS_j$ as 
\bea \label{eq:assembleSj}
\MS_j = \sum_{i=1}^N \barMS_{i,j}, \quad j=1,\dots,l_g.
\eea
The rationale for \eqref{eq:definebarS} and \eqref{eq:definebarS2} can be understood from the complementarity condition of the KKT system \eqref{eq:sdpKKT}, and more deeply from the dual perspective of sums-of-squares (SOS) polynomials \cite{Blekherman12Book-sdpandConvexAlgebraicGeometry} (precisely, we are assembling a SOS polynomial in $(\vxx,\vtheta)$ from $N$ SOS polynomials, each only involves the variables $(\vxx,\theta_i)$). Since this is less relevant for the purpose of this paper (and it is only used for warmstart), we only state the assembling procedure as in \eqref{eq:assembleSv} and \eqref{eq:assembleSj} without diving too deep into the theory of sums of squares. The interested reader is encouraged to refer to the dual SOS perspective in \cite{lasserre10book-momentsOpt}.

\subsubsection{Semi-proximal ADMM}
\label{sec:app-admmplus}
After obtaining $\MX^0$ from primal heuristics such as {\gnc} \cite{Yang20ral-gnc} or {\ransac} \cite{Fischler81}, and $\MS^0$ from solving \eqref{eq:correlativerelax} and performing the assembly procedure in Section \ref{sec:app-cssr}, we use the {semi-proximal alternating direction method of multipliers} ({\admmplus}) proposed in \cite{Sun15siopt-admmplus} to refine both the primal and the dual initializations $(\MX^0,\MS^0)$. The full {\admmplus} algorithm, for solving a standard SDP \eqref{eq:primalSDP}-\eqref{eq:dualSDP}, is presented in Algorithm~\ref{alg:admmplus}. As we can see, at each iteration of {\admmplus}, the major computation involves solving a linear system (\cf \eqref{eq:admmpluslinsolve1} and \eqref{eq:admmpluslinsolve2}) and performing a projection onto the product of positive semidefinite cones $\calK$ (\cf \eqref{eq:admmplusprojpsd}). Since $\calA$ is typically sparse in our examples, Cholesky factorization of $\calA\calAadj$ can be done efficiently and needs to be performed only once. {\admmplus} is a globally convergent algorithm for solving the SDP \eqref{eq:primalSDP}-\eqref{eq:dualSDP} and the interested reader can refer to \cite{Sun15siopt-admmplus} for a detailed study. Notably, \cite{Sun15siopt-admmplus} shows that {\admmplus} is typically $2$ to $3$ times faster than a conventional ADMM. In our implementation, we use the function \texttt{admmplus} in {\sdpnal} \cite{Yang2015mpc-sdpnalplus} to refine $(\MX^0,\MS^0)$ and warmstart {\strideplus}. Although one can directly pass $(\MX^0,\MS^0)$ to {\stride}, empirically we found it is beneficial to refine $(\MX^0,\MS^0)$ using {\admmplus} because the refined initial points will have higher quality that promotes the convergence of {\stride}. In our experiments, we run {\admmplus} for a maximum of $20,000$ iterations, or until $\max\{ \pfeas,\dfeas \}$ is below a threshold (\eg $1\ee{-6}$).

\let\oldnl\nl
\setlength{\textfloatsep}{5pt}%
\begin{algorithm}[ht!]
\nonl 
Given $ \MX^{0} = \MS^0 \in \bbX $, $\sigma > 0$, and $ \gamma \in (0, 2) $. Iterate the following steps for $ k = 0,1,\dots $.
\\
Compute 
\begin{equation} \label{eq:admmpluslinsolve1}
\hatvy^{k+1} =   \big(\calA\calAadj \big)\inv \left( \frac{1}{\sigma} \vb - \calA \left(\frac{1}{\sigma} \MX^{k} + \MS^{k} - \MC \right)\right).
\end{equation}
\\
Compute
\begin{subequations} 
\begin{equation}
\MM^{k+1} = \MX^{k} + \sigma \parentheses{ \calAadj (\hatvy^{k+1}) - \MC },
\end{equation}
\begin{equation}\label{eq:admmplusprojpsd}
\MS^{k+1} = \frac{1}{\sigma} \parentheses{\Pi_{\calK} (\MM^{k+1}) - \MM^{k+1} }.
\end{equation}
\end{subequations}
\\
Compute 
\begin{equation}\label{eq:admmpluslinsolve2}
	\vy^{k+1} = \big(\calA\calAadj \big)\inv \parentheses{ \frac{1}{\sigma}\vb - \calA \parentheses{ \frac{1}{\sigma}\MX^{k} + \MS^{k+1} - \MC } }.
\end{equation}
\\
Compute 
\begin{equation}
	\MX^{k+1} = \MX^{k} + \gamma \sigma \parentheses{ \MS^{k+1} + \calAadj( \vy^{k+1} ) - \MC }.
\end{equation}
\caption{{\admmplus} \label{alg:admmplus}}
\end{algorithm}


\subsection{Local Search and Nonlinear Programming}
\label{sec:app-local-search}
Recall the local search step \eqref{eq:nlpinlocalsearch} applies a nonlinear programming (NLP) algorithm to solve the \eqref{eq:binaryTLS} problem given an initial point. Since \eqref{eq:binaryTLS} is a polynomial optimization, it is straightforward to implement NLP using {\fmincon} in Matlab. However, here we show that it is possible to exploit the smooth manifold structure of \eqref{eq:binaryTLS} and solve it more efficiently with Riemannian optimization \cite{Absil07book} (\eg using {\manopt} \cite{manopt}). First, we can model the vector of binary variables $\vtheta$ as an \emph{oblique manifold} of size $N \times 1$ (an oblique manifold contains matrices with unit-norm rows). Second, from Examples \ref{ex:singlerotation}-\ref{ex:category}, we know the geometric model $\vxx$ contains 2D and 3D rotations, which are both smooth manifolds. However, $\vxx$ can also contain translation $\vt$ and shape parameters $\vc$ that do not live on smooth manifolds. Fortunately, we can drop some constraints so that they both live on smooth manifolds. For example, in Examples \ref{ex:pointcloud}, \ref{ex:mesh}, and \ref{ex:category}, we can relax $\vt \in \ball^3_T$ to $\vt \in \Real{3}$, with the rationale that when the SDP iterate $\MX^k$ is close to optimal, $\norm{\vt} \leq T$ should be naturally satisfied (from rounding \eqref{eq:roundingrestate}) even without explicit constraint. Similarly, we relax $\vt \in \ball^3_T \cap \calC_\alpha$ in Example \ref{ex:absolutepose} to $\vt \in \Real{3}$, and relax $\vc \in \bbR^{K}_{+} \cap \ball^K_T$ in Example \ref{ex:category} to $\vc \in \bbR^{K}_{++}$ (matrices with strictly positive entries live on a smooth manifold). Note that these modifications will not affect the global convergence of {\stride} because \eqref{eq:accept-reject} will reject the NLP solution if it violates the constraints that have been dropped.

\section{Extra Experimental Results}
\label{sec:supp-experiments}

In this section, we report extra experimental results. 

\subsection{Mesh Registration}
\renewcommand{\mpwfour}{5.0cm}
\renewcommand{\myhspace}{\hspace{-4mm}}
\newcommand{\mpwfive}{4cm}
\begin{figure*}[t]
	\begin{center}
	\begin{minipage}{\textwidth}
	\begin{tabular}{cccc}%
		   \myhspace
			\begin{minipage}{\mpwfour}%
			\centering%
			\includegraphics[width=\columnwidth]{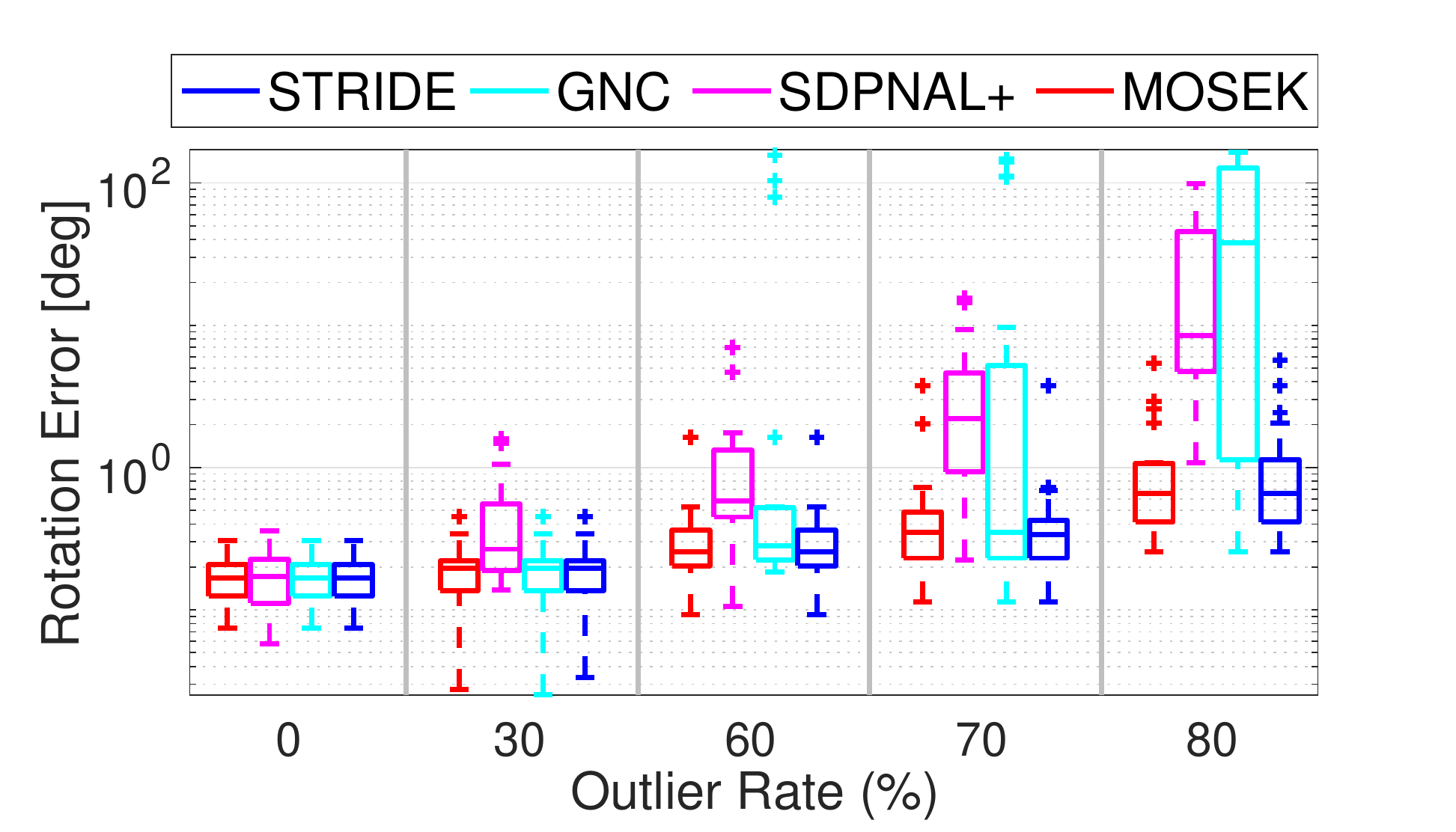}
			\end{minipage}
		&  \myhspace \hspace{-4mm}
			\begin{minipage}{\mpwfour}%
			\centering%
			\includegraphics[width=\columnwidth]{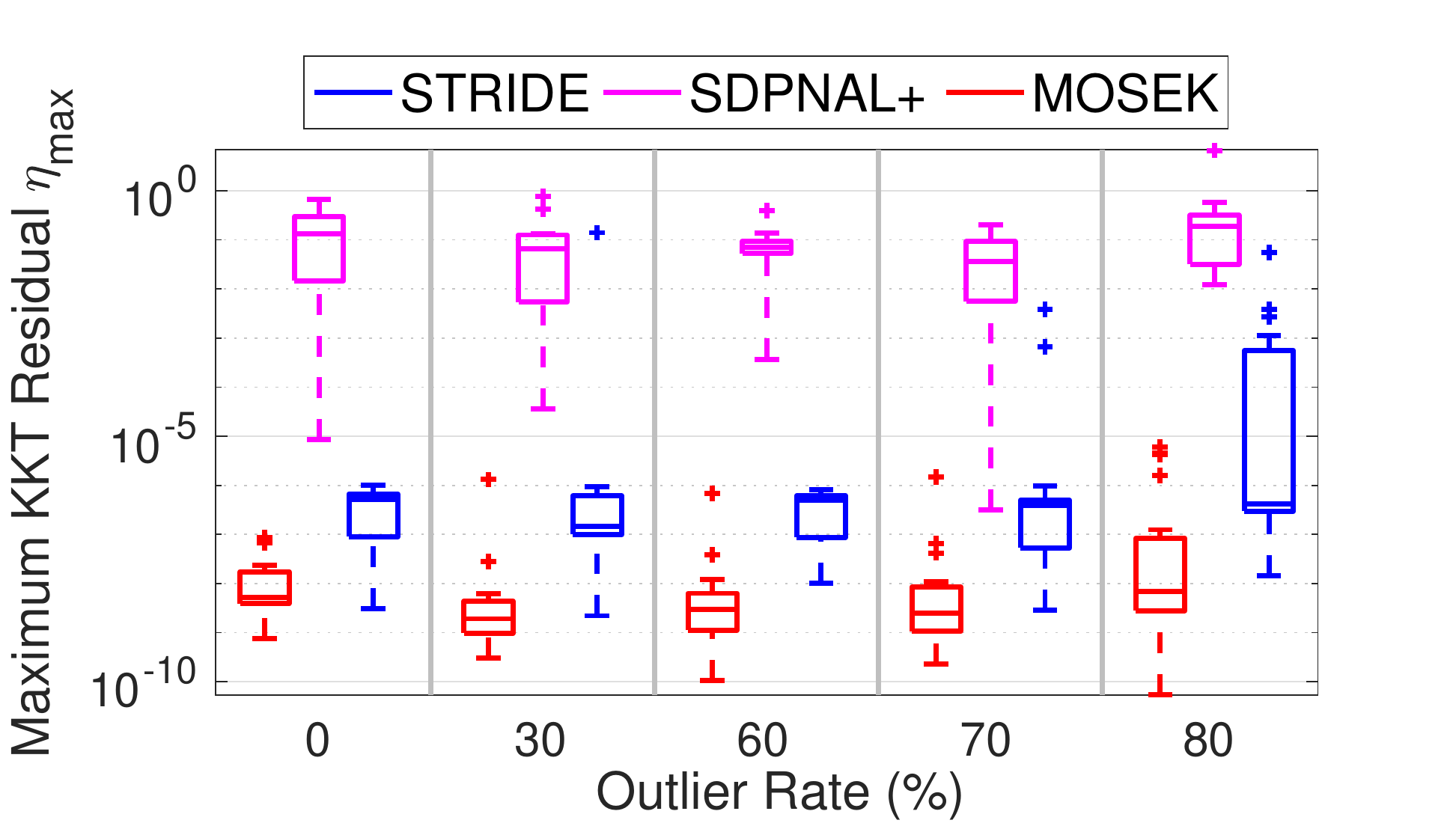}
			\end{minipage}
		&  \myhspace \hspace{-4mm}
			\begin{minipage}{\mpwfour}%
			\centering%
			\includegraphics[width=\columnwidth]{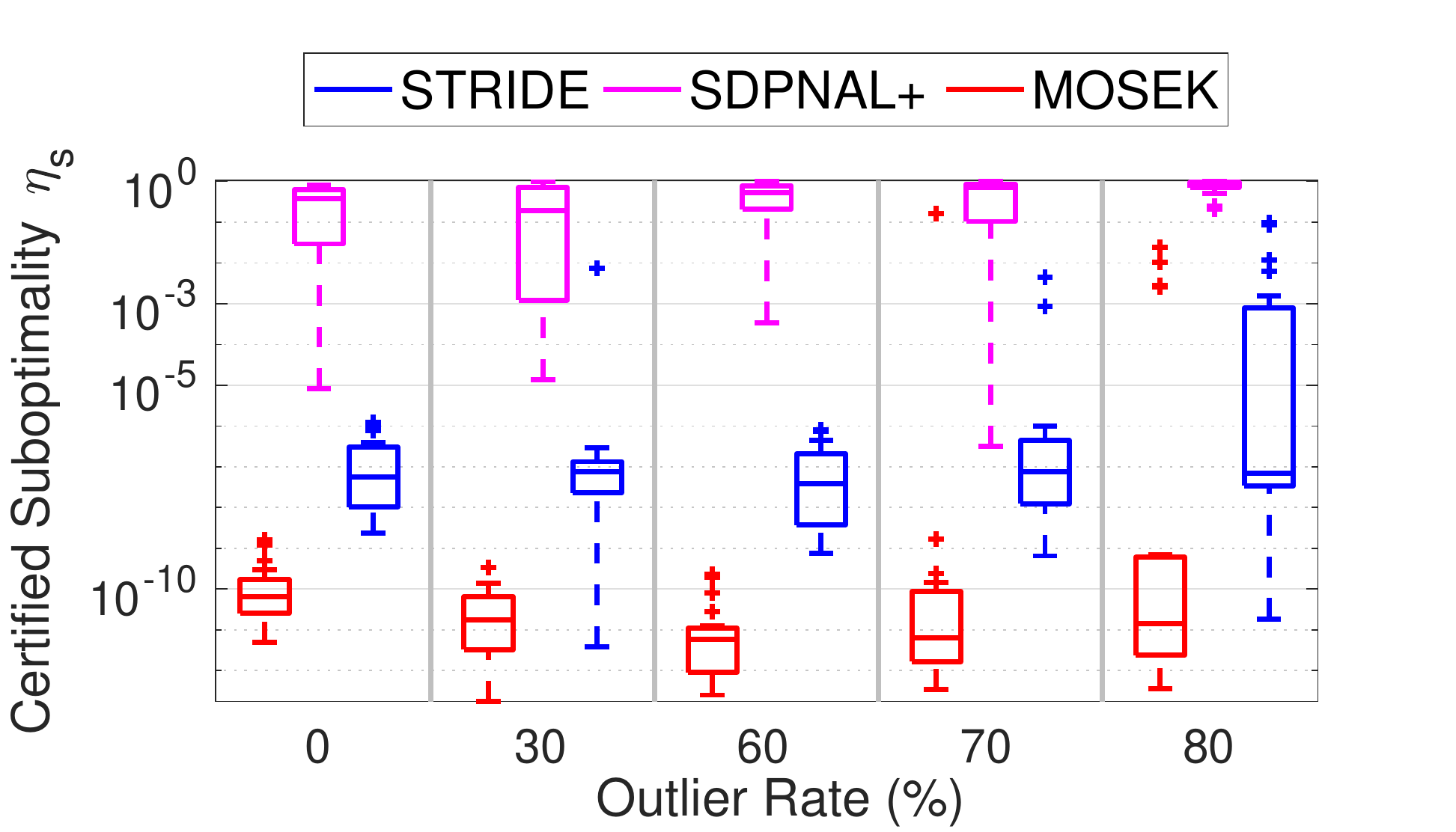}
			\end{minipage}
		&  \myhspace \hspace{-4mm}
			\begin{minipage}{\mpwfour}%
			\centering%
			\includegraphics[width=\columnwidth]{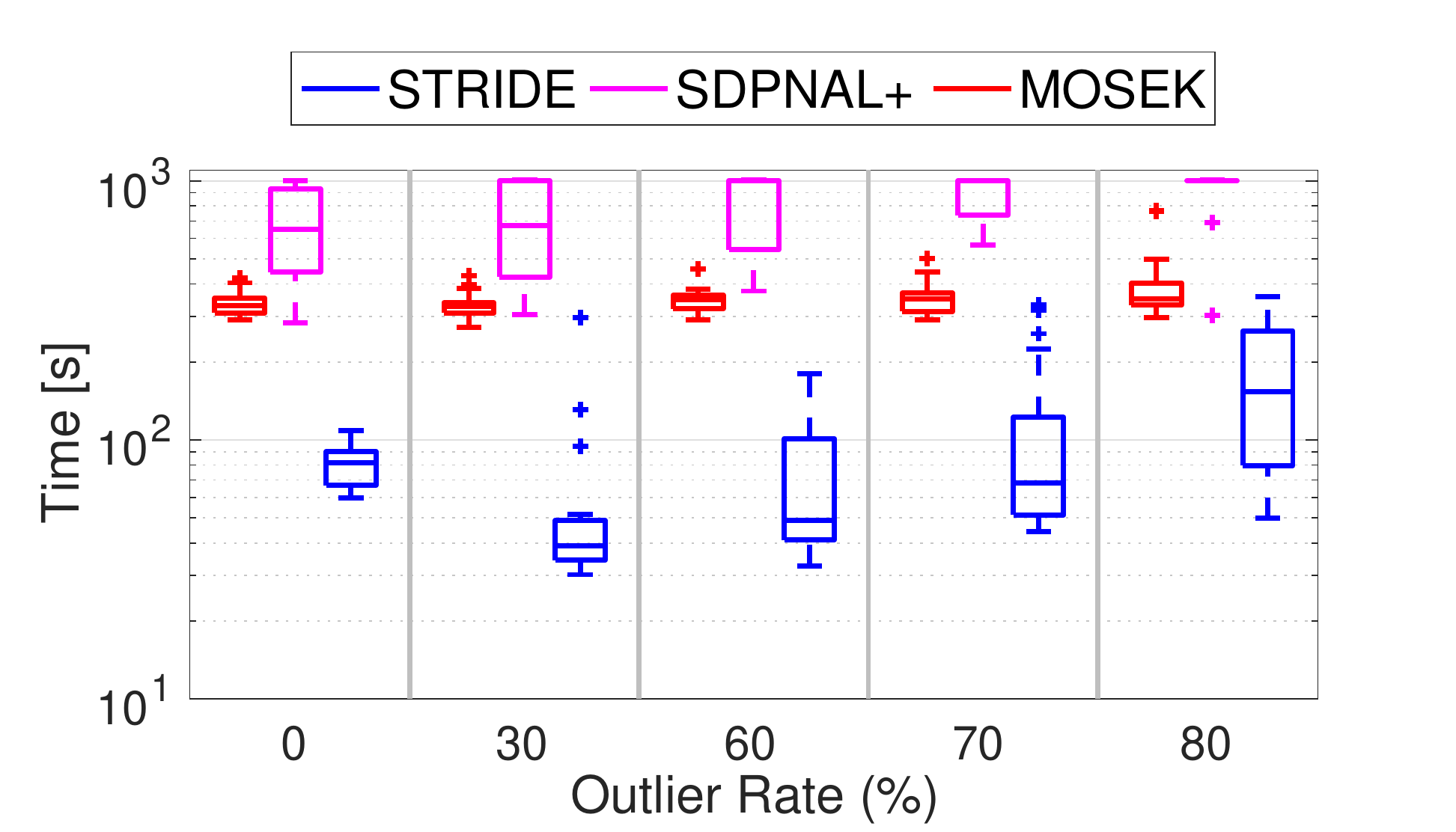}
			\end{minipage} \\
		\multicolumn{4}{c}{\subcapsize (a) $N=20$, $n_1=273$, $m=21,897$}
		\\
		   \myhspace
			\begin{minipage}{\mpwfour}%
			\centering%
			\includegraphics[width=\columnwidth]{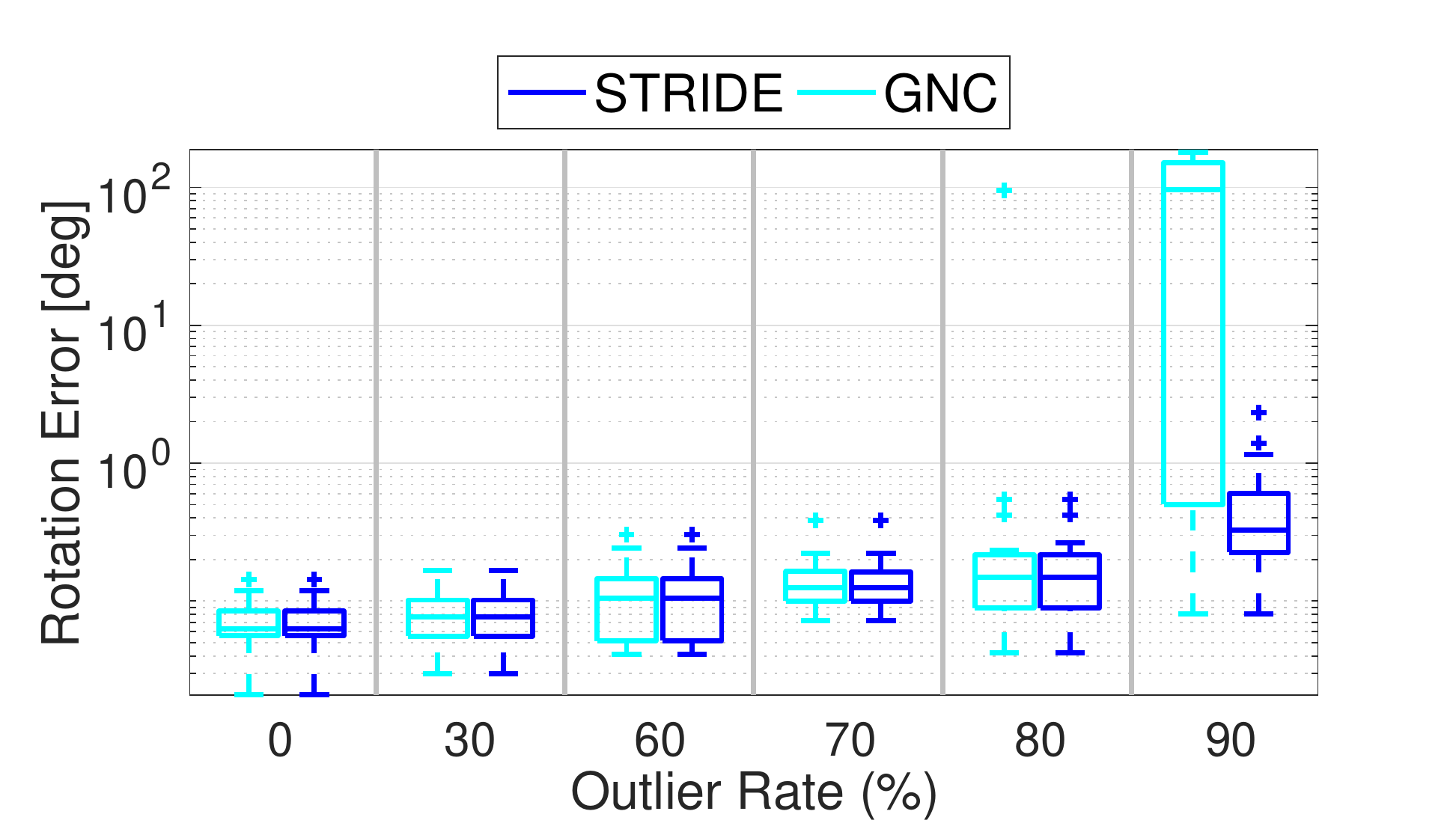}
			\end{minipage}
		&  \myhspace \hspace{-4mm}
			\begin{minipage}{\mpwfour}%
			\centering%
			\includegraphics[width=\columnwidth]{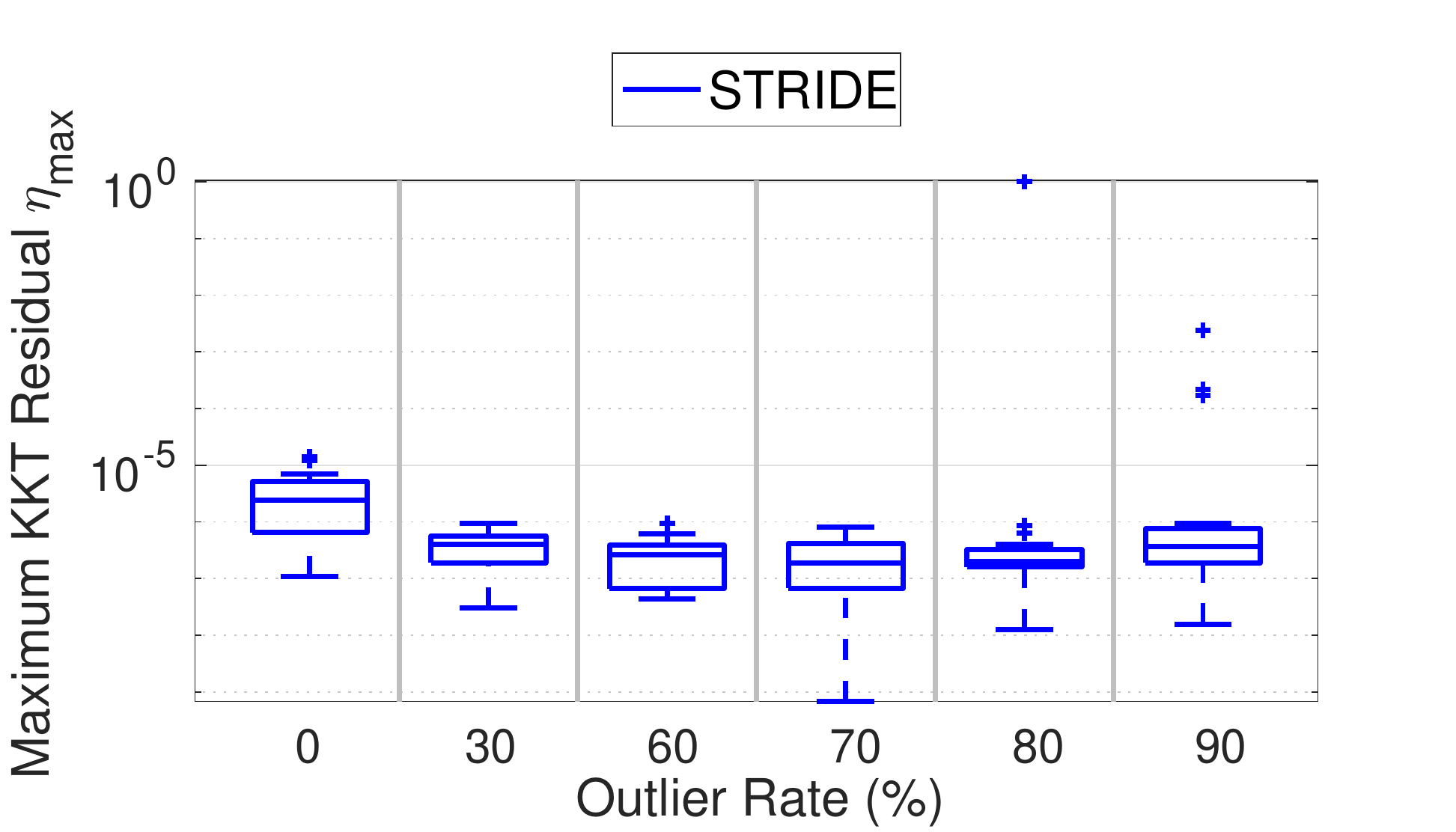}
			\end{minipage}
		&  \myhspace \hspace{-4mm}
			\begin{minipage}{\mpwfour}%
			\centering%
			\includegraphics[width=\columnwidth]{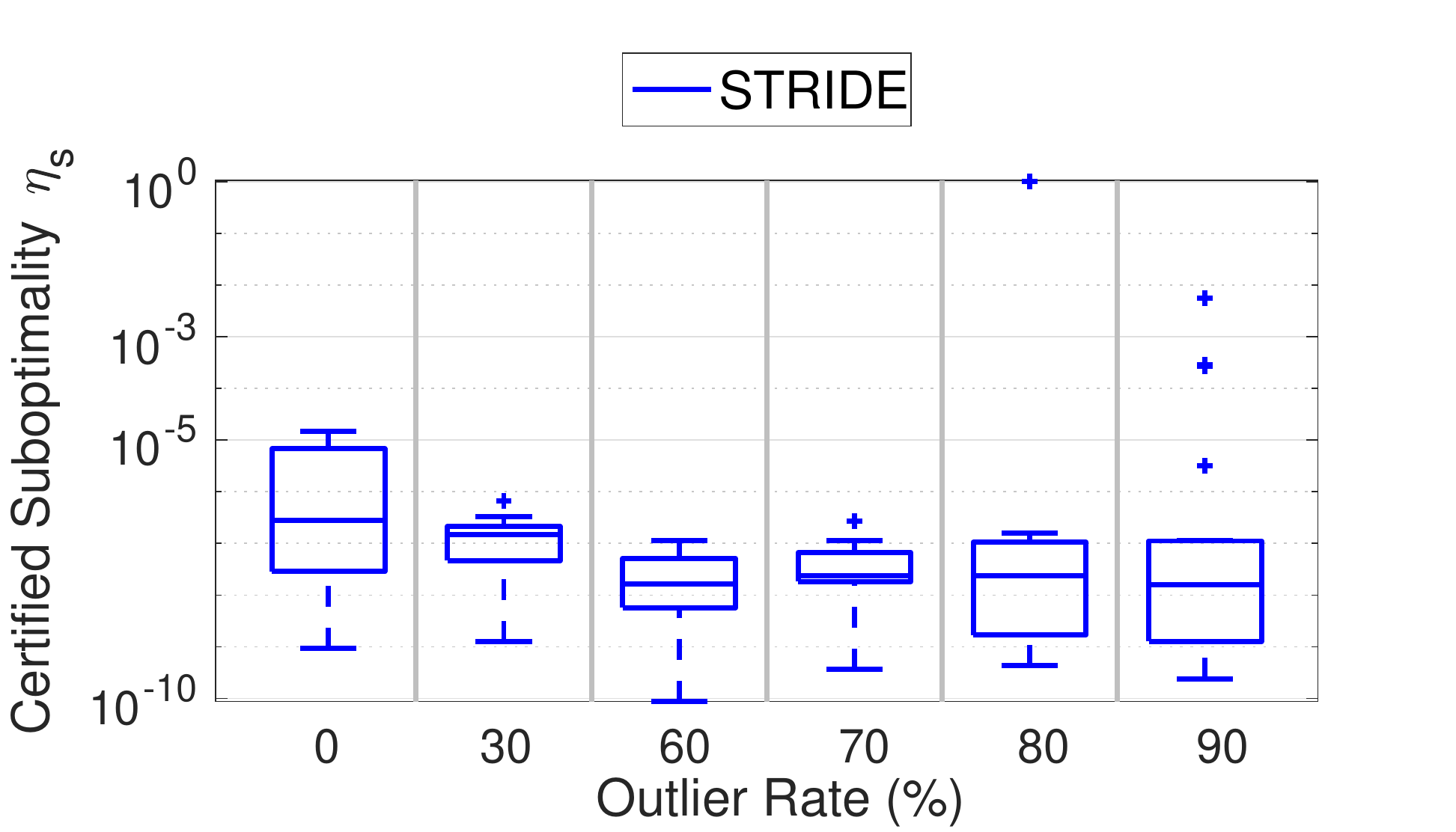}
			\end{minipage}
		&  \myhspace \hspace{-4mm}
			\begin{minipage}{\mpwfour}%
			\centering%
			\includegraphics[width=\columnwidth]{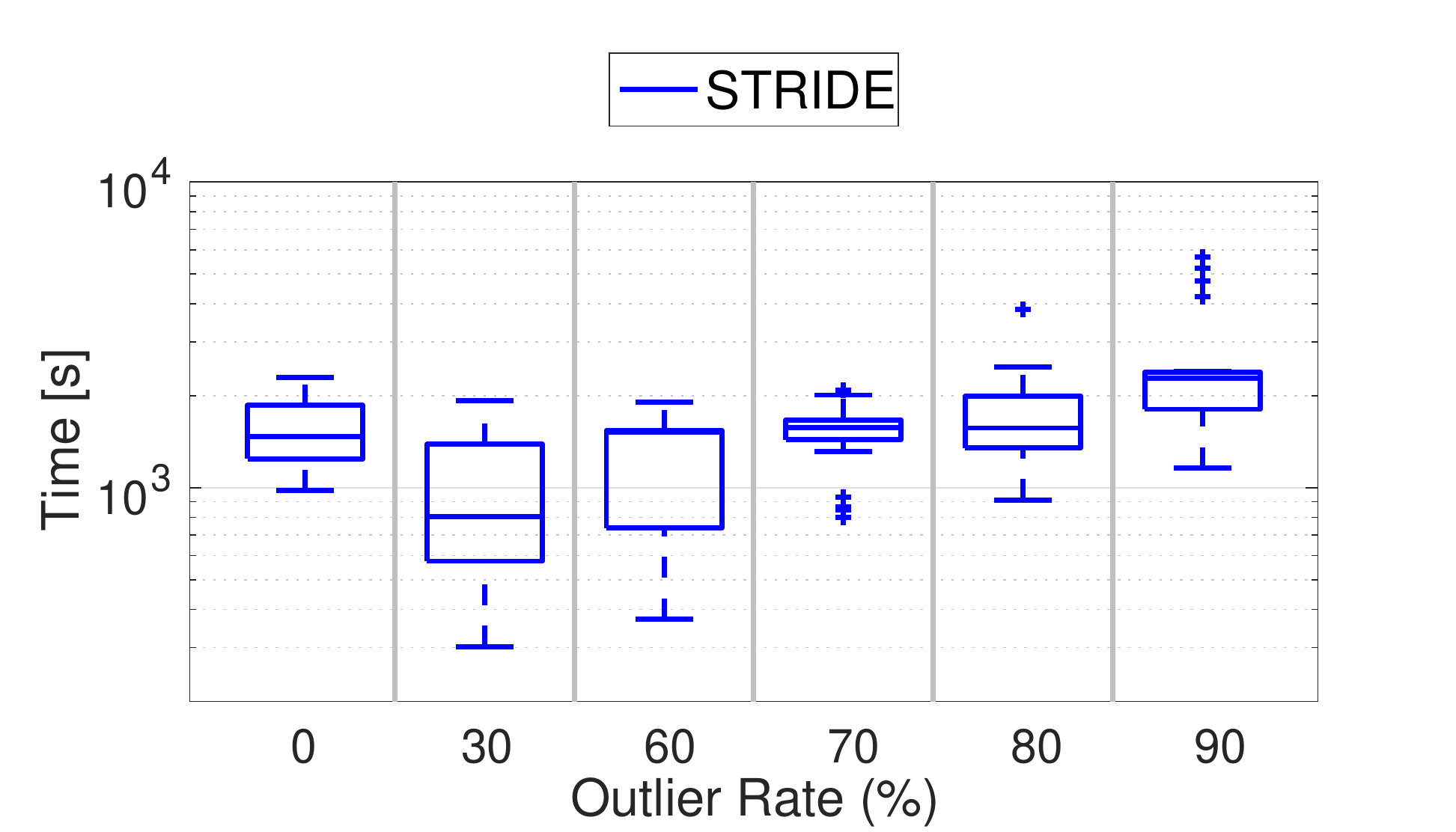}
			\end{minipage} \\
		\multicolumn{4}{c}{\subcapsize (b) $N=100$, $n_1=1313$, $m=485,417$}
	\end{tabular}
	\end{minipage}

	\begin{minipage}{\textwidth}
	\begin{tabular}{ccccc}%
		   \myhspace
			\begin{minipage}{\mpwfive}%
			\centering%
			\includegraphics[width=0.8\columnwidth]{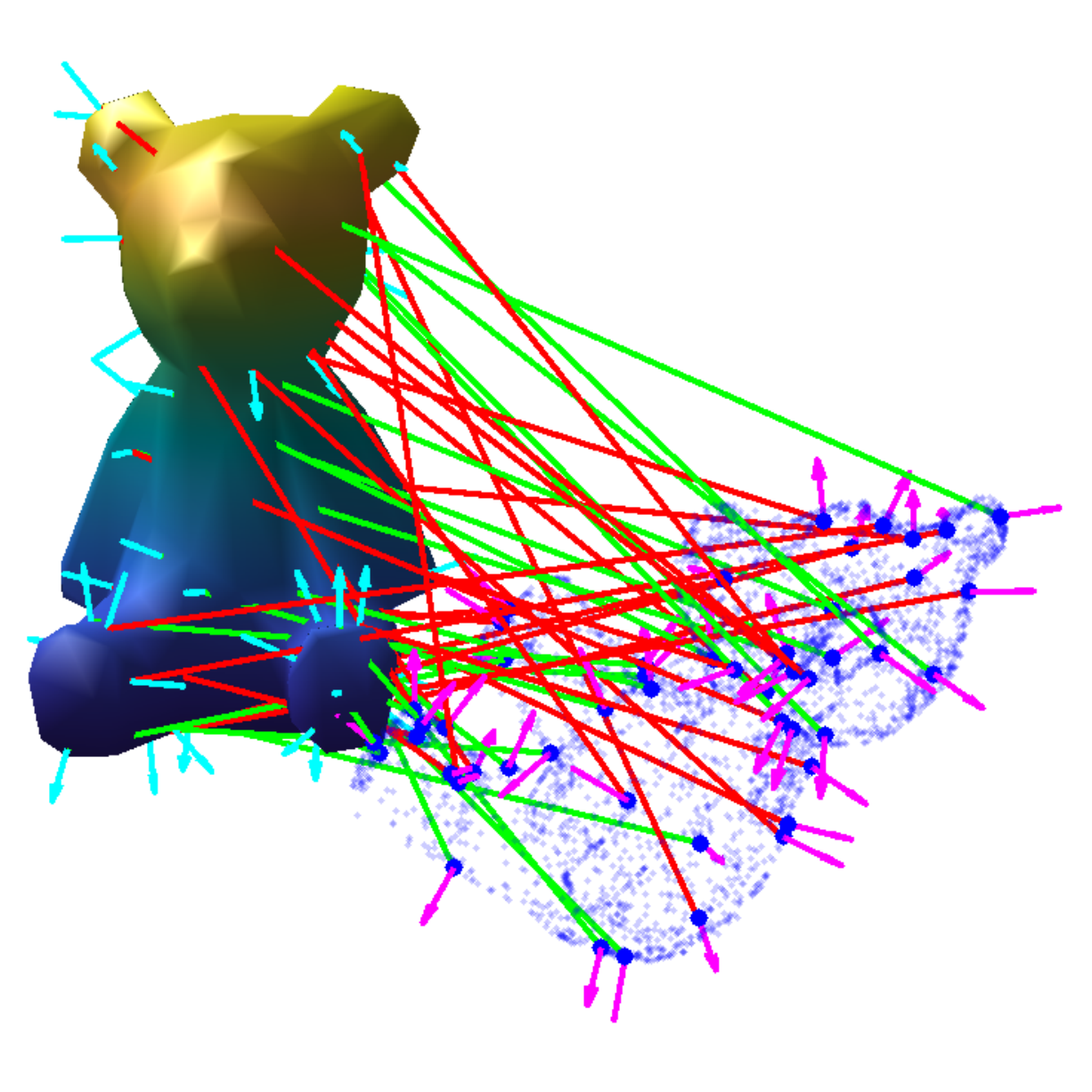}
			\end{minipage}
		&  \myhspace \hspace{-4mm}
			\begin{minipage}{\mpwfive}%
			\centering%
			\includegraphics[width=0.8\columnwidth]{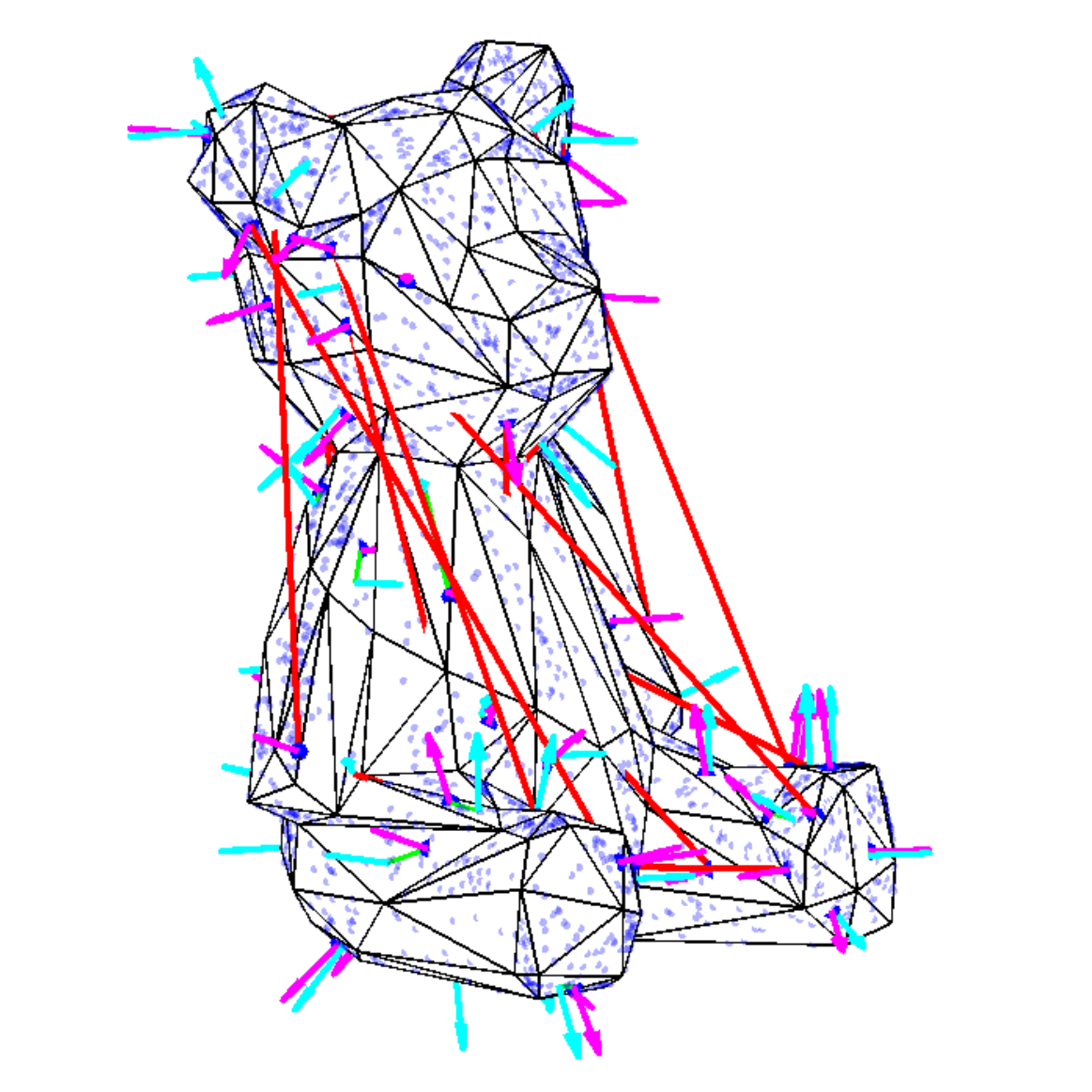}
			\end{minipage}
		&  \myhspace \hspace{-4mm}
			\begin{minipage}{\mpwfive}%
			\centering%
			\includegraphics[width=0.8\columnwidth]{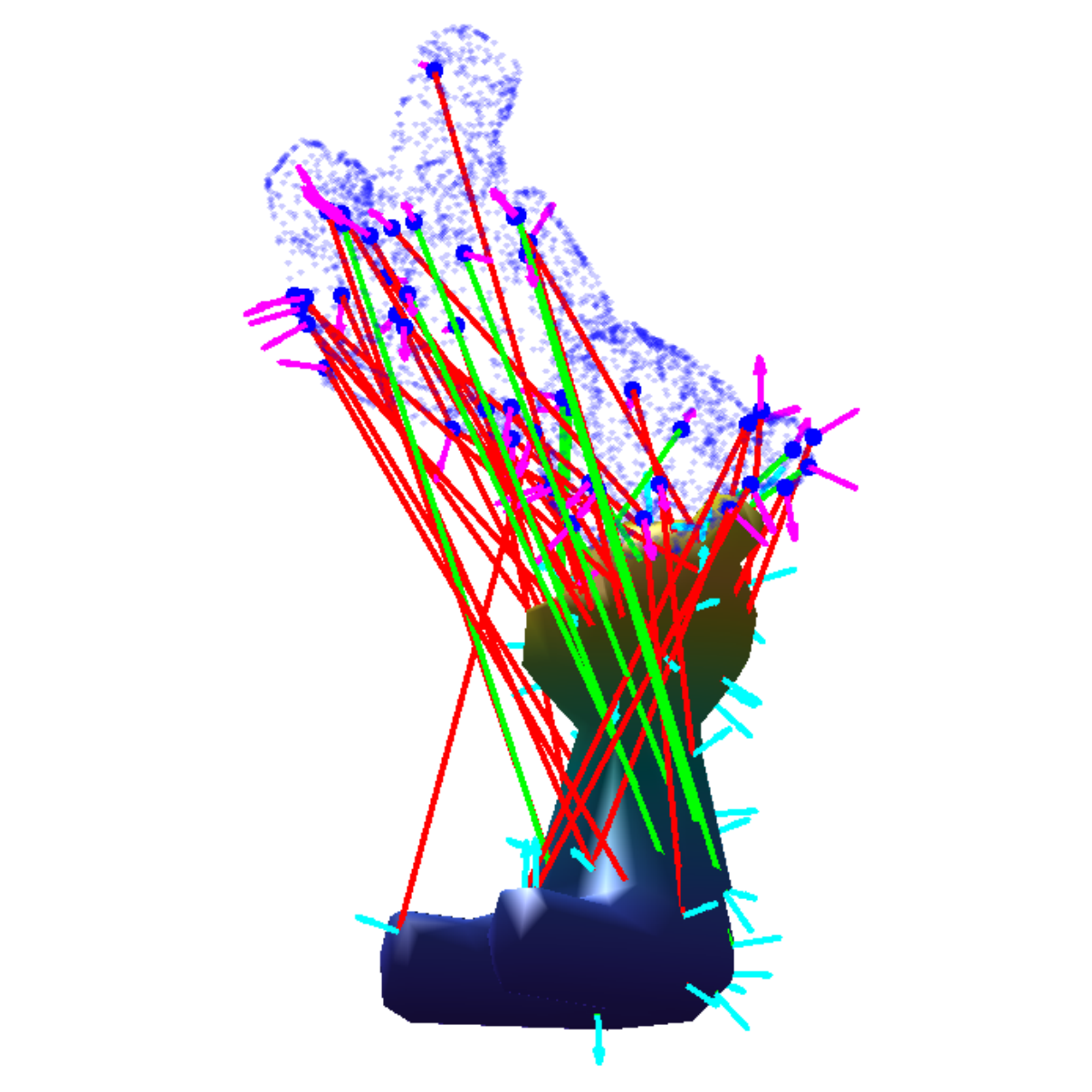}
			\end{minipage}
		&  \myhspace \hspace{-4mm}
			\begin{minipage}{\mpwfive}%
			\centering%
			\includegraphics[width=0.8\columnwidth]{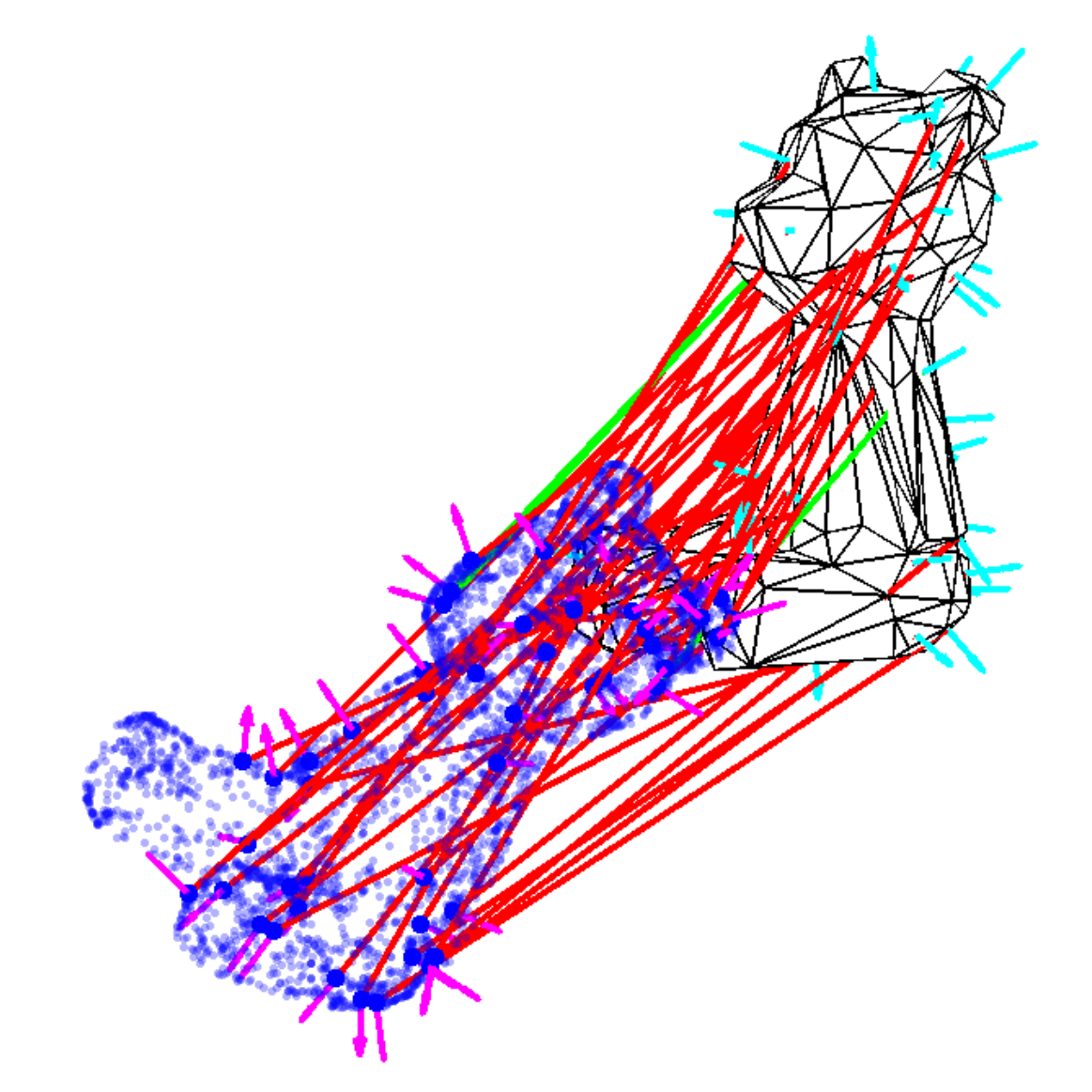}
			\end{minipage}
		&  \myhspace \hspace{-4mm}
			\begin{minipage}{\mpwfive}%
			\centering%
			\includegraphics[width=0.8\columnwidth]{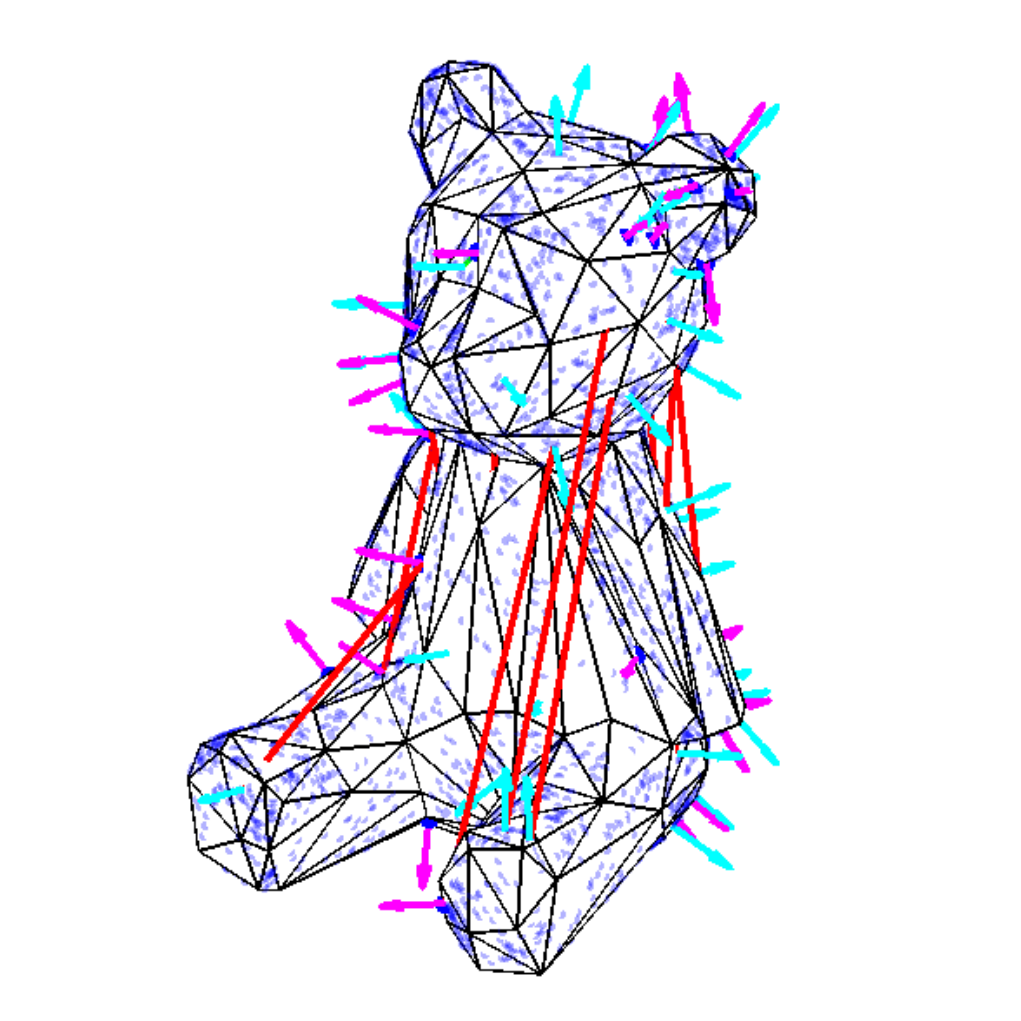}
			\end{minipage} \\
		\multicolumn{2}{c}{\hspace{-4mm} \begin{minipage}{8cm} \subcapsize (c-1) Registration using {\teddybear} \cite{Kaskman19-homebrewedDB}. An example where the globally optimal estimate returned by {\gnc} is certified by {\stride}. $N=50$, $50\%$ outliers.  \end{minipage} \vspace{-2mm} }
		&
		\multicolumn{3}{c}{\hspace{-6mm} \begin{minipage}{10cm} \subcapsize (c-2) An example where {\gnc} converges to a local minimum (middle column), but {\stride} escapes it and obtains a certifiably optimal estimate (right column). $N=50$, $70\%$ outliers. \end{minipage}}
	\end{tabular}
	\end{minipage}
	\caption{Mesh Registration (Example \ref{ex:mesh}).
	\label{fig:exp-mr-results}} 
	\vspace{-8mm} 
	\end{center}
\end{figure*}
\revise{
{\bf Setup}. 
We first simulate a random mesh by sampling a set of $N$ 3D planes $\{\vq_i, \vv_i\}_{i=1}^N$, where $\vv_i$ is the unit normal of the plane (by sampling a random 3D direction) and $\vq_i \sim \calN(\zero,\eye_3)$ is an arbitrary point on the plane. We then generate a random point on each plane via $\vq_i' = \vq_i + \vw_i \times \vv_i$, where $\vw_i \sim \calN(\zero,\eye_3)$ is a random point and ``$\times$'' denotes the vector cross product. After this, we generate a random groundtruth transformation $(\MRgt,\vtgt)$, and transform $(\vq_i',\vv_i)$ to be $\vp_i = \MRgt \vq_i' + \vtgt + \vvarepsilon_{pi}$ and $\vu_i = \normalize(\MRgt \vv_i + \vvarepsilon_{ni})$, if $(\vp_i,\vu_i)$ is an inlier, where $\vvarepsilon_{pi},\vvarepsilon_{ni} \sim \calN(\zero,0.01^2 \eye_3)$ are random Gaussian noise, and $\normalize(\vv) \triangleq \vv / \norm{\vv}$ normalizes a vector to have unit norm. If $(\vp_i,\vu_i)$ is an outlier, then $\vp_i$ is a random 3D point and $\vu_i$ is a random 3D direction. Given the mesh $\{\vq_i, \vv_i\}_{i=1}^N$ and the noisy point cloud with normals $\{\vp_i,\vu_i\}_{i=1}^N$, we seek the best transformation $(\MRstar,\vtstar)$ to \emph{align the point cloud to the mesh} using the residual defined in Example \ref{ex:mesh}. After $(\MRstar,\vtstar)$ is found, its \emph{inverse} transformation is used to compute the estimation errors \wrt $(\MRgt,\vtgt)$ (recall $(\MRgt,\vtgt)$ is generated to {align the mesh to the point cloud}). We test $N=20$ and $N=100$ with increasing outlier rates.

{\bf Results}. Fig.~\ref{fig:exp-mr-results}(a)-(b) plots the evaluation metrics for $N=20$ and $N=100$, respectively. The results are mostly the same as point cloud registration in Fig. \ref{fig:exp-pcr-results}(a)-(b), except that when $N=20$, the relaxation is not always tight at $70\%$ and $80\%$ outlier rates (from the $\subopt$ plot of {\mosek} we see one inexact run at $70\%$ and three inexact runs at $80\%$).

{\bf Mesh registration with {\teddybear}}. We perform mesh registration using the {\teddybear} mesh model from the {\homebrew} dataset \cite{Kaskman19-homebrewedDB}. From the {\teddybear} mesh, we generate a noisy point cloud by densely sampling points on each face of the mesh with additive Gaussian noise, and transform the point cloud using a random rigid transformation. We use the $\texttt{pcnormals}$ function in Matlab to estimate surface normals for each point in the point cloud. We then randomly sample $N=50$ point-to-face correspondences with outliers, and use {\stride} to estimate the rigid transformation. Fig.~\ref{fig:exp-mr-results}(c-1) shows an instance with $50\%$ outliers, where {\gnc} successfully returns the globally optimal solution and {\stride} computes a certificate of optimality ($\subopt = 2.5\ee{-8}$). Fig.~\ref{fig:exp-mr-results}(c-2) shows an instance with $70\%$ outliers, where {\gnc} converges to a suboptimal solution but {\stride} escapes the local minimum and finds the globally optimal solution with a certificate of optimality ($\subopt = 1.1\ee{-7}$).
}

\subsection{Robustness of the TLS Estimator}
\revise{
	Here we show that the accuracy of the TLS
estimator increases with the number of inliers and is comparable
with a least-squares solution computed from the inliers only.
We perform an experiment in single rotation averaging with outlier rate fixed at $80\%$ but number of measurements $N$ increased from $N=30$ to $N=100$. At each $N$, we perform 20 random simulations and compute the rotation estimation error w.r.t. groundtruth. Fig. \ref{fig:estimationcontract}(c) shows that all TLS solutions are certified as globally optimal. Fig. \ref{fig:estimationcontract}(a) shows that, as $N$ increases and the number of inliers increases, the estimation error in general decreases (both in terms of the average estimation error and the quantiles as shown by the boxplot). This demonstrates the empirical robustness of the TLS estimator against outliers and its capability in exploiting information from the inliers.

	Using the same experimental setup, we compare the rotation error between the TLS estimator and the least squares (LS) estimator (after running TLS, we discard the measurements deemed as outliers by TLS and run LS on the remaining inliers). Fig. \ref{fig:estimationcontract}(b) shows that the TLS estimator is exactly the same as the LS estimator after discarding outliers (up to numerical inaccuracies, the rotation errors are shown in degrees). This demonstrates that the outliers do not affect the TLS solution, and the TLS estimator is truly robust against outliers.
}
\begin{figure*}[h]
\begin{center}
	\begin{minipage}{\textwidth}
	\begin{tabular}{ccc}%
			\begin{minipage}{5.5cm}%
			\centering%
			\includegraphics[width=\columnwidth]{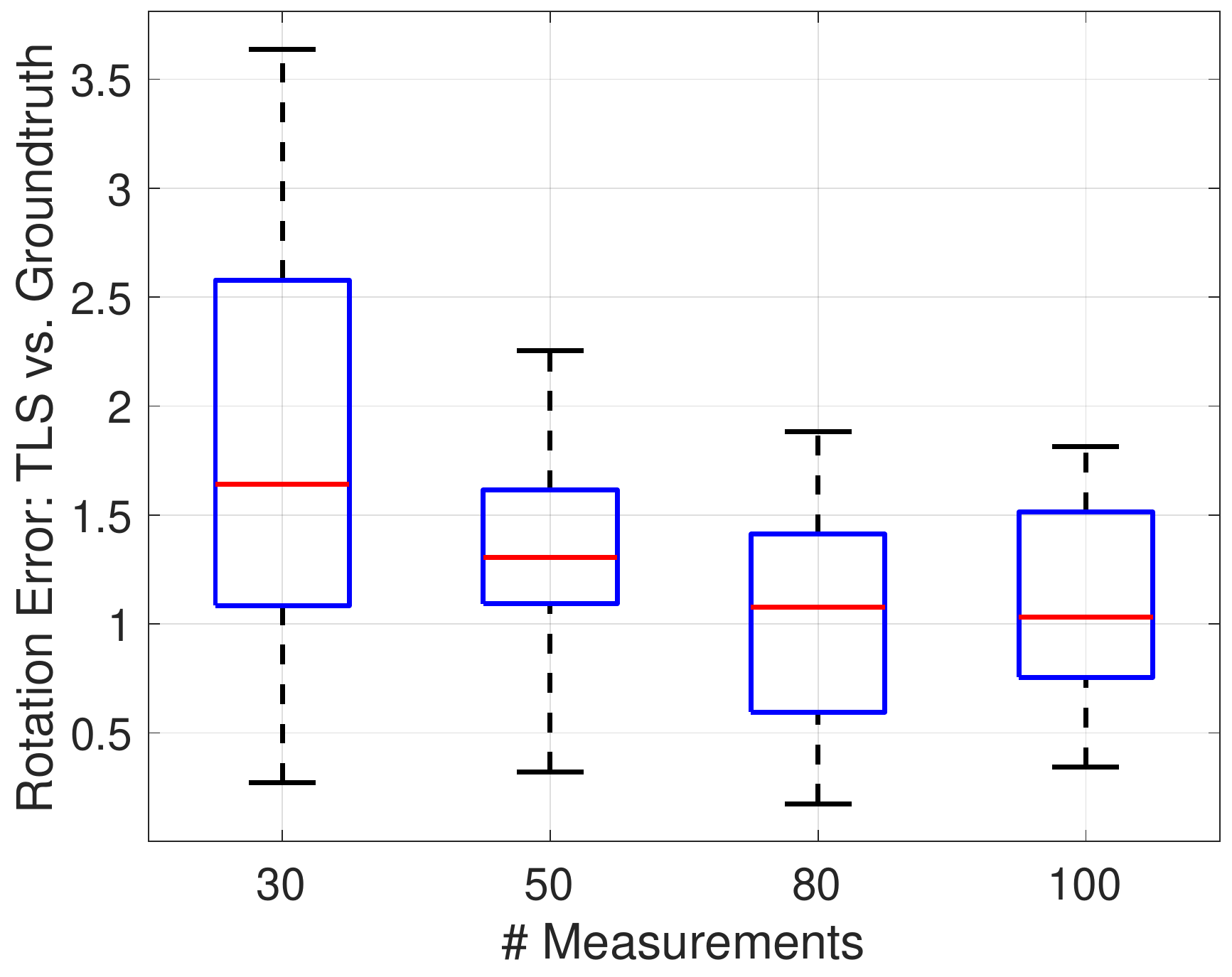} \\
			{\small (a) TLS vs. Groundtruth}
			\end{minipage}
		&  
			\begin{minipage}{5.5cm}%
			\centering%
			\includegraphics[width=\columnwidth]{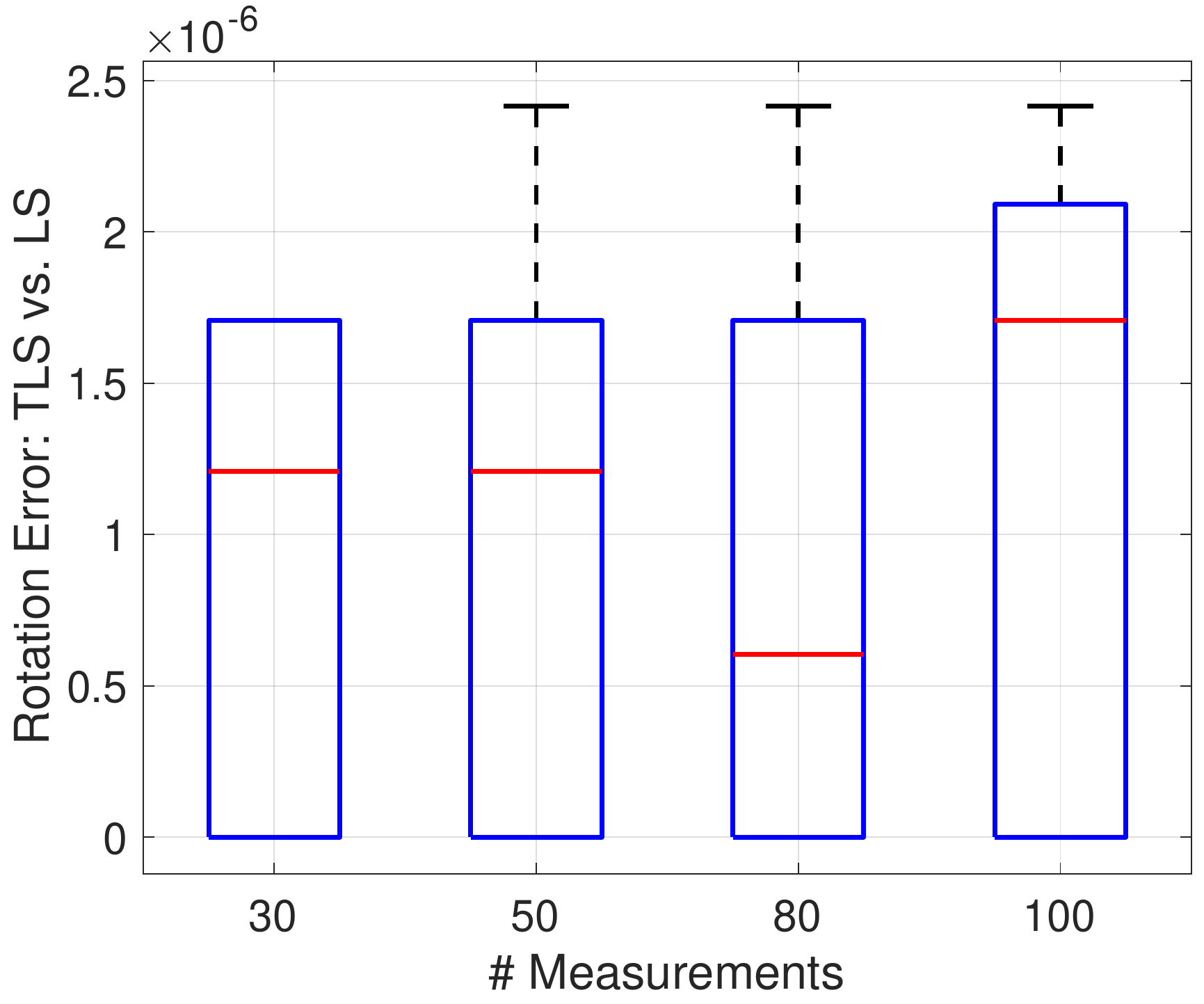} \\
			{\small (b) TLS vs. LS }
			\end{minipage}
		& 
			\begin{minipage}{5.5cm}%
			\centering%
			\includegraphics[width=\columnwidth]{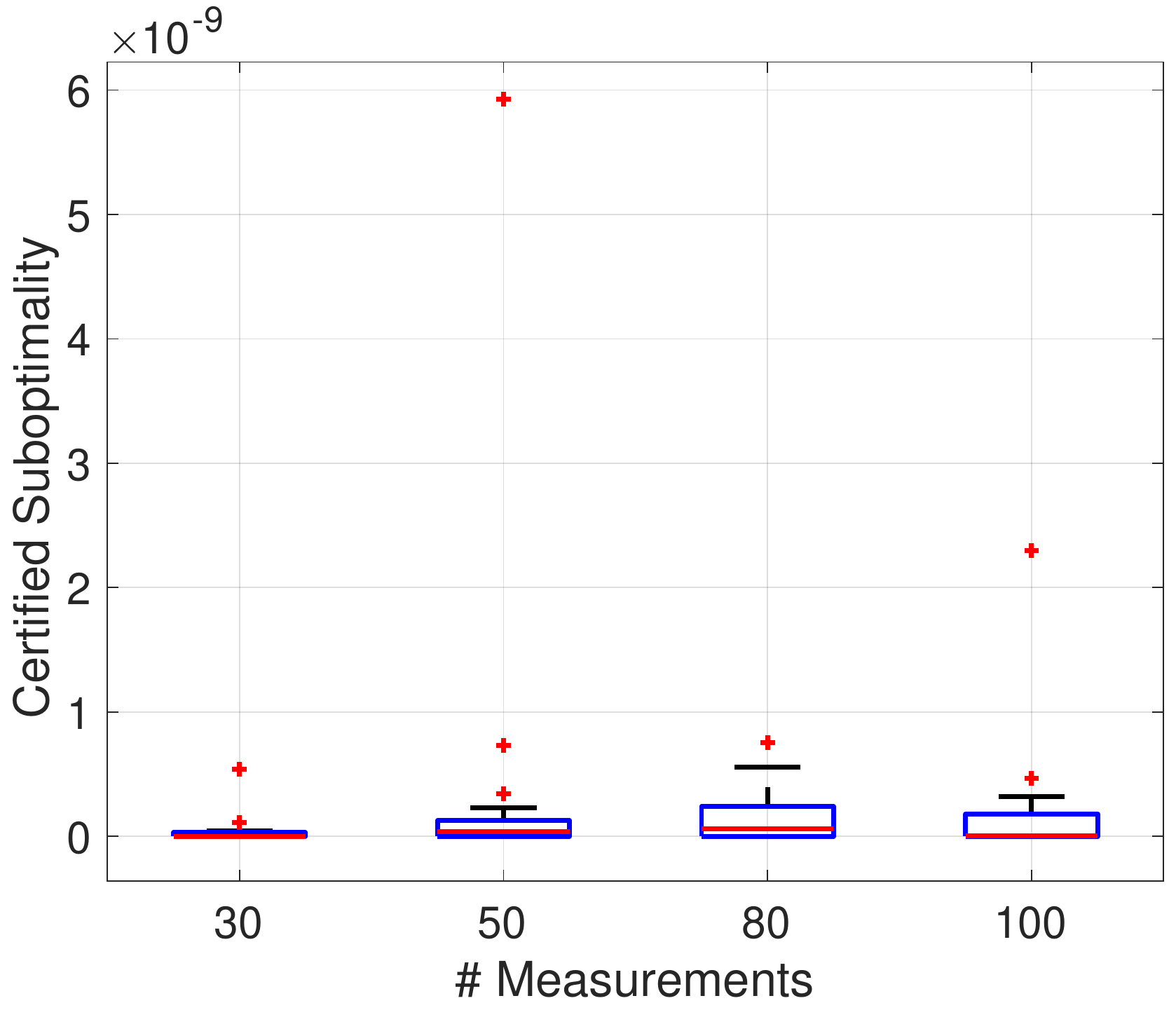} \\
			{\small (c) Certified Suboptimality}
			\end{minipage}
	\end{tabular}
	\end{minipage} 
	 
	\caption{Rotation estimation error under increasing number of measurements for single rotation averaging with $80\%$ fixed outlier rate (thus increasing number of inliers). (a) Rotation error between TLS estimate and groundtruth rotation. (b) Rotation error between TLS estimate and least squares (LS) estimate. (c) All TLS solutions are certified as optimal. Rotation errors shown in degrees. \label{fig:estimationcontract}}
\end{center}
\end{figure*}

\subsection{Scaling the Noise Bound}
\revise{
	We perform an experiment to investigate how the optimal solutions change when the noise bound $\beta$ is varied, using the single rotation averaging problem with $N=50$ measurements and $50\%$ outlier rate.
	Fig. \ref{fig:scaleboundsra} plots the rotation estimation error and certified suboptimality w.r.t. different scaling on the original noise bound $\beta$ (all the measurements are the same at each random simulation, only the noise bound $\beta$ is chosen differently). We can see that (1) there is a wide range of $\beta$ that leads to certifiably optimal and accurate estimation, and (2) when the noise bound $\beta$ is slightly perturbed (\eg decreased to $90\%$ or increased to $110\%$), the optimal solution remains the same for most problem instances, as shown by the similar boxplots in Fig. \ref{fig:scaleboundsra}(a) at horizontal locations $0.9$, $1$, and $1.1$ (in fact, $15$ out of the $20$ runs have exactly the same solutions).  
}

\begin{figure*}[h]
\begin{center}
	\begin{minipage}{\textwidth}
	\centering
	\begin{tabular}{cc}%
			\begin{minipage}{5.5cm}%
			\centering%
			\includegraphics[width=\columnwidth]{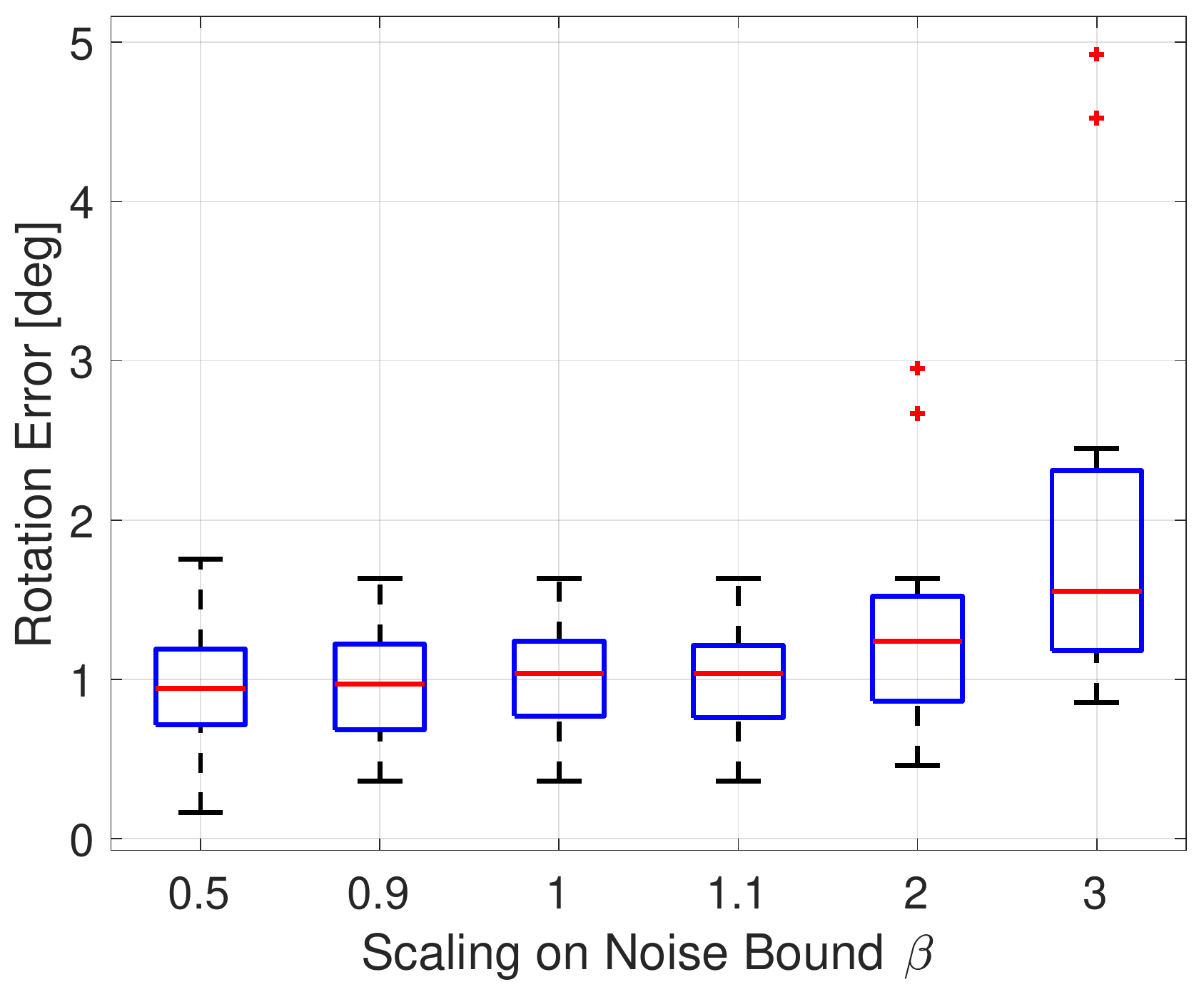} \\
			{\small (a) Rotation estimation error}
			\end{minipage}
		&  
			\begin{minipage}{6cm}%
			\centering%
			\includegraphics[width=\columnwidth]{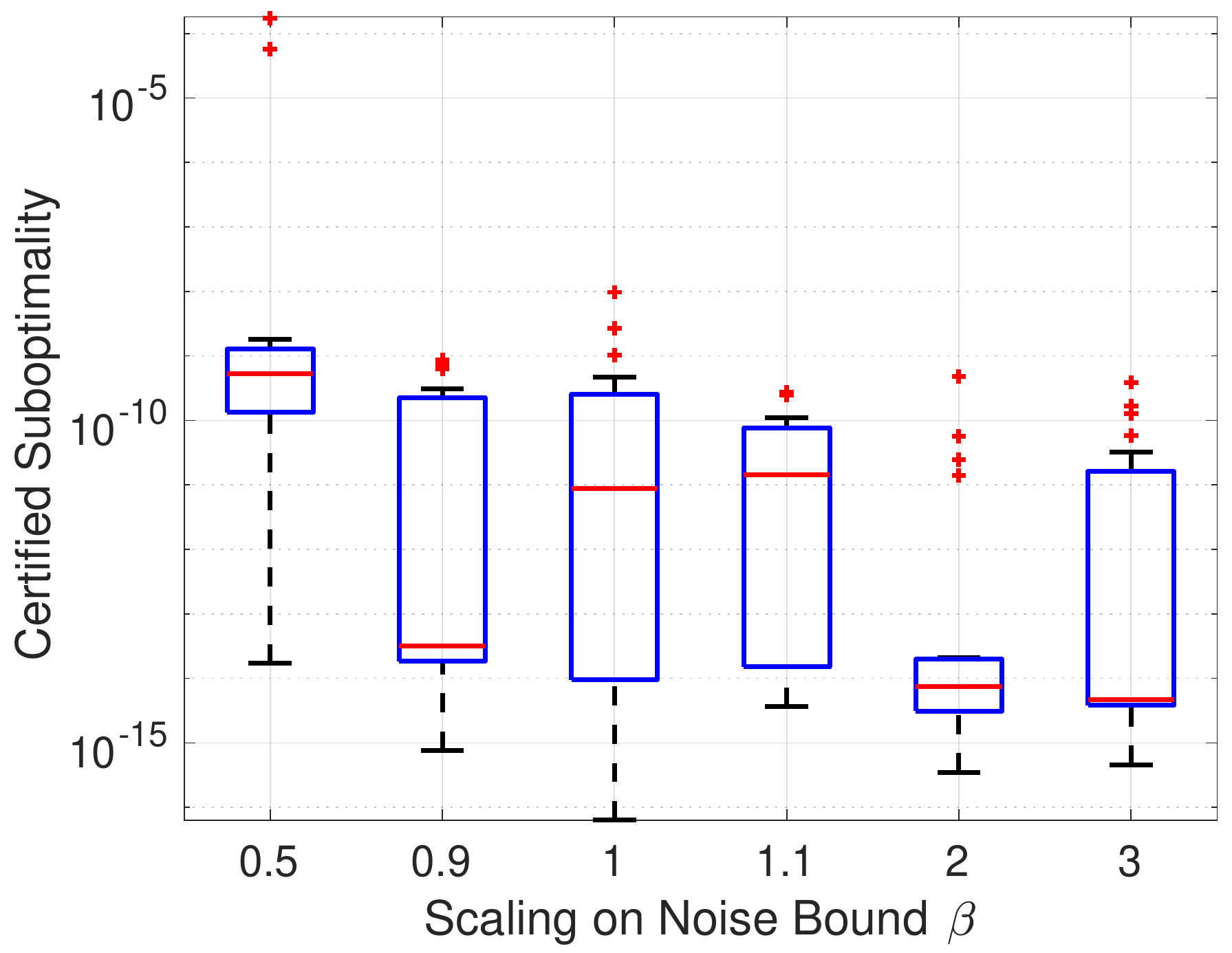} \\
			{\small (b) Certified suboptimality }
			\end{minipage}
	\end{tabular}
	\end{minipage} 
	 
	\caption{(a) Rotation estimation error and (b) certified suboptimality w.r.t. scaling on the noise bound $\beta$ in single rotation averaging with $N=50$ and $50\%$ outlier rate. \label{fig:scaleboundsra}}
\end{center}
\end{figure*}

\subsection{Point Cloud Registration on {\threedmatch}}

We provide $10$ extra scan matching results by {\stride} on the {\threedmatch} dataset \cite{Zeng17cvpr-3dmatch} in Fig. \ref{fig:supp-3dmatch}. {\stride} returned the globally optimal transformation estimates in all cases.

\renewcommand{\myhspace}{\hspace{-3mm}}
\renewcommand{\mpwtwo}{9.0cm}
\begin{figure*}[t]
	\begin{center}
	\begin{minipage}{\textwidth}
	\begin{tabular}{cc}%
		\myhspace
		\begin{minipage}{\mpwtwo}%
		\centering%
		\includegraphics[width=0.9\columnwidth]{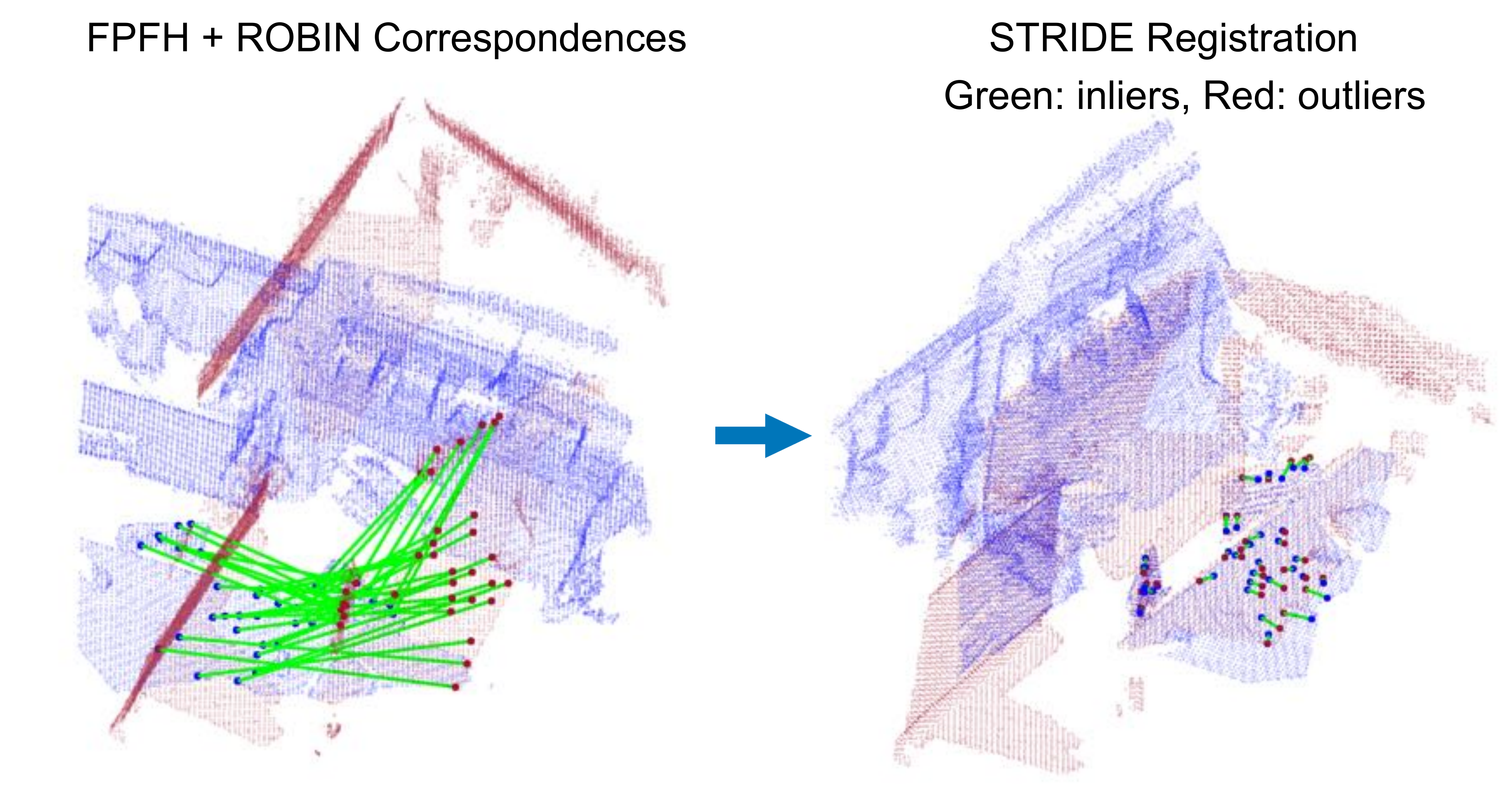}
		\\
		{\subcapsize (a) $\MR$ error: $2.8^{\circ}$, $\vt$ error: $1.2\ee{-1}$,  $\subopt = 6.5\ee{-8}$, time: $129$ [s]}
		\end{minipage}

		& 
		\myhspace
		\begin{minipage}{\mpwtwo}%
		\centering%
		\includegraphics[width=0.9\columnwidth]{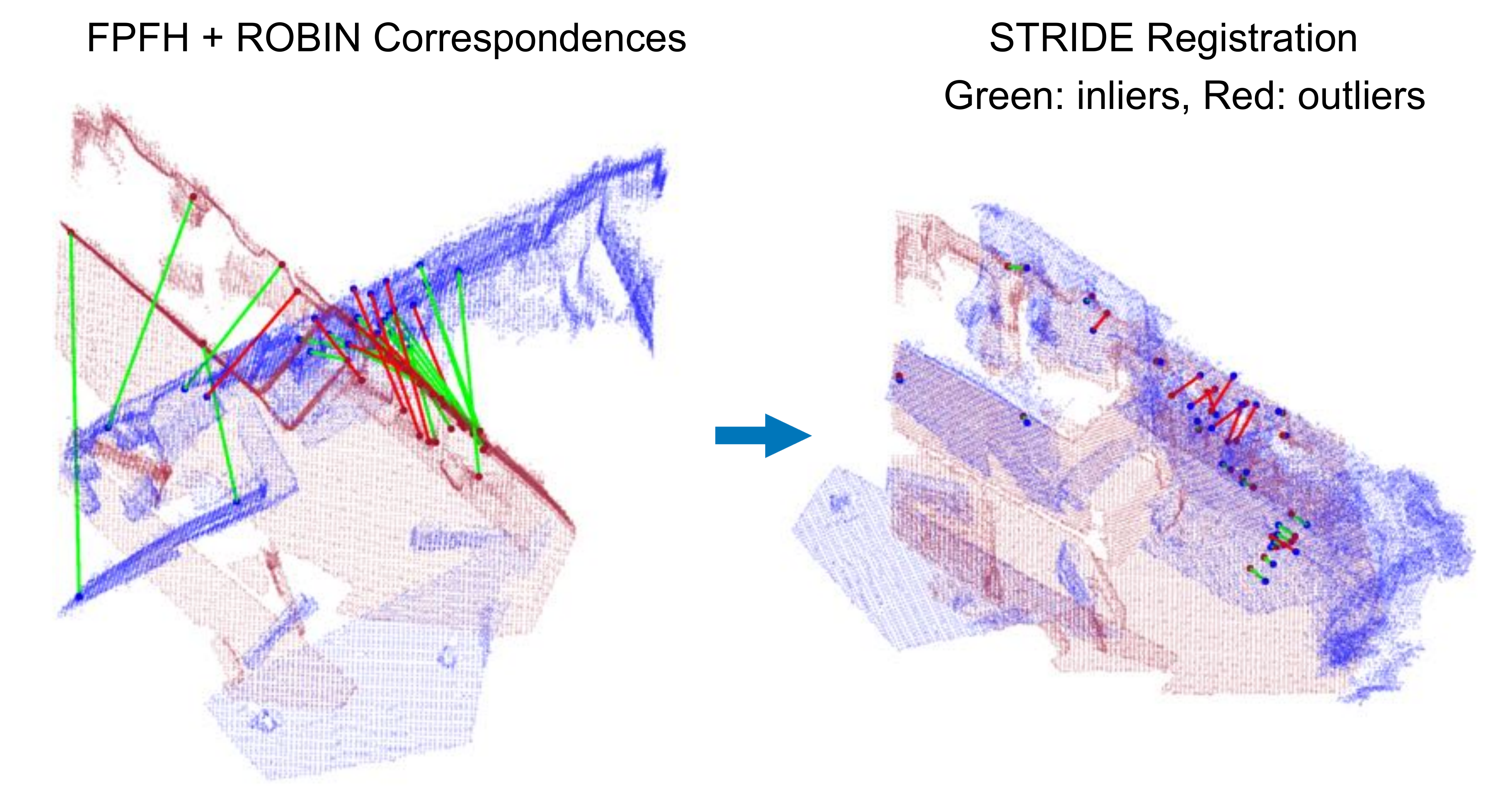}
		\\
		{\subcapsize (b) $\MR$ error: $3.0^{\circ}$, $\vt$ error: $1.9\ee{-1}$,  $\subopt = 1.2\ee{-8}$, time: $228$ [s]}
		\end{minipage}

		\\

		\myhspace
		\begin{minipage}{\mpwtwo}%
		\centering%
		\includegraphics[width=0.9\columnwidth]{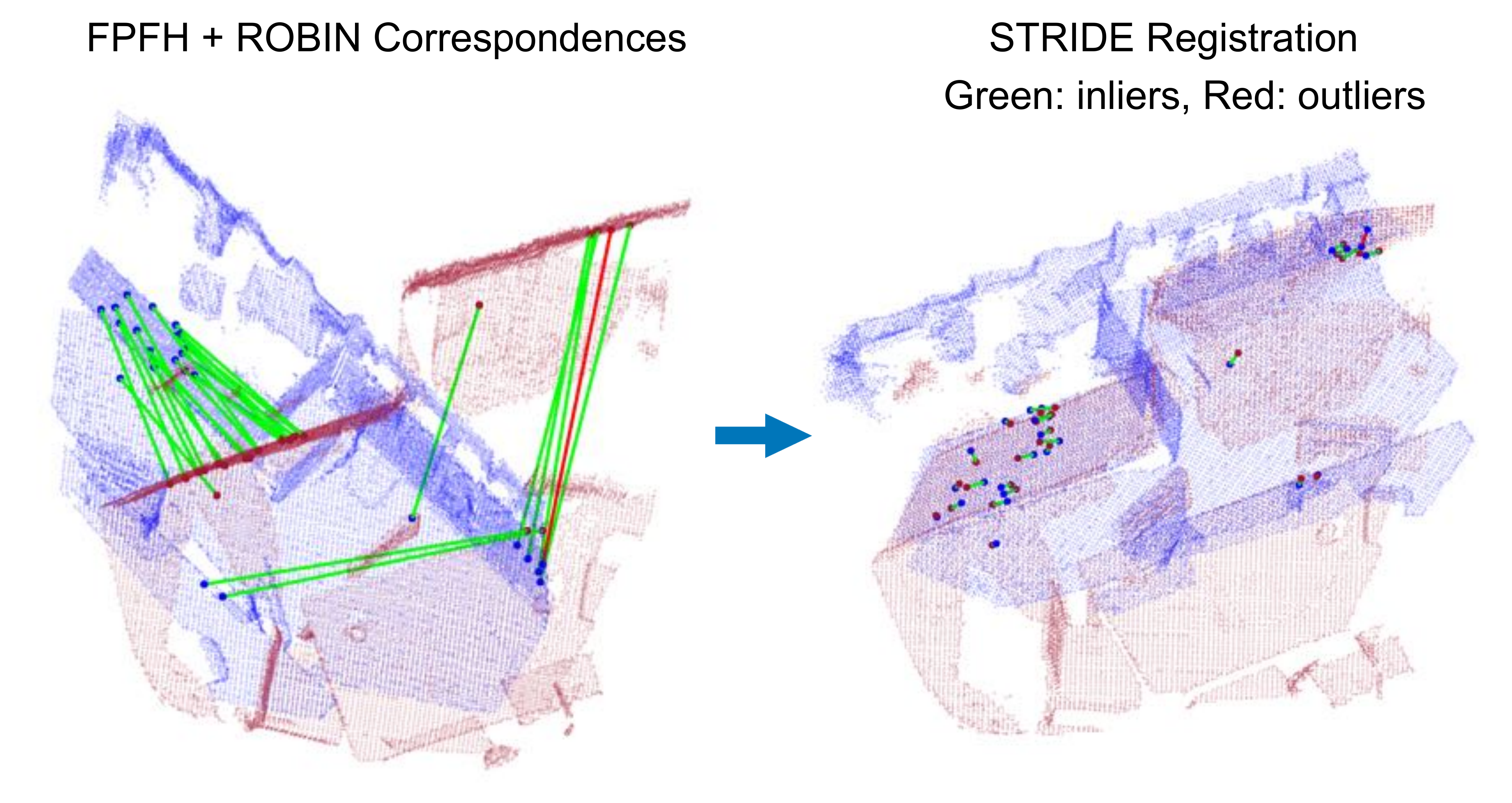}
		\\
		{\subcapsize (c) $\MR$ error: $4.3^{\circ}$, $\vt$ error: $1.7\ee{-1}$,  $\subopt = 2.1\ee{-8}$, time: $117$ [s]}
		\end{minipage}

		& 
		\myhspace
		\begin{minipage}{\mpwtwo}%
		\centering%
		\includegraphics[width=0.9\columnwidth]{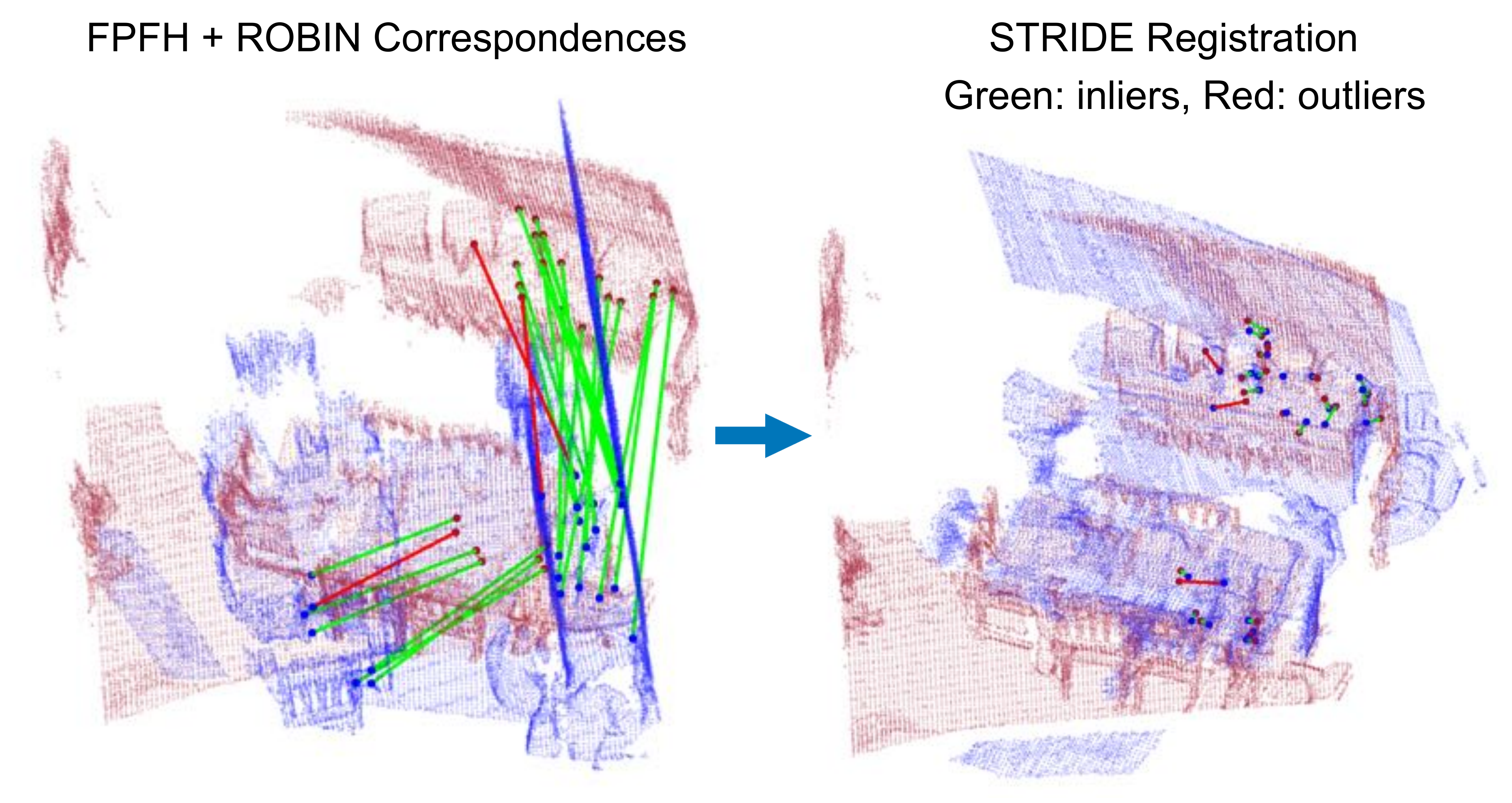}
		\\
		{\subcapsize (d) $\MR$ error: $2.3^{\circ}$, $\vt$ error: $1.5\ee{-1}$,  $\subopt = 2.2\ee{-8}$, time: $134$ [s]}
		\end{minipage}

		\\

		\myhspace
		\begin{minipage}{\mpwtwo}%
		\centering%
		\includegraphics[width=0.9\columnwidth]{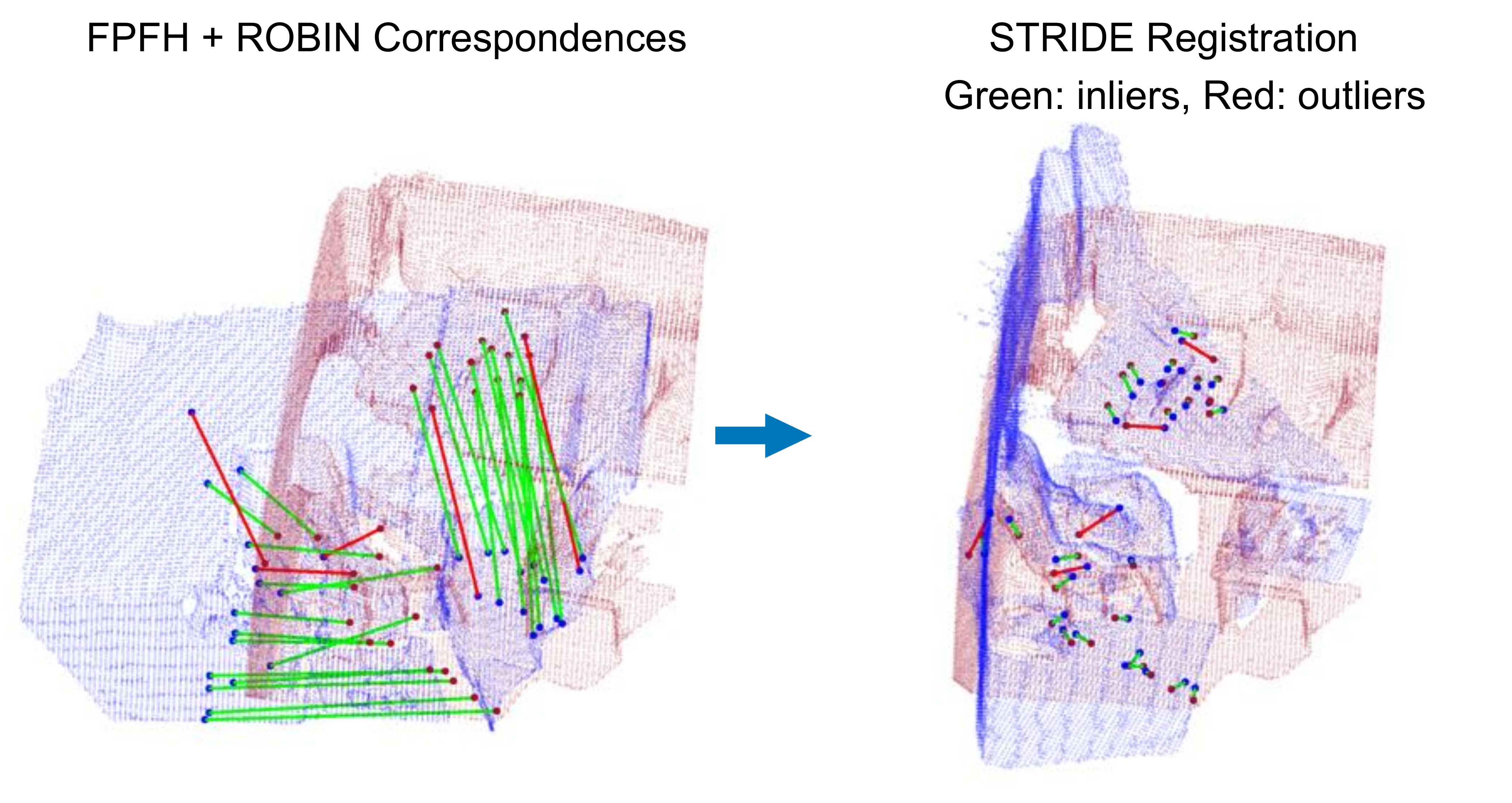}
		\\
		{\subcapsize (e) $\MR$ error: $4.5^{\circ}$, $\vt$ error: $2.3\ee{-1}$,  $\subopt = 1.3\ee{-8}$, time: $176$ [s]}
		\end{minipage}

		& 
		\myhspace
		\begin{minipage}{\mpwtwo}%
		\centering%
		\includegraphics[width=0.9\columnwidth]{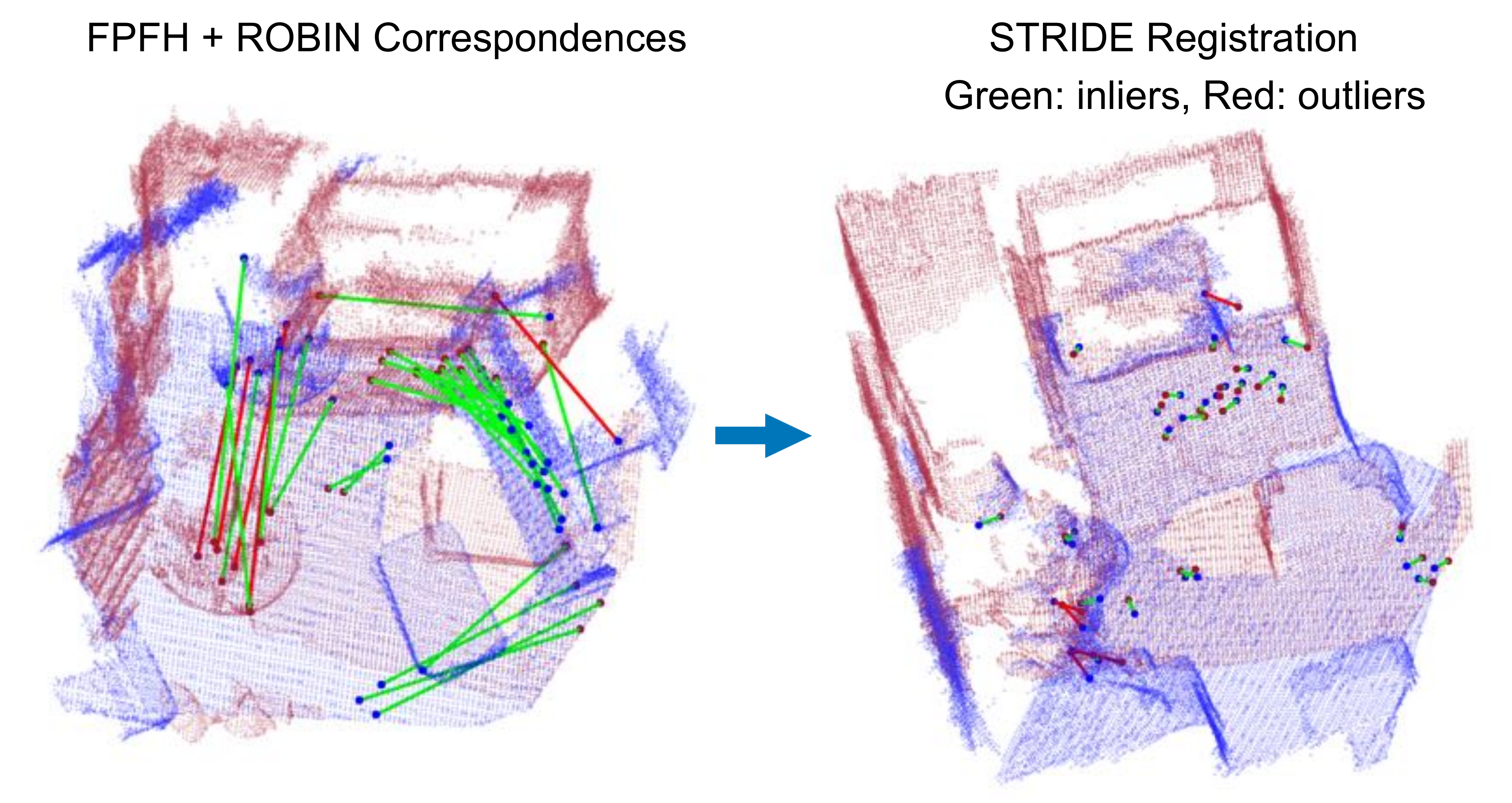}
		\\
		{\subcapsize (f) $\MR$ error: $3.1^{\circ}$, $\vt$ error: $1.6\ee{-1}$,  $\subopt = 2.4\ee{-8}$, time: $187$ [s]}
		\end{minipage}

		\\

		\myhspace
		\begin{minipage}{\mpwtwo}%
		\centering%
		\includegraphics[width=0.9\columnwidth]{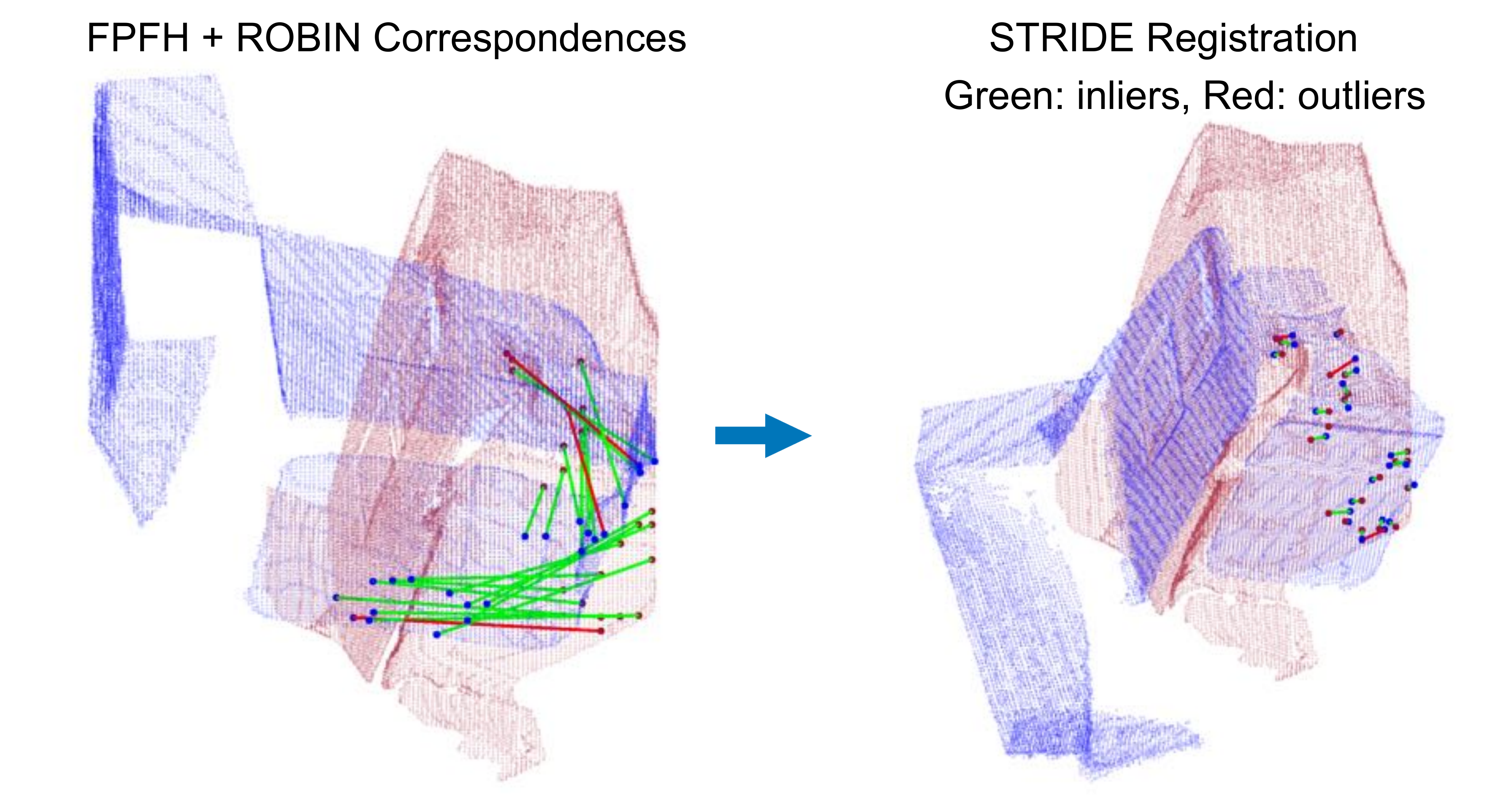}
		\\
		{\subcapsize (g) $\MR$ error: $2.4^{\circ}$, $\vt$ error: $8.9\ee{-2}$,  $\subopt = 2.4\ee{-8}$, time: $117$ [s]}
		\end{minipage}

		& 
		\myhspace
		\begin{minipage}{\mpwtwo}%
		\centering%
		\includegraphics[width=0.9\columnwidth]{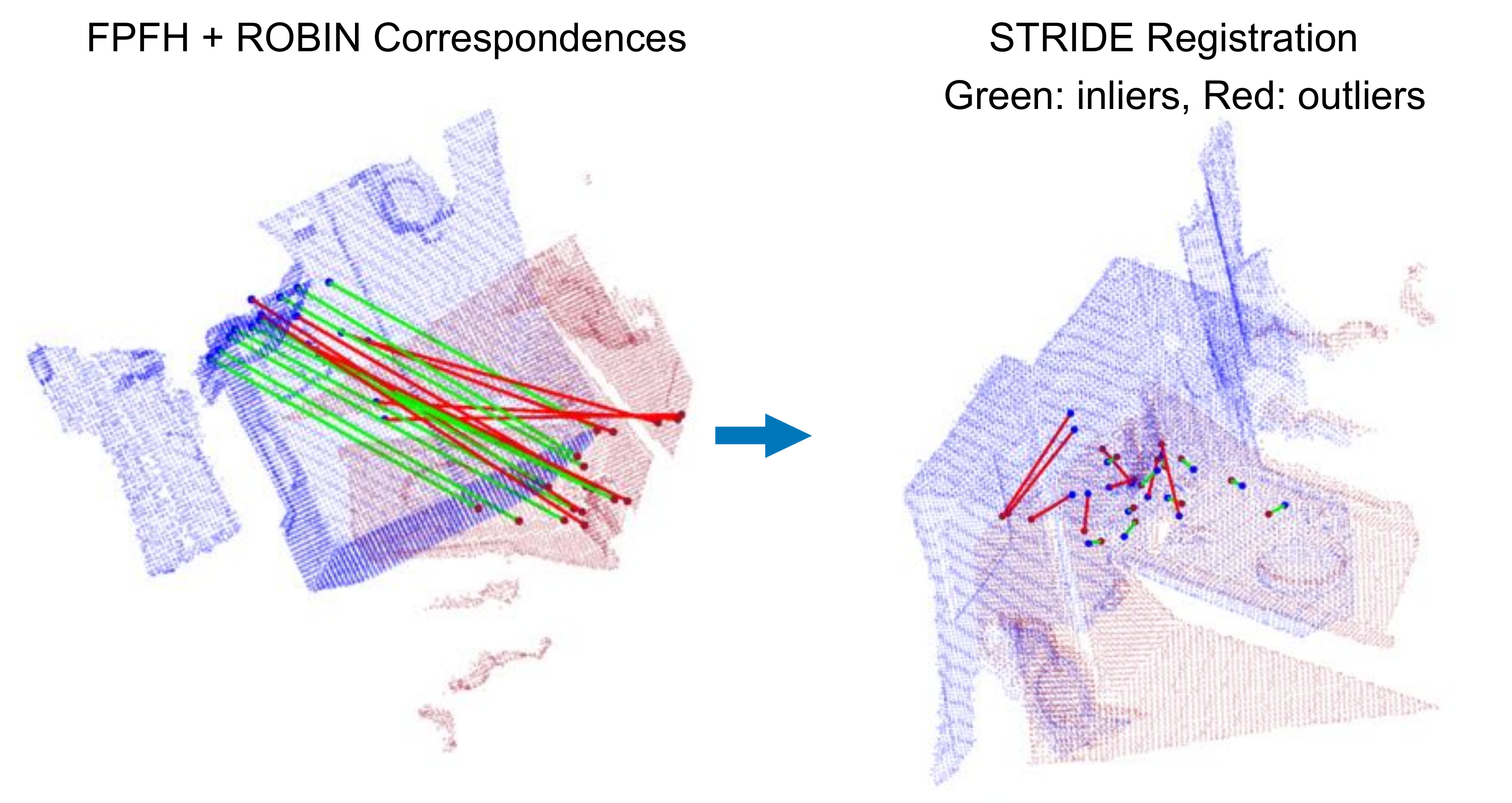}
		\\
		{\subcapsize (h) $\MR$ error: $3.2^{\circ}$, $\vt$ error: $1.5\ee{-1}$,  $\subopt = 4.5\ee{-8}$, time: $81$ [s]}
		\end{minipage}
		\\

		\myhspace
		\begin{minipage}{\mpwtwo}%
		\centering%
		\includegraphics[width=0.9\columnwidth]{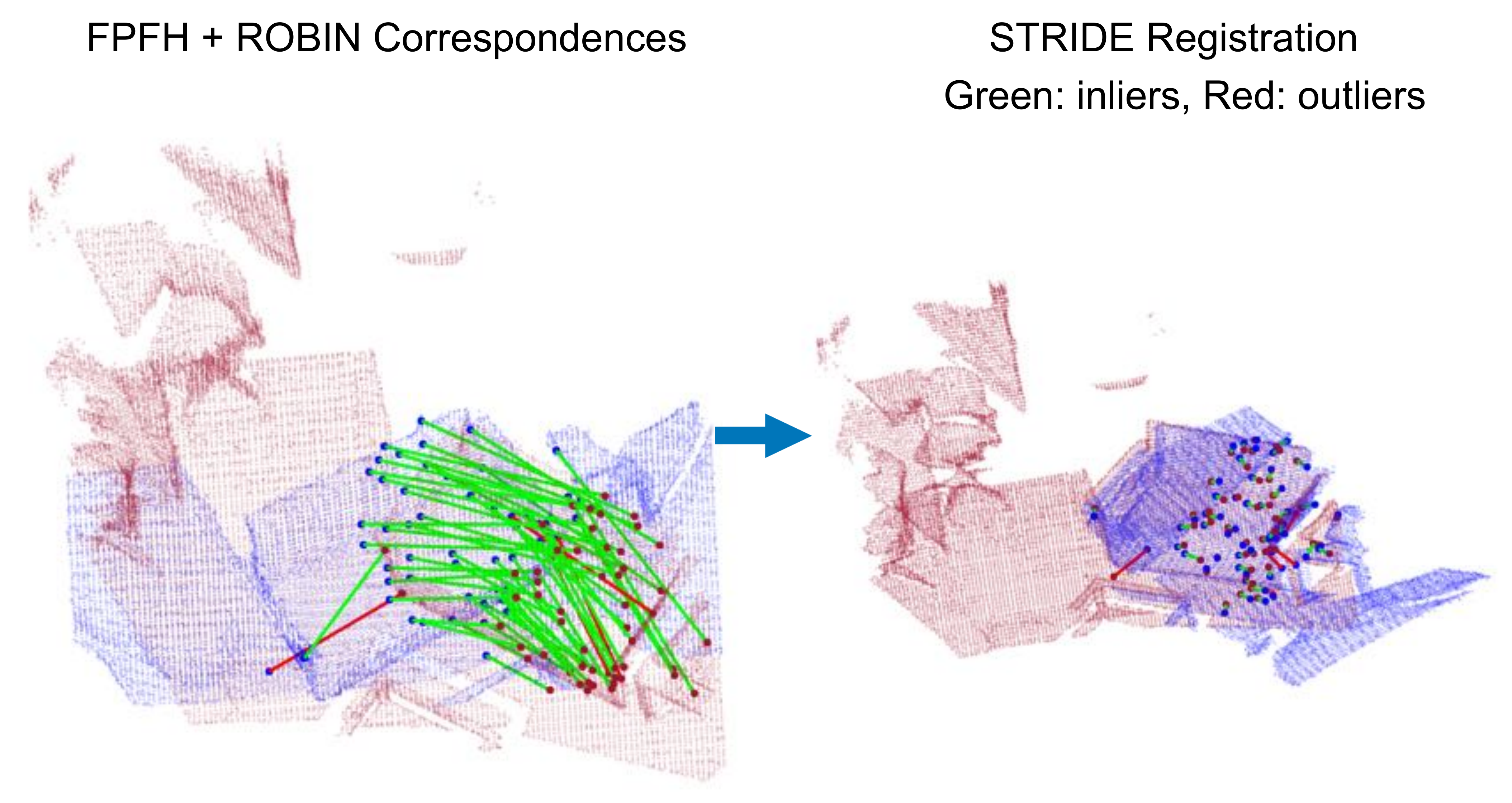}
		\\
		{\subcapsize (i) $\MR$ error: $3.5^{\circ}$, $\vt$ error: $1.3\ee{-1}$,  $\subopt = 1.7\ee{-6}$, time: $874$ [s]}
		\end{minipage}

		& 
		\myhspace
		\begin{minipage}{\mpwtwo}%
		\centering%
		\includegraphics[width=0.9\columnwidth]{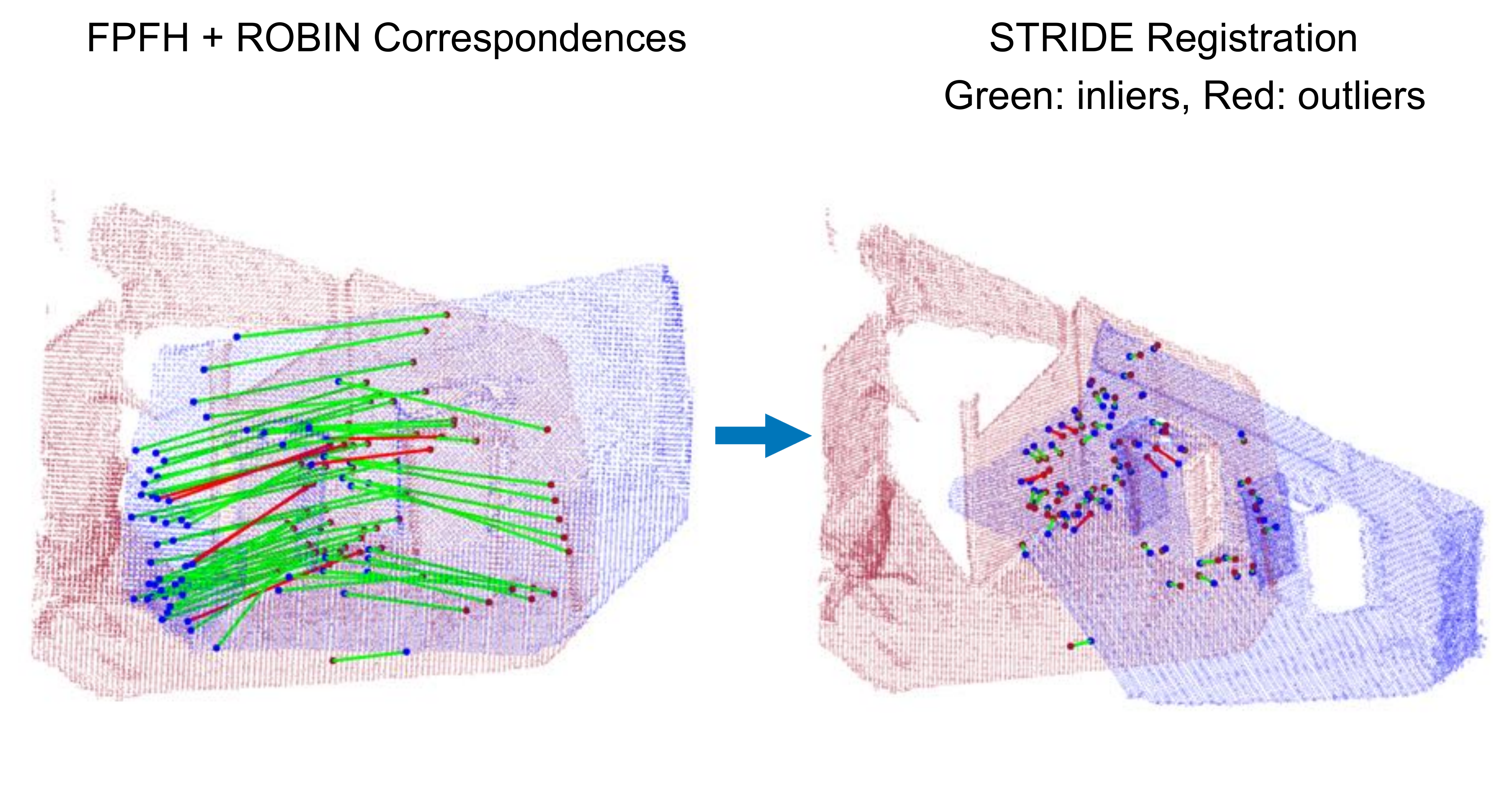}
		\\
		{\subcapsize (j) $\MR$ error: $3.4^{\circ}$, $\vt$ error: $1.9\ee{-1}$,  $\subopt = 3.8\ee{-10}$, time: $1560$ [s]}
		\end{minipage}
	\end{tabular}
	\end{minipage} 
	\caption{Extra scan matching results on {\threedmatch} \cite{Zeng17cvpr-3dmatch}.
	\label{fig:supp-3dmatch}} 
	\vspace{-7mm} 
	\end{center}
\end{figure*}

\subsection{Absolute Pose Estimation on {\speed}}
We provide extra satellite pose estimation results by {\stride} on the {\speed} dataset \cite{Sharma19arxiv-SPEED} in Fig. \ref{fig:supp-speed-results}. In all six image instances with $2$-$5$ outliers, {\stride} returned accurate pose estimates with global optimality certificates.

\renewcommand{\mpwfour}{4.2cm}
\begin{figure*}[t]
	\begin{center}
	\begin{minipage}{\textwidth}
	\begin{tabular}{cccc}%
	$2$ outliers & $3$ outliers & $4$ outliers & $5$ outliers \\
			\begin{minipage}{\mpwfour}%
			\centering%
			\includegraphics[width=\columnwidth]{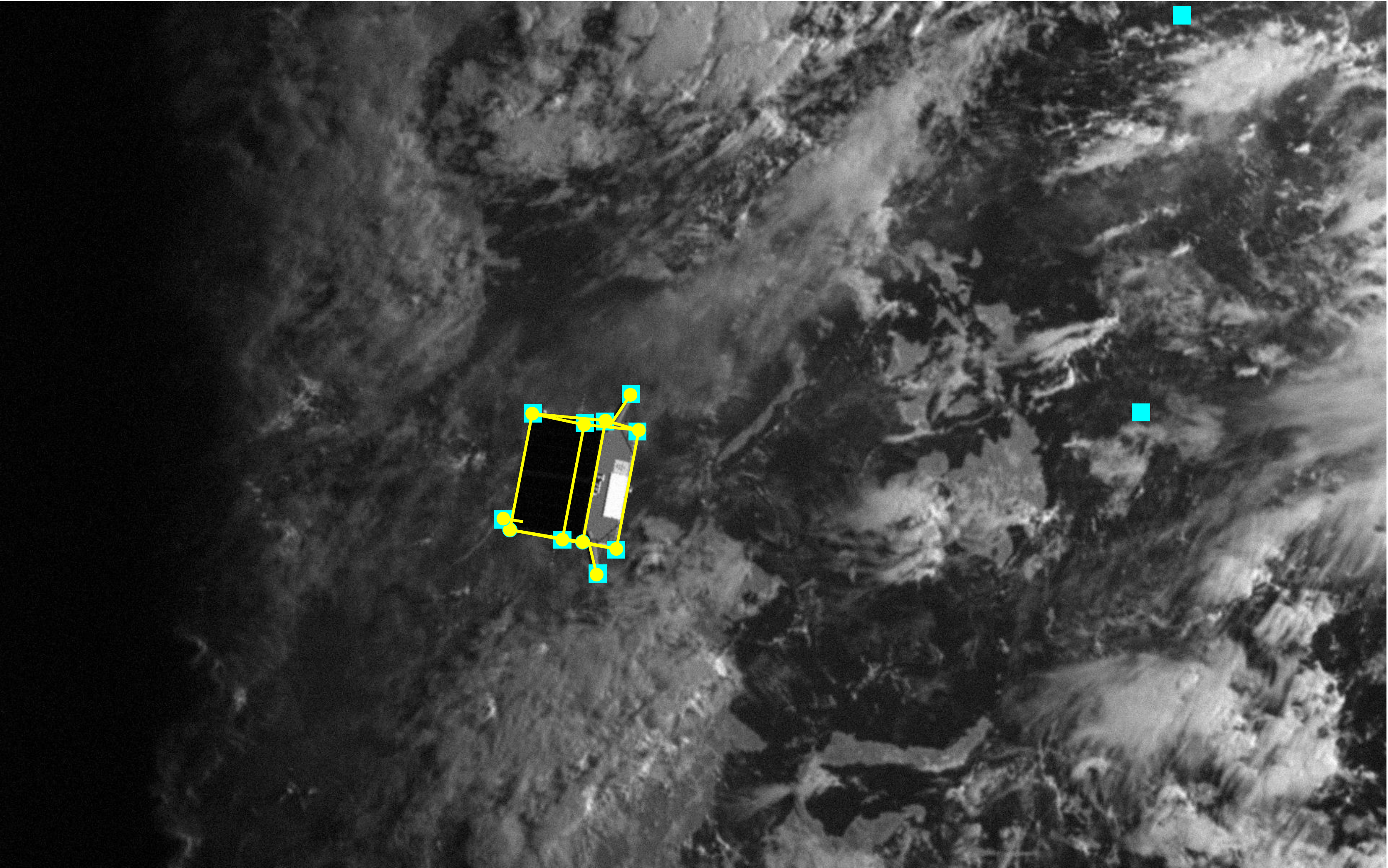}\\
			 $\MR$ error: $0.48^{\circ}$, $\vt$ error: $0.05$ \\ 
			 $\subopt=2.9\ee{-8}$, time: $51$ [s]
			\end{minipage}
		&  
			\begin{minipage}{\mpwfour}%
			\centering%
			\includegraphics[width=\columnwidth]{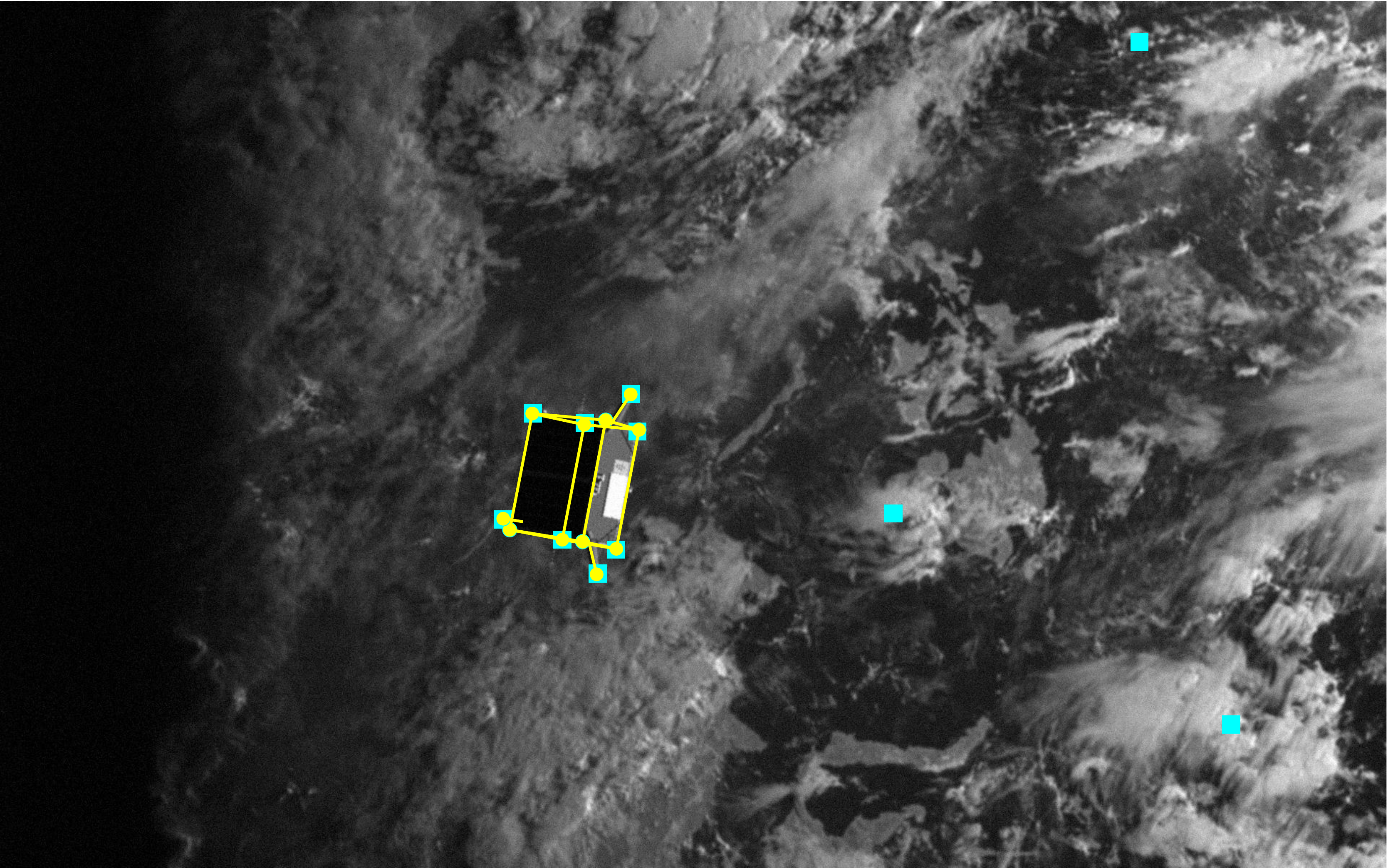}\\
			 $\MR$ error: $0.37^{\circ}$, $\vt$ error: $0.07$ \\ 
			 $\subopt=1.3\ee{-7}$, time: $47$ [s]
			\end{minipage}
		&  
			\begin{minipage}{\mpwfour}%
			\centering%
			\includegraphics[width=\columnwidth]{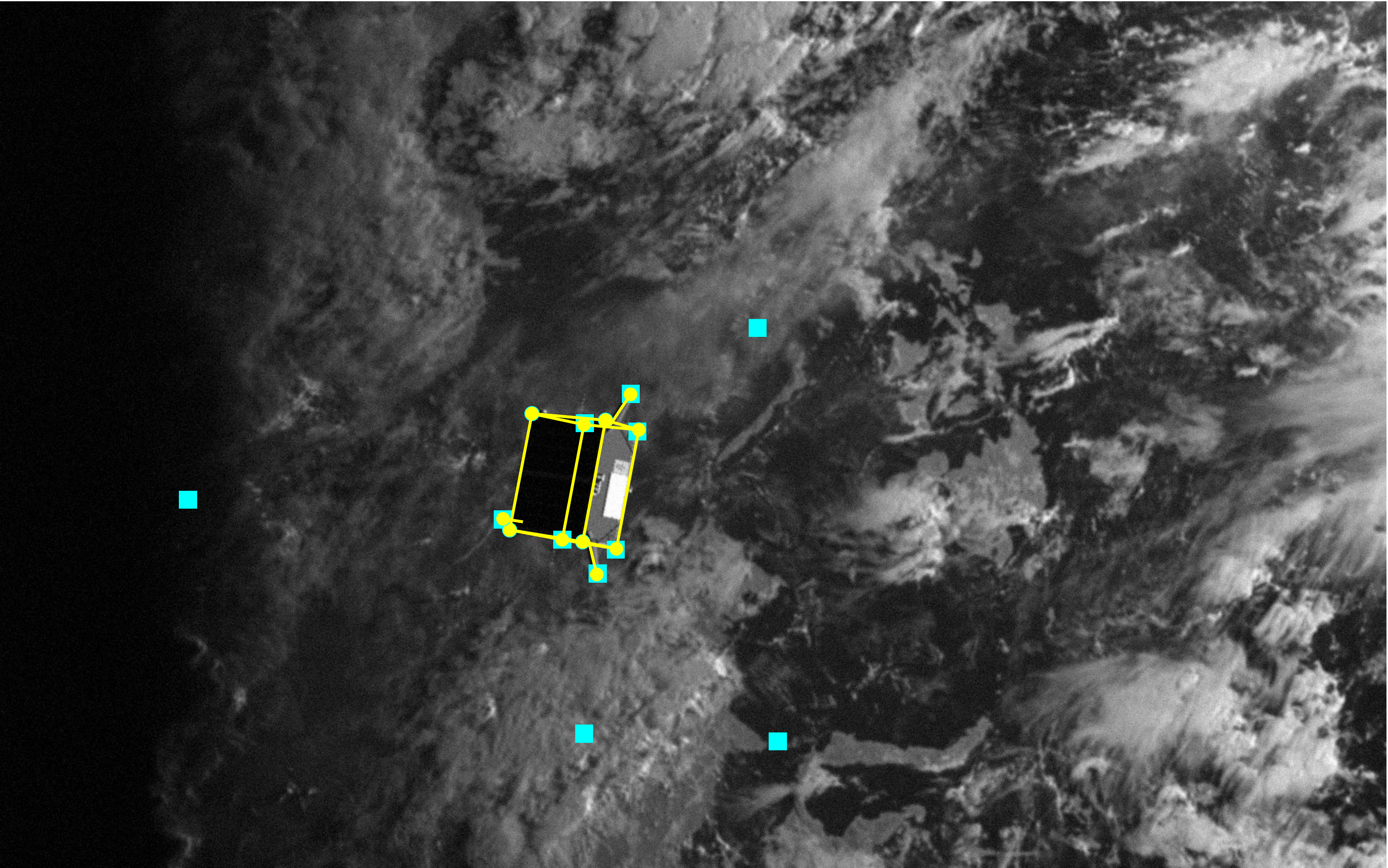}\\
			 $\MR$ error: $0.37^{\circ}$, $\vt$ error: $0.07$ \\ 
			 $\subopt=1.8\ee{-8}$, time: $109$ [s]
			\end{minipage}
		&  
			\begin{minipage}{\mpwfour}%
			\centering%
			\includegraphics[width=\columnwidth]{ape_img_1_5-eps-converted-to.pdf}\\
			 $\MR$ error: $0.34^{\circ}$, $\vt$ error: $0.07$ \\ 
			 $\subopt=1.4\ee{-9}$, time: $107$ [s]
			\end{minipage} \vspace{1mm}
		\\

			\begin{minipage}{\mpwfour}%
			\centering%
			\includegraphics[width=\columnwidth]{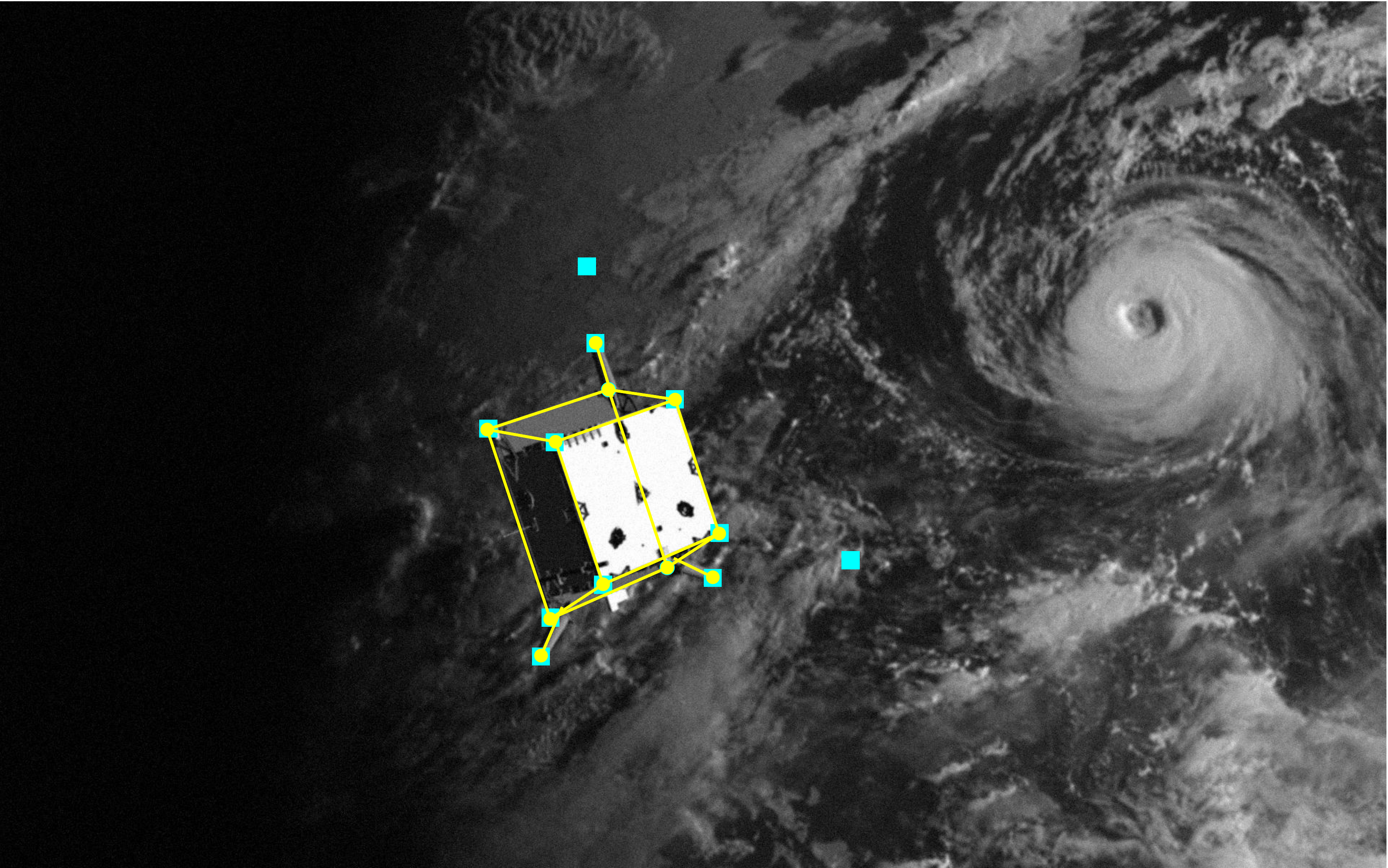}\\
			 $\MR$ error: $0.21^{\circ}$, $\vt$ error: $0.007$ \\ 
			 $\subopt=5.1\ee{-8}$, time: $46$ [s]
			\end{minipage}
		&  
			\begin{minipage}{\mpwfour}%
			\centering%
			\includegraphics[width=\columnwidth]{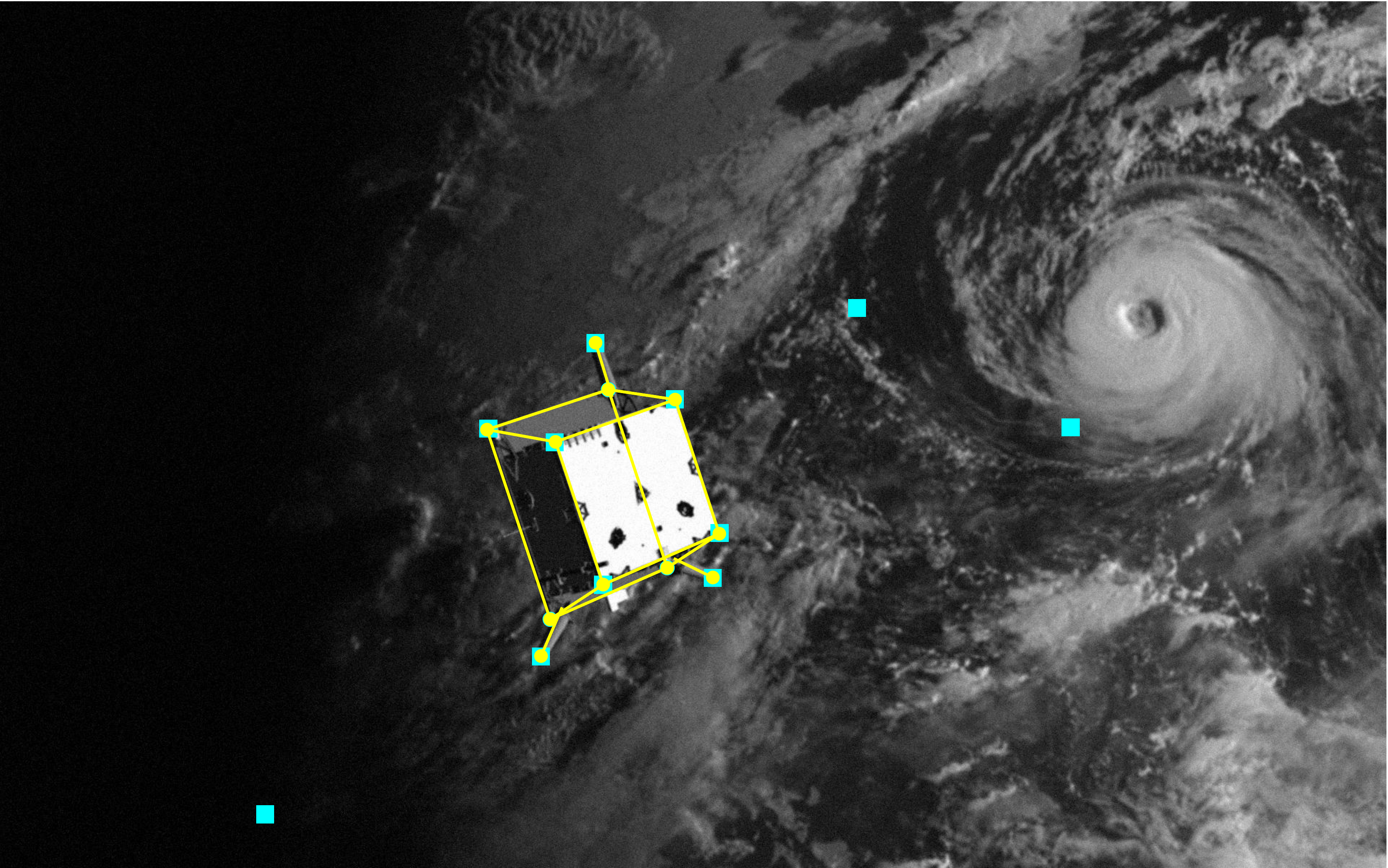}\\
			 $\MR$ error: $0.23^{\circ}$, $\vt$ error: $0.02$ \\ 
			 $\subopt=5.0\ee{-8}$, time: $49$ [s]
			\end{minipage}
		&  
			\begin{minipage}{\mpwfour}%
			\centering%
			\includegraphics[width=\columnwidth]{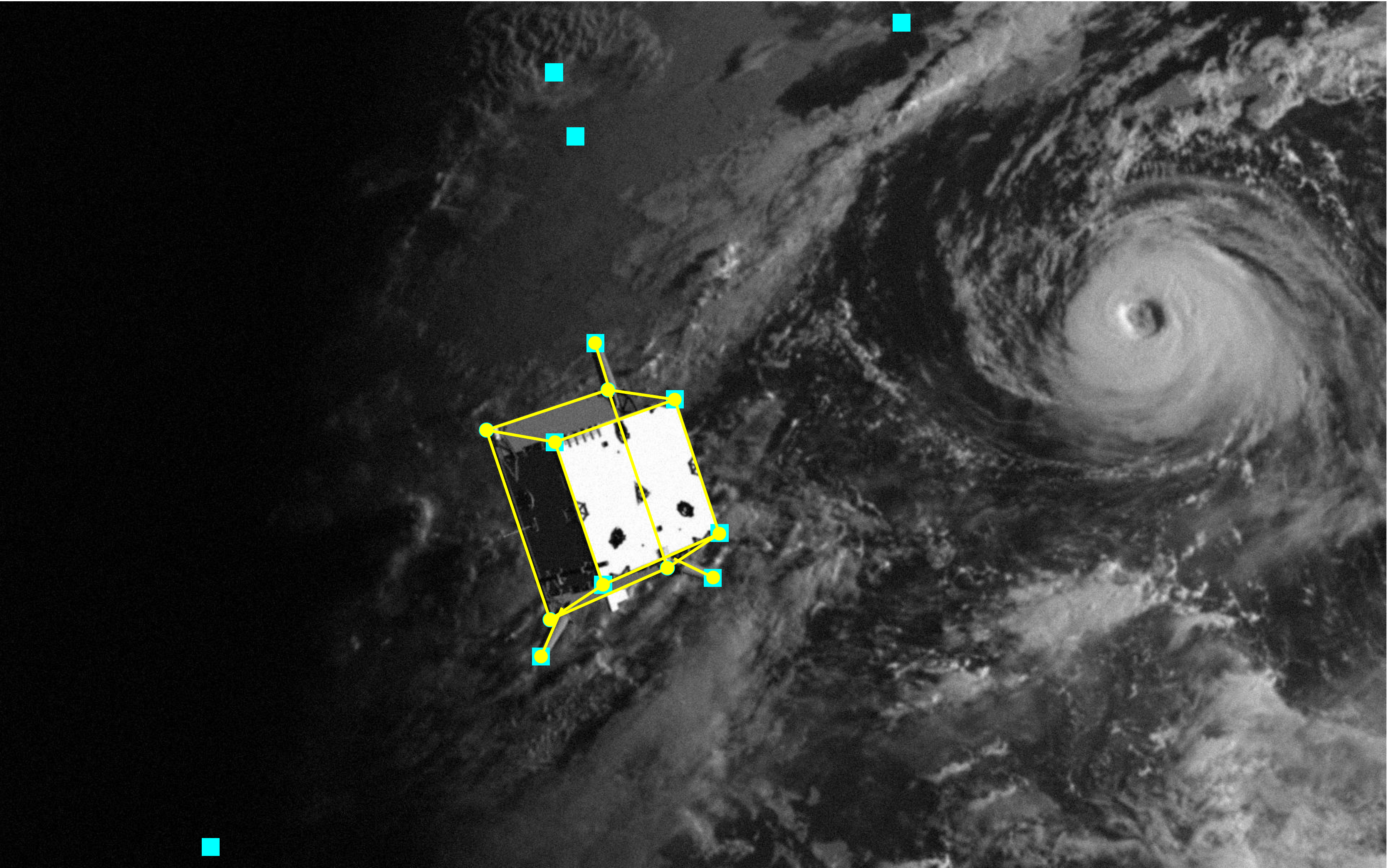}\\
			 $\MR$ error: $0.07^{\circ}$, $\vt$ error: $0.02$ \\ 
			 $\subopt=3.6\ee{-7}$, time: $132$ [s]
			\end{minipage}
		&  
			\begin{minipage}{\mpwfour}%
			\centering%
			\includegraphics[width=\columnwidth]{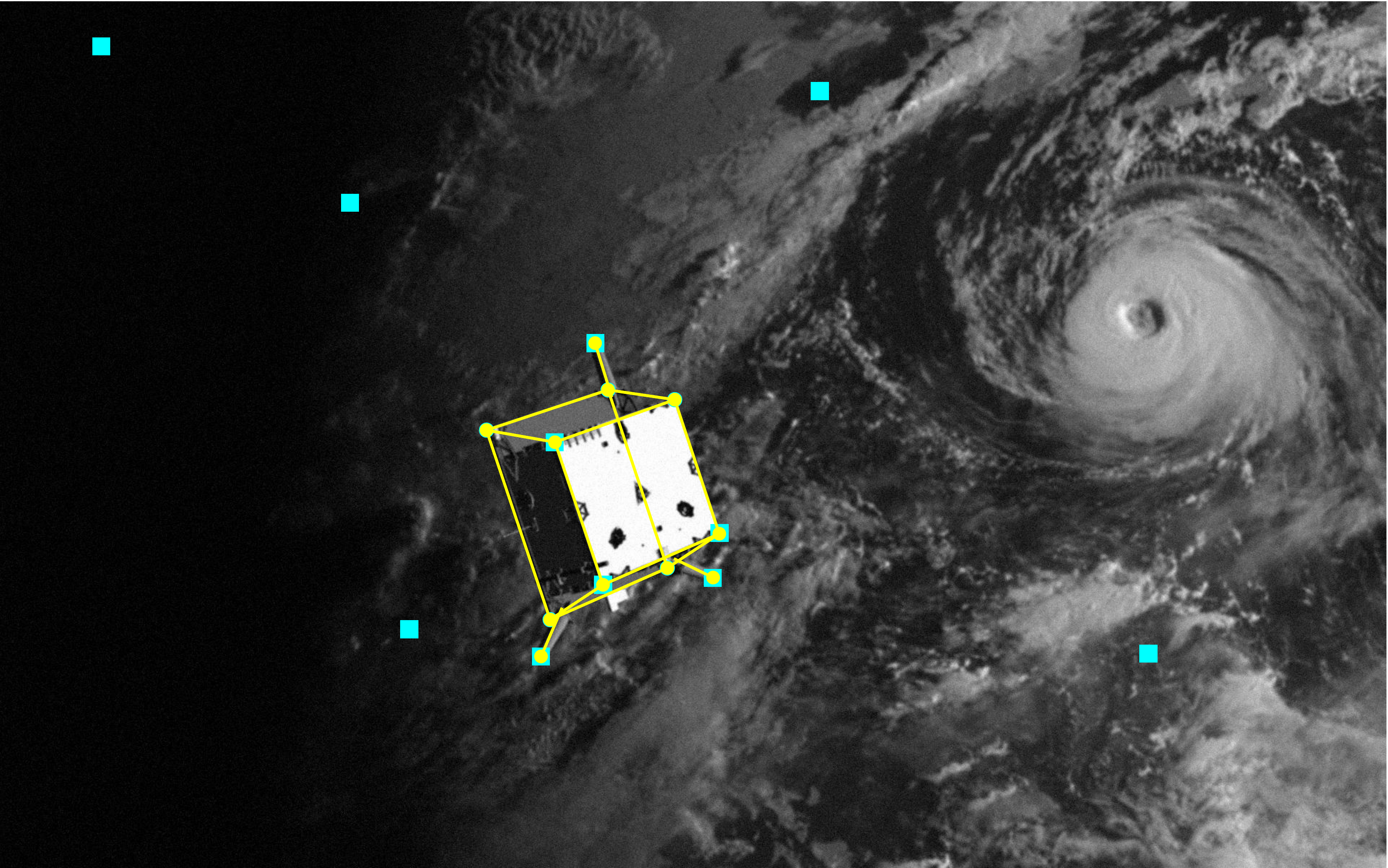}\\
			 $\MR$ error: $0.06^{\circ}$, $\vt$ error: $0.02$ \\ 
			 $\subopt=1.3\ee{-7}$, time: $95$ [s]
			\end{minipage} \vspace{1mm}
		\\

			\begin{minipage}{\mpwfour}%
			\centering%
			\includegraphics[width=\columnwidth]{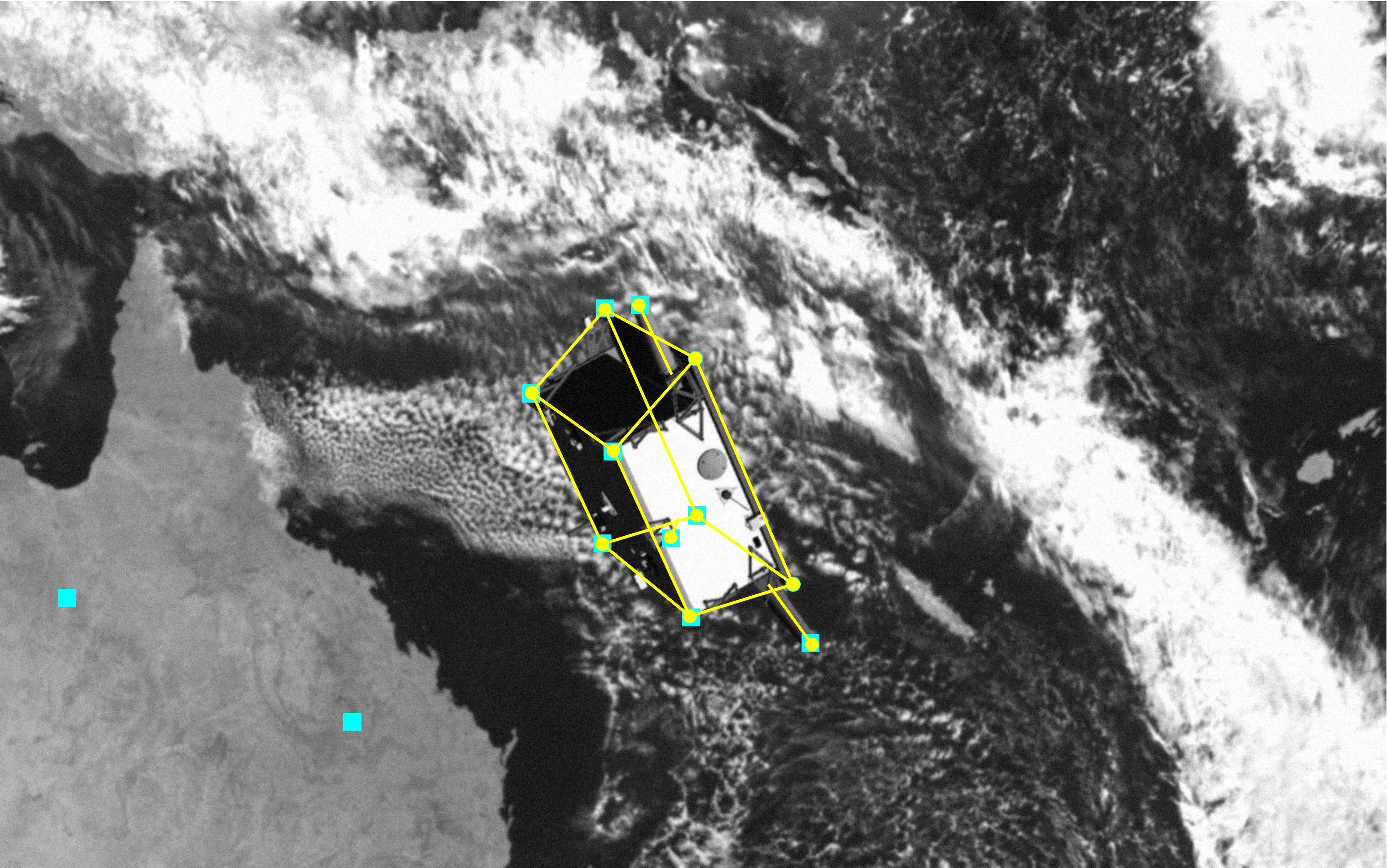}\\
			 $\MR$ error: $0.28^{\circ}$, $\vt$ error: $0.02$ \\ 
			 $\subopt=3.9\ee{-8}$, time: $46$ [s]
			\end{minipage}
		&  
			\begin{minipage}{\mpwfour}%
			\centering%
			\includegraphics[width=\columnwidth]{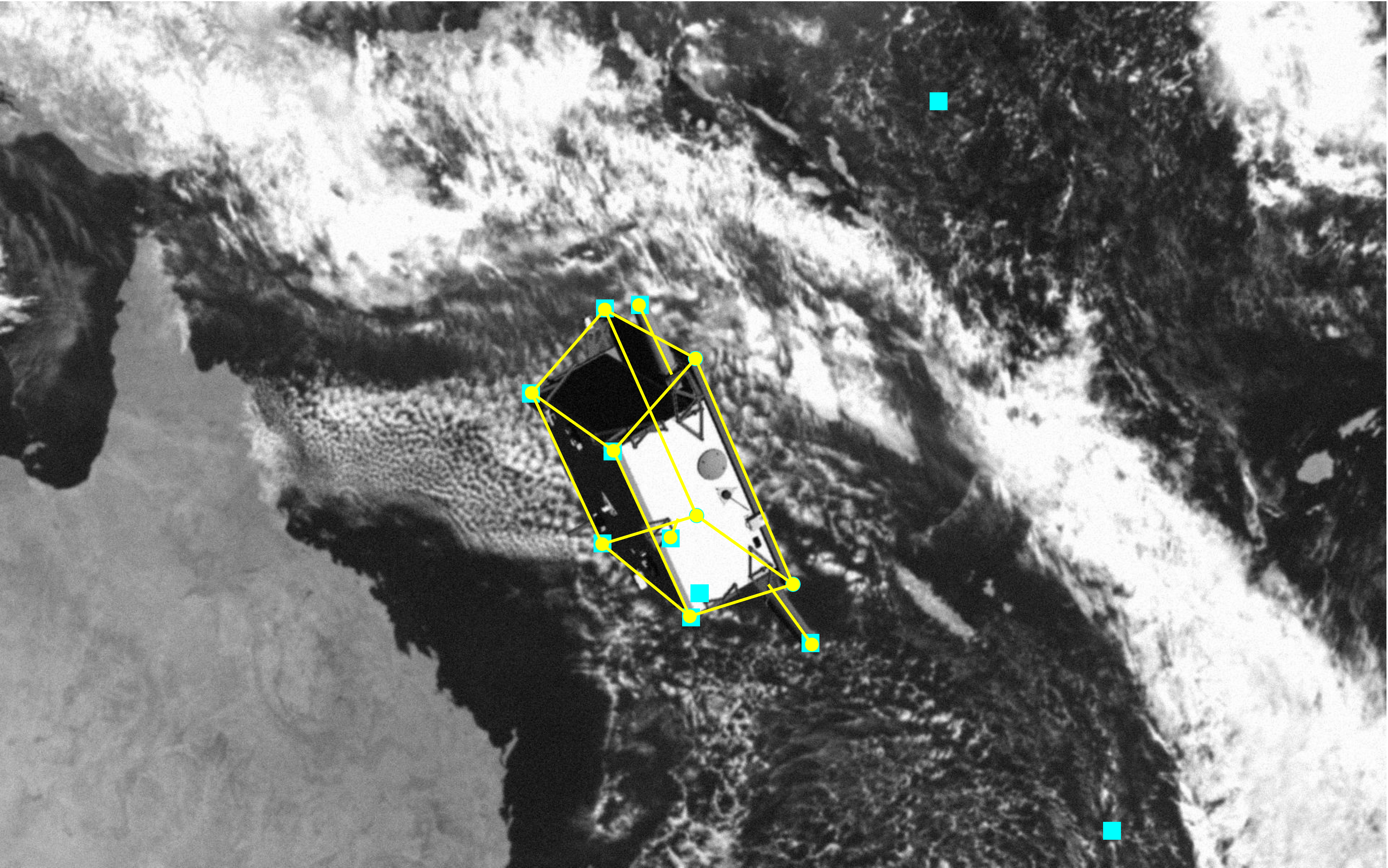}\\
			 $\MR$ error: $0.30^{\circ}$, $\vt$ error: $0.02$ \\ 
			 $\subopt=5.1\ee{-8}$, time: $47$ [s]
			\end{minipage}
		&  
			\begin{minipage}{\mpwfour}%
			\centering%
			\includegraphics[width=\columnwidth]{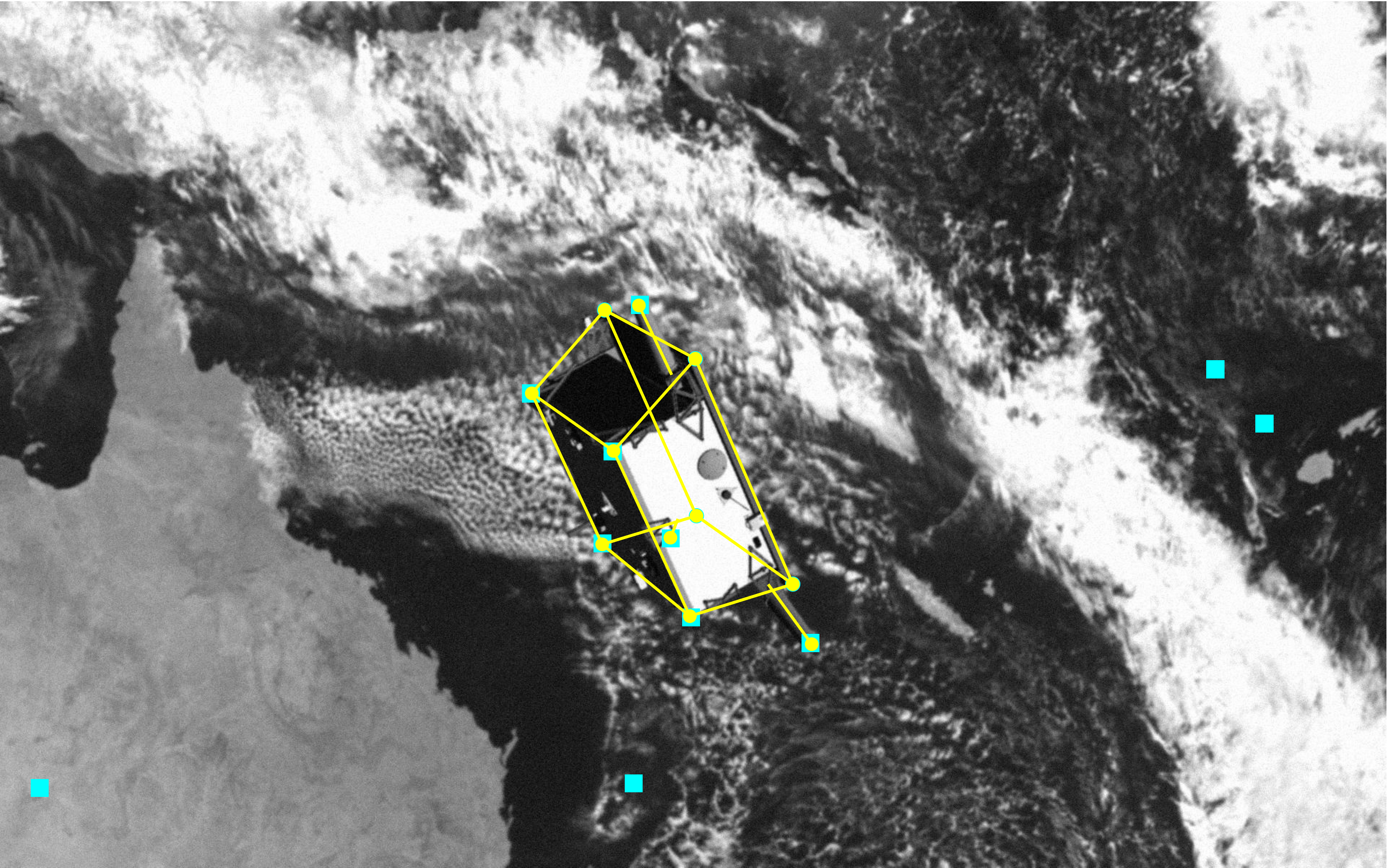}\\
			 $\MR$ error: $0.37^{\circ}$, $\vt$ error: $0.04$ \\ 
			 $\subopt=8.0\ee{-8}$, time: $93$ [s]
			\end{minipage}
		&  
			\begin{minipage}{\mpwfour}%
			\centering%
			\includegraphics[width=\columnwidth]{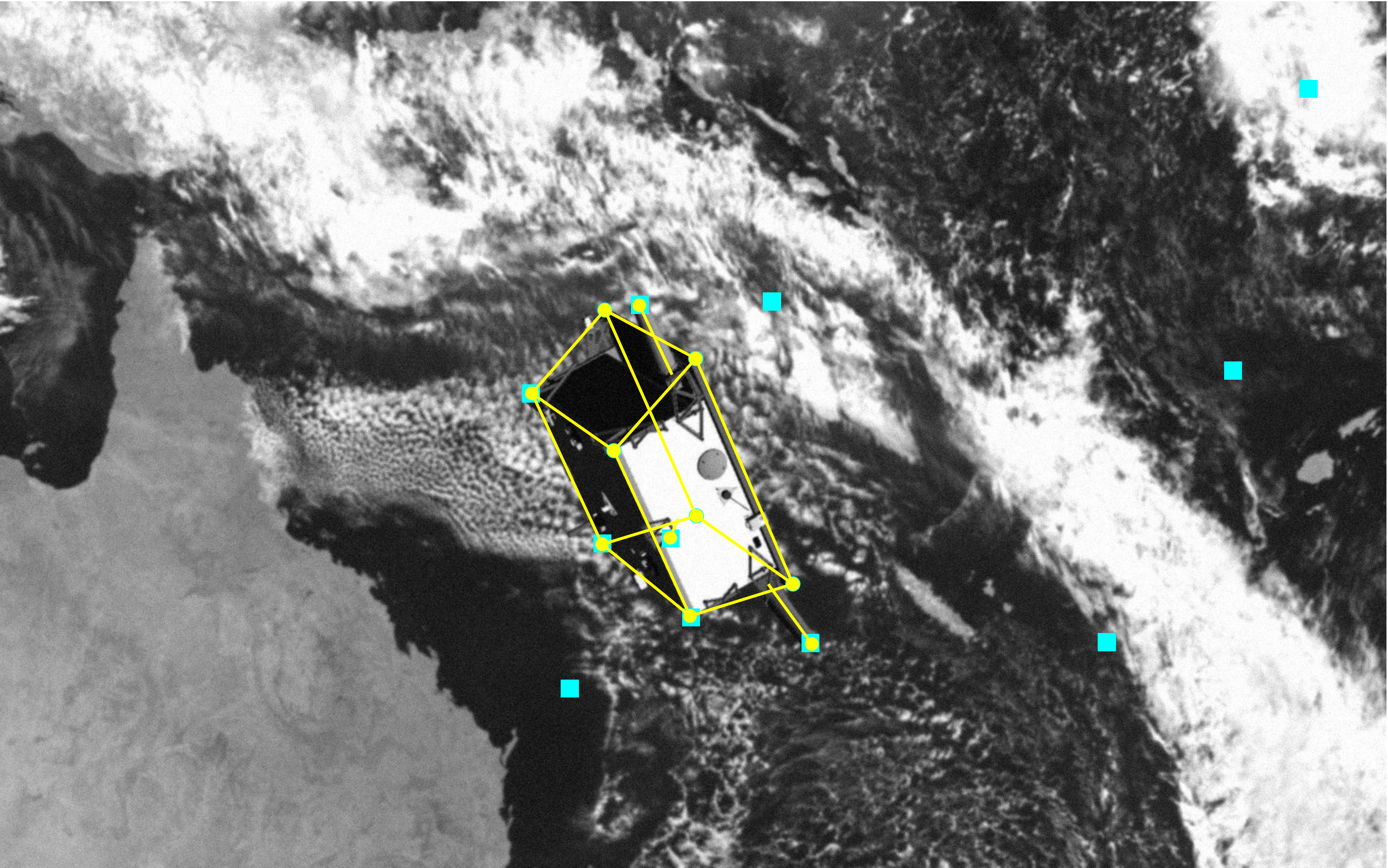}\\
			 $\MR$ error: $0.43^{\circ}$, $\vt$ error: $0.04$ \\ 
			 $\subopt=1.6\ee{-6}$, time: $155$ [s]
			\end{minipage} \vspace{1mm}
		\\

			\begin{minipage}{\mpwfour}%
			\centering%
			\includegraphics[width=\columnwidth]{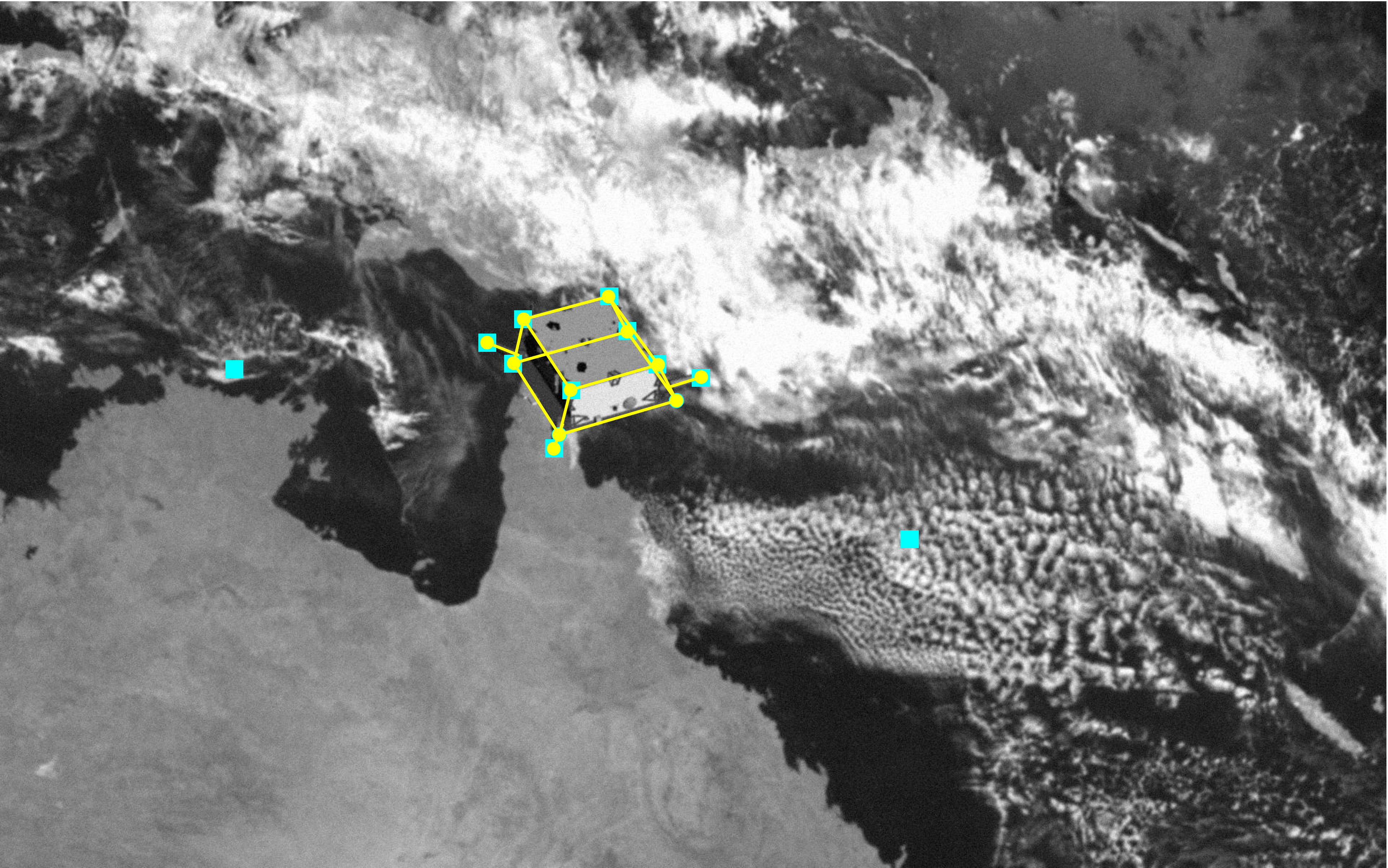}\\
			 $\MR$ error: $0.64^{\circ}$, $\vt$ error: $0.06$ \\ 
			 $\subopt=1.7\ee{-7}$, time: $52$ [s]
			\end{minipage}
		&  
			\begin{minipage}{\mpwfour}%
			\centering%
			\includegraphics[width=\columnwidth]{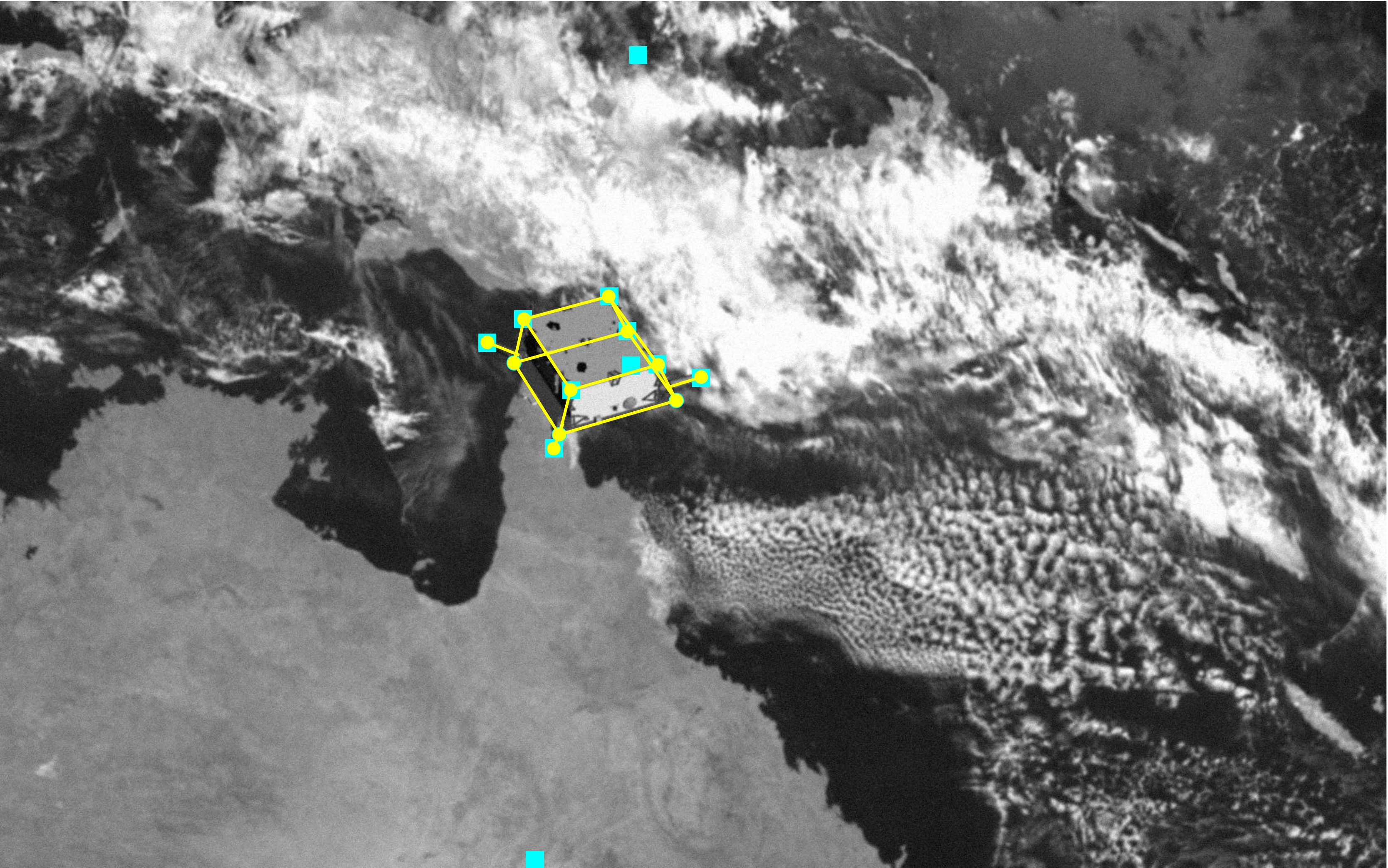}\\
			 $\MR$ error: $0.68^{\circ}$, $\vt$ error: $0.04$ \\ 
			 $\subopt=8.7\ee{-8}$, time: $47$ [s]
			\end{minipage}
		&  
			\begin{minipage}{\mpwfour}%
			\centering%
			\includegraphics[width=\columnwidth]{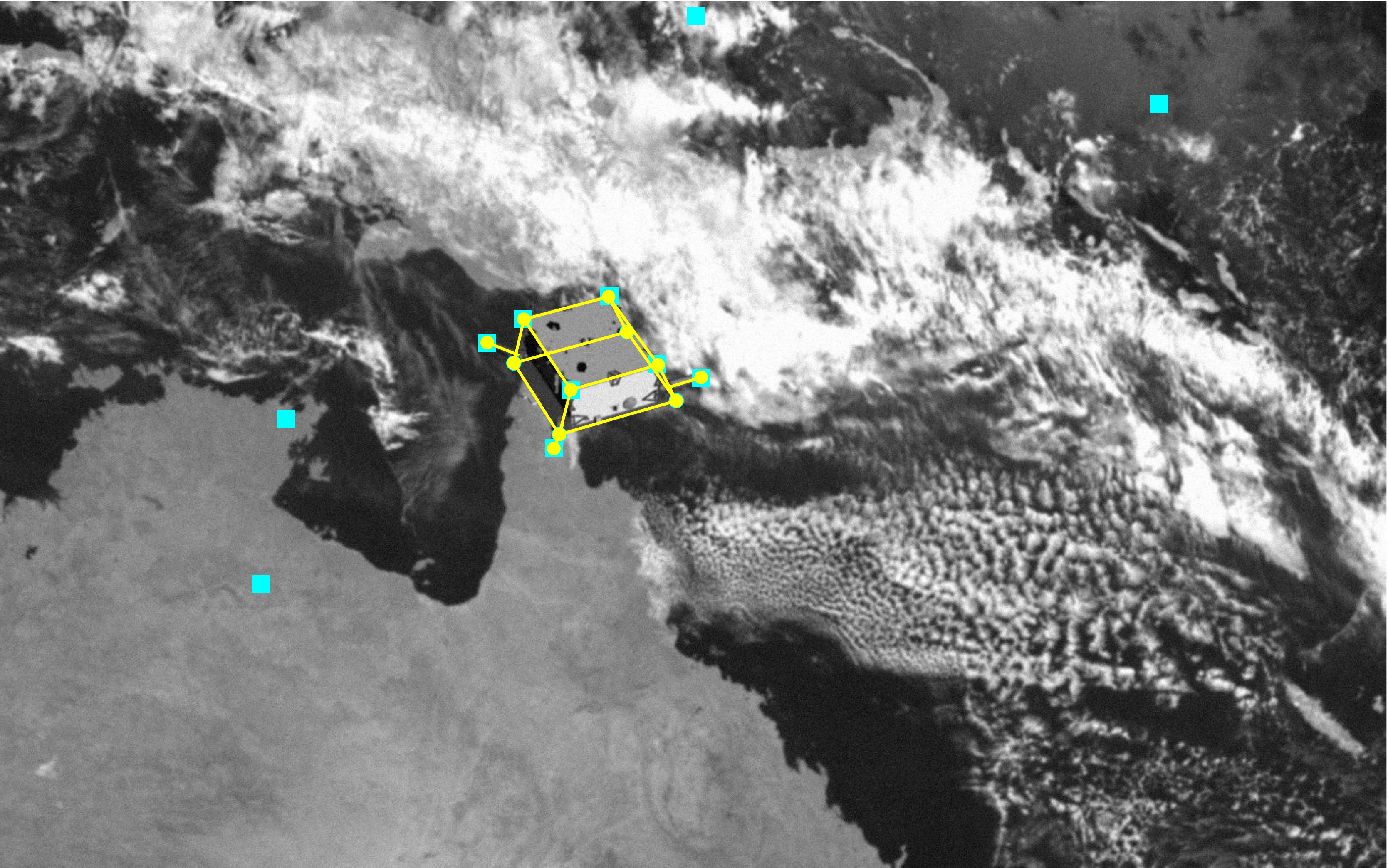}\\
			 $\MR$ error: $0.55^{\circ}$, $\vt$ error: $0.04$ \\ 
			 $\subopt=3.5\ee{-8}$, time: $127$ [s]
			\end{minipage}
		&  
			\begin{minipage}{\mpwfour}%
			\centering%
			\includegraphics[width=\columnwidth]{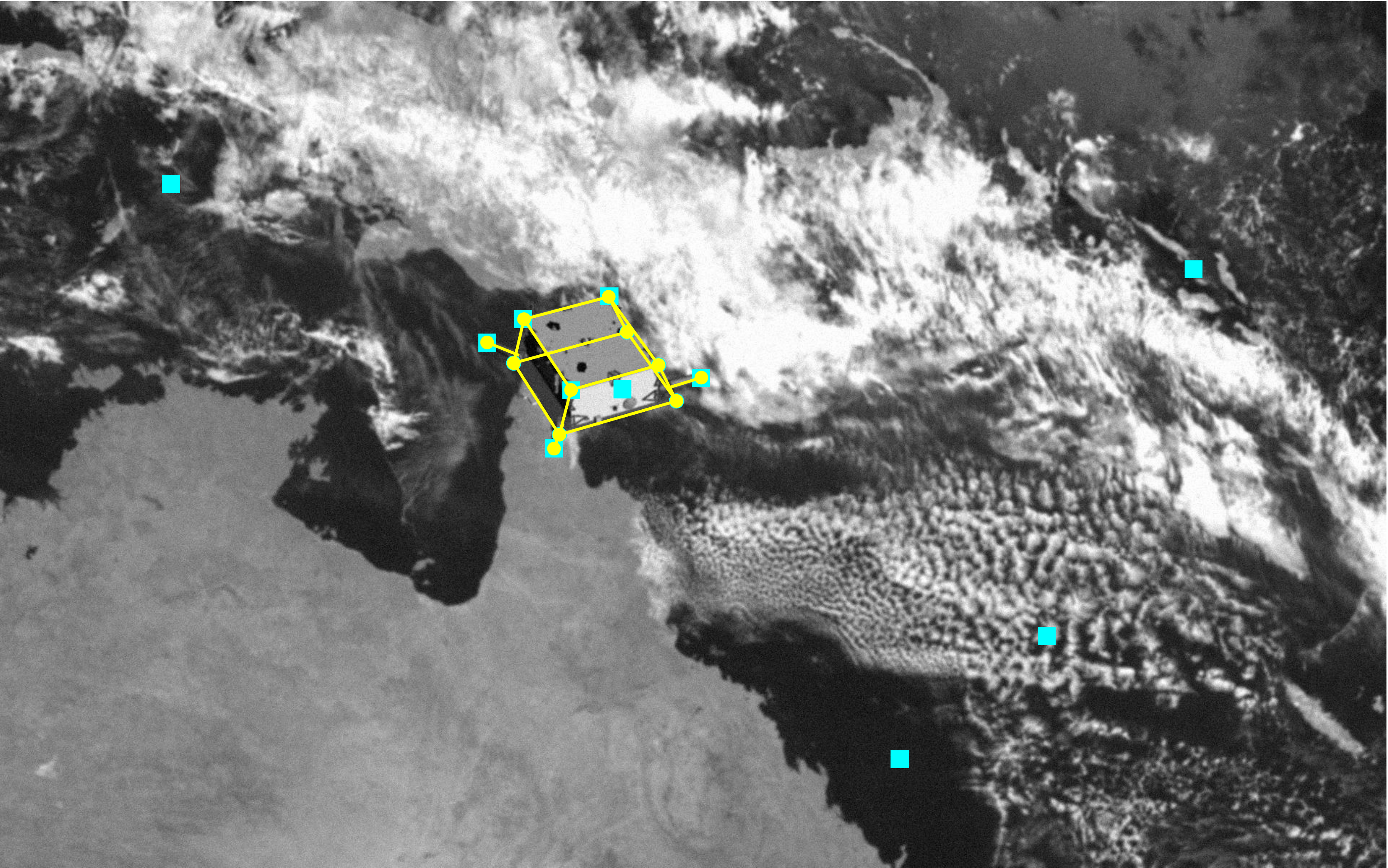}\\
			 $\MR$ error: $0.45^{\circ}$, $\vt$ error: $0.06$ \\ 
			 $\subopt=4.3\ee{-8}$, time: $64$ [s]
			\end{minipage} \vspace{1mm}
		\\

			\begin{minipage}{\mpwfour}%
			\centering%
			\includegraphics[width=\columnwidth]{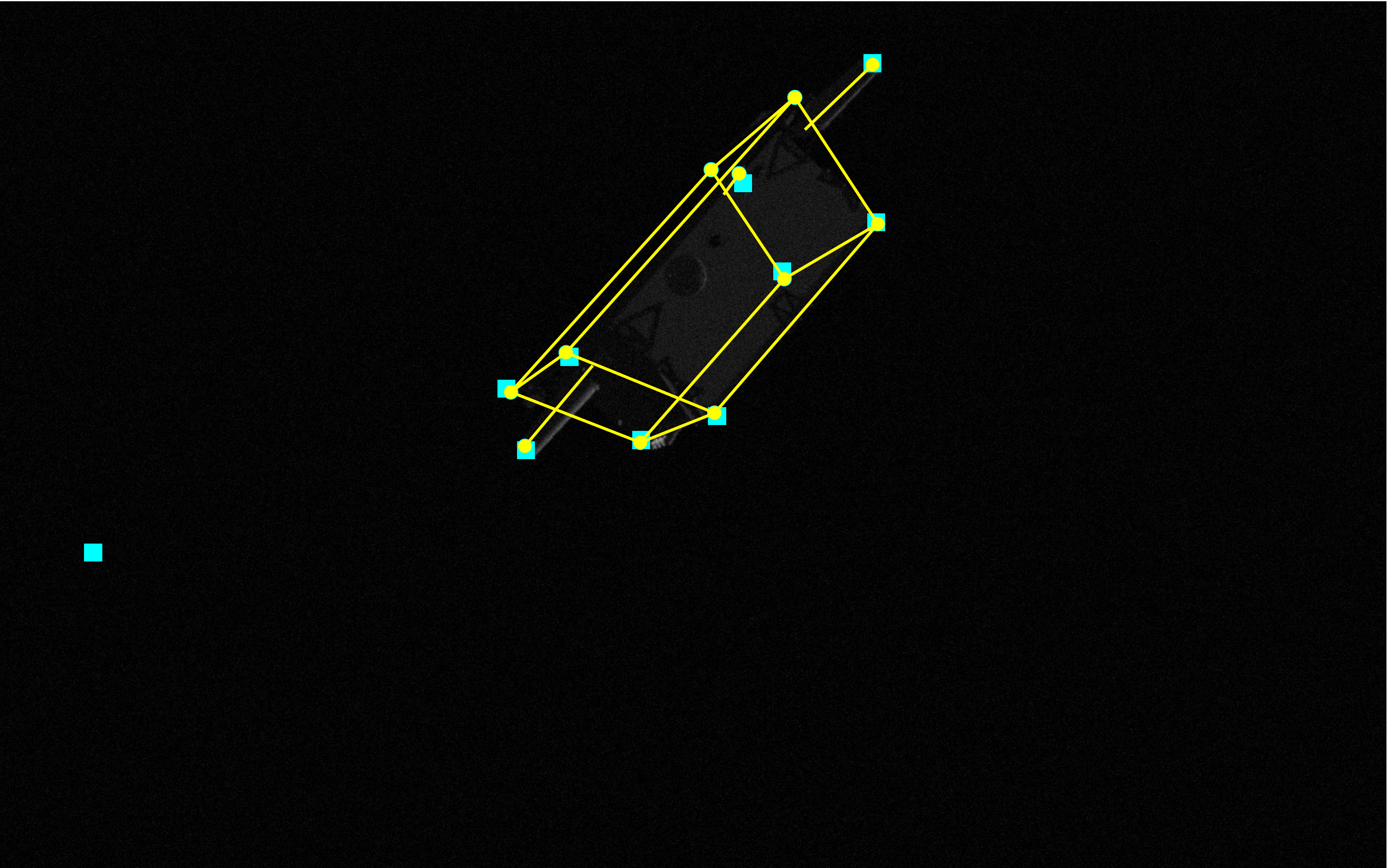}\\
			 $\MR$ error: $1.70^{\circ}$, $\vt$ error: $0.02$ \\ 
			 $\subopt=1.8\ee{-8}$, time: $47$ [s]
			\end{minipage}
		&  
			\begin{minipage}{\mpwfour}%
			\centering%
			\includegraphics[width=\columnwidth]{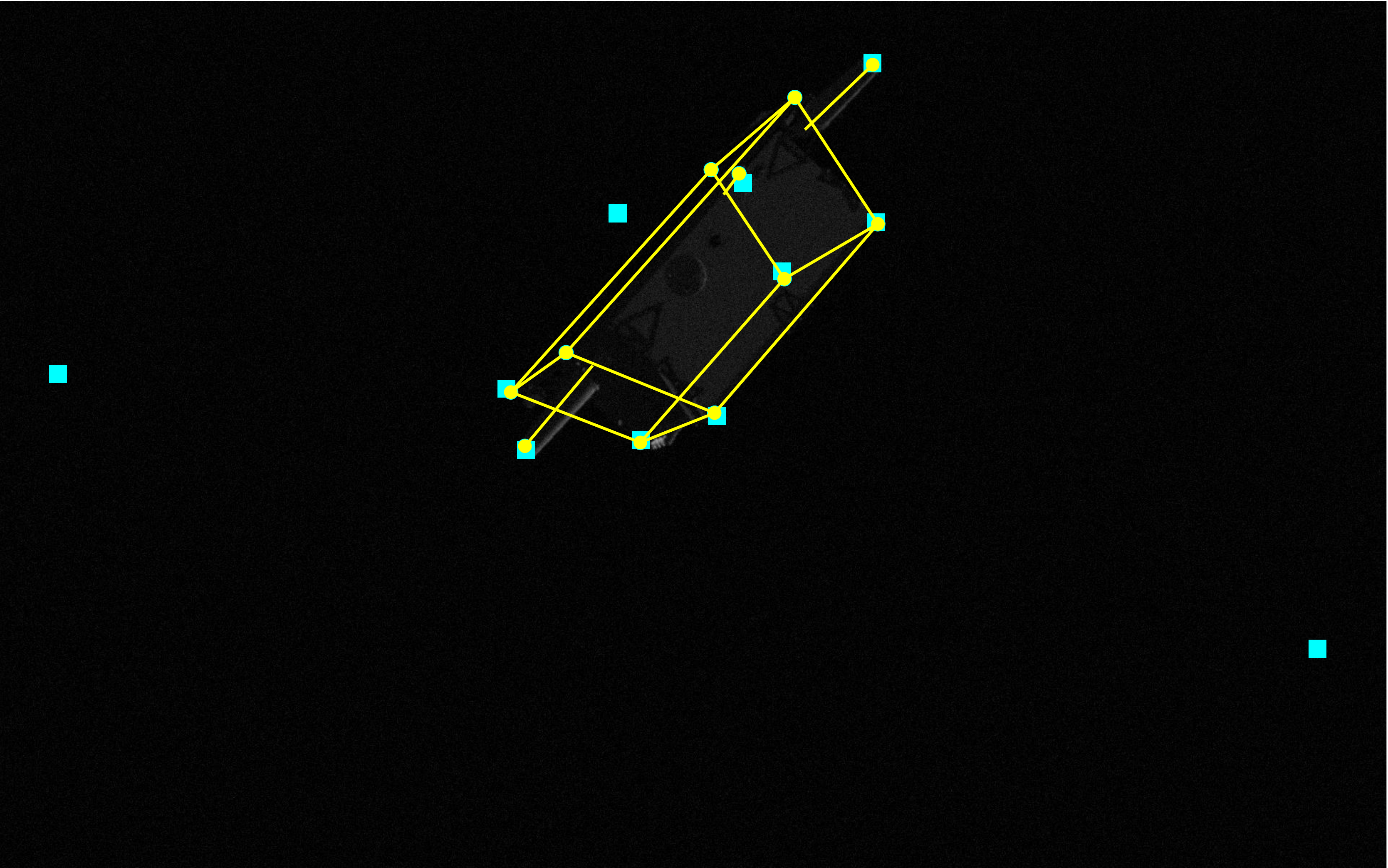}\\
			 $\MR$ error: $1.70^{\circ}$, $\vt$ error: $0.02$ \\ 
			 $\subopt=6.0\ee{-8}$, time: $48$ [s]
			\end{minipage}
		&  
			\begin{minipage}{\mpwfour}%
			\centering%
			\includegraphics[width=\columnwidth]{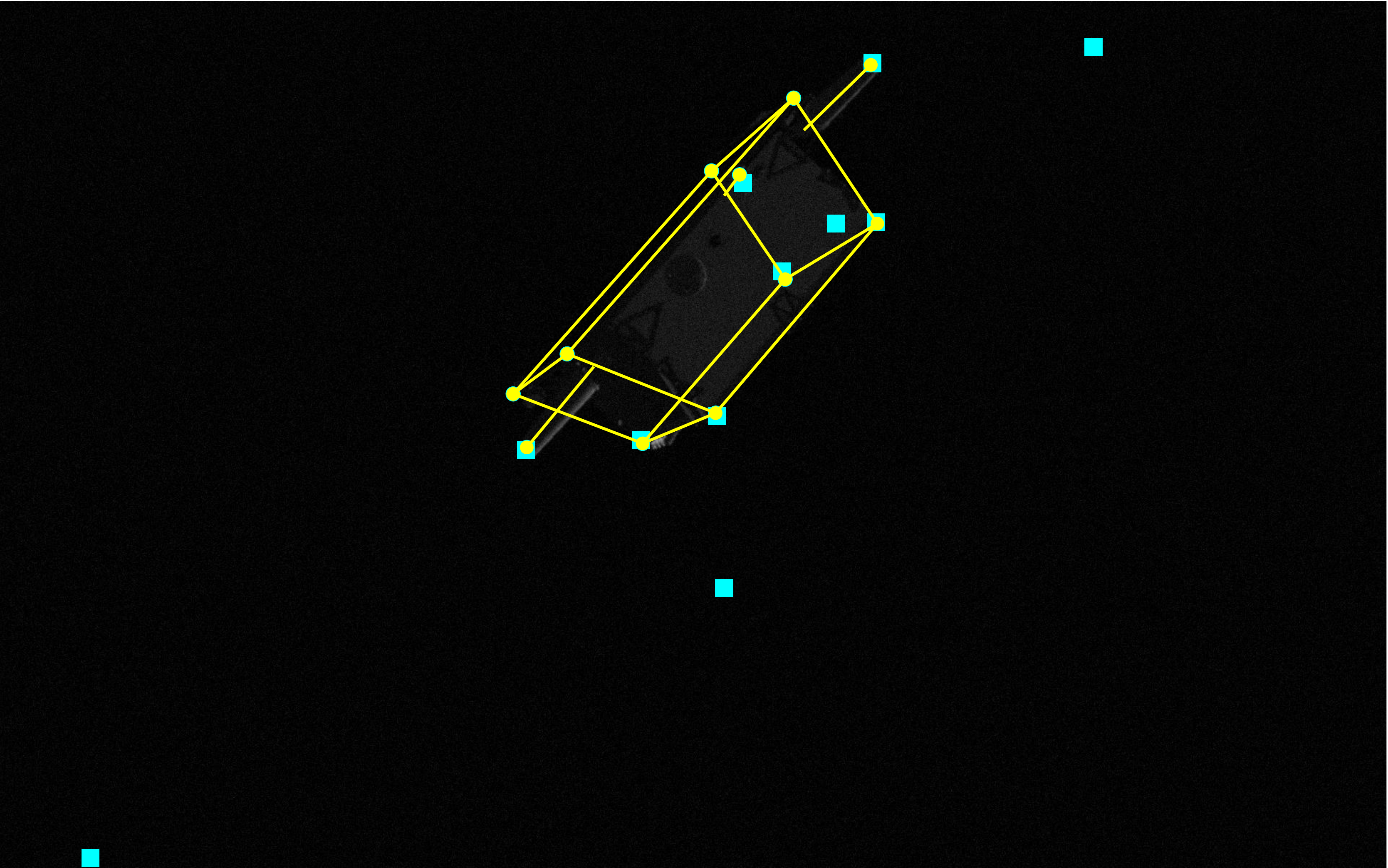}\\
			 $\MR$ error: $1.74^{\circ}$, $\vt$ error: $0.04$ \\ 
			 $\subopt=6.8\ee{-8}$, time: $48$ [s]
			\end{minipage}
		&  
			\begin{minipage}{\mpwfour}%
			\centering%
			\includegraphics[width=\columnwidth]{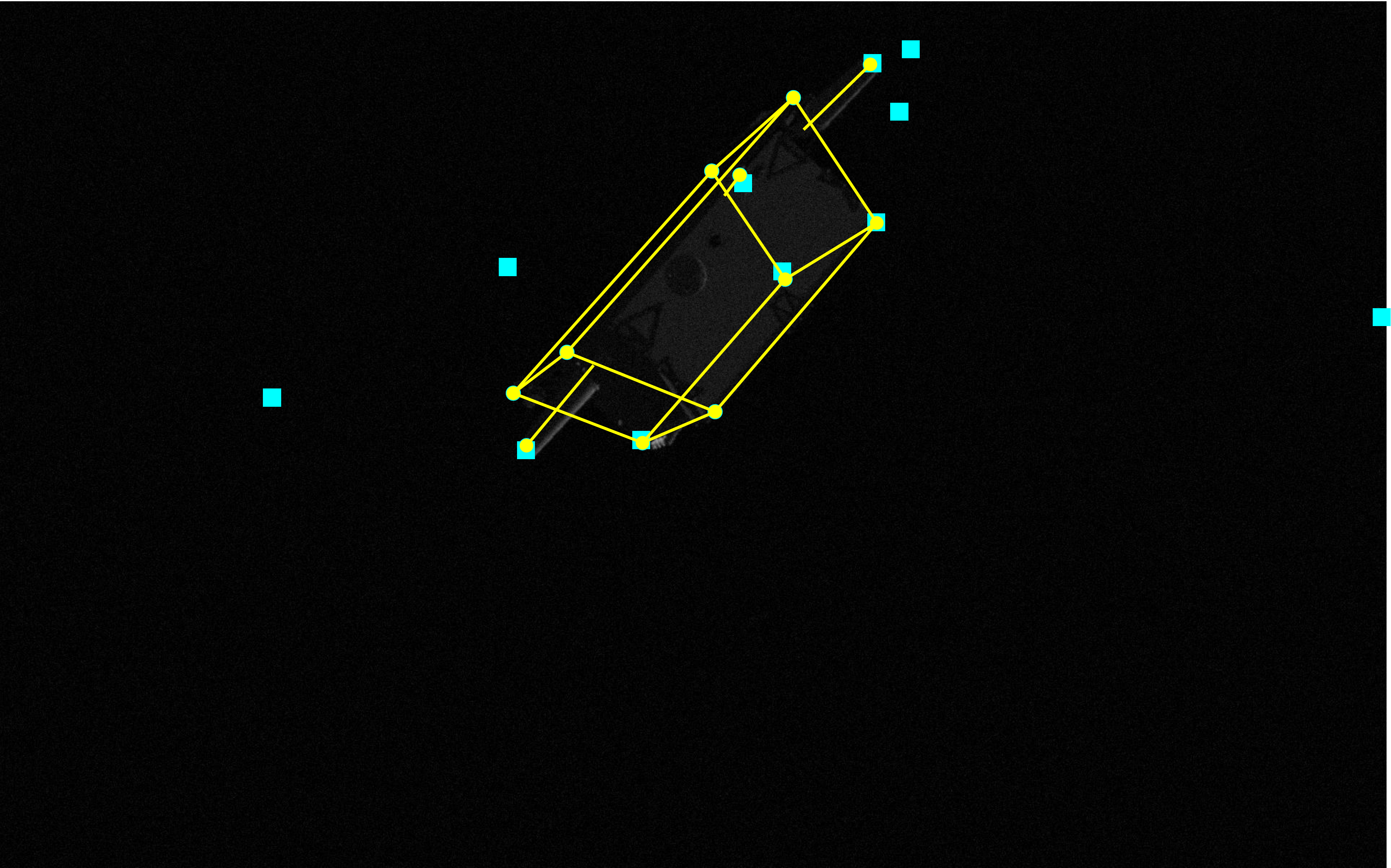}\\
			 $\MR$ error: $1.90^{\circ}$, $\vt$ error: $0.05$ \\ 
			 $\subopt=4.6\ee{-5}$, time: $159$ [s]
			\end{minipage} \vspace{1mm}
		\\

			\begin{minipage}{\mpwfour}%
			\centering%
			\includegraphics[width=\columnwidth]{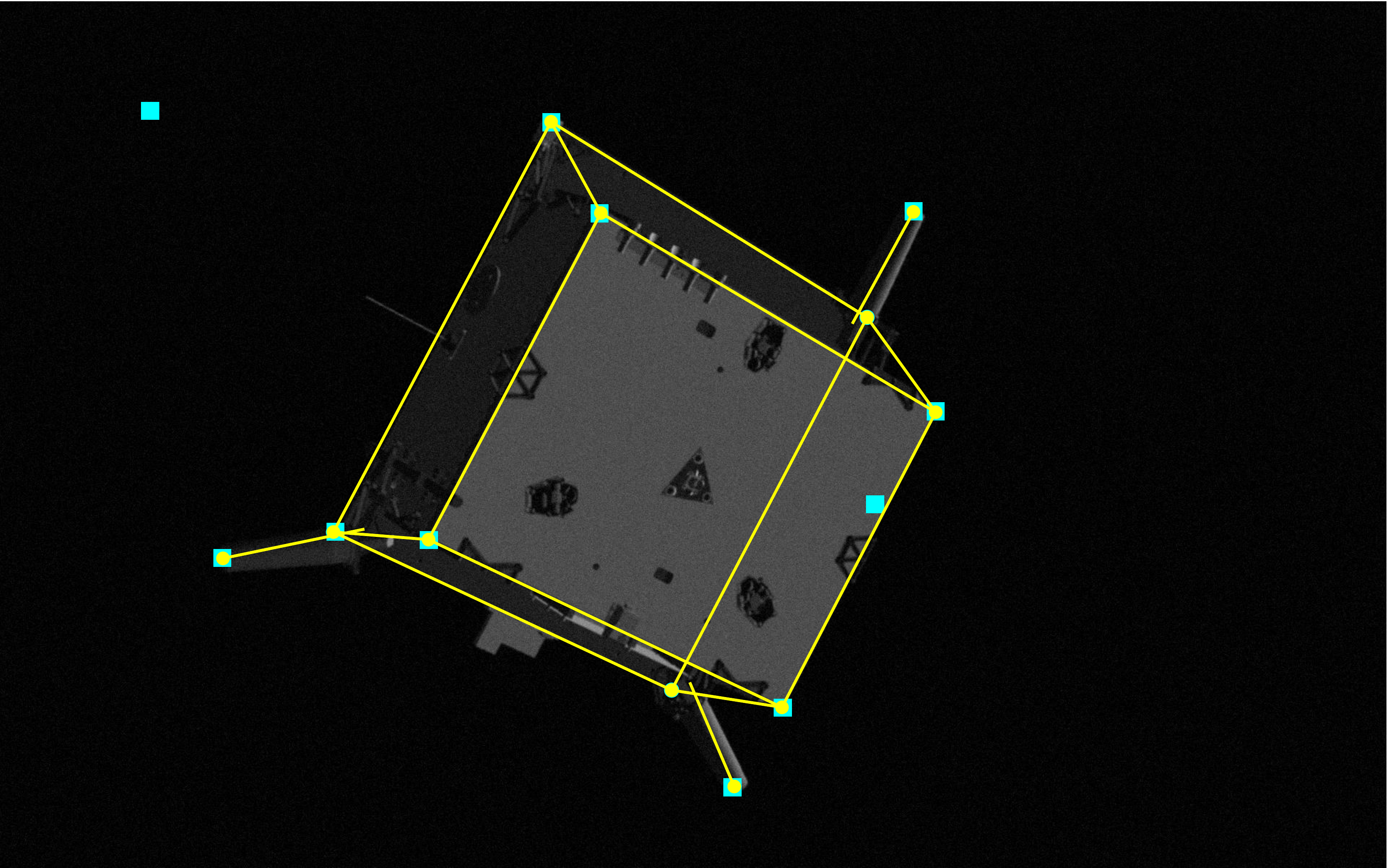}\\
			 $\MR$ error: $0.21^{\circ}$, $\vt$ error: $0.003$ \\ 
			 $\subopt=5.7\ee{-10}$, time: $44$ [s]
			\end{minipage}
		&  
			\begin{minipage}{\mpwfour}%
			\centering%
			\includegraphics[width=\columnwidth]{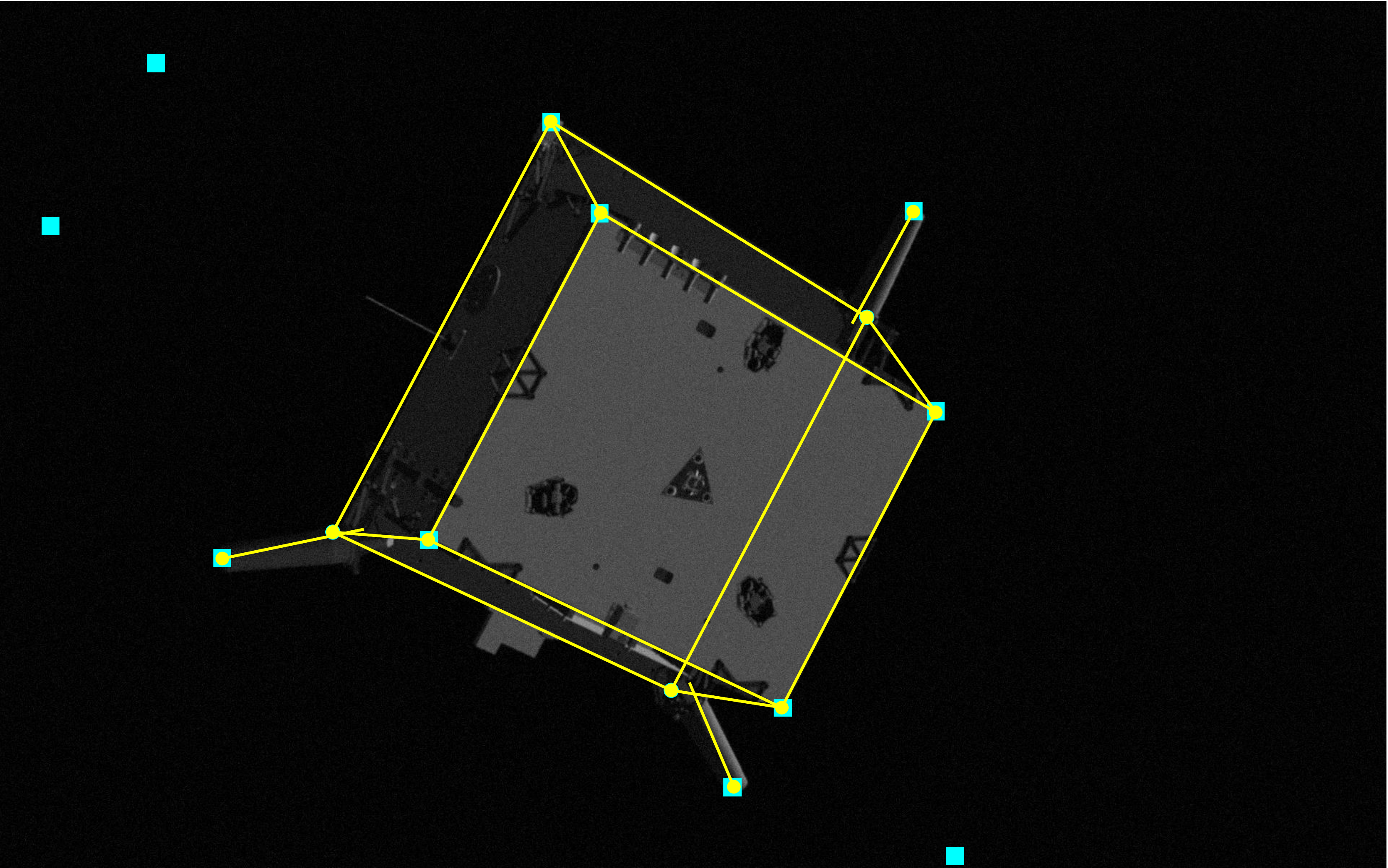}\\
			 $\MR$ error: $0.28^{\circ}$, $\vt$ error: $0.006$ \\ 
			 $\subopt=4.6\ee{-8}$, time: $47$ [s]
			\end{minipage}
		&  
			\begin{minipage}{\mpwfour}%
			\centering%
			\includegraphics[width=\columnwidth]{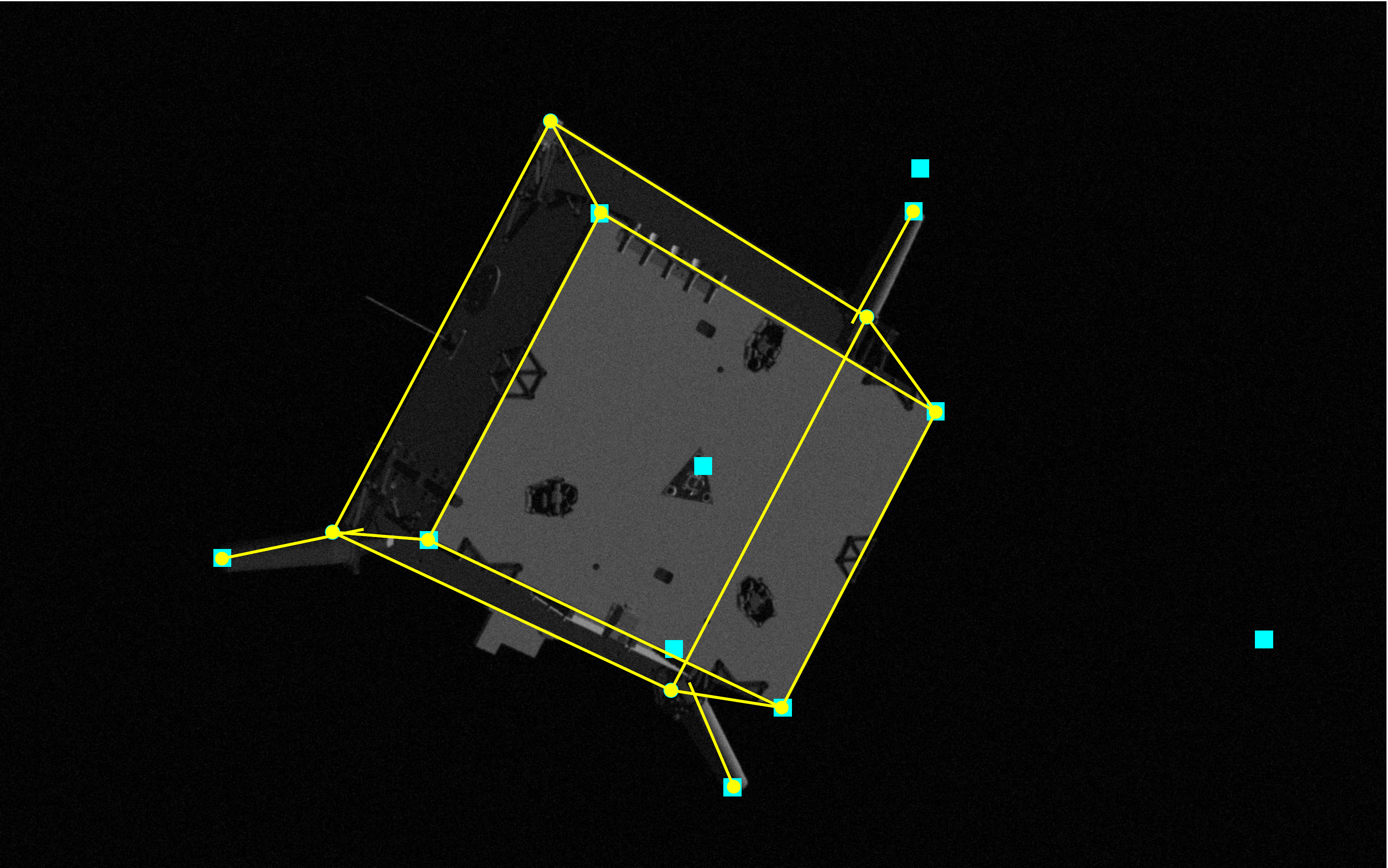}\\
			 $\MR$ error: $0.32^{\circ}$, $\vt$ error: $0.008$ \\ 
			 $\subopt=4.6\ee{-8}$, time: $49$ [s]
			\end{minipage}
		&  
			\begin{minipage}{\mpwfour}%
			\centering%
			\includegraphics[width=\columnwidth]{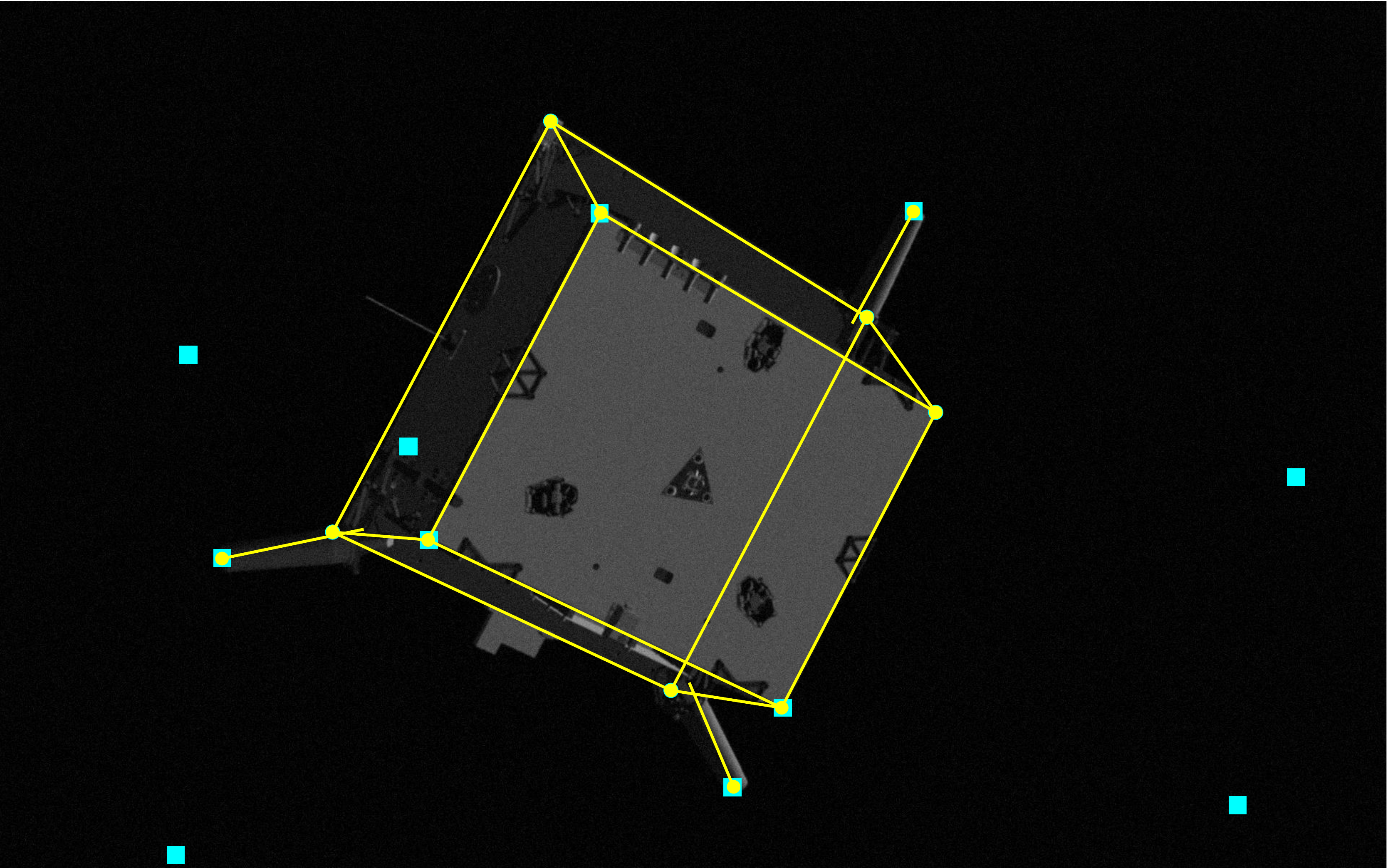}\\
			 $\MR$ error: $0.32^{\circ}$, $\vt$ error: $0.008$ \\ 
			 $\subopt=1.8\ee{-6}$, time: $131$ [s]
			\end{minipage} \vspace{1mm}
		
	\end{tabular}
	\end{minipage}
	\caption{Extra satellite pose estimation results on {\speed}.
	\label{fig:supp-speed-results}} 
	\vspace{-4mm} 
	\end{center}
\end{figure*}

\subsection{Vehicle Pose and Shape Estimation on {\apollo}}

We provide vehicle pose and shape estimation results by {\stride} on the {\apollo} dataset \cite{Wang19pami-apolloscape} in Fig. \ref{fig:supp-apollo}, whose first row also includes the four examples presented in Fig. \ref{fig:exp-catreg-results}(c-1). We provide details of each problem instance such as $N$, $n_1$ and $m$, as well as evaluation metrics such as $(\MR,\vt)$ errors, relative suboptimality $\subopt$, and {\stride}'s computation time. In all cases, {\stride} returned accurate pose and shape estimates with global optimality certificates.

\renewcommand{\mpwfour}{5cm}
\renewcommand{\myhspace}{\hspace{-4mm}}
\begin{figure*}[t]
	\begin{center}
	\begin{minipage}{\textwidth}
	\begin{tabular}{cccc}%
		   \myhspace \hspace{-4mm}
			\begin{minipage}{\mpwfour}%
			\centering%
			\includegraphics[width=0.95\columnwidth]{001_05_59_26_171206_034625454_Camera_5.jpg} \\
			$\MR$ error: $1.75^{\circ}$, $\vt$ error: $0.09$ \\
			$\subopt=1.5\ee{-7}$, time: $4899$ [s] \\
			$n_1 = 1080, m=320,275$\\
			$N=59$, \# outliers: $26$
			\end{minipage}
		&  \myhspace \hspace{-3mm}
			\begin{minipage}{\mpwfour}%
			\centering%
			\includegraphics[width=0.95\columnwidth]{002_03_63_27_171206_035907482_Camera_5.jpg}\\
			$\MR$ error: $0.51^{\circ}$, $\vt$ error: $0.01$ \\
			$\subopt=1.35\ee{-9}$, time: $4174$ [s]\\
			$n_1 = 1152, m=364,037$\\
			$N=63$, \# outliers: $27$
			\end{minipage}
		&  \myhspace \hspace{-3mm}
			\begin{minipage}{\mpwfour}%
			\centering%
			\includegraphics[width=0.95\columnwidth]{003_07_65_30_171206_041139639_Camera_5.jpg}\\
			$\MR$ error: $1.07^{\circ}$, $\vt$ error: $0.02$ \\
			$\subopt=1.4\ee{-10}$, time: $4507$ [s]\\
			$n_1 = 1188, m=386,968$\\
			$N=65$, \# outliers: $30$
			\end{minipage}
		&  \myhspace \hspace{-3mm}
			\begin{minipage}{\mpwfour}%
			\centering%
			\includegraphics[width=0.95\columnwidth]{004_03_66_30_171206_070423775_Camera_5.jpg}\\
			$\MR$ error: $1.74^{\circ}$, $\vt$ error: $0.03$ \\
			$\subopt=1.6\ee{-9}$, time: $4446$ [s]\\
			$n_1 = 1206, m=398,696$\\
			$N=66$, \# outliers: $30$
			\end{minipage}
			\vspace{1mm}
		\\

		   \myhspace \hspace{-4mm}
			\begin{minipage}{\mpwfour}%
			\centering%
			\includegraphics[width=0.95\columnwidth]{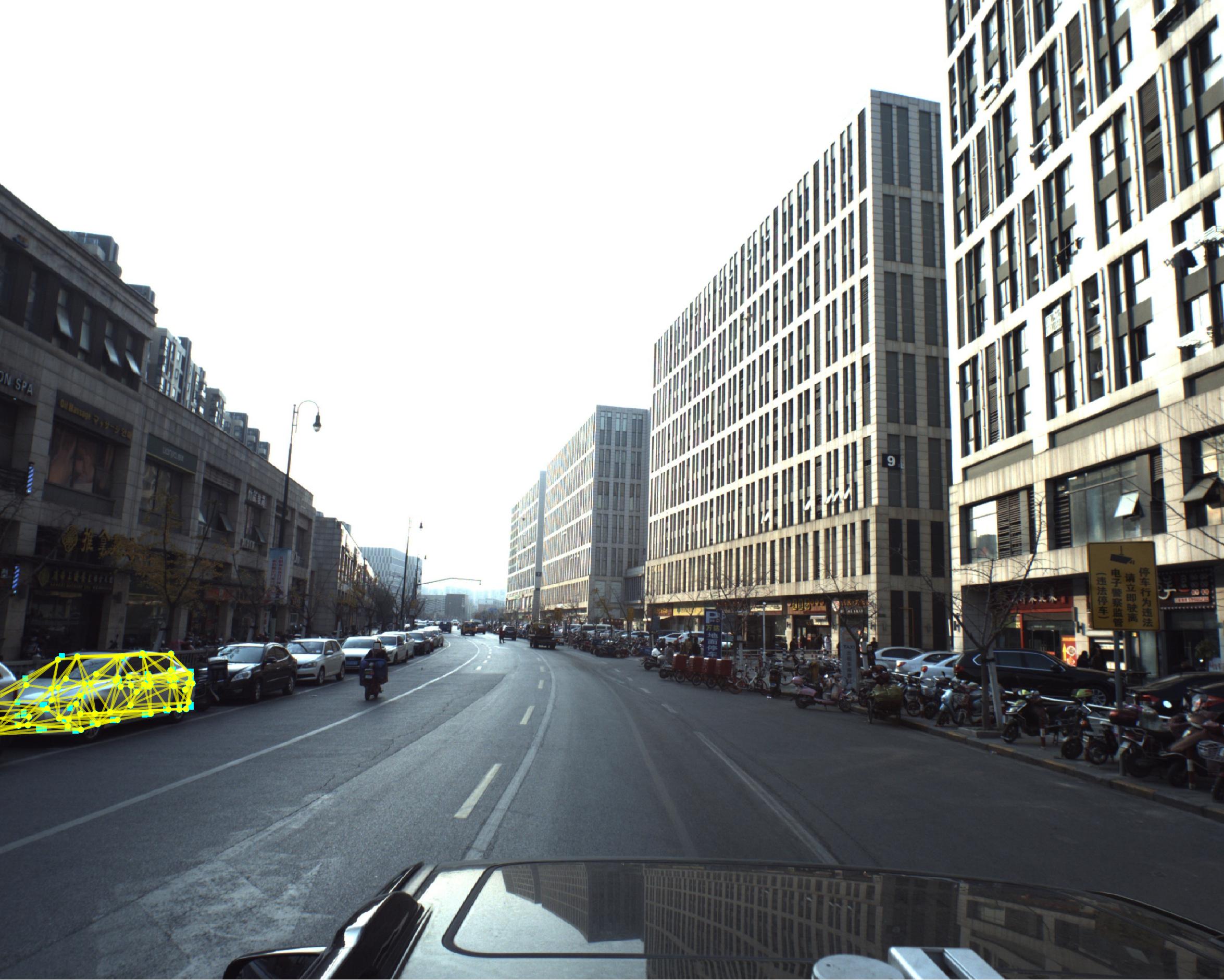} \\
			$\MR$ error: $2.02^{\circ}$, $\vt$ error: $0.04$ \\
			$\subopt=5.6\ee{-10}$, time: $4096$ [s] \\
			$n_1 = 1116, m=341,806$\\
			$N=61$, \# outliers: $29$
			\end{minipage}
		&  \myhspace \hspace{-3mm}
			\begin{minipage}{\mpwfour}%
			\centering%
			\includegraphics[width=0.95\columnwidth]{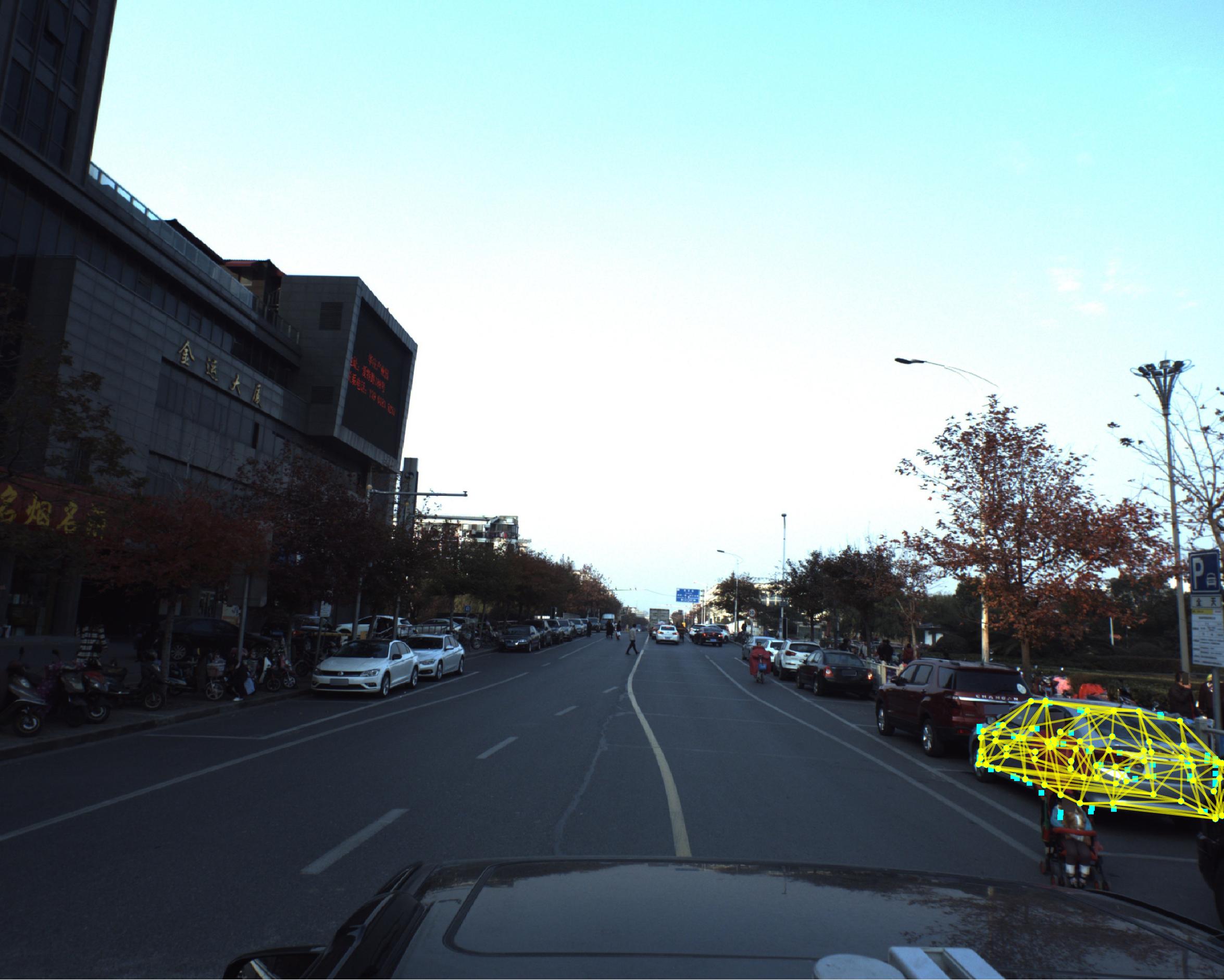}\\
			$\MR$ error: $2.48^{\circ}$, $\vt$ error: $0.05$ \\
			$\subopt=1.9\ee{-5}$, time: $7255$ [s]\\
			$n_1 = 1116, m=341,806$\\
			$N=61$, \# outliers: $25$
			\end{minipage}
		&  \myhspace \hspace{-3mm}
			\begin{minipage}{\mpwfour}%
			\centering%
			\includegraphics[width=0.95\columnwidth]{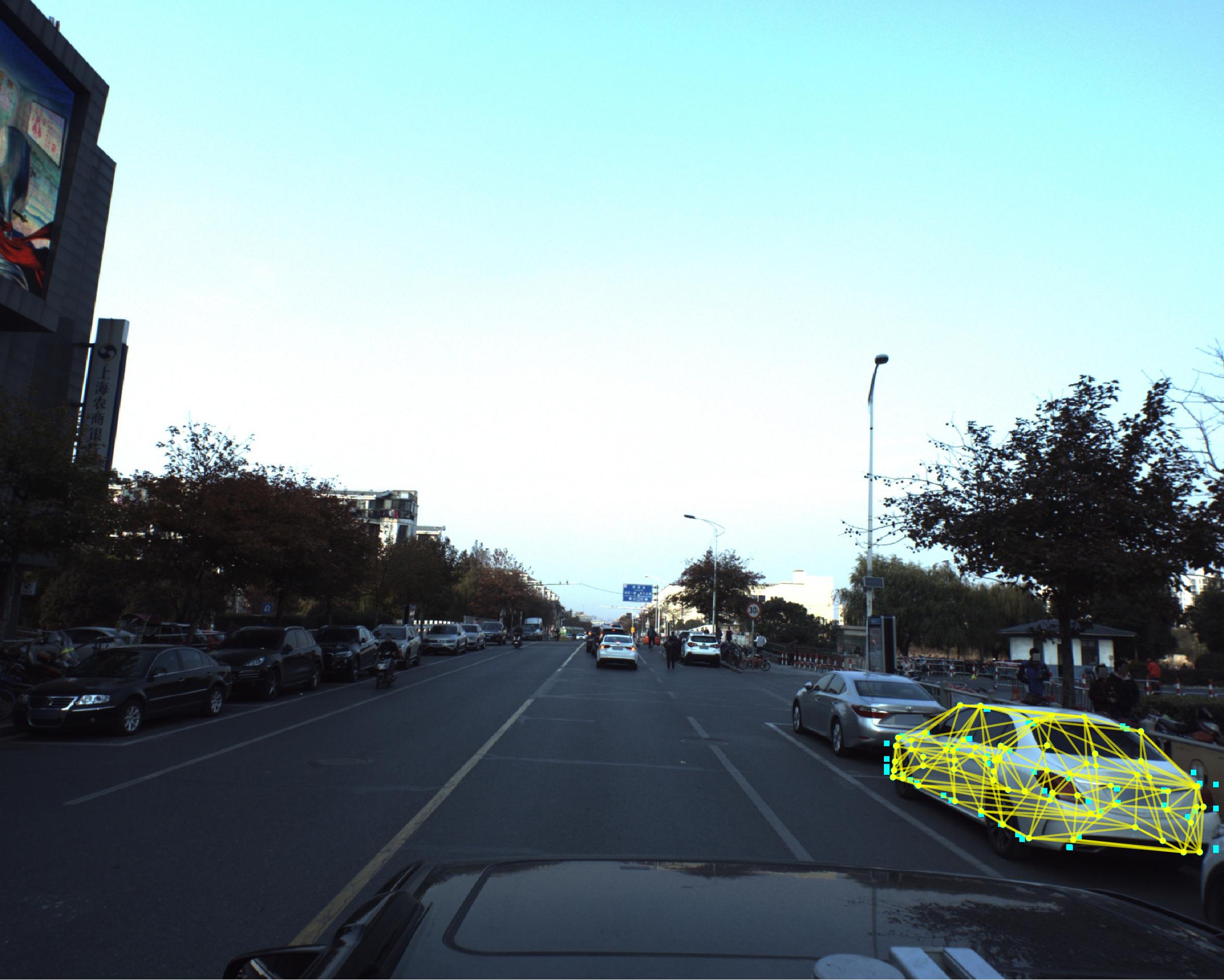}\\
			$\MR$ error: $1.89^{\circ}$, $\vt$ error: $0.05$ \\
			$\subopt=2.0\ee{-7}$, time: $5277$ [s]\\
			$n_1 = 1170, m=375,415$\\
			$N=64$, \# outliers: $26$
			\end{minipage}
		&  \myhspace \hspace{-3mm}
			\begin{minipage}{\mpwfour}%
			\centering%
			\includegraphics[width=0.95\columnwidth]{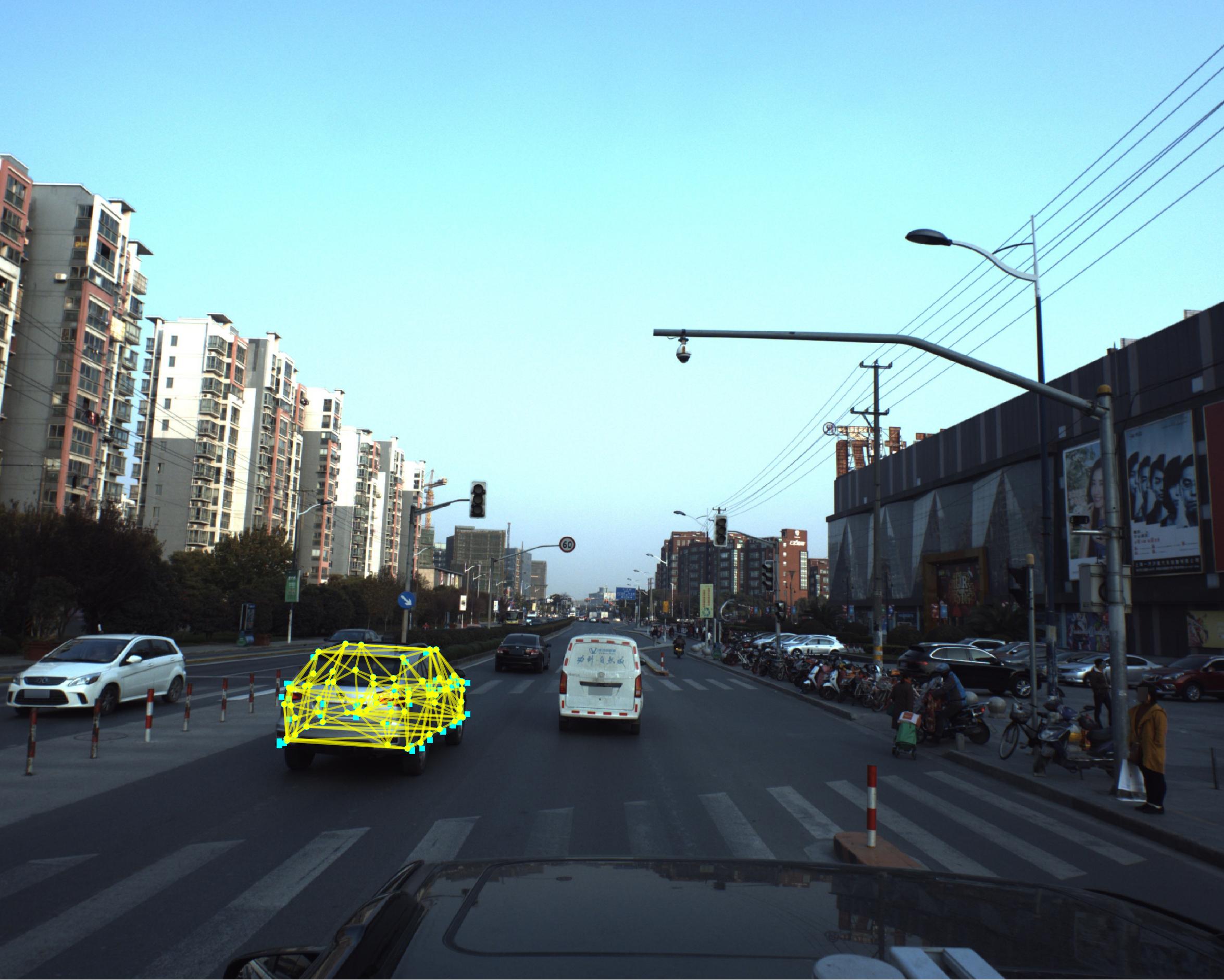}\\
			$\MR$ error: $1.83^{\circ}$, $\vt$ error: $0.04$ \\
			$\subopt=3.5\ee{-9}$, time: $3445$ [s]\\
			$n_1 = 1062, m=309,772$\\
			$N=58$, \# outliers: $23$
			\end{minipage}
			\vspace{1mm}
		\\

		\myhspace \hspace{-4mm}
			\begin{minipage}{\mpwfour}%
			\centering%
			\includegraphics[width=0.95\columnwidth]{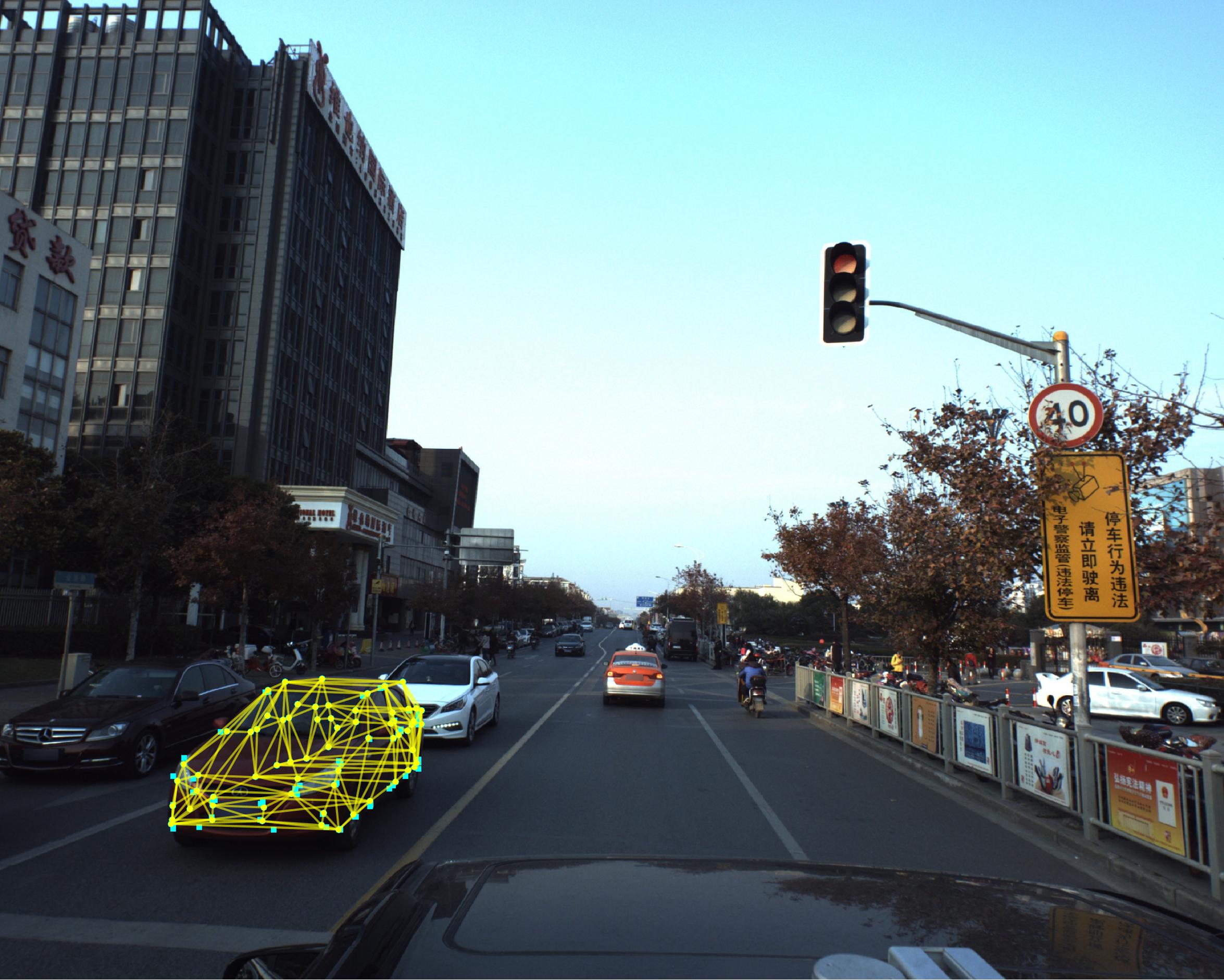}\\
			$\MR$ error: $1.51^{\circ}$, $\vt$ error: $0.03$ \\
			$\subopt=6.6\ee{-10}$, time: $4144$ [s]\\
			$n_1 = 1170, m=375,415$\\
			$N=64$, \# outliers: $29$
			\end{minipage}
		&  \myhspace \hspace{-3mm}
			\begin{minipage}{\mpwfour}%
			\centering%
			\includegraphics[width=0.95\columnwidth]{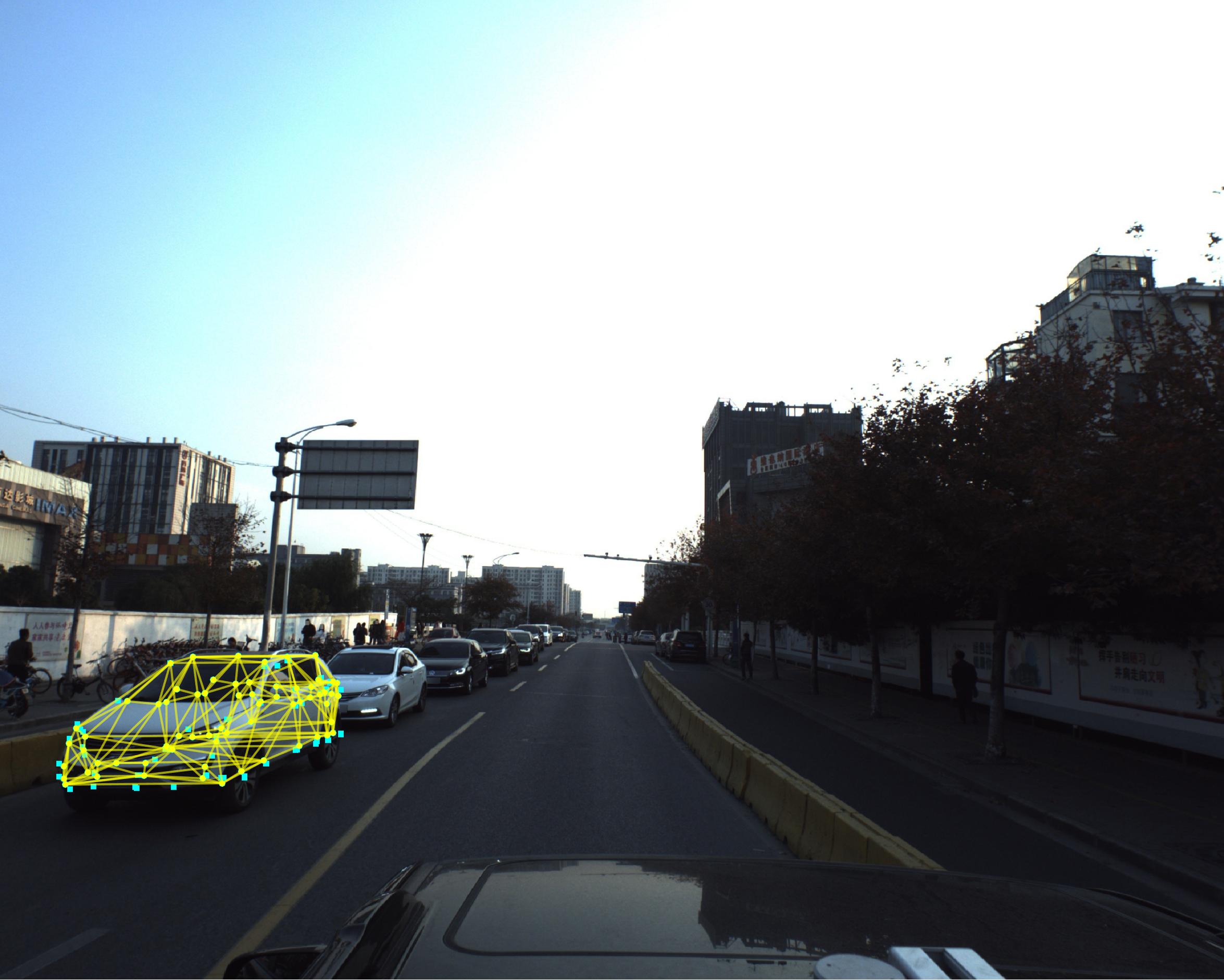}\\
			$\MR$ error: $2.03^{\circ}$, $\vt$ error: $0.09$ \\
			$\subopt=5.1\ee{-10}$, time: $4424$ [s]\\
			$n_1 = 1206, m=398,696$\\
			$N=66$, \# outliers: $30$
			\end{minipage}
		&  \myhspace \hspace{-3mm}
			\begin{minipage}{\mpwfour}%
			\centering%
			\includegraphics[width=0.95\columnwidth]{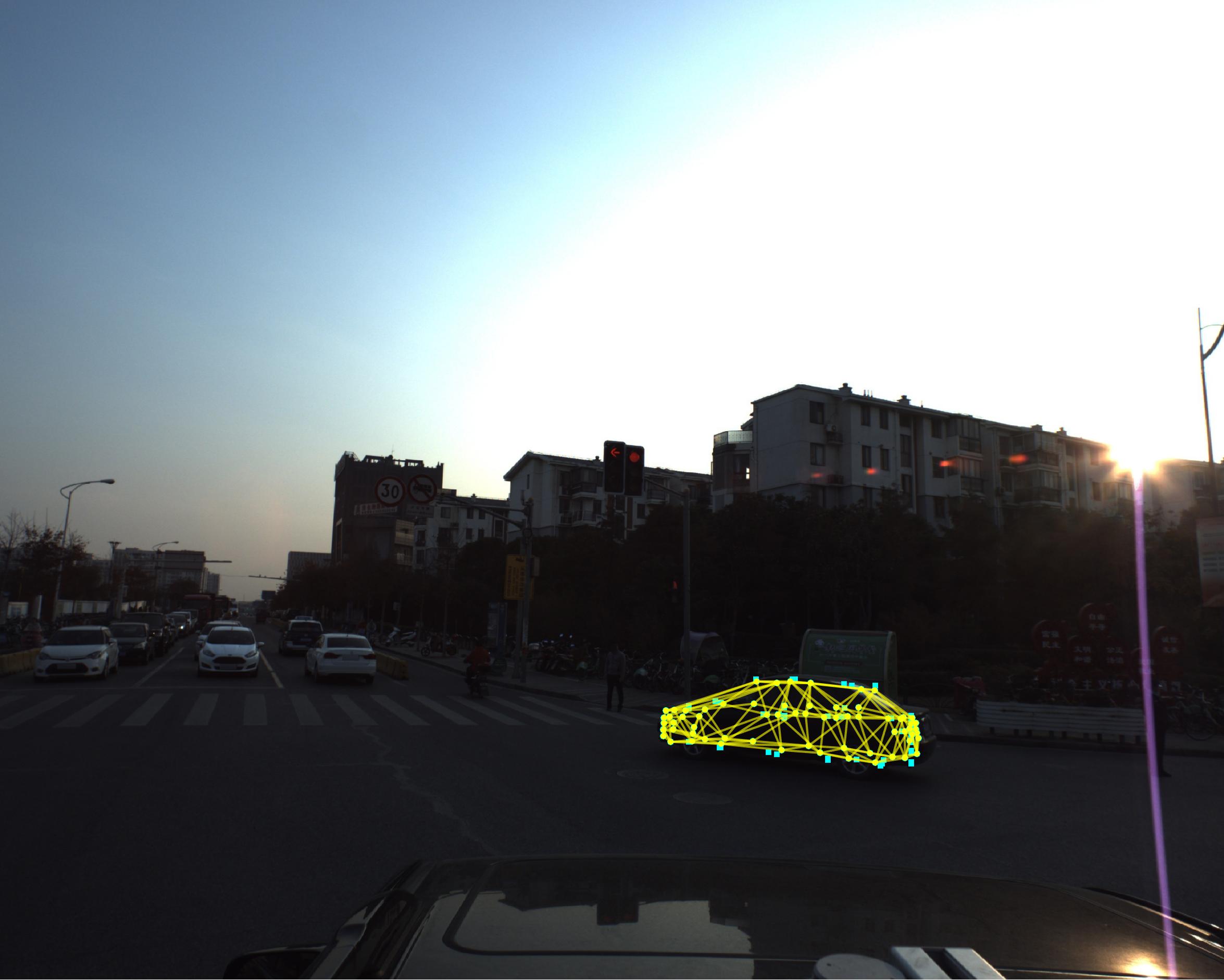}\\
			$\MR$ error: $2.36^{\circ}$, $\vt$ error: $0.08$ \\
			$\subopt=1.1\ee{-3}$, time: $6604$ [s]\\
			$n_1 = 1098, m=330,953$\\
			$N=60$, \# outliers: $23$
			\end{minipage}
		&  \myhspace \hspace{-3mm}
			\begin{minipage}{\mpwfour}%
			\centering%
			\includegraphics[width=0.95\columnwidth]{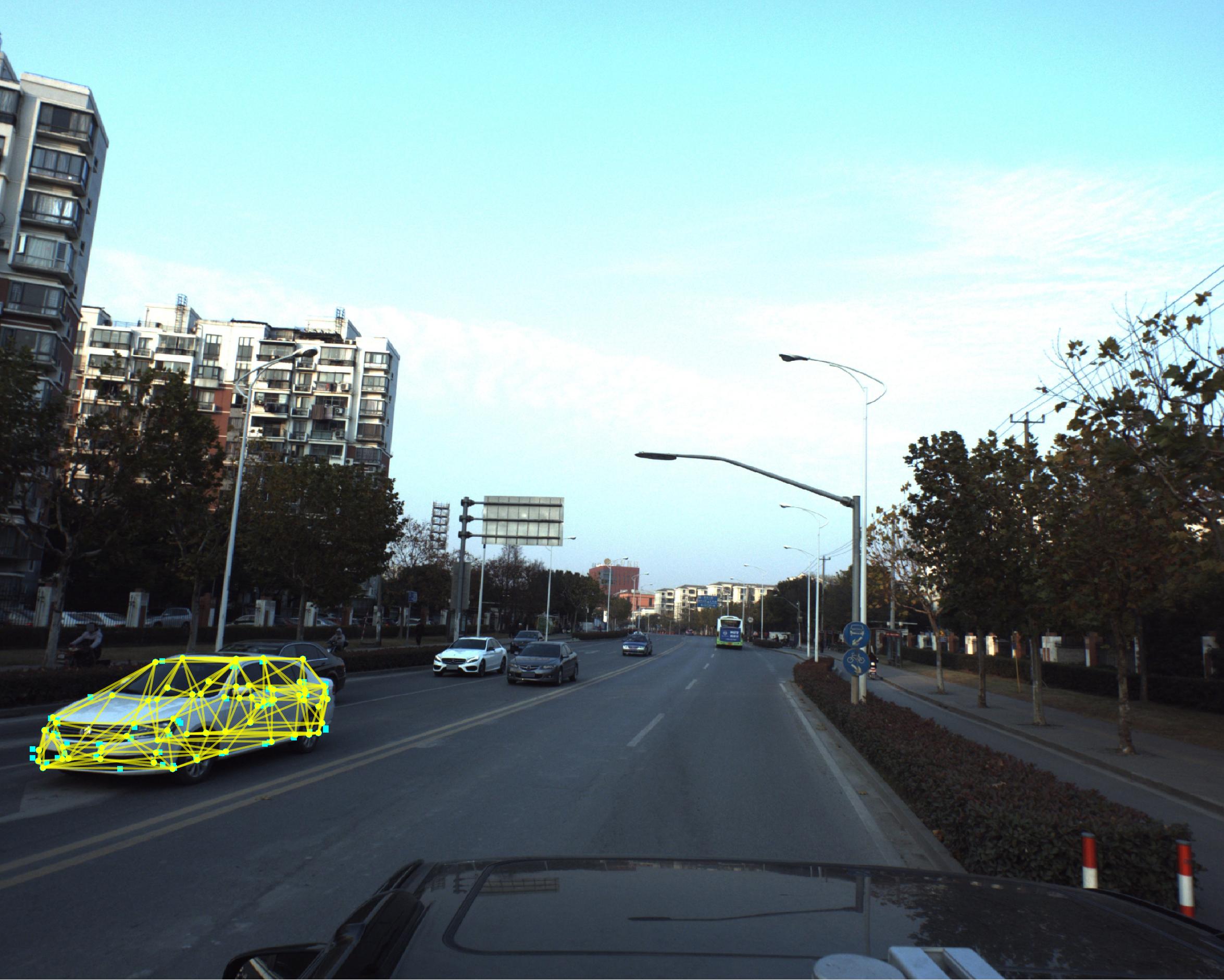}\\
			$\MR$ error: $2.80^{\circ}$, $\vt$ error: $0.09$ \\
			$\subopt=1.5\ee{-9}$, time: $4367$ [s]\\
			$n_1 = 1188, m=386,968$\\
			$N=65$, \# outliers: $28$
			\end{minipage}
			\vspace{1mm}
		\\

		\myhspace \hspace{-4mm}
			\begin{minipage}{\mpwfour}%
			\centering%
			\includegraphics[width=0.95\columnwidth]{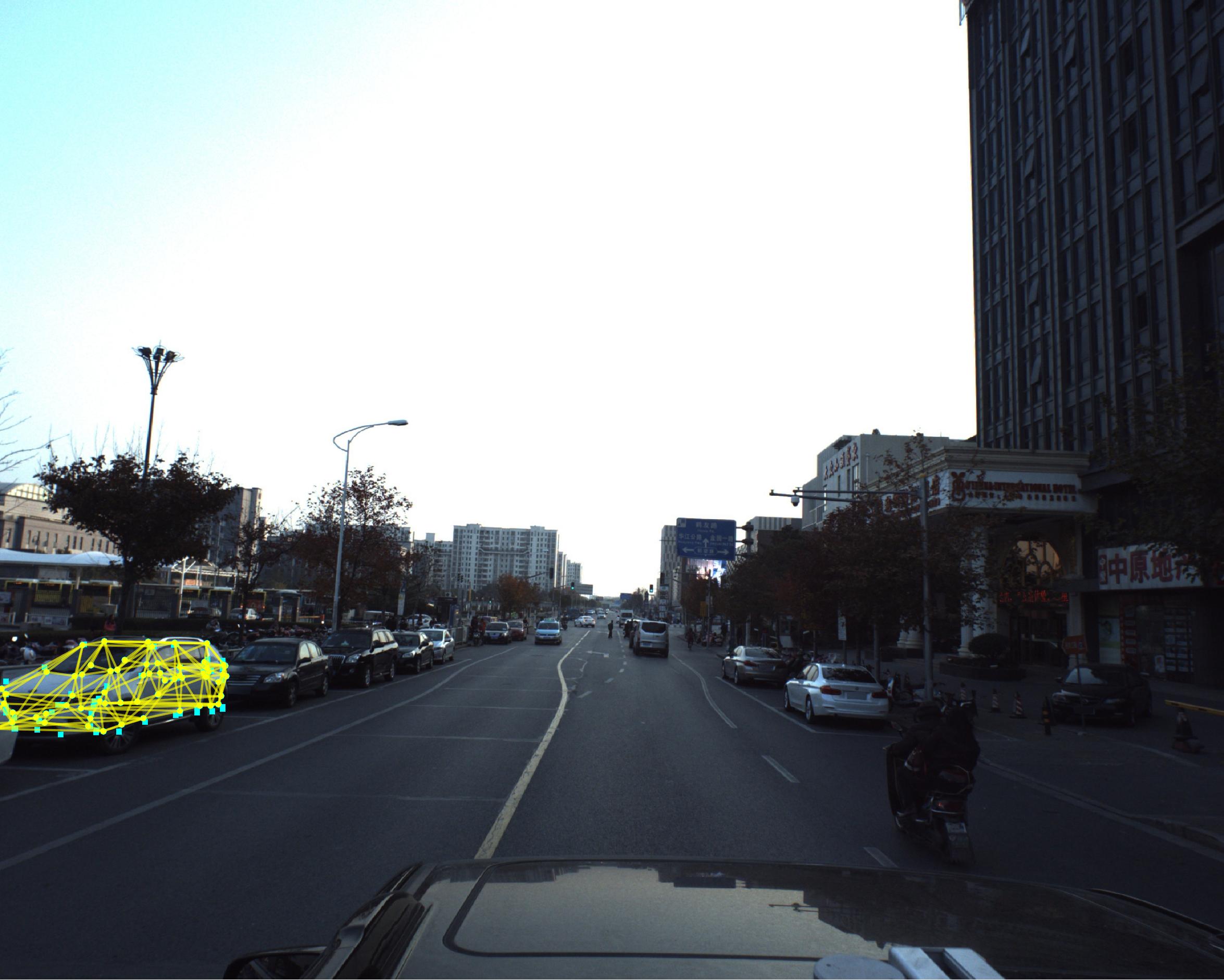}\\
			$\MR$ error: $2.12^{\circ}$, $\vt$ error: $0.10$ \\
			$\subopt=3.2\ee{-10}$, time: $4482$ [s]\\
			$n_1 = 1152, m=364,037$\\
			$N=63$, \# outliers: $28$
			\end{minipage}
		&  \myhspace \hspace{-3mm}
			\begin{minipage}{\mpwfour}%
			\centering%
			\includegraphics[width=0.95\columnwidth]{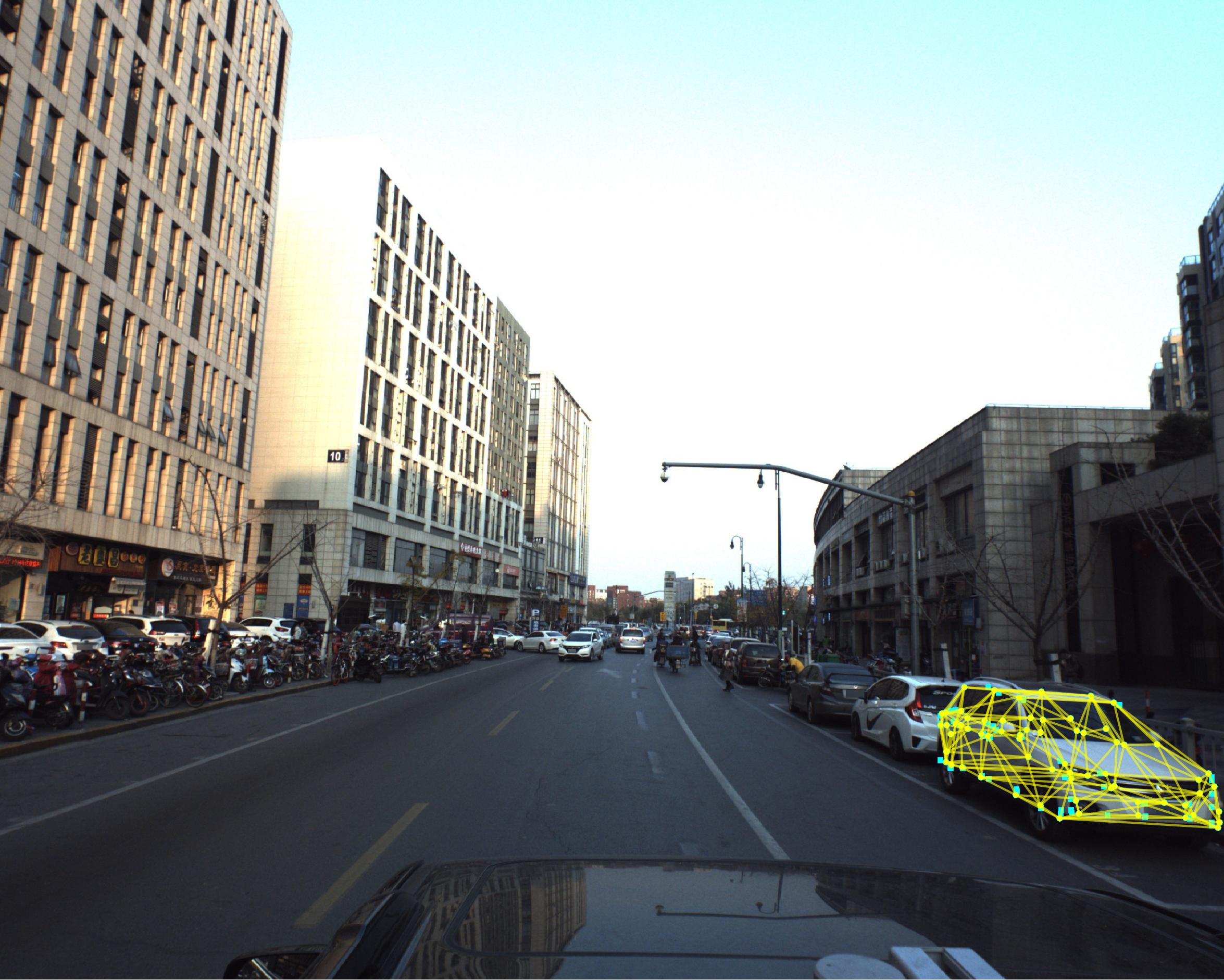}\\
			$\MR$ error: $2.62^{\circ}$, $\vt$ error: $0.04$ \\
			$\subopt=1.2\ee{-10}$, time: $4524$ [s]\\
			$n_1 = 1152, m=364,037$\\
			$N=63$, \# outliers: $29$
			\end{minipage}
		&  \myhspace \hspace{-3mm}
			\begin{minipage}{\mpwfour}%
			\centering%
			\includegraphics[width=0.95\columnwidth]{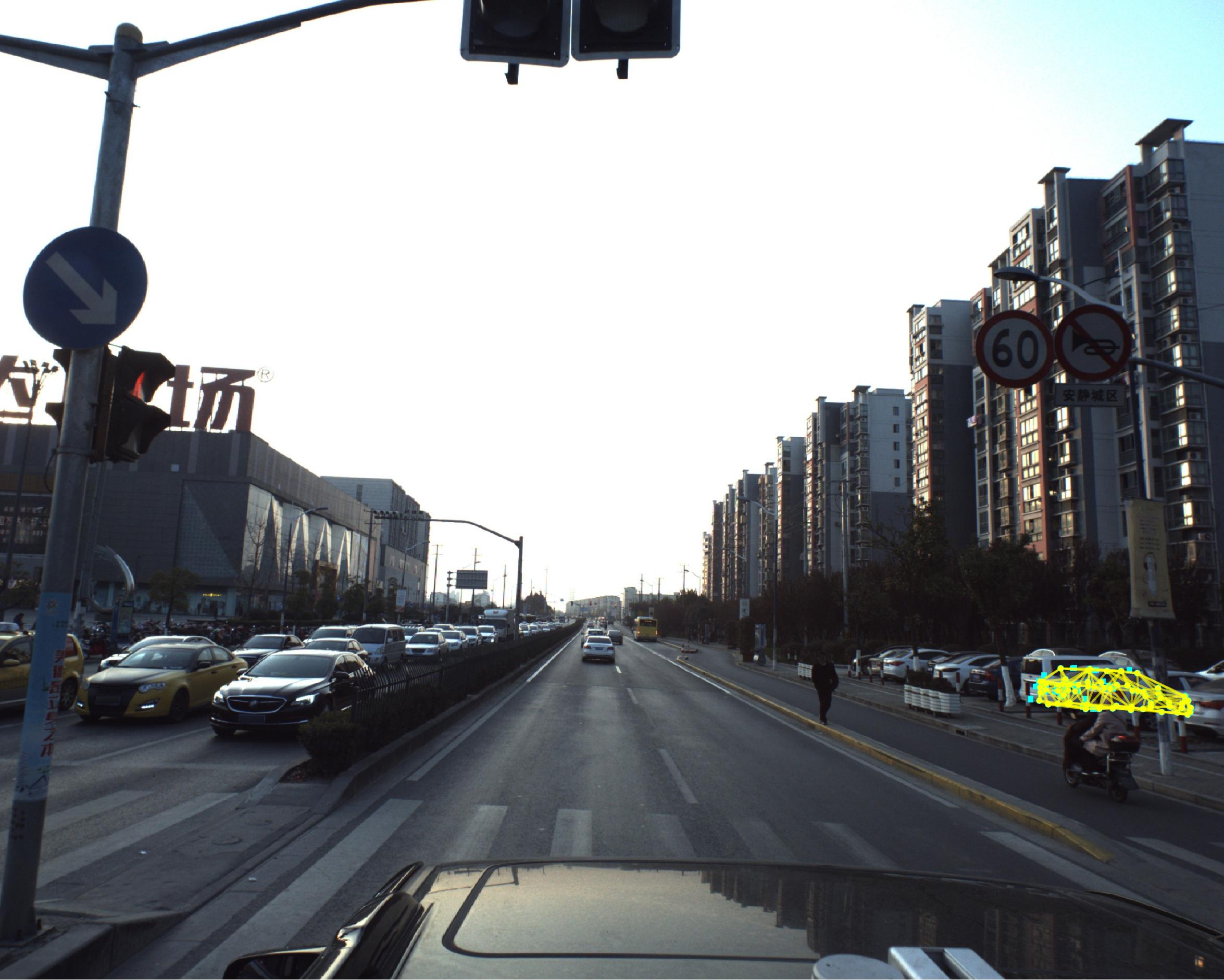}\\
			$\MR$ error: $2.62^{\circ}$, $\vt$ error: $0.07$ \\
			$\subopt=1.8\ee{-7}$, time: $2344$ [s]\\
			$n_1 = 684, m=129,634$\\
			$N=37$, \# outliers: $16$
			\end{minipage}
		&  \myhspace \hspace{-3mm}
			\begin{minipage}{\mpwfour}%
			\centering%
			\includegraphics[width=0.95\columnwidth]{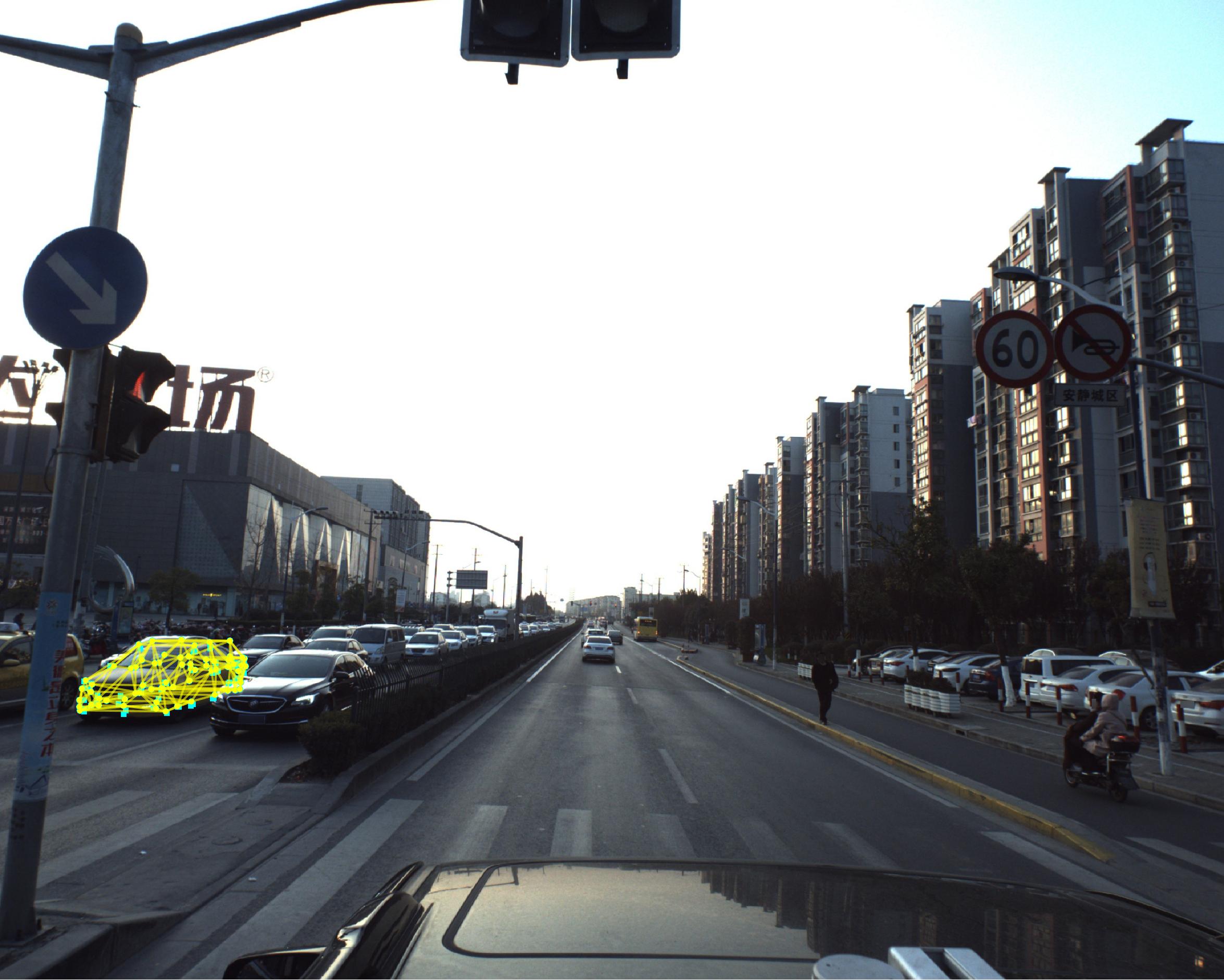}\\
			$\MR$ error: $2.19^{\circ}$, $\vt$ error: $0.09$ \\
			$\subopt=3.1\ee{-10}$, time: $5090$ [s]\\
			$n_1 = 1170, m=375,415$\\
			$N=64$, \# outliers: $28$
			\end{minipage}
	\end{tabular}
	\end{minipage} 
	\caption{Extra vehicle pose and shape estimation on {\apollo} \cite{Wang19pami-apolloscape}.
	\label{fig:supp-apollo}} 
	\vspace{-4mm} 
	\end{center}
\end{figure*}

\bibliographystyle{ieee}
\bibliography{myRefs,../../../references/refs}

%
\begin{IEEEbiography}[{\includegraphics[width=1in,height=1.25in,keepaspectratio]{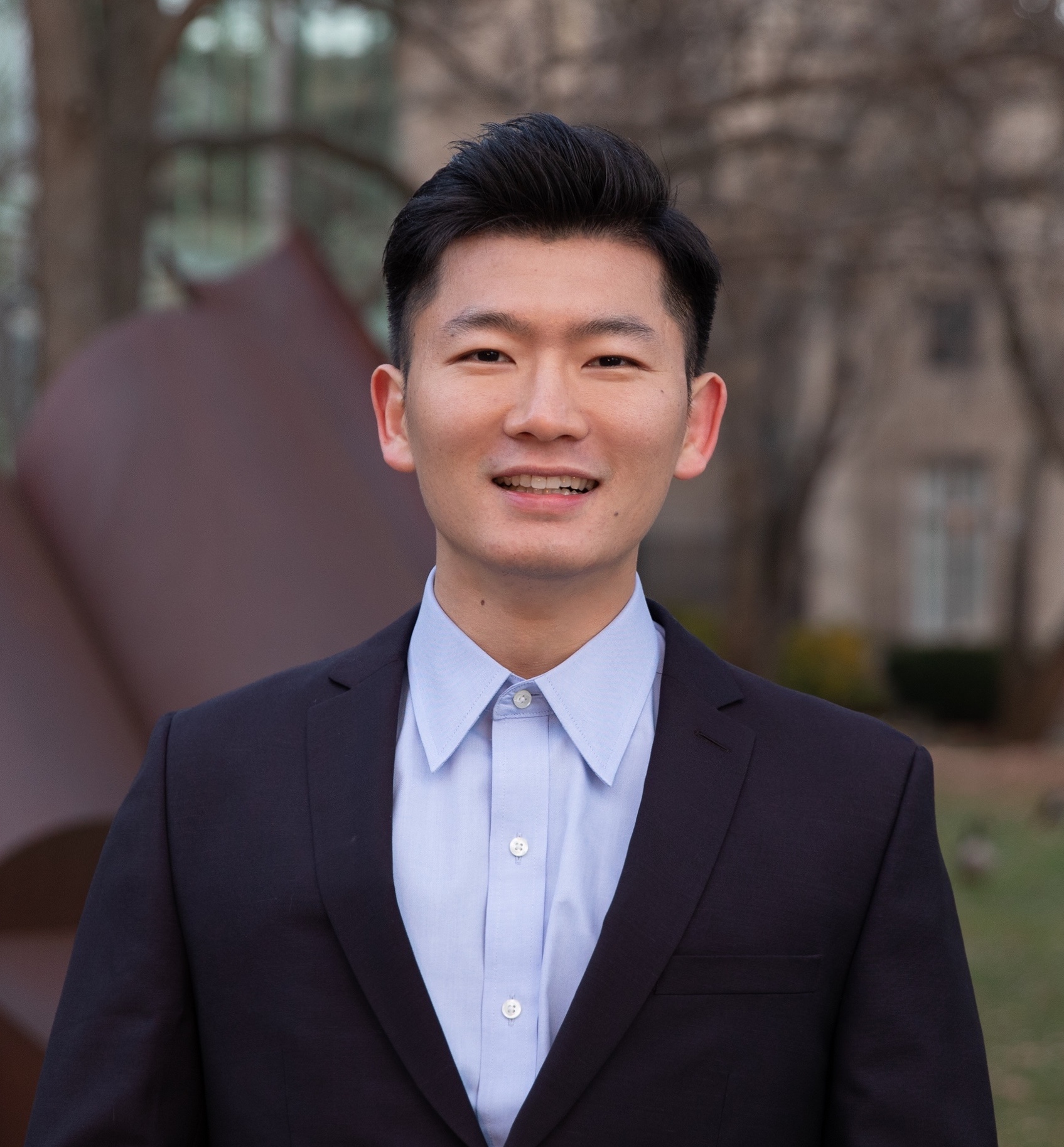}}]{Heng Yang} is a Ph.D.~candidate in the Laboratory for
  Information \& Decision Systems at the
  Massachusetts Institute of Technology. He has obtained a B.S.~degree in Mechanical
  Engineering (with honors) from the Tsinghua University, Beijing, China, in 2015; and an S.M.~degree
  in Mechanical Engineering from MIT in 2017.
  His research interests include large-scale convex optimization, semidefinite relaxation, robust estimation, and machine learning, applied to computer vision, robotics, and trustworthy autonomy. 
  Heng Yang is a recipient of the Best Paper Award in Robot Vision at the 2020 IEEE International Conference on Robotics and Automation (ICRA), a Best Paper Award Honorable Mention from the 2020 IEEE Robotics and Automation Letters (RA-L), and a Best Paper Award Finalist at the 2021 Robotics: Science and Systems (RSS) conference. He is a Class of 2021 RSS Pioneer. 
\end{IEEEbiography}
\begin{IEEEbiography}[{\includegraphics[width=1in,height=1.25in,clip,keepaspectratio]{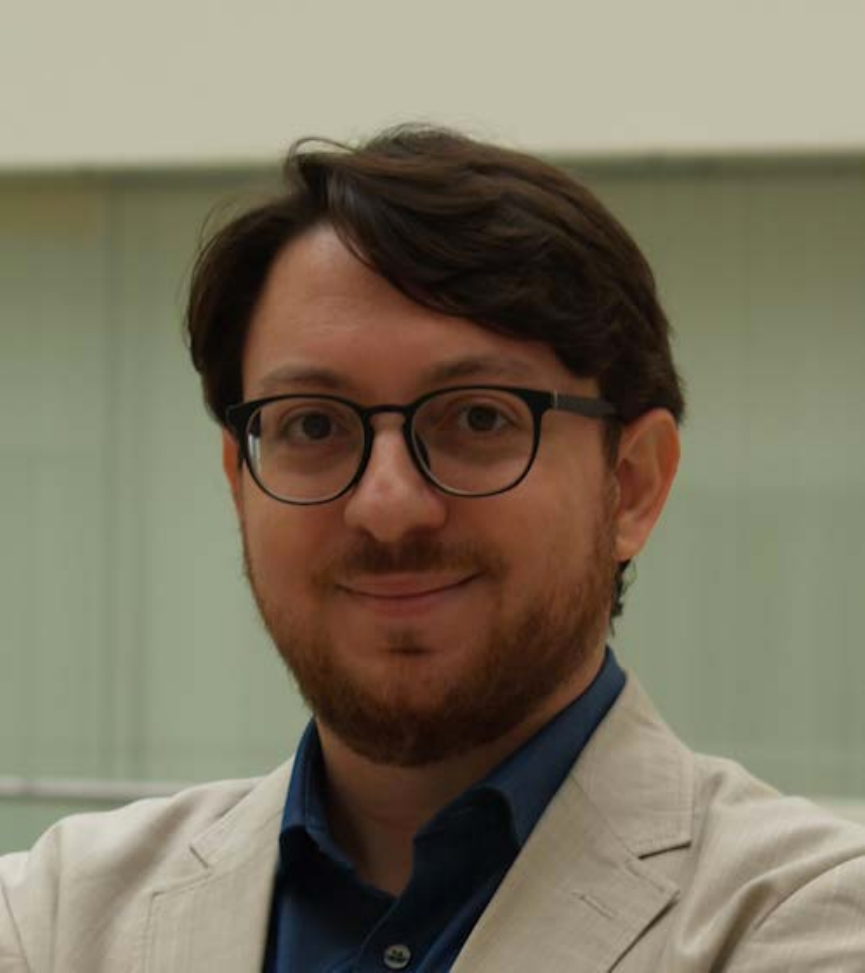}}]{Luca Carlone}
  is the Leonardo Career Development Associate Professor in the Department of Aeronautics and Astronautics at the Massachusetts Institute of Technology, and a Principal Investigator in the Laboratory for Information \& Decision Systems (LIDS).
  He has obtained a B.S. degree in mechatronics from the Polytechnic University of Turin, Italy, in 2006;
  an S.M. degree in mechatronics from the Polytechnic University of Turin, Italy, in 2008;
  an S.M. degree in automation engineering from the Polytechnic University of Milan, Italy, in 2008;
  and a Ph.D. degree in robotics also from the Polytechnic University of Turin in 2012.
  He joined LIDS as a postdoctoral associate (2015) and later as a Research Scientist (2016), after spending two years as a postdoctoral fellow at the Georgia Institute of Technology (2013-2015).
  His research interests include nonlinear estimation, numerical and distributed optimization, and probabilistic inference, applied to sensing, perception, and decision-making in single and multi-robot systems. His work includes seminal results on certifiably correct algorithms for localization and mapping, as well as approaches for visual-inertial navigation and distributed mapping.
  He is a recipient of the Best Student Paper Award at IROS 2021, the Best Paper Award in Robot Vision at ICRA 2020, a 2020 Honorable Mention from the IEEE Robotics and Automation Letters, a Track Best Paper award at the 2021 IEEE Aerospace Conference, the 2017 Transactions on Robotics King-Sun Fu Memorial Best Paper Award, the Best Paper Award at WAFR 2016, the Best Student Paper Award at the 2018 Symposium on VLSI Circuits, and he was best paper finalist at RSS 2015 and RSS 2021. He is also a recipient of the AIAA Aeronautics and Astronautics Advising Award (2022), the NSF CAREER Award (2021), the RSS Early Career Award (2020), the Google Daydream (2019) and the Amazon Research Award (2020, 2022), and the MIT AeroAstro Vickie Kerrebrock Faculty Award (2020).
\end{IEEEbiography}









\end{document}